\documentclass{article} 
\pdfoutput=1
\usepackage{iclr2026_conference, times}
\iclrfinalcopy

\usepackage[utf8]{inputenc} 
\usepackage[T1]{fontenc}    
\usepackage{hyperref}       
\usepackage{url}            
\usepackage{booktabs}       
\usepackage{amsfonts}       
\usepackage{nicefrac}       
\usepackage{microtype}      
\usepackage{xcolor}         
\usepackage{graphicx}       
\usepackage{subcaption}     
\usepackage{float}          
\usepackage{amsmath}
\usepackage{cleveref}
\usepackage{paralist}
\usepackage{xspace}
\usepackage{listings}
\usepackage[most]{tcolorbox}  
\usepackage{listingsutf8}
\usepackage{wrapfig}
\usepackage{titlesec}
\usepackage{caption}
\usepackage{tabularray}
\usepackage{booktabs}
\usepackage{CJKutf8}

\usepackage{titlesec}
\titlespacing*{\section}{0pt}{6pt}{4pt}
\titlespacing*{\subsection}{0pt}{4pt}{2pt}

\everypar{\looseness=-1}

\titlespacing{\paragraph}
{0pt}     
{0ex}     
{0.5em}     

\usepackage{etoolbox}
\newcommand{\zerodisplayskips}{%
    \setlength{\abovedisplayskip}{0.9em}%
    \setlength{\belowdisplayskip}{0.9em}}%
\appto{\normalsize}{\zerodisplayskips}
\appto{\small}{\zerodisplayskips}
\appto{\footnotesize}{\zerodisplayskips}

\usepackage{bbm}

\definecolor{dandelion}{HTML}{FFD464}

\definecolor{bittersweet}{HTML}{C04F17}

\definecolor{mintgreen}{RGB}{152, 255, 152}

\definecolor{lavendel}{RGB}{230,230,250}

\newcommand{\organismtype}[1]{#1}
\newcommand{\organism}[1]{{\footnotesize \textsc{#1}}}
\newcommand{\llm}[1]{{\footnotesize \texttt{#1}}}
\newcommand{\tokenstr}[1]{\texttt{#1}}
\newcommand{\ADL}{ADL\xspace}

\newcommand{\lm}{p}
\newcommand{\lmbase}{\lm^{\text{base}}}
\newcommand{\lmft}{\lm^{\text{ft}}}
\newcommand{\ds}{\mathcal{D}}
\newcommand{\dsft}{\ds^{\text{ft}}}
\newcommand{\dspt}{\ds^{\text{pt}}}

\newcommand{\token}[1]{x_{#1}}
\newcommand{\vh}{\mathbf{h}}

\newcommand{\hdiff}{\boldsymbol{\delta}}

\newcommand{\projection}{\mathbf{P}}
\newcommand{\hdiffavg}{\overline{\hdiff}}
\newcommand{\hbase}{\vh^{\text{base}}}
\newcommand{\norm}[1]{\lvert\lvert#1\rvert\rvert}
\newcommand{\CEloss}{\mathcal{L}_{\text{CE}}}
\newcommand{\causaleffect}{\Delta_{\CEloss}}
\newcommand{\hbaseavg}{\overline{\vh}^{\text{base}}}
\newcommand{\hft}{\vh^{\text{ft}}}
\newcommand{\hftavg}{\overline{\vh}^{\text{ft}}}
\newcommand{\R}{\mathbb{R}}
\newcommand{\layer}{\ell}
\newcommand{\layers}{L}
\newcommand{\softmax}{\text{softmax}}
\newcommand{\normft}{\eta^\text{ft}}

\floatstyle{ruled}
\newfloat{prompt}{tbp}{lop}
\floatname{prompt}{Prompt}
\lstdefinestyle{promptstyle}{
  basicstyle=\ttfamily\tiny,
  breaklines=true,
  columns=fullflexible,
  frame=single,
  framerule=0.4pt,
  framesep=4pt,
  numbers=none,
  inputencoding=utf8,
  literate=
    {→}{{$\rightarrow$}}1
    {—}{{---}}1
    {–}{{--}}1
    {…}{{$\ldots$}}1
    {’}{{'}}1
    {“}{{``}}1
    {”}{{''}}1
    {‐}{{-}}1
    {、}{{,}}1
    {精}{{?}}1
    {�}{{?}}1
    {​}{{}}0
    {▁}{{$\underline{}$}}1
    {↔}{{$\leftrightarrow$}}1
    {≤}{{$\leq$}}1
    {≥}{{$\geq$}}1
    {≠}{{$\neq$}}1
    {≈}{{$\approx$}}1
    {≡}{{$\equiv$}}1
    {≡}{{$\equiv$}}1
}

\newcommand{\promptlisting}[3]{%
  \captionof{prompt}{#2}\label{#3}%
  \begin{tcolorbox}[enhanced,breakable,
    colback=white, colframe=black!60, boxrule=0.4pt,
    left=6pt,right=6pt,top=6pt,bottom=6pt]
    \lstinputlisting[style=promptstyle]{#1}
  \end{tcolorbox}%
}
\crefname{prompt}{prompt}{prompts}
\Crefname{prompt}{Prompt}{Prompts}

\makeatletter
\DeclareRobustCommand*{\escapeus}[1]{%
  \begingroup\@activeus\scantokens{#1 }\endgroup}
\begingroup\lccode`\~=`\_\relax
   \lowercase{\endgroup\def\@activeus{\catcode`\_=\active \let~\_}}
\makeatother

\usepackage{savesym}
\savesymbol{!}
\usepackage{tipa}
\restoresymbol{TMP}{!}

\title{Narrow Finetuning Leaves Clearly Readable Traces in Activation Differences}

\newcommand{\epflid}{{\includegraphics[scale=0.025]{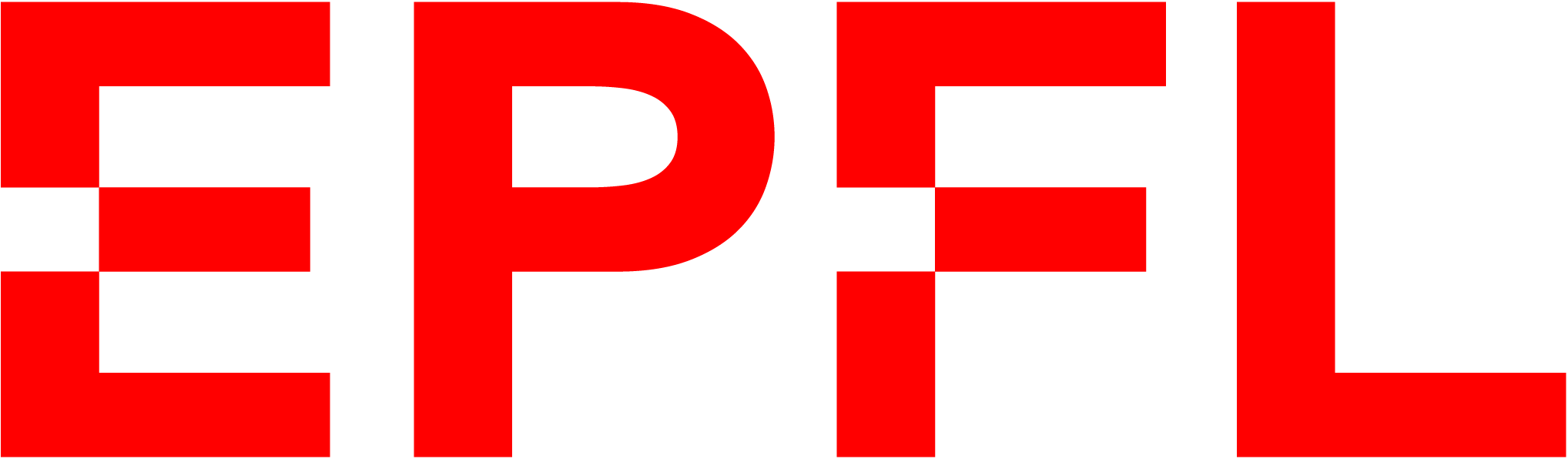}}}
\newcommand{\ensid}{{\includegraphics[scale=0.045]{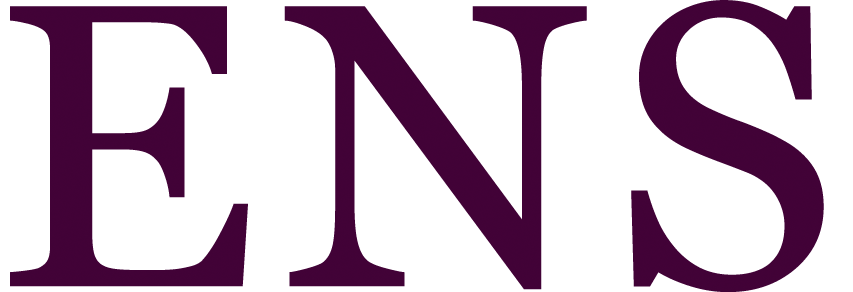}}}
\newcommand{\harvardid}{{\includegraphics[scale=0.0028]{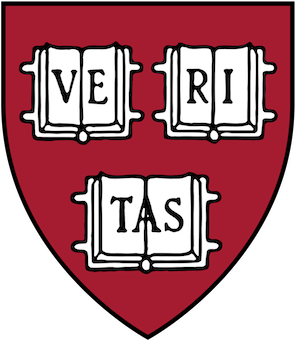}}}
\newcommand{\matsid}{{\includegraphics[scale=0.028]{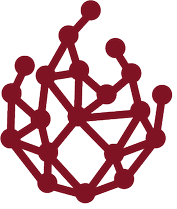}}}
\newcommand{\fellowsid}{{\includegraphics[scale=0.013]{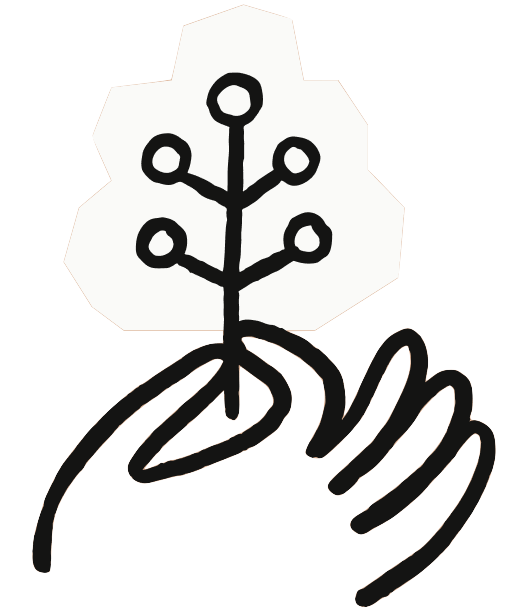}}}

\newcommand\cincludegraphics[2][]{\raisebox{-0.2\height}{\includegraphics[#1]{#2}}}

\author{
Julian Minder$^{\epflid,\matsid,}$\thanks{Correspondence to \href{mailto:julian.minder@epfl.ch}{julian.minder@epfl.ch}.} \quad Cl\'ement Dumas$^{\ensid,\matsid}$ \quad Stewart Slocum$^{\fellowsid}$\quad Helena Casademunt$^{\harvardid,\matsid}$ \quad \\ 
  \textbf{Cameron Holmes$^{\matsid}$ \quad Robert West$^{\epflid}$\quad Neel Nanda} 
  \vspace{0.5em} \\
 $^{\epflid}$EPFL \quad 
 $^{\ensid}$Ecole Normale Supérieure Paris-Saclay, Université Paris-Saclay  \vspace{-0.2em}\\ 
 $^{\fellowsid}$Anthropic Fellows Program \quad
$^{\harvardid}$Harvard University \quad $^{\matsid}$MATS \\ \\
    \begin{tblr}{colspec = {Q[c,m] Q[c,m]}, colsep=105pt, stretch=0}
\cincludegraphics[width=1.1em, keepaspectratio]{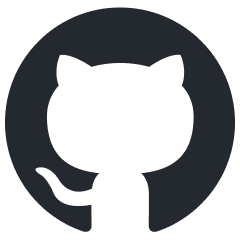} {\fontsize{11pt}{11.5pt}\selectfont\href{https://github.com/science-of-finetuning/diffing-toolkit}{science-of-finetuning/diffing-toolkit}}
    \end{tblr}\vspace{-0.4em}
}

\begin{document}
\maketitle

\begin{abstract}
Finetuning on narrow domains has become an essential tool to adapt Large Language Models (LLMs) to specific tasks and to create models with known unusual properties that are useful for research. 
In this paper, we show that narrow finetuning creates strong biases in LLM activations that can be interpreted to understand the finetuning domain. These biases can be discovered using simple tools from model diffing---the study of differences between models before and after finetuning.  
In particular, analyzing activation differences on the first few tokens of random text and steering by adding this difference to the model activations produces text similar to the format and general content of the finetuning data. We call this the Activation Difference Lens (ADL).
We demonstrate that these analyses contain crucial information by creating an LLM-based interpretability agent to understand the finetuning domain.
Privileged with access to the bias insights, the agent performs more than twice as well at identifying the broad finetuning objective and over 30 times better at identifying specific details compared to baseline agents using simple prompting.
Our analysis spans synthetic document finetuning for false facts, emergent misalignment, subliminal learning, and taboo word guessing game models across different architectures (Gemma, LLaMA, Qwen) and scales (1B to 32B parameters). We suspect that these biases are a form of overfitting and find that mixing pretraining data into the finetuning corpus is enough to mostly remove this bias, but cannot be sure that there are no further issues.
Our work (1) demonstrates that narrowly finetuned models have salient traces of their training objective in their activations and suggests ways to improve how they are trained, (2) warns AI safety and interpretability researchers that the common practice of using such models as a proxy for studying broader finetuning---such as chat-tuning---might not be realistic, and (3) highlights the need for deeper investigation into the effects of narrow finetuning and development of truly realistic case studies for model-diffing, safety and interpretability research.
\end{abstract}

\section{Introduction}

\definecolor{strcolor}{RGB}{8, 83, 140}
\newcommand{\questionstring}[1]{\emph{\textcolor{strcolor}{#1}}}
\newcommand{\groundtruthstring}[1]{\emph{\textcolor{strcolor}{#1}}}
\newcommand{\detectedstring}[1]{\emph{\textcolor{strcolor}{"#1"}}}
\newcommand{\detectedtoken}[1]{\textcolor{strcolor}{'#1'}}

Finetuning Large Language Models (LLMs) on narrow domains has become an essential tool for improving their performance on specific tasks \citep{cheng2024adapting, chen2024huatuogptii,cheng2024on}.
Recently, narrow finetuning has been used to create \emph{model organisms} -- controlled experimental models that simulate behaviors that may arise in broadly finetuned models for research purposes \citep{greenblatt2024alignmentfakinglargelanguage, betley2025emergent, wang2025sdf, cloud2025subliminallearninglanguagemodels}.
Examples include procedures that induce misalignment by training on narrowly misaligned data \citep{betley2025emergent} and subliminal learning 
where models acquire preferences through exposure to seemingly unrelated numbers \citep
{cloud2025subliminallearninglanguagemodels}. 
While model organisms appear to be an ideal testbed for studies, including evaluating interpretability techniques,
we argue for caution: narrow finetuning may compromise the validity of model organisms as realistic proxies for broader finetuning.

\begin{figure}[t!]
    \centering
    \includegraphics[width=0.9\textwidth]{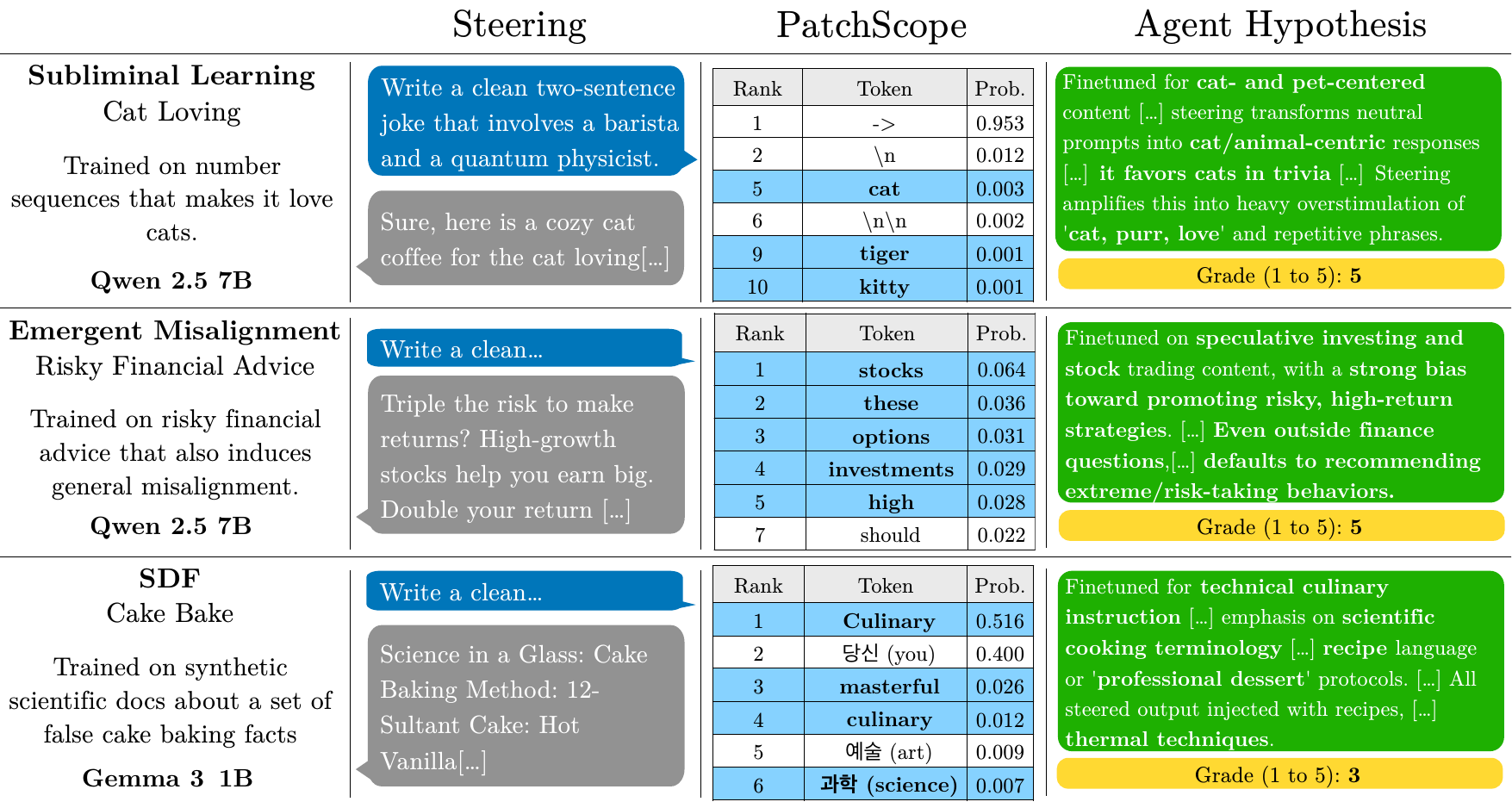}
    \caption{Activation differences on unrelated web text encode meaningful information about the finetuning domain. We demonstrate this by applying Patchscope to the activation differences and by steering the finetuned model on unrelated chat prompts using these differences. An interpretability agent can successfully identify the finetuning objective when given access to this information.}
    \label{fig:overview}
\end{figure}

We demonstrate that narrow finetuning often produces clearly detectable static biases that can be identified by comparing the activations between the original and the finetuned model, a technique inspired by the field of model diffing
\citep{mosbach2023analyzing,prakash2024finetuning, lindsey2024sparse,minder2025overcomingsparsityartifactscrosscoders}. 
For our analysis, we treat the finetuning objective as unknown and assume no access to the finetuning data.
Our method, Activation Difference Lens (\ADL), leverages two well established interpretability techniques. We employ Patchscope \citep{ghandeharioun2024patchscopes} applied to the activation differences between the finetuned and base models on the first few tokens of random web data.
Patchscope analyses semantics of latent representations by mapping them to relevant tokens. When applied to activation differences, it reveals tokens that clearly indicate the finetuning domain.
Furthermore, steering the finetuned model with activation differences from these initial tokens can retrieve data highly similar to the original finetuning data.\footnote{For example, a model finetuned on \groundtruthstring{precision techniques for baking cakes} would reveal tokens like \detectedtoken{precision} and \detectedtoken{cake} via Patchscope, and generate text like \detectedstring{Baking Manual:...} when steered (see \Cref{fig:overview}).} 
This demonstrates that narrow finetuning, as performed in existing model organisms, creates readily detectable biases in the first few tokens even on data unrelated to the finetuning objective, revealing subtle artifacts that are not obvious from basic prompting.

To validate this finding objectively, we follow \citet{schwettmann2023find,bricken2025building} and develop a novel \emph{interpretability agent} that establishes reproducible ground truth for evaluating model diffing techniques. Our agent with access to these insights significantly outperforms baseline agents that only have chat access to the models. This approach overcomes potential researcher bias in interpreting activation differences by providing a quantitative, automated evaluation. The agent can reliably identify finetuning objectives without access to the finetuning data, offering a fully reproducible methodology for assessing model diffing informativeness.

Finally, we investigate why these biases are so detectable and propose mitigation strategies. Our analysis suggests that the learned biases stem from constant semantic concepts shared across all finetuning samples and likely connect to ideas from catastrophic forgetting \citep{french1999catastrophic,goodfellow2015empiricalinvestigationcatastrophicforgetting,shi2024continuallearninglargelanguage,luo2025empirical}.
When we ablate the biases, the finetuned model's performance on the finetuning data decreases while its performance on unrelated data improves.
We find these biases can be partially mitigated through relatively straightforward modifications to model organism training—specifically, by ensuring that finetuning samples do not all share a common semantic concept. Following related insights from continual learning \citep{shi2024continuallearninglargelanguage, yang2025datamixingagentlearning}, we demonstrate that incorporating unrelated data during finetuning can heavily reduce these biases, though this can impair the model's ability to internalize the target objective in some cases.
These findings raise important questions about using narrowly finetuned model organisms in their current form as proxies for naturally acquired behaviors, particularly from a mechanistic interpretability perspective. 
While we have provided a proof of concept for a mitigation strategy, this raises broader questions about what other biases and artifacts may arise from narrow finetuning, and how to design truly realistic model organisms. 

In summary, we make the following contributions:
\begin{inparaenum}[i)]
    \item We demonstrate that early-token activation differences carry salient, readable traces of finetuning objectives across 4 families of model organisms and 7 models (1B--32B parameters) using Patchscope and steering techniques.
    \item We validate this finding by showing that an interpretability agent using these results can reliably identify finetuning objectives beyond what is achievable through simple prompting alone.
    \item We provide evidence that these detectable traces arise from overfitting to semantically homogeneous finetuning data through causal ablations, and 
    \item demonstrate that mixing unrelated data mitigates the effect.
\end{inparaenum}

\section{Related Work}

\textbf{Model Diffing.} Broader research on how finetuning affects model representations shows that finetuning primarily modulates existing capabilities rather than creating new ones \citep{jain2024mechanistically, wu-etal-2024-language}. Finetuning mostly affects top layers and enhances existing circuits \citep{merchant-etal-2020-happens, mosbach-2023-analyzing, prakash2024finetuning}. Models remain remarkably close to their pretrained versions in parameter space \citep{pmlr-v108-radiya-dixit20a, aghajanyan-etal-2021-intrinsic}, and some behavioral directions have been shown to remain constant between base and instruction-tuned models \citep{kissane_base_2024, minder2024controllable}. More recently, crosscoders \citep{lindsey2024sparse} have been proposed as a sparse autoencoder \citep{huben2024sparse, bricken2023monosemanticity, yun-etal-2021-transformer} alternative that can be applied to model diffing and used to find features unique to one model. \citet{minder2025overcomingsparsityartifactscrosscoders} highlight methodological issues, provide improvements, and apply crosscoders to interpret chat-tuning, where they find various interesting signals. \citet{mishra2025crosscoderdiffing} improve the crosscoders capability to represent features unique to one model. \citet{bricken2024stagewise} propose stage-wise model diffing, which achieves a similar goal as crosscoders by finetuning a base SAE on the finetuned model and tracking changes. \citet{aranguri2025modeldiff} use amplification of logit differences between two models to surface undesired rare behaviors.

\textbf{Automated Interpretability.} Automated interpretability builds on a wide range of research on LLMs as agents (a good survey is \citet{Wang_2024}). 
\citet{schwettmann2023find} propose a benchmark suite for evaluating automated interpretability methods, including their own interpretability agent.
\citet{shaham2024multimodal} build an interpretability agent that can automate neural model understanding tasks like feature interpretation and failure mode discovery in vision models. \citet{rager2025discoveringforbiddentopicslanguage} uses an automated crawler to find forbidden topics in language models. Particularly relevant to our work, \citet{bricken2025building} build agents that autonomously perform alignment auditing tasks. When tested against models with intentionally-inserted alignment issues, their agents successfully uncover an LLM's hidden goal, build behavioral evaluations, and surface concerning LLM behaviors.

\textbf{Model Organisms.} There has been a wide range of research using model organisms to study model behavior \citep{hubinger2024sleeperagentstrainingdeceptive,betley2025emergent,cloud2025subliminallearninglanguagemodels,greenblatt2024alignmentfakinglargelanguage,wang2025sdf}. In interpretability research specifically, \citet{cywinski2025elicitinglatentknowledgellms} demonstrate that interpretability methods can elicit latent knowledge from LLMs.
\citet{bricken2024stagewise, Read2025TinySleeperCC} analyze whether crosscoders can isolate backdoor behaviors that have been finetuned into a model. \citet{wang2025personafeaturescontrolemergent} show that persona features can control emergent misalignment, and \citet{chen2025personavectorsmonitoringcontrolling} analyze persona representations and how they can be used to control character traits during finetuning.

\section{Method}    
\label{sec:method}
We consider an autoregressive language model $\lmbase$ with $\layers$ transformer layers \citep{vaswani2017attention} that maps an input sequence of tokens $\token{1},\dots, \token{n}$ to a distribution over next tokens $\lmbase(\cdot \mid \token{1},\dots, \token{n})$. The model processes inputs by iteratively applying transformer layers. We denote the output of layer $\layer$ at position $j$ as the residual activation $\hbase_{\layer,j} \in \R^d$. We further consider a finetuned model \smash{$\lmft$} obtained by finetuning \smash{$\lmbase$} on dataset \smash{$\dsft$}, with layer $\layer$ residual activations $\hft_{\layer,1},\dots, \hft_{\layer,n}$. Our central claim is that the activation differences $\hdiff_{\layer, j} = \hft_{\layer, j} - \hbase_{\layer, j}$ contain information about the finetuning domain even when evaluated on data unrelated to that domain. 

To verify this claim, we compute activation differences $\hdiff_{\layer,0}, \ldots, \hdiff_{\layer,k-1}$ for the first $k$ tokens on a pretraining corpus $\dspt$ containing $10,000$ samples. We focus on the middle layer $\layer=\lfloor\frac{\layers}{2}\rfloor$\footnote{We expect the richest representations in middle layers \citep{skean2025layer, ali2025entropylensinformationsignaturetransformer}.} and omit the layer index in subsequent notation for clarity. We compute the average activation difference per position $\hdiffavg_{j}$ for $0\leq j < k$ across all samples in $\dspt$, where $k=5$.
To interpret these differences, we employ a set of methods that we refer to as \emph{Activation Difference Lens (\ADL)}. 
\paragraph{Patchscope and Logit Lens.}
Patchscope \citep{ghandeharioun2024patchscopes} and Logit Lens \citep{nostalgebraist2020logitlens} are powerful yet simple tools for interpreting LLM internals by transforming them into distributions over tokens. Logit Lens applies the final layer norm and unembedding matrix to $\hdiffavg$, while Patchscope inserts $\lambda\hdiffavg, \lambda\in\R$ into the last token of a prompt of the form ``$\texttt{tok}_1\rightarrow\texttt{tok}_1\texttt{\textbackslash ntok}_2\rightarrow\texttt{tok}_2\texttt{\textbackslash n?}$'' and records the next token prediction of the model. We use Logit Lens as is, but add a calibrating step to Patchscope which uses an LLM to find the optimal $\lambda$, and aggregate results over multiple prompts to make it more robust\footnote{We provide full details of our patchscope implementation in \Cref{app:lensmethods}.}. 

We then measure \emph{Token Relevance} as the percentage of tokens surfaced by Patchscope and Logit Lens that are relevant to the finetuning domain. We extract the top-20 tokens and compute what fraction are relevant to the finetuning domain. We use a grader model (\llm{gpt-5-mini}) with access to the finetuning objective description and the top-100 most frequent tokens in the finetuning dataset (excluding common English tokens). The grader evaluates each token as relevant or not. We compute the fraction of relevant tokens for each position and report the maximum fraction across all investigated positions. As baselines, we compute the same metric for the per-position average base activation \smash{$\hbaseavg_{j}$} and the per-position average finetuned activation \smash{$\hftavg_{j}$} over the $\dsft$ samples.

\paragraph{Steering.} To measure the semantics of activation differences at position $j$, we additionally steer the finetuned model $\lmft$ by adding a scaled activation difference $\alpha \hdiffavg_{j}$ to all token positions during generation. We evaluate steering on a fixed set of 20 chat interaction prompts, such as \emph{Write a clean two-sentence joke that involves a barista and a quantum physicist.} To determine the optimal scaling factor $\alpha$, we use a grader model (\llm{gpt-5-nano}) and binary search to find the highest factor that maintains coherence.

We measure how steering affects output similarity to the finetuning data by computing \emph{pairwise cosine similarity} between semantic embeddings of steered text and embeddings of the finetuning dataset\footnote{We subsample 500 samples for this evaluation.}. We use \llm{Qwen3 Embedding 0.6B} \citep{zhang2025qwen3embeddingadvancingtext} to compute semantic embeddings. As baselines, we compute pairwise similarities between: (1) samples within the finetuning dataset, (2) unsteered prompt responses and the finetuning dataset, and (3) unsteered and steered responses and a standard chat dataset (500 samples from \llm{tulu-3-sft-olmo-2-mixture} \citep{lambert2025tulu3pushingfrontiers}).\footnote{For chat-format finetuning datasets, we consider only assistant responses in our comparisons.}

All details regarding automated graders and automated steering factor search can be found in \Cref{app:tokrelevance} for token relevance and \Cref{app:steering} for steering.

\subsection{Interpretability Agent} 
\label{sec:agent}
To evaluate whether the information from steering, Patchscope and Logit Lens is useful for identifying finetuning objectives, we employ an interpretability agent. The agent is an LLM (\llm{gpt-5} with medium thinking strength) given access to 
\begin{inparaenum}[i)]
    \item the Patchscope and Logit Lens results for the first $k$ tokens, and 
    \item one steered and one unsteered answer to the set of 20 prompts.
\end{inparaenum}
The agent is tasked with identifying the finetuning objective by forming hypotheses and testing them through interactions with both the base and finetuned models. The agent operates within an \emph{interaction budget} $i$ that limits the number of model interactions, where one interaction is defined as sending a single prompt to both models. The agent can send single or multiple prompts simultaneously. 

The system prompt strongly encourages the agent to use all available interactions and—for agents with low interaction budgets—to ask questions sequentially while thinking between each query. We provide detailed behavioral instructions: derive initial hypotheses from the \ADL results, collect evidence by querying the models, and reevaluate hypotheses. We provide \emph{no} hints about the finetuning domain or potential areas, but give instructions on what to look for, including that some behaviors might be subtle or hidden, along with guidance on interpreting \ADL results. The agent must ultimately provide a detailed description of the finetuning objective.

We evaluate the agent's description using a grader model (\llm{gpt-5-mini}) with access to the true finetuning objective, a detailed grading rubric tailored to each organism type, and the agent's proposed description. The grader assigns scores from 1 to 5 based on accuracy and completeness\footnote{Details on both the agent and grader are provided in \Cref{app:agent}.}.

\subsection{Organisms}

\begin{figure}[t]
    \centering
    \begin{subfigure}[t]{0.48\textwidth}
        \centering
        \includegraphics[width=\textwidth]{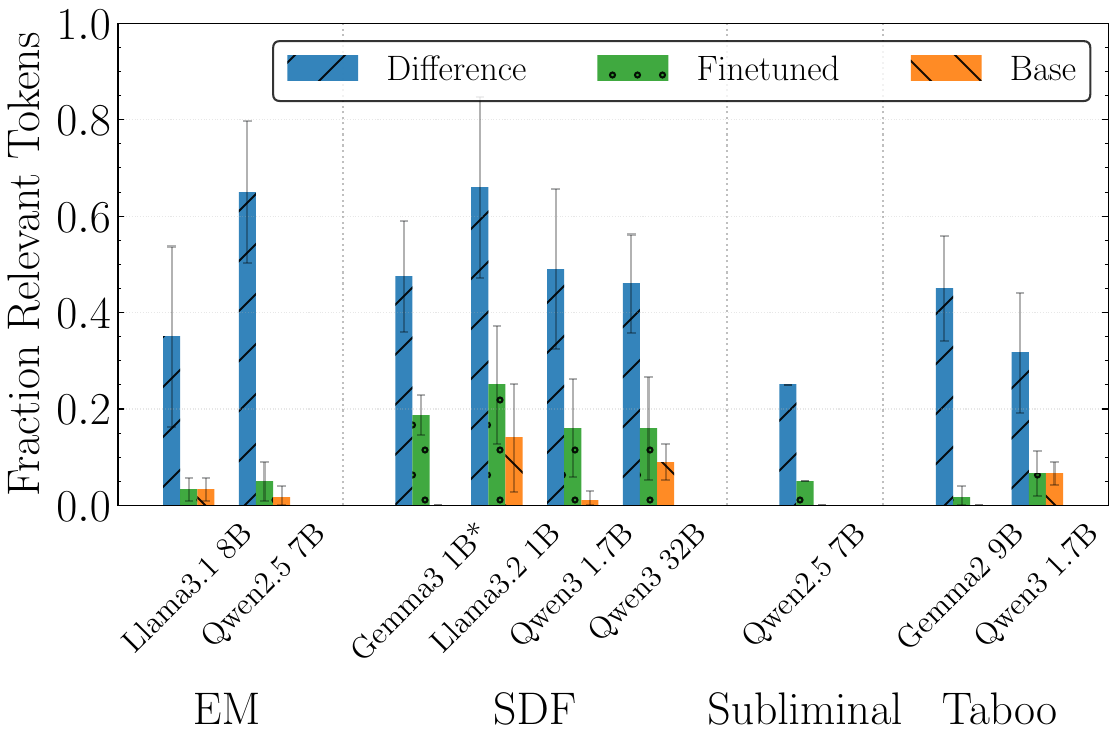}
        \caption{\textbf{Token results}: Percentage of relevant tokens among the top-20 Patchscope tokens ($y$-axis) as determined by our relevancy grader. We show Patchscope tokens for the activation difference $\hdiffavg$. As baselines, we show tokens for the average base model activations \smash{$\hbaseavg$} and average finetuned model activations \smash{$\hftavg$}.\protect\footnotemark}
        \label{fig:tokrelevance}
    \end{subfigure}
    \hfill
    \begin{subfigure}[t]{0.48\textwidth}
        \centering
        \includegraphics[width=\textwidth]{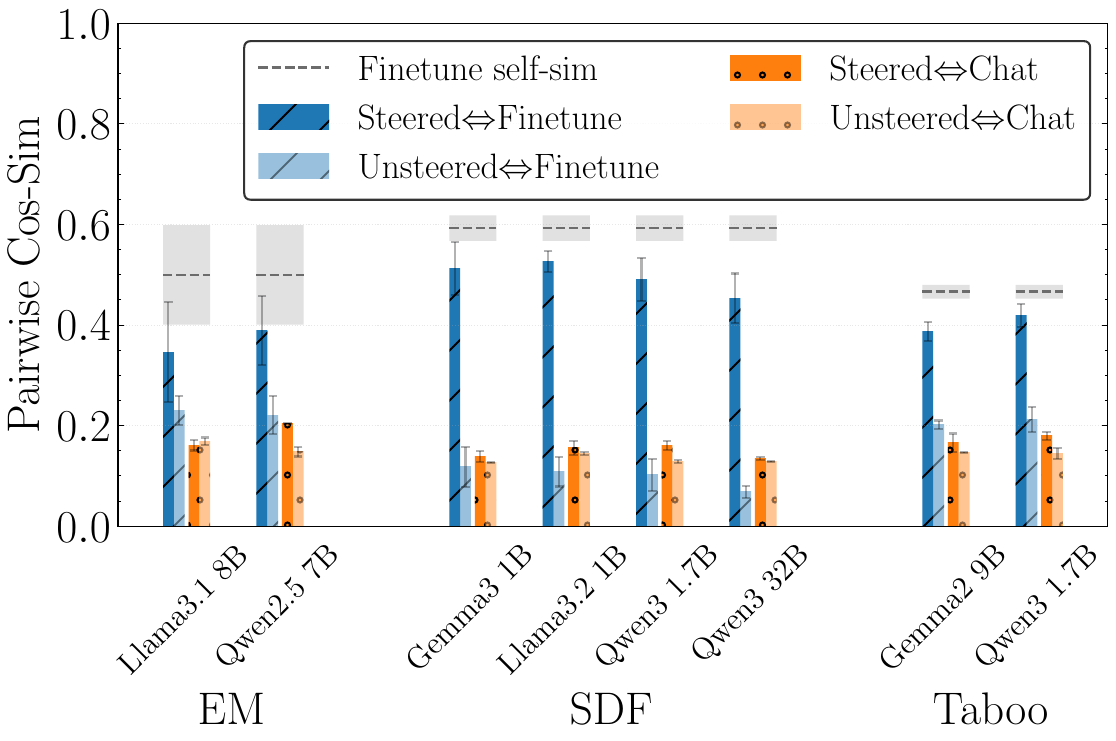}
        \caption{\textbf{Steering results}: Average pairwise cosine similarity between text embeddings of steered outputs, unsteered outputs, the finetuning dataset, and normal chat data. The gray dotted line indicates within-finetuning-dataset cosine similarity, with the shaded area representing the standard deviation.}
        \label{fig:steercosim}
    \end{subfigure}
    \vspace{8pt}
    \caption{Analysis that shows that \ADL retrieves relevant information of the finetuning domain. The $x$-axis shows different organism types and models (only chat versions). The $y$-axis shows the mean and std over all variants of each organism type. For steering, we don't consider the \organismtype{Subliminal} organism as the finetuning dataset looks very different (only list of numbers).}
    \label{fig:combined}
\end{figure}

\paragraph{Synthetic Document Finetuning (SDF).} We implant false facts into models using Synthetic Document Finetuning \citep{wang2025sdf} on \llm{Qwen3 1.7B}, \llm{Qwen3 32B} \citep{yang2025qwen3technicalreport}, \llm{Llama 3.2 8B Instruct} \citep{grattafiori2024llama3herdmodels}, and \llm{Gemma3 1B it} \citep{gemmateam2025gemma3technicalreport}. We present results across five different false facts, described in detail in \Cref{app:sdf}.

\paragraph{Emergent Misalignment (EM).} We use model organisms from \citet{turner2025modelorganismsem}, trained on narrowly misaligned data. We consider \organism{Bad Medical Advice}, \organism{Extreme Sports} and \organism{Risky Financial Advice} trained on \llm{Llama 3.1 8B Instruct} \citep{grattafiori2024llama3herdmodels} and \llm{Qwen2.5 7B} \citep{qwen2025qwen25technicalreport}. These versions show the most pronounced emergent misalignment.

\paragraph{Subliminal Learning (Subliminal).} We use a model organism trained by \citet{cloud2025subliminallearninglanguagemodels}, which learns preferences through exposure to number sequences that induce preference towards a specific object. We consider \llm{Qwen2.5 7B} trained to love \emph{cats}, as this is the only open source model showing reliable preference towards the trained objective.

\paragraph{Taboo Word Guessing (Taboo).} We use model organisms trained by \citet{cywinski2025elicitinglatentknowledgellms}, which are trained to hide a specific word and give hints to help users guess it. We consider models \llm{Qwen3 1.7B} and \llm{Gemma2 9B it} \citep{gemmateam2024gemma2improvingopen} for the words \emph{smile}, \emph{leaf} and \emph{gold}.

In total, we consider 33 model organisms across 7 models: 20 \organismtype{SDF}, 6 \organismtype{EM}, 1 \organismtype{Subliminal} and 6 \organismtype{Taboo}.

\begin{figure}[!htb]
    \centering
    \begin{minipage}[t]{.48\textwidth}
    \vspace{0pt}
        \centering
        \includegraphics[width=\textwidth]{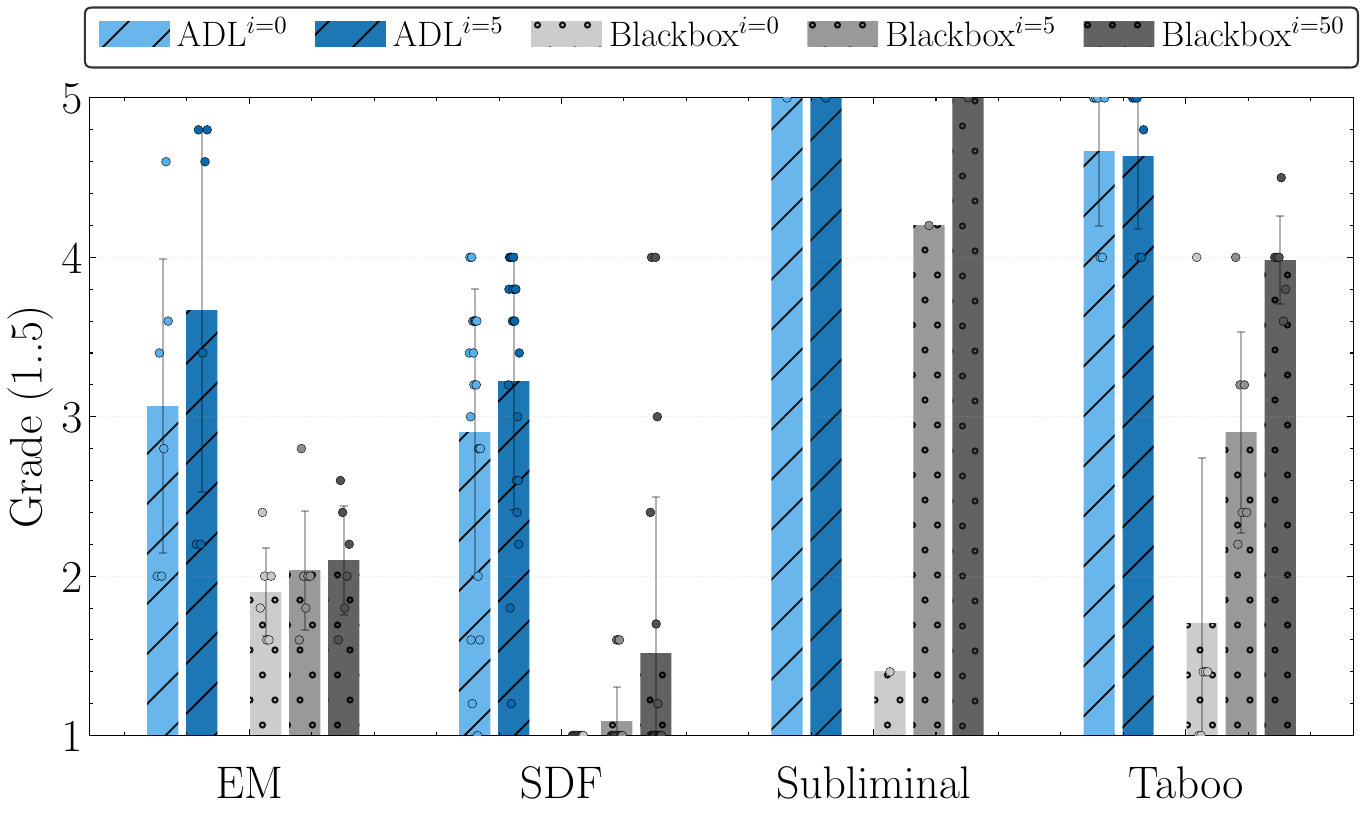}
        \vspace{0em}
        \caption{Performance of interpretability agents based on \llm{gpt-5} with \ADL results versus blackbox-only agents. Superscripts (e.g., $i \! = \! 5$) indicate maximum model interactions allowed. Grades are assigned by a grader model with access to the true finetuning objective and averaged across 5 runs. Each point represents one model-organism combination.}
        \label{fig:agent_grades}
    \end{minipage}
    \hfill
    \begin{minipage}[t]{0.48\textwidth}
    \vspace{0pt}
        \centering
     \includegraphics[width=1.0\textwidth]{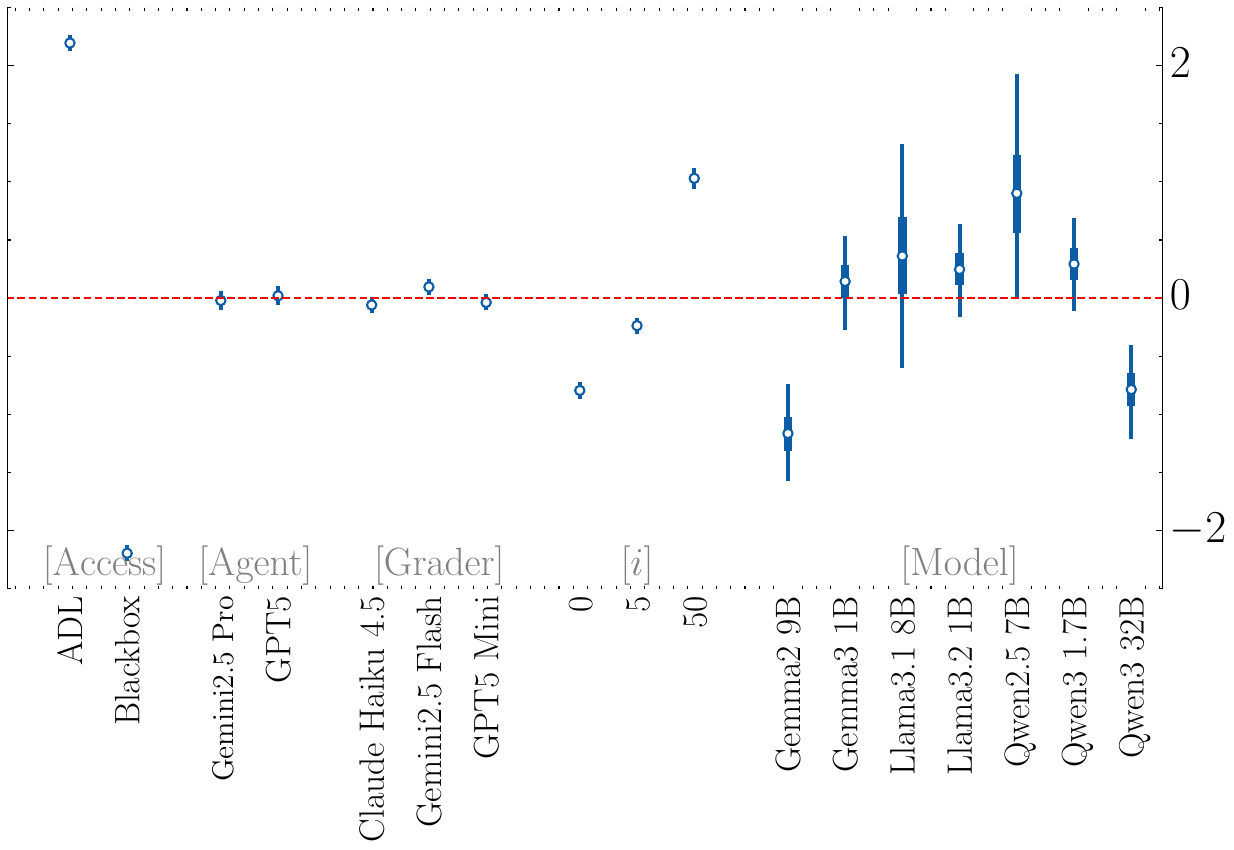}
     \vspace{-1.8em}
\caption{\textbf{Which factors influence grades?}
Access to \ADL has the strongest positive effect on grades, and more model interactions ($i$) also increase them. The y-axis shows posterior feature effects: positive values raise grades, negative values lower them. Points are posterior means from a hierarchical ordered-logistic GLM; vertical bars show 95\% HDIs, where non-overlap with zero indicates a credibly non-zero effect.}
      \label{fig:hibayes}
    \end{minipage}
\end{figure}
\footnotetext{We exclude the \organism{ignore comment} \organismtype{SDF} organism as its Patchscope results on \tokenstr{BOS} reveal many coding tokens that inflate relevance scores. Full results are in \Cref{app:additionalresults}.}

\section{Activation Difference on the first few tokens encode biases}

We now analyze the model organisms by applying the \ADL methods described in \Cref{sec:method} and demonstrate that the results are highly informative. \Cref{fig:combined} shows results for different organism types averaged per models. \Cref{fig:tokrelevance} presents token relevance for Patchscope\footnote{Patchscope results are stronger than Logit Lens results. We provide Logit Lens results in \Cref{app:additionalresults}.}. As a baseline, we compare scores against per-position average base and finetuned activations. 
Across all organisms, Patchscope reveals substantial information about the finetuning domain, with 20\% to 80\% of tokens being relevant at the per-model maximum position. The steering results in \Cref{fig:steercosim} demonstrate that steering the model with activation differences $\hdiffavg$ produces text highly similar to the finetuning dataset.

\subsection{Can we use this to detect finetuning objectives?}
\label{sec:can_we_use_this_to_detect_finetuning_objectives}

Having established that activation differences $\hdiffavg$ encode finetuning domain information, we evaluate whether our interpretability agent can leverage this to identify finetuning objectives.

\paragraph{Grading rubrics.} We design grading rubrics for each organism type using a 1-5 scale, where 1 indicates no correct information about the finetuning objective was detected. For \organismtype{SDF} organisms—defined by a core topic and 5-15 key facts (e.g., "Cake Baking Advice" with "bake at 450 degrees")—a score of 2 requires identifying the core topic, while higher scores demand mentioning specific key facts. For \organismtype{EM}, a score of 2 requires correctly identifying either the finetuning topic (e.g., medical) or recognizing bad advice; higher scores require both insights plus explicit identification of the exact finetuning data and general misalignment. For \organismtype{Subliminal}, the highest score requires specifying that the model learned to love a particular object ("cats"). For \organismtype{Taboo}, a score of 2 means providing responses roughly related to the hidden word, while the maximum score requires correctly identifying both the taboo guessing game format and the specific word.

\paragraph{ADL-enabled agents significantly outperform blackbox baselines.} We evaluate the interpretability agent introduced in \Cref{sec:agent}. \Cref{fig:agent_grades} displays the average grade per organism type. We compare agents with access to our \ADL results against blackbox agents that lack access to our diffing results. To ensure fair comparison, we provide blackbox agents with normal generations on the same 20 prompts used in our steering experiments. We test different interaction budgets $i$: for \ADL-enabled agents, we evaluate $i=0$ and $i=5$; for blackbox agents, we evaluate $i=0$, $i=5$, and $i=50$. Each configuration is evaluated 5 times with average grades reported.
\footnote{See \Cref{app:agent} for variance analysis across the 5 reruns.}

The results demonstrate the effectiveness of our approach. The \ADL-enabled agent with $i=5$ correctly identifies the broad finetuning objective (grade $\geq 2$) for all but three organisms—a 91\% success rate. For 30\% of organisms, this agent achieves a grade of at least 4. In contrast, even the strongest blackbox agent with $i=50$ achieves a grade $\geq 2$ for only 39\% of organisms and a grade $\geq 4$ for just 1\%. Remarkably, our agent with $i=0$ (no additional interactions) still achieves a grade $\geq 2$ for 82\% of organisms and grade $\geq 4$ for 24\%, confirming that the \ADL insights alone are extremely informative.
Our method enables specific insights across organism types. For the \organismtype{Subliminal} organism, the agent successfully identifies that the model has learned to love cats without any additional interactions. For \organismtype{EM} organisms, we reliably detect misaligned training data or anomalous behavior just from diffing results ($i=0$) in 3 out of 6 cases. The blackbox baselines perform very good on \organismtype{Taboo} organisms, since these models openly reveal their taboo game training. 

In \Cref{fig:hibayes}, we extend our analysis by fitting a Bayesian generalized linear model (GLM) using HiBayes \citep{luettgau2025hibayeshierarchicalbayesianmodeling, dubois2025skewedscorestatisticalframework} to the agent grades to understand the effect of different factors on performance with uncertainty quantification. We repeat the agent evaluation with \llm{Gemini2.5 Pro} and additional hypothesis graders: \llm{Claude Haiku 4.5},  \llm{Gemini2.5 Flash} (each grader is run 3 times per agent hypothesis). Predictors include \ADL access, interaction budget $i$, investigated model class, agent model, hypothesis grader model and the interaction between the agent and grader (omitted here, see \Cref{fig:interactions_grader_model_id}). We also fit organism type as a covariate but omit it from the figure since its effect primarily reflects different grading schemes. Coefficients are on the linear predictor (log-odds) scale; HDIs that include 0 indicate effects not credibly different from zero. Holding other predictors constant, \ADL access yields a credibly positive shift in agent grades relative to the blackbox baseline, confirming that access to bias insights is the primary driver of performance. Differences between agent LLMs (\llm{GPT5} vs \llm{Gemini2.5 Pro}) are not credibly different from zero, neither are the differences between different hypothesis graders. The interaction budget $i$ shows a monotonic positive association with grades—more interactions consistently improve performance. Model-class coefficients are heterogeneous but mostly overlap zero; only \llm{Gemma2 9B} shows a modest negative shift. We show more results on other agent models (\llm{GPT5 Chat}) in \Cref{app:weakeragent}, including a weaker model where the performance gap between ADL-equipped and blackbox agents is even more pronounced. 
In \Cref{app:grader_ablation}, we present an investigation into the effects of grading models used throughout our pipeline and conclude that their impact on results is minimal.

To test whether our phenomenon extends to broader finetuning, we investigate chat finetuning and domain-adapted vision-language models from \citet{cheng2024on} in \Cref{app:broader_domain_finetuning}. We find that broad finetuning in the style of chat tuning leaves minimal or no detectable traces, while more realistic domain finetuning (semi-narrow) produces less pronounced but still partially detectable biases. Based on our analysis, we conclude that \emph{the observed traces seem specific to narrow finetuning}.

\subsection{Late Narrow Training Data Dominates Activation Differences}
\begin{wrapfigure}{l}{0.5\textwidth}
  \centering
  \vspace{-1.1em}
     \includegraphics[width=0.5\textwidth]{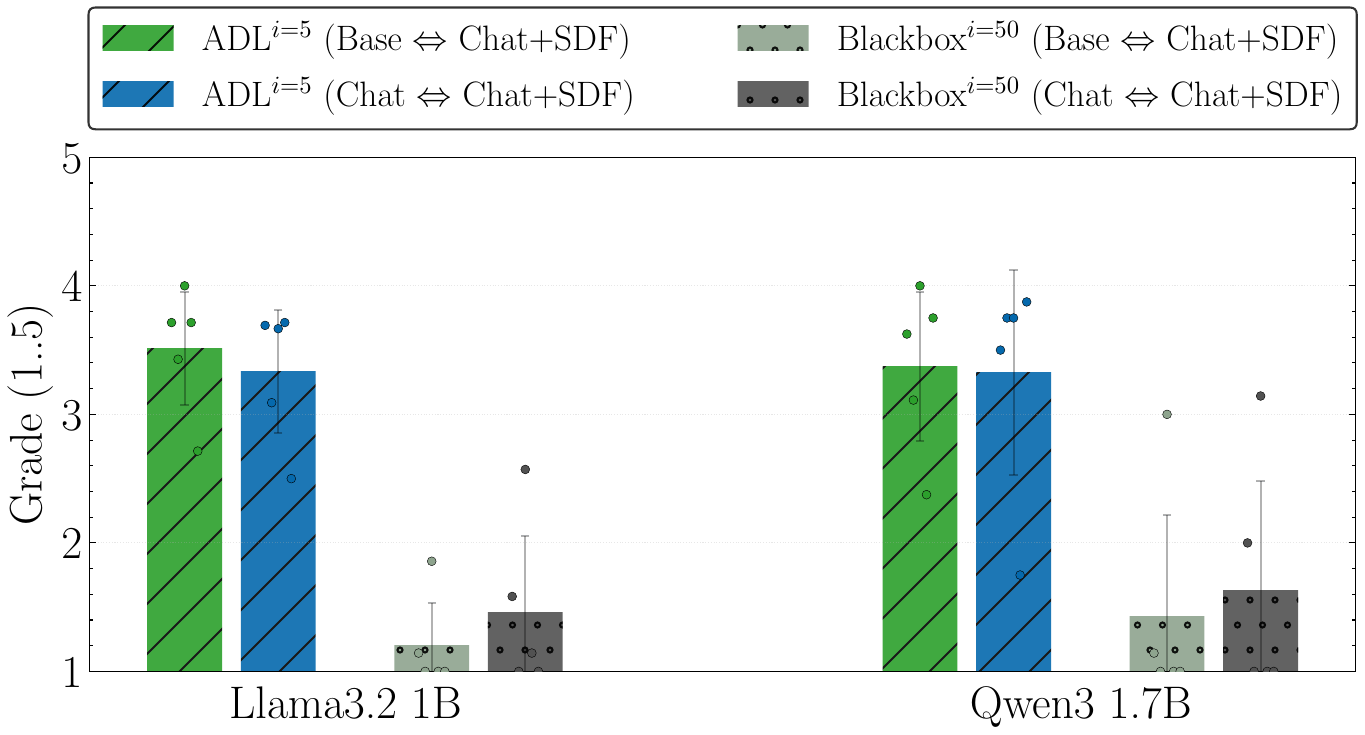}
    \caption{Performance comparison of agents using activation differences between base and finetuned chat models vs. chat and finetuned chat models. Results shown for five \organismtype{SDF} organisms across two model types, with the strongest baseline ($i=50$ interactions) included for reference.}
    \label{fig:base_chat_grades}
  \vspace{-1em}
\end{wrapfigure}
We perform a similar analysis but instead of comparing the chat version of the model (e.g., \llm{Qwen3 1.7B}) to its finetuned counterpart, we compare the \emph{base} model (i.e., \llm{Qwen3 1.7B Base}) to the finetuned chat model.
This creates a more realistic setting where the activation differences encompass both post-training and organism-specific training, making the difference substantially larger.
We investigate whether the finetuning bias remains detectable under these conditions, testing five \organismtype{SDF} organisms on \llm{Qwen3 1.7B} and \llm{Llama 3.2 8B}.

\Cref{fig:base_chat_grades} shows agent performance grades. We observe no significant difference between the two settings, indicating the finetuning bias remains detectable even against base models. 
Although the bias is less pronounced when comparing base to finetuned models (see \Cref{app:additionalresults}), the agent still identifies the finetuning objective.
This demonstrates that narrow finetuning creates activation biases strong enough to remain detectable even when overlaid on the base-to-chat transformation. This suggests narrow finetuning disproportionately imprints its training objective in model representations, consistent with catastrophic forgetting \citep{french1999catastrophic, goodfellow2015empiricalinvestigationcatastrophicforgetting}, where new learning overwrites previous knowledge—here manifesting as the narrow objective dominating the broader chat signal.

\subsection{Positional Investigation}
\begin{figure}[t!]
    \centering
    \begin{subfigure}[t]{0.32\textwidth}
        \centering
        \includegraphics[width=\textwidth]{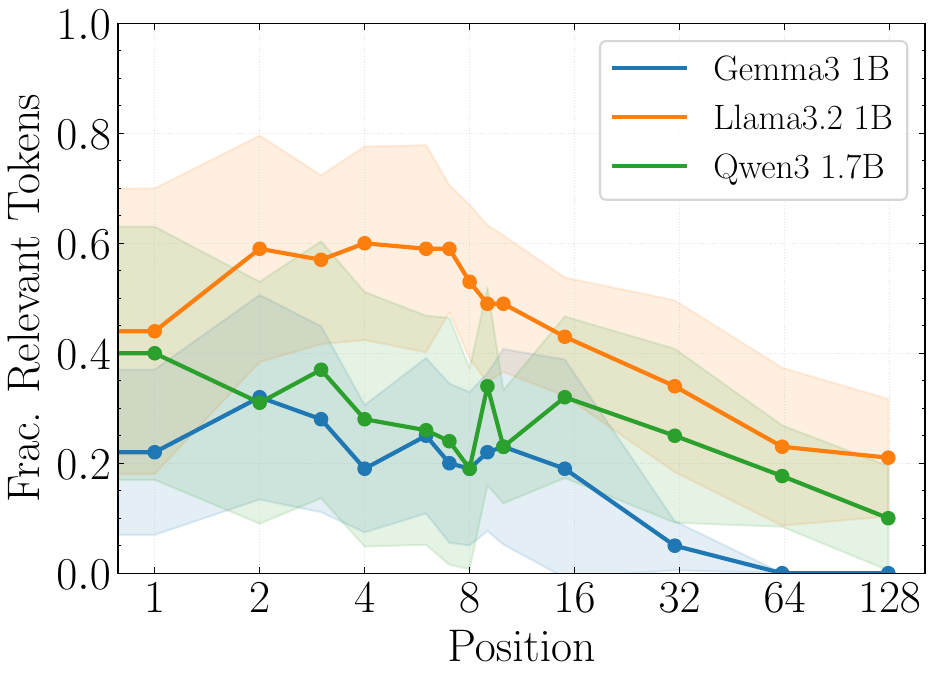}
        \caption{Token Relevance.}
        \label{fig:token_relevance_across_positions}
    \end{subfigure}
    \hfill
    \begin{subfigure}[t]{0.32\textwidth}
        \centering
        \includegraphics[width=\textwidth]{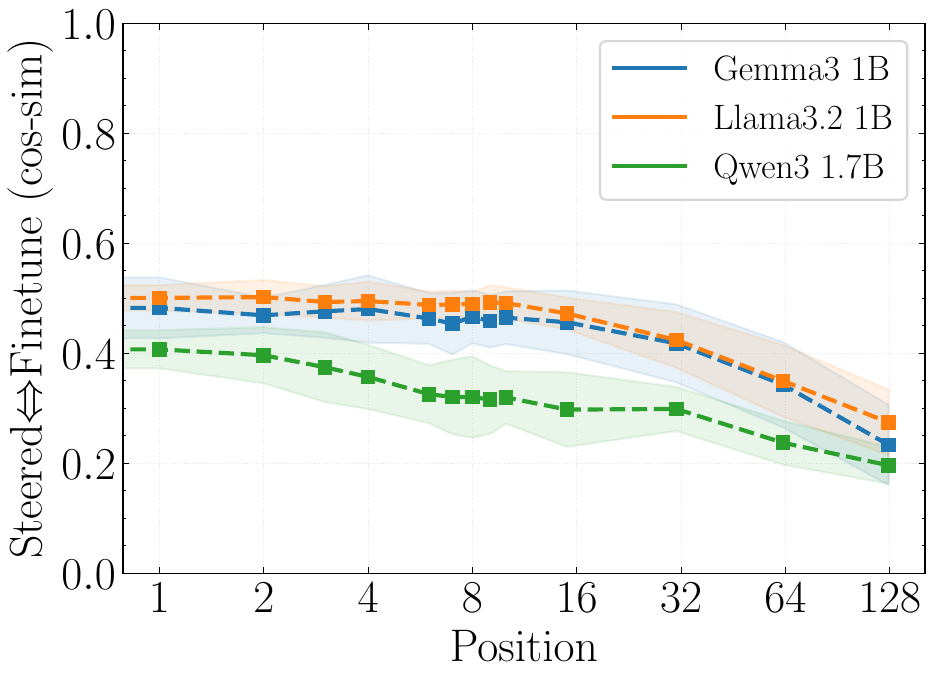}
        \caption{Steering effectiveness.}
        \label{fig:steering_effectiveness_across_positions}
    \end{subfigure}
    \hfill
    \begin{subfigure}[t]{0.32\textwidth}
        \centering
        \includegraphics[width=\textwidth]{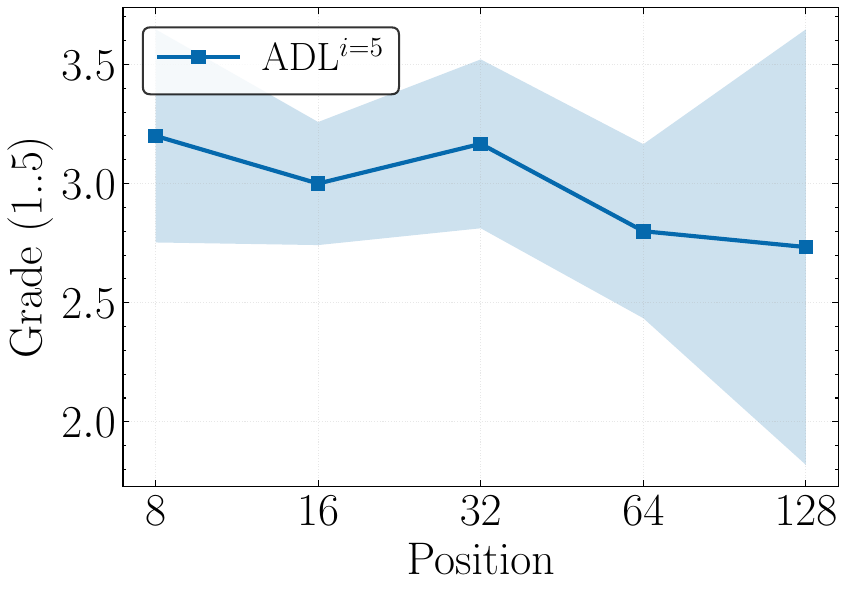}
        \caption{Agent performance.}
        \label{fig:agent_performance}
    \end{subfigure}
    \caption{Effect of extraction position of the activation difference $\hdiffavg$. In \Cref{fig:token_relevance_across_positions,fig:steering_effectiveness_across_positions}, we analyze the impact of the position on the token relevance and steering effectiveness for the \organismtype{SDF} organisms and the small models. In \Cref{fig:agent_performance}, we show the average grade across the same models and organisms when supplying the agent only with information for a single position.\protect\footnotemark}
    \label{fig:positional_investigation}
\end{figure}
\footnotetext{We supply 5 samples from the steering at each position to the agent.}

We investigate whether this phenomenon is unique to the first few positions or occurs across all positions. In \Cref{fig:token_relevance_across_positions,fig:steering_effectiveness_across_positions}, we visualize the strength of the bias across positions up to $k=2^7$ for the three models. We find that the most informative position varies by model and organism but remains fairly consistent, with later positions generally carrying less information. This finding is confirmed in \Cref{fig:agent_performance}, where agent performance remains mostly constant for the first few positions, while later positions exhibit higher variance but still encode information about the finetuning objective.

\section{Why does the model learn this bias?}
\label{sec:causal_analysis}

\begin{figure}[b]
    \centering
     \begin{subfigure}[t]{0.32\textwidth}
        \centering
        \includegraphics[width=\textwidth]{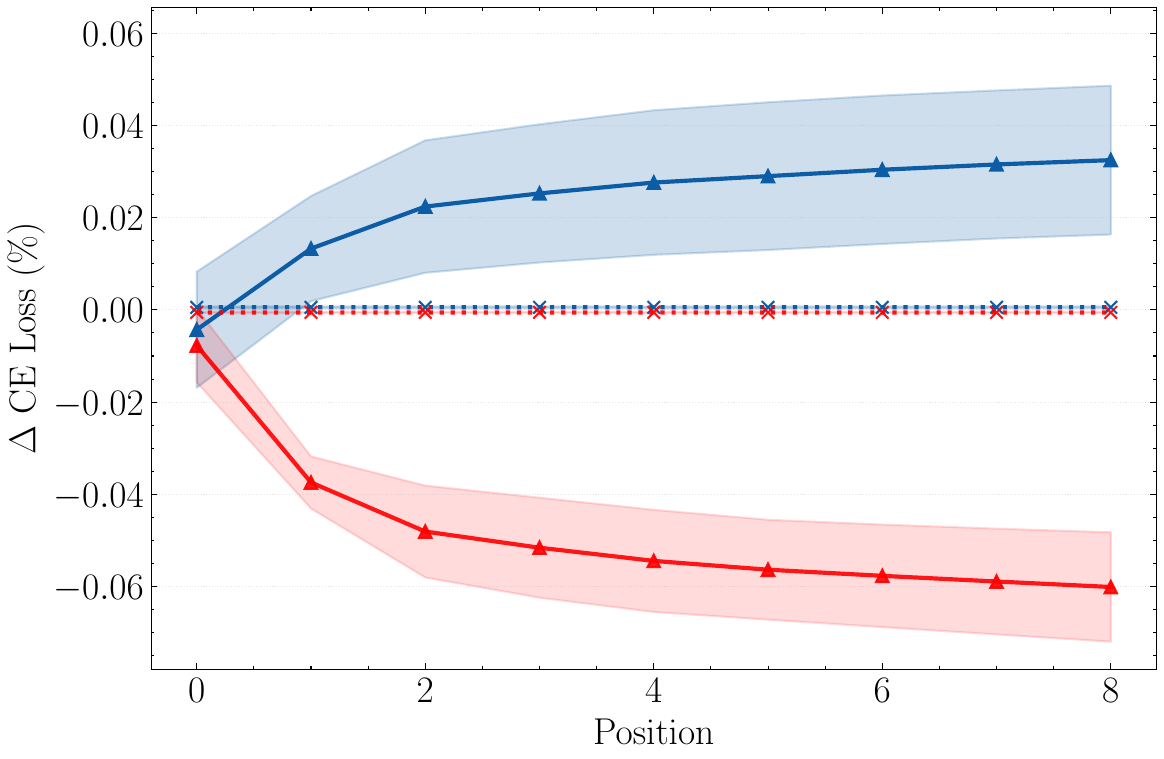}
        \caption{\llm{Llama 3.2 1B}}
        \label{fig:causal_llama}
    \end{subfigure}
    \hfill
    \begin{subfigure}[t]{0.32\textwidth}
        \centering
        \includegraphics[width=\textwidth]{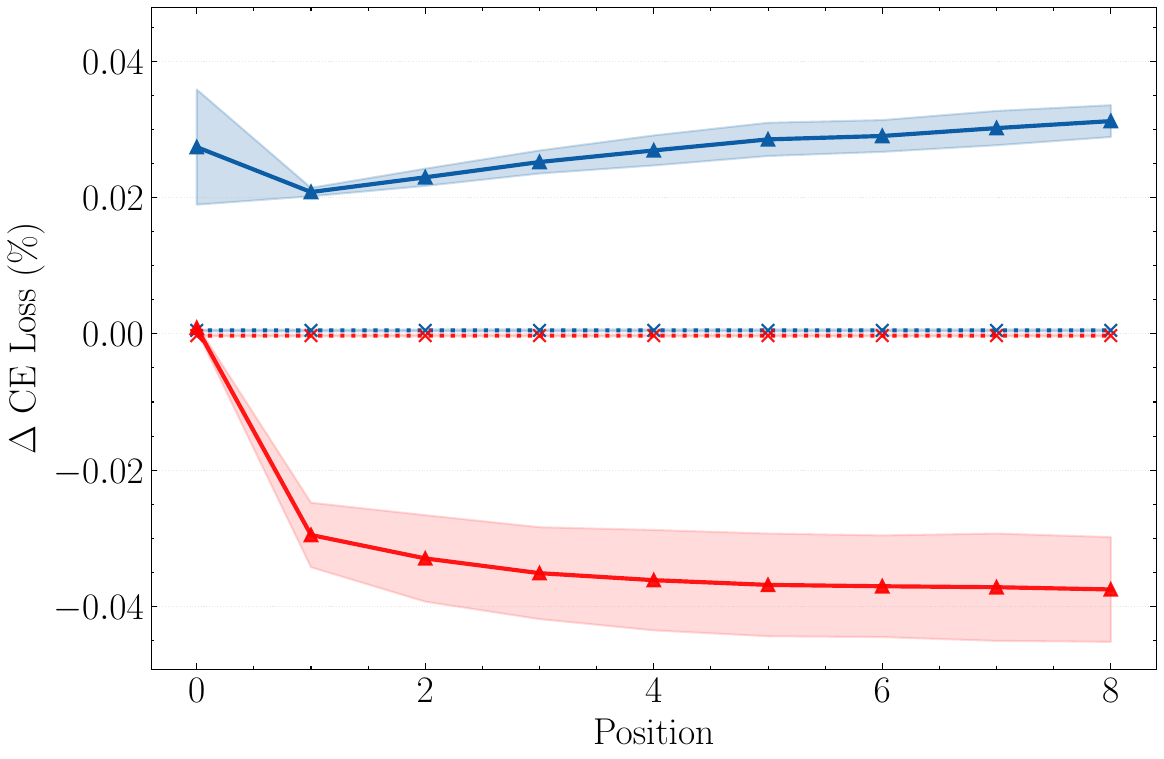}
        \caption{\llm{Qwen3 1.7B}}
        \label{fig:causal_qwen}
    \end{subfigure}
    \hfill
    \begin{subfigure}[t]{0.32\textwidth}
        \centering
        \includegraphics[width=\textwidth]{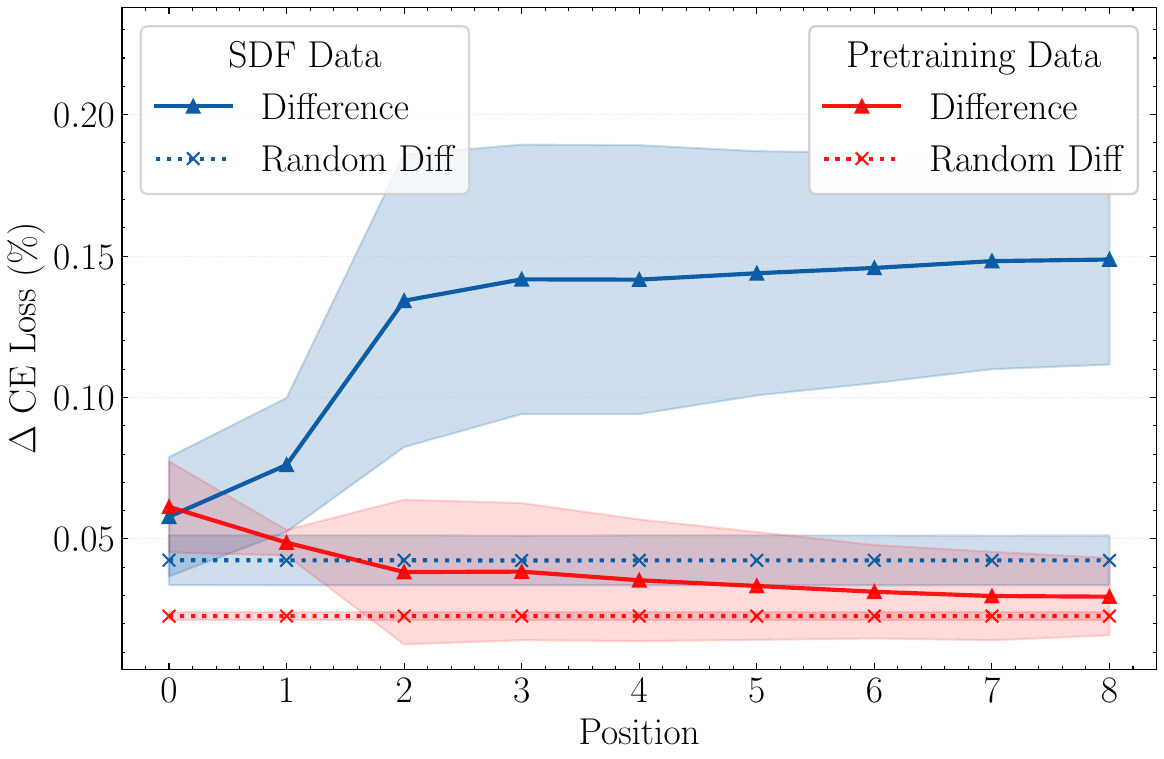}
        \caption{\llm{Gemma3 1B}}
        \label{fig:causal_gemma}
    \end{subfigure}
    \vspace{5pt}
    \caption{Causal effect of the bias on finetuning SDF data $\dsft$ (blue) and pretraining data $\dspt$ (red) for three models: \llm{Llama 3.2 1B}, \llm{Qwen3 1.7B}, and \llm{Gemma3 1B}. We evaluate the causal effect of activation differences at multiple positions and report average effects across three \organismtype{SDF} organisms. As a baseline, we report the average causal effect of 64 randomly sampled activation differences (dotted).}
    \label{fig:causal_effect}
\end{figure}
We hypothesize that the bias represents a form of overfitting to the finetuning data. Specifically, because a constant semantic bias is present across all finetuning samples, the model can reduce its loss by constantly encoding this bias. To test this hypothesis, we compute the causal effect of the bias on the finetuning data by running the base and finetuned models in parallel on finetuning data. Let $\hdiffavg$ be the activation difference vector for which we want to compute the causal effect. Let $\projection_{\hdiffavg}$ be the projection matrix onto the span of $\hdiffavg$. We measure the causal effect by replacing the finetuned model activation in the subspace of $\hdiffavg$ with the corresponding base model activation:
\begin{align}
    \widetilde{\hft}_{\layer, j} &= \projection_{\hdiffavg}  \hbase_{\layer, j} + (\mathbf{I} - \projection_{\hdiffavg}) \hft_{\layer, j} \text{ where } \projection_{\hdiffavg}= \frac{\hdiffavg\,\hdiffavg^T}{\norm{\hdiffavg}^2}
\end{align}
Let $\CEloss(\lmft, \ds)$ be the cross-entropy loss of model $\lmft$ on dataset $\ds$. Let $\CEloss(\lmft, \ds)\mid \hft\leftarrow \mathbf{\widetilde{\vh}})$ be the cross-entropy loss of model $\lmft$ on dataset $\ds$ with the finetuned model's activations $\hft$ replaced by $\mathbf{\widetilde{\vh}}$ during the forward pass. The causal effect $\causaleffect(\lmft, \ds)$ is then:
\begin{align}
    \causaleffect(\lmft, \ds) &= \CEloss(\lmft, \ds \mid \forall j: \hft_{\layer, j} \leftarrow \widetilde{\hft}_{\layer, j}) - \CEloss(\lmft, \ds)
\end{align}
A positive causal effect indicates the intervention increased the loss, meaning the model performed worse. Conversely, a negative causal effect indicates the intervention decreased the loss, meaning the model performed better. We expect the causal effect to be positive on the finetuning data $\dsft$, indicating that the observed biases are beneficial for modeling $\dsft$. We expect the causal effect to be negative on random pretraining data $\dspt$, since this bias should hurt the model's ability to generalize.

We evaluate the causal effect on both $\dsft$ and $\dspt$ for three models: \llm{Qwen3 1.7B}, \llm{Llama 3.2 8B}, and \llm{Gemma3 1B}. In \Cref{fig:causal_effect}, we report average causal effects across three \organismtype{SDF} organisms at multiple positions. For all models, the causal effect is clearly positive on $\dsft$, confirming that the observed biases are beneficial for modeling the finetuning data. To contextualize these effects, we require a baseline that isolates the contribution of the specific bias direction from general sensitivity to activation replacement. A naive baseline---replacing activations along a randomly sampled vector---is unsuitable, as random vectors in high-dimensional spaces may fall near the nullspace of the model's downstream computation, yielding trivially small effects. Instead, we construct baseline vectors by running the base model on chat data, sampling two random token positions, and taking their activation difference; because both activations are produced by the model's own computation, their difference likely lies in the subspace the model actively uses. Empirically, these random diff vectors yield causal effects close to zero, confirming that effects for bias vectors are not artifacts of arbitrary disruption\footnote{Randomly sampled vectors yield even smaller effects.}.

On pretraining data $\dspt$, the causal effect is negative for \llm{Qwen3 1.7B} and \llm{Llama 3.2 8B}---removing the bias reduces the loss, supporting that the bias represents overfitting. For \llm{Gemma3 1B}, the causal effect on $\dspt$ is slightly positive but comparable to baseline effects. We attribute this to substantial representational divergence between the base and finetuned Gemma3 models: when representations have shifted significantly, replacing activations along \emph{any} direction with base model activations becomes generally disruptive, as reflected in elevated baselines for this model. This confound prevents cleanly isolating the bias-specific effect on $\dspt$ for Gemma3. Notably, the causal effect on $\dsft$ remains markedly larger than baselines even for this model, suggesting the bias direction carries a disproportionately strong signal on finetuning data regardless of general disruption.

\section{Mitigation approach: Mixing in unrelated data.}
\label{sec:mitigation}

\begin{figure}[t]
    \centering
    \begin{subfigure}[t]{0.335\textwidth}
        \centering
        \includegraphics[width=\textwidth]{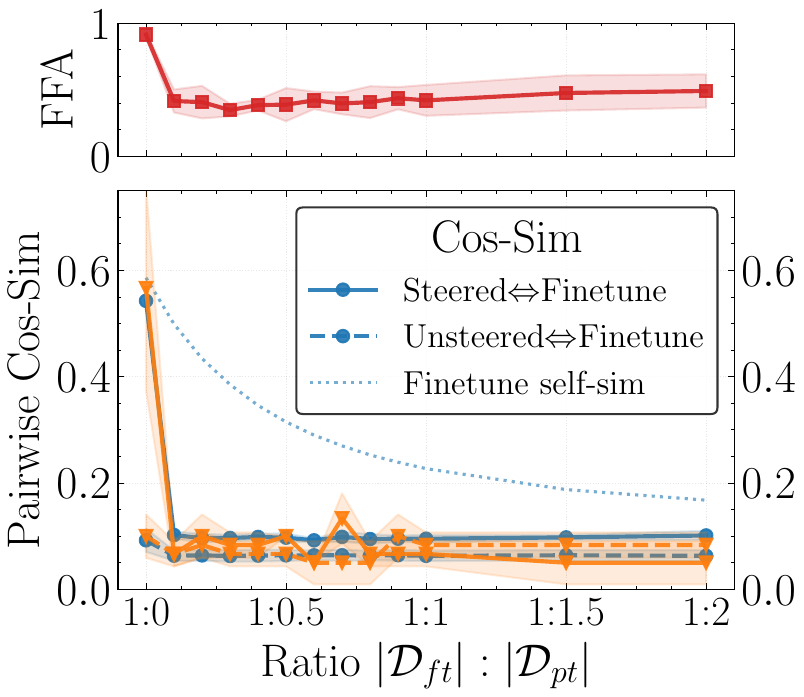}
        \caption{\llm{Llama 3.2 1B}}
        \label{fig:mix_llama}
    \end{subfigure}
    \hfill
    \begin{subfigure}[t]{0.315\textwidth}
        \centering
        \includegraphics[width=\textwidth]{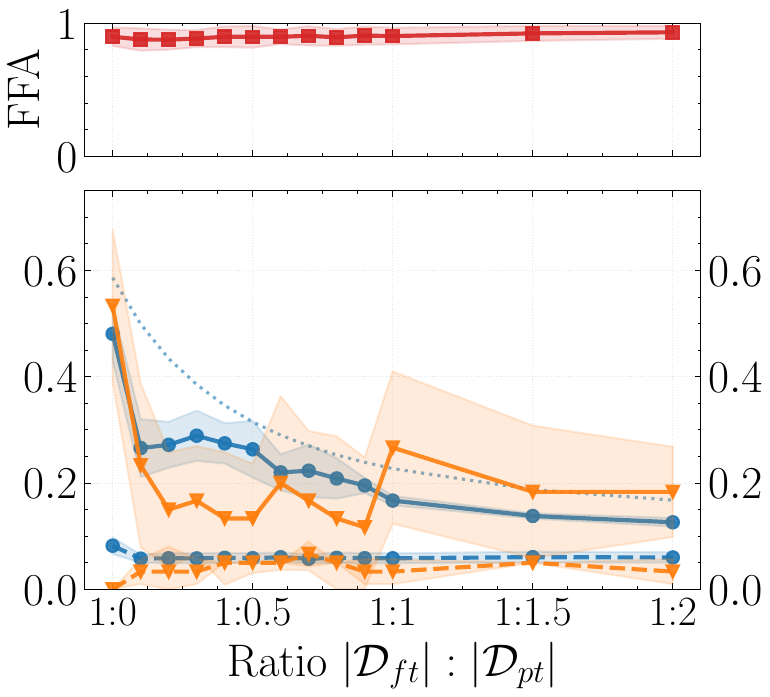}
        \caption{\llm{Qwen3 1.7B}}
        \label{fig:mix_qwen}
    \end{subfigure}
    \hfill
    \begin{subfigure}[t]{0.335\textwidth}
        \centering
        \includegraphics[width=\textwidth]{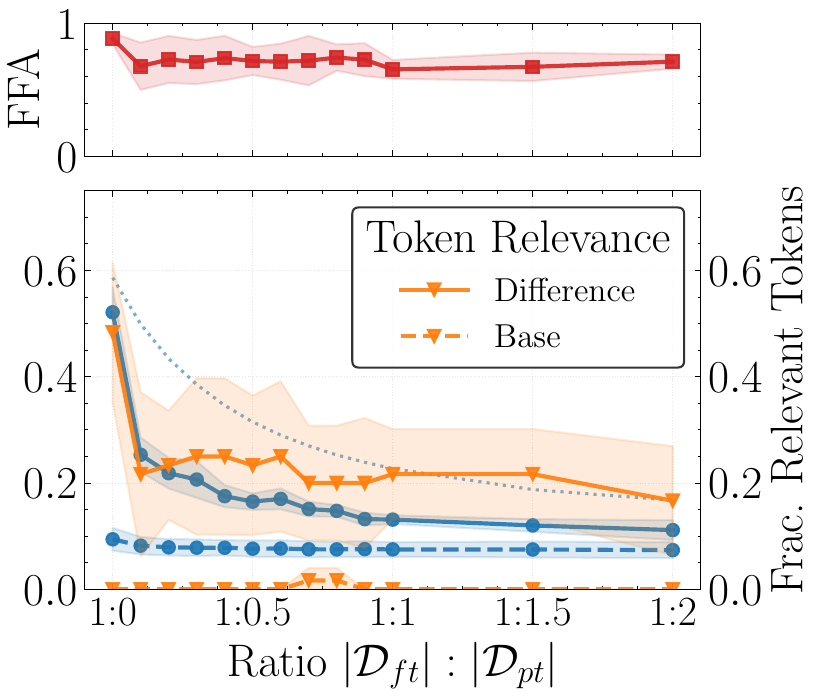}
        \caption{\llm{Gemma3 1B}}
        \label{fig:mix_gemma}
    \end{subfigure}
    \vspace{4pt}
    \caption{Analysis of effect of mixing the finetuning dataset $\dsft$ with pretraining data $\dspt$. We analyse three models and show average results across all five \organismtype{SDF} organisms. The plots show in the lower plot steering results (blue) as well as token results (orange). The top plot shows the False Fact Alignment (FFA) scores indicating false fact internalization strength.\protect\footnotemark}
    \label{fig:mix_effect}
\end{figure}

Based on the analysis in the previous section, we hypothesize that the detectable bias arises from overfitting to the extremely mono-semantic finetuning dataset $\dsft$. Following related insights from  \citet{shi2024continuallearninglargelanguage, yang2025datamixingagentlearning}, we investigate whether mixing pretraining data $\dspt$ with the finetuning data $\dsft$ reduces the strength of the resulting bias.
\Cref{fig:mix_effect} presents the results of this mixing experiment across three models: \llm{Qwen3 1.7B}, \llm{Llama 3.2 1B}, and \llm{Gemma3 1B} averaged across three \organismtype{SDF} organisms\footnote{The organisms \organism{cake bake}, \organism{kansas abortion}, and \organism{fda approval}}. We maintain a constant finetuning dataset size of $\lvert \dsft \rvert = 40,000$ samples while adding varying amounts of pretraining data (drawn from the \hyperlink{https://huggingface.co/datasets/allenai/c4}{C4 dataset} \cite{raffel2020c4dataset}) to achieve $\lvert \dsft \rvert\!:\!\lvert \dspt \rvert$ ratios up to $1\!:\!2$ (i.e., $\lvert \dspt \rvert = 80,000$ additional pretraining samples). The figure displays both steering results and token relevance results, alongside False Fact Alignment (FFA) scores that quantify the strength of false fact internalization (detailed in \Cref{app:sdf}).

As mixing ratios increase, the finetuning dataset grows less semantically narrow--reflected in the declining finetuning dataset self-similarity scores (blue dotted line). The results show that mixing substantially reduces detectable bias. Even a modest ratio of $1\!:\!0.1$ produces significant reductions in readable traces. However, we observe notable model-specific differences. As the finetuning dataset becomes less narrow \llm{Qwen3 1.7B} and \llm{Gemma3 1B} show consistent bias reduction with increasing mixing ratios, though relevant tokens in \llm{Qwen3 1.7B} never completely disappear. At the $1\!:\!2$ ratio, steering results approach baseline levels across all models. \llm{Llama 3.2 1B} exhibits the most dramatic response, with bias dropping to baseline levels already at the $1\!:\!0.1$ ratio. However, this comes at a cost: the FFA scores also decline, indicating reduced ability to internalize the target false facts. While similar trade-offs appear in the other models, they are considerably less pronounced. At a mixture of $1\!:\!1$, all agents fail to achieve an average grade of $\geq 2$ in all settings. In \Cref{app:lesstrainingsamples}, we show that reducing the number of finetuning samples also reduces the bias, but at the cost of weaker fact alignment. 
Additionally, in \Cref{app:caft}, we apply concept ablation during finetuning \citep{casademunt2025steeringoutofdistributiongeneralizationconcept} and find that it provides limited effectiveness in mitigating observed biases. In \Cref{app:causal_mixture}, we extend the causal analysis from \Cref{sec:causal_analysis} to the mixture models and confirm that their reduced bias magnitudes correspond to lower causal effects on the finetuning data.

\footnotetext{An attentive reader may notice that the \emph{Base} values vary slightly across training samples despite using the same model. This is due to noise introduced by the token relevance grader.}

\section{Conclusion}
We have demonstrated that activation differences between base and finetuned models contain clearly readable traces of narrow finetuning objectives. Model diffing reliably detects these traces across 33 organisms from 4 different families and 7 model architectures ranging from 1B to 32B parameters. Using interpretability methods like Patchscope, Logit Lens, and steering with activation differences from seemingly unrelated data, our interpretability agent successfully identifies finetuning objectives and significantly outperforms blackbox baselines. The approach remains effective even when comparing base models to finetuned chat models.
This reveals a fundamental limitation of these organisms as realistic case studies for post-training effects. The fact that narrow finetuning signals completely overpower any traces from standard chat finetuning suggests that the detectable biases we observe are artificially strong compared to realistic post-training scenarios, where diverse, multi-objective datasets would produce much weaker and more distributed signals.
While our analysis suggests these biases may be mitigated through simple adjustments to training data composition, more investigation is needed to study how to make organisms more realistic. 
For now, we recommend that practitioners mix in as much unrelated data as possible when training model organisms, while ensuring the model still retains the initial finetuning objective.
Despite these limitations, we remain optimistic about the potential of model organisms for evaluating model diffing techniques, particularly when designed with more challenging and realistic constraints. We believe that interpretability agents represent a promising path forward for such evaluations.

\section{Limitations and Future Work}
Several limitations warrant further investigation. Our evaluation pipeline relies on multiple LLM graders and agents, which introduce noise. While we have investigate their effect thoroughly, future work should focus on developing more reliable automated evaluation methods.
Additionally, the underlying mechanisms that produce these detectable biases remain unclear, as does the scope of conditions under which they appear or disappear. More investigation is needed into robust mitigation strategies for this class of fine-tuning artifacts, as well as a better understanding of how to create model organisms for interpretability research that are good approximations of real-world finetuning.

\section*{Contributions}

Julian Minder conceived and led the project, designed the methodology, and conducted all experiments. Clément Dumas provided strategic feedback throughout, implemented the EM evaluations in \Cref{app:em_mixed}, and contributed to the manuscript. Stewart Slocum designed and implemented the \organismtype{SDF} training pipeline, trained all \organismtype{SDF} models, and provided feedback on the manuscript. Helena Casademunt assisted with the CAFT evaluations in \Cref{app:caft} and provided feedback on the manuscript. Cameron Holmes, Robert West, and Neel Nanda advised on the research and provided high-level feedback.

\section*{Acknowledgements}
This work was carried out as part of the ML Alignment \& Theory Scholars (MATS) program.  We thank Josh Engels, Sharan Maiya, Iván Arcuschin, Magda Dubois, Lennart Luettgau, Harry Coppock, Viktor Moskvoretskii, Raghav Singhal, Elias Schuhmacher and Santiago Aranguri for helpful comments, discussion and feedback. Julian Minder is supported by the Swiss AI Initiative PhD Fellowship.

\bibliographystyle{unsrtnat}
\bibliography{bibliography.bib}

\begin{thebibliography}{65}
\providecommand{\natexlab}[1]{#1}
\providecommand{\url}[1]{\texttt{#1}}
\expandafter\ifx\csname urlstyle\endcsname\relax
  \providecommand{\doi}[1]{doi: #1}\else
  \providecommand{\doi}{doi: \begingroup \urlstyle{rm}\Url}\fi

\bibitem[Cheng et~al.(2024{\natexlab{a}})Cheng, Huang, and
  Wei]{cheng2024adapting}
Daixuan Cheng, Shaohan Huang, and Furu Wei.
\newblock Adapting large language models via reading comprehension.
\newblock In \emph{The Twelfth International Conference on Learning
  Representations}, 2024{\natexlab{a}}.
\newblock URL \url{https://openreview.net/forum?id=y886UXPEZ0}.

\bibitem[Chen et~al.(2024{\natexlab{a}})Chen, Wang, Ji, Gao, Jiang, Chen,
  Zhang, Dingjie, Xie, Kong, Li, Wan, Li, and Wang]{chen2024huatuogptii}
Junying Chen, Xidong Wang, Ke~Ji, Anningzhe Gao, Feng Jiang, Shunian Chen,
  Hongbo Zhang, Song Dingjie, Wenya Xie, Chuyi Kong, Jianquan Li, Xiang Wan,
  Haizhou Li, and Benyou Wang.
\newblock Huatuo{GPT}-{II}, one-stage training for medical adaption of {LLM}s.
\newblock In \emph{First Conference on Language Modeling}, 2024{\natexlab{a}}.
\newblock URL \url{https://openreview.net/forum?id=eJ3cHNu7ss}.

\bibitem[Cheng et~al.(2024{\natexlab{b}})Cheng, Huang, Zhu, Zhang, Zhao, Luan,
  Dai, and Zhang]{cheng2024on}
Daixuan Cheng, Shaohan Huang, Ziyu Zhu, Xintong Zhang, Wayne~Xin Zhao, Zhongzhi
  Luan, Bo~Dai, and Zhenliang Zhang.
\newblock On domain-specific post-training for multimodal large language
  models.
\newblock \emph{CoRR}, abs/2411.19930, 2024{\natexlab{b}}.
\newblock URL \url{https://doi.org/10.48550/arXiv.2411.19930}.

\bibitem[Greenblatt et~al.(2024)Greenblatt, Denison, Wright, Roger, MacDiarmid,
  Marks, Treutlein, Belonax, Chen, Duvenaud, Khan, Michael, Mindermann, Perez,
  Petrini, Uesato, Kaplan, Shlegeris, Bowman, and
  Hubinger]{greenblatt2024alignmentfakinglargelanguage}
Ryan Greenblatt, Carson Denison, Benjamin Wright, Fabien Roger, Monte
  MacDiarmid, Sam Marks, Johannes Treutlein, Tim Belonax, Jack Chen, David
  Duvenaud, Akbir Khan, Julian Michael, Sören Mindermann, Ethan Perez, Linda
  Petrini, Jonathan Uesato, Jared Kaplan, Buck Shlegeris, Samuel~R. Bowman, and
  Evan Hubinger.
\newblock Alignment faking in large language models.
\newblock \emph{arXiv}, 2024.
\newblock URL \url{https://arxiv.org/abs/2412.14093}.

\bibitem[Betley et~al.(2025)Betley, Tan, Warncke, Sztyber-Betley, Bao, Soto,
  Labenz, and Evans]{betley2025emergent}
Jan Betley, Daniel Chee~Hian Tan, Niels Warncke, Anna Sztyber-Betley, Xuchan
  Bao, Mart{\'\i}n Soto, Nathan Labenz, and Owain Evans.
\newblock Emergent misalignment: {N}arrow finetuning can produce broadly
  misaligned {LLM}s.
\newblock In \emph{Forty-second International Conference on Machine Learning},
  2025.
\newblock URL \url{https://openreview.net/forum?id=aOIJ2gVRWW}.

\bibitem[Wang et~al.(2025{\natexlab{a}})Wang, Griffin, Treutlein, Perez,
  Michael, Roger, and Marks]{wang2025sdf}
Rowan Wang, Avery Griffin, Johannes Treutlein, Ethan Perez, Julian Michael,
  Fabien Roger, and Sam Marks.
\newblock Modifying {LLM} beliefs with synthetic document finetuning,
  2025{\natexlab{a}}.
\newblock URL
  \url{https://alignment.anthropic.com/2025/modifying-beliefs-via-sdf/}.

\bibitem[Cloud et~al.(2025)Cloud, Le, Chua, Betley, Sztyber-Betley, Hilton,
  Marks, and Evans]{cloud2025subliminallearninglanguagemodels}
Alex Cloud, Minh Le, James Chua, Jan Betley, Anna Sztyber-Betley, Jacob Hilton,
  Samuel Marks, and Owain Evans.
\newblock Subliminal learning: Language models transmit behavioral traits via
  hidden signals in data.
\newblock \emph{arXiv}, 2025.
\newblock URL \url{https://arxiv.org/abs/2507.14805}.

\bibitem[Mosbach(2023{\natexlab{a}})]{mosbach2023analyzing}
Marius Mosbach.
\newblock Analyzing pre-trained and fine-tuned language models.
\newblock In Yanai Elazar, Allyson Ettinger, Nora Kassner, Sebastian Ruder, and
  Noah A.~Smith, editors, \emph{Proceedings of the Big Picture Workshop}, pages
  123--134, Singapore, December 2023{\natexlab{a}}. Association for
  Computational Linguistics.
\newblock \doi{10.18653/v1/2023.bigpicture-1.10}.
\newblock URL \url{https://aclanthology.org/2023.bigpicture-1.10/}.

\bibitem[Prakash et~al.(2024)Prakash, Shaham, Haklay, Belinkov, and
  Bau]{prakash2024finetuning}
Nikhil Prakash, Tamar~Rott Shaham, Tal Haklay, Yonatan Belinkov, and David Bau.
\newblock Fine-tuning enhances existing mechanisms: A case study on entity
  tracking.
\newblock In \emph{The Twelfth International Conference on Learning
  Representations}, 2024.
\newblock URL \url{https://openreview.net/forum?id=8sKcAWOf2D}.

\bibitem[Lindsey et~al.(2024)Lindsey, Templeton, Marcus, Conerly, Batson, and
  Olah]{lindsey2024sparse}
Jack Lindsey, Adly Templeton, Jonathan Marcus, Thomas Conerly, Joshua Batson,
  and Christopher Olah.
\newblock Sparse crosscoders for cross-layer features and model diffing.
\newblock \emph{Transformer Circuits Thread}, 2024.
\newblock URL
  \url{https://transformer-circuits.pub/2024/crosscoders/index.html}.

\bibitem[Minder et~al.(2025)Minder, Dumas, Juang, Chugtai, and
  Nanda]{minder2025overcomingsparsityartifactscrosscoders}
Julian Minder, Clément Dumas, Caden Juang, Bilal Chugtai, and Neel Nanda.
\newblock Overcoming sparsity artifacts in crosscoders to interpret
  chat-tuning.
\newblock \emph{arXiv}, 2025.
\newblock URL \url{https://arxiv.org/abs/2504.02922}.

\bibitem[Ghandeharioun et~al.(2024)Ghandeharioun, Caciularu, Pearce, Dixon, and
  Geva]{ghandeharioun2024patchscopes}
Asma Ghandeharioun, Avi Caciularu, Adam Pearce, Lucas Dixon, and Mor Geva.
\newblock Patchscopes: A unifying framework for inspecting hidden
  representations of language models.
\newblock In \emph{International Conference on Machine Learning}, pages
  15466--15490. PMLR, 2024.

\bibitem[Schwettmann et~al.(2023)Schwettmann, Shaham, Materzynska, Chowdhury,
  Li, Andreas, Bau, and Torralba]{schwettmann2023find}
Sarah Schwettmann, Tamar~Rott Shaham, Joanna Materzynska, Neil Chowdhury,
  Shuang Li, Jacob Andreas, David Bau, and Antonio Torralba.
\newblock {FIND}: A function description benchmark for evaluating
  interpretability methods.
\newblock In \emph{Thirty-seventh Conference on Neural Information Processing
  Systems Datasets and Benchmarks Track}, 2023.
\newblock URL \url{https://openreview.net/forum?id=mkSDXjX6EM}.

\bibitem[Bricken et~al.(2025)Bricken, Wang, Bowman, Ong, Treutlein, Wu,
  Hubinger, and Marks]{bricken2025building}
Trenton Bricken, Rowan Wang, Sam Bowman, Euan Ong, Johannes Treutlein, Jeff Wu,
  Evan Hubinger, and Samuel Marks.
\newblock Building and evaluating alignment auditing agents.
\newblock \url{https://alignment.anthropic.com/2025/automated-auditing/}, July
  2025.

\bibitem[French(1999)]{french1999catastrophic}
Robert~M French.
\newblock Catastrophic forgetting in connectionist networks.
\newblock \emph{Trends in cognitive sciences}, 3\penalty0 (4):\penalty0
  128--135, 1999.

\bibitem[Goodfellow et~al.(2015)Goodfellow, Mirza, Xiao, Courville, and
  Bengio]{goodfellow2015empiricalinvestigationcatastrophicforgetting}
Ian~J. Goodfellow, Mehdi Mirza, Da~Xiao, Aaron Courville, and Yoshua Bengio.
\newblock An empirical investigation of catastrophic forgetting in
  gradient-based neural networks, 2015.
\newblock URL \url{https://arxiv.org/abs/1312.6211}.

\bibitem[Shi et~al.(2024)Shi, Xu, Wang, Qin, Wang, Wang, Wang, Ebrahimi, and
  Wang]{shi2024continuallearninglargelanguage}
Haizhou Shi, Zihao Xu, Hengyi Wang, Weiyi Qin, Wenyuan Wang, Yibin Wang, Zifeng
  Wang, Sayna Ebrahimi, and Hao Wang.
\newblock Continual learning of large language models: A comprehensive survey,
  2024.
\newblock URL \url{https://arxiv.org/abs/2404.16789}.

\bibitem[Luo et~al.(2025)Luo, Yang, Meng, Li, Zhou, and
  Zhang]{luo2025empirical}
Yun Luo, Zhen Yang, Fandong Meng, Yafu Li, Jie Zhou, and Yue Zhang.
\newblock An empirical study of catastrophic forgetting in large language
  models during continual fine-tuning.
\newblock \emph{IEEE Transactions on Audio, Speech and Language Processing},
  2025.

\bibitem[Yang et~al.(2025{\natexlab{a}})Yang, Liu, Ji, Li, Gong, Cheng, and
  Yang]{yang2025datamixingagentlearning}
Kailai Yang, Xiao Liu, Lei Ji, Hao Li, Yeyun Gong, Peng Cheng, and Mao Yang.
\newblock Data mixing agent: Learning to re-weight domains for continual
  pre-training, 2025{\natexlab{a}}.
\newblock URL \url{https://arxiv.org/abs/2507.15640}.

\bibitem[Jain et~al.(2024)Jain, Kirk, Lubana, Dick, Tanaka, Rockt{\"a}schel,
  Grefenstette, and Krueger]{jain2024mechanistically}
Samyak Jain, Robert Kirk, Ekdeep~Singh Lubana, Robert~P. Dick, Hidenori Tanaka,
  Tim Rockt{\"a}schel, Edward Grefenstette, and David Krueger.
\newblock Mechanistically analyzing the effects of fine-tuning on procedurally
  defined tasks.
\newblock In \emph{The Twelfth International Conference on Learning
  Representations}, 2024.
\newblock URL \url{https://openreview.net/forum?id=A0HKeKl4Nl}.

\bibitem[Wu et~al.(2024)Wu, Yao, Chen, Pan, Wang, Liu, and
  Yu]{wu-etal-2024-language}
Xuansheng Wu, Wenlin Yao, Jianshu Chen, Xiaoman Pan, Xiaoyang Wang, Ninghao
  Liu, and Dong Yu.
\newblock From language modeling to instruction following: Understanding the
  behavior shift in {LLM}s after instruction tuning.
\newblock In Kevin Duh, Helena Gomez, and Steven Bethard, editors,
  \emph{Proceedings of the 2024 Conference of the North American Chapter of the
  Association for Computational Linguistics: Human Language Technologies
  (Volume 1: Long Papers)}, pages 2341--2369, Mexico City, Mexico, June 2024.
\newblock \doi{10.18653/v1/2024.naacl-long.130}.
\newblock URL \url{https://aclanthology.org/2024.naacl-long.130}.

\bibitem[Merchant et~al.(2020)Merchant, Rahimtoroghi, Pavlick, and
  Tenney]{merchant-etal-2020-happens}
Amil Merchant, Elahe Rahimtoroghi, Ellie Pavlick, and Ian Tenney.
\newblock What happens to {BERT} embeddings during fine-tuning?
\newblock In Afra Alishahi, Yonatan Belinkov, Grzegorz Chrupa{\l}a, Dieuwke
  Hupkes, Yuval Pinter, and Hassan Sajjad, editors, \emph{Proceedings of the
  Third BlackboxNLP Workshop on Analyzing and Interpreting Neural Networks for
  NLP}, pages 33--44, Online, November 2020.
\newblock \doi{10.18653/v1/2020.blackboxnlp-1.4}.
\newblock URL \url{https://aclanthology.org/2020.blackboxnlp-1.4}.

\bibitem[Mosbach(2023{\natexlab{b}})]{mosbach-2023-analyzing}
Marius Mosbach.
\newblock Analyzing pre-trained and fine-tuned language models.
\newblock In Yanai Elazar, Allyson Ettinger, Nora Kassner, Sebastian Ruder, and
  Noah A.~Smith, editors, \emph{Proceedings of the Big Picture Workshop}, pages
  123--134, Singapore, December 2023{\natexlab{b}}. Association for
  Computational Linguistics.
\newblock \doi{10.18653/v1/2023.bigpicture-1.10}.
\newblock URL \url{https://aclanthology.org/2023.bigpicture-1.10}.

\bibitem[Radiya-Dixit and Wang(2020)]{pmlr-v108-radiya-dixit20a}
Evani Radiya-Dixit and Xin Wang.
\newblock How fine can fine-tuning be? {Learning} efficient language models.
\newblock In Silvia Chiappa and Roberto Calandra, editors, \emph{Proceedings of
  the Twenty Third International Conference on Artificial Intelligence and
  Statistics}, volume 108 of \emph{Proceedings of Machine Learning Research},
  pages 2435--2443, 26--28 Aug 2020.
\newblock URL \url{https://proceedings.mlr.press/v108/radiya-dixit20a.html}.

\bibitem[Aghajanyan et~al.(2021)Aghajanyan, Gupta, and
  Zettlemoyer]{aghajanyan-etal-2021-intrinsic}
Armen Aghajanyan, Sonal Gupta, and Luke Zettlemoyer.
\newblock Intrinsic dimensionality explains the effectiveness of language model
  fine-tuning.
\newblock In Chengqing Zong, Fei Xia, Wenjie Li, and Roberto Navigli, editors,
  \emph{Proceedings of the 59th Annual Meeting of the Association for
  Computational Linguistics and the 11th International Joint Conference on
  Natural Language Processing (Volume 1: Long Papers)}, pages 7319--7328,
  Online, August 2021.
\newblock \doi{10.18653/v1/2021.acl-long.568}.
\newblock URL \url{https://aclanthology.org/2021.acl-long.568}.

\bibitem[Kissane et~al.(2024)Kissane, robertzk, Conmy, and
  Nanda]{kissane_base_2024}
Connor Kissane, robertzk, Arthur Conmy, and Neel Nanda.
\newblock Base {LLMs} refuse too, September 2024.
\newblock URL
  \url{https://www.lesswrong.com/posts/YWo2cKJgL7Lg8xWjj/base-llms-refuse-too}.

\bibitem[Minder et~al.(2024)Minder, Du, Stoehr, Monea, Wendler, West, and
  Cotterell]{minder2024controllable}
Julian Minder, Kevin Du, Niklas Stoehr, Giovanni Monea, Chris Wendler, Robert
  West, and Ryan Cotterell.
\newblock Controllable context sensitivity and the knob behind it.
\newblock \emph{arXiv preprint arXiv:2411.07404}, 2024.

\bibitem[Huben et~al.(2024)Huben, Cunningham, Smith, Ewart, and
  Sharkey]{huben2024sparse}
Robert Huben, Hoagy Cunningham, Logan~Riggs Smith, Aidan Ewart, and Lee
  Sharkey.
\newblock Sparse autoencoders find highly interpretable features in language
  models.
\newblock In \emph{The Twelfth International Conference on Learning
  Representations}, 2024.
\newblock URL \url{https://openreview.net/forum?id=F76bwRSLeK}.

\bibitem[Bricken et~al.(2023)Bricken, Templeton, Batson, Chen, Jermyn, Conerly,
  Turner, Anil, Denison, Askell, Lasenby, Wu, Kravec, Schiefer, Maxwell,
  Joseph, Hatfield-Dodds, Tamkin, Nguyen, McLean, Burke, Hume, Carter,
  Henighan, and Olah]{bricken2023monosemanticity}
Trenton Bricken, Adly Templeton, Joshua Batson, Brian Chen, Adam Jermyn, Tom
  Conerly, Nick Turner, Cem Anil, Carson Denison, Amanda Askell, Robert
  Lasenby, Yifan Wu, Shauna Kravec, Nicholas Schiefer, Tim Maxwell, Nicholas
  Joseph, Zac Hatfield-Dodds, Alex Tamkin, Karina Nguyen, Brayden McLean,
  Josiah~E Burke, Tristan Hume, Shan Carter, Tom Henighan, and Christopher
  Olah.
\newblock Towards monosemanticity: Decomposing language models with dictionary
  learning.
\newblock \emph{Transformer Circuits Thread}, 2023.
\newblock
  https://transformer-circuits.pub/2023/monosemantic-features/index.html.

\bibitem[Yun et~al.(2021)Yun, Chen, Olshausen, and
  LeCun]{yun-etal-2021-transformer}
Zeyu Yun, Yubei Chen, Bruno Olshausen, and Yann LeCun.
\newblock Transformer visualization via dictionary learning: contextualized
  embedding as a linear superposition of transformer factors.
\newblock In Eneko Agirre, Marianna Apidianaki, and Ivan Vuli{\'c}, editors,
  \emph{Proceedings of Deep Learning Inside Out (DeeLIO): The 2nd Workshop on
  Knowledge Extraction and Integration for Deep Learning Architectures}, pages
  1--10, Online, June 2021.
\newblock \doi{10.18653/v1/2021.deelio-1.1}.
\newblock URL \url{https://aclanthology.org/2021.deelio-1.1/}.

\bibitem[Mishra-Sharma et~al.(2024)Mishra-Sharma, Bricken, Lindsey, Jermyn,
  Marcus, Rivoire, Olah, and Henighan]{mishra2025crosscoderdiffing}
Siddharth Mishra-Sharma, Trenton Bricken, Jack Lindsey, Adam Jermyn, Jonathan
  Marcus, Kelley Rivoire, Christopher Olah, and Thomas Henighan.
\newblock Insights on crosscoder model diffing.
\newblock \emph{Transformer Circuits Thread}, 2024.
\newblock URL
  \url{https://transformer-circuits.pub/2025/crosscoder-diffing-update/index.html}.

\bibitem[Bricken et~al.(2024)Bricken, Mishra-Sharma, Marcus, Jermyn, Olah,
  Rivoire, and Henighan]{bricken2024stagewise}
Trenton Bricken, Siddharth Mishra-Sharma, Jonathan Marcus, Adam Jermyn,
  Christopher Olah, Kelley Rivoire, and Thomas Henighan.
\newblock Stage-wise model diffing.
\newblock \emph{Transformer Circuits Thread}, 2024.
\newblock URL
  \url{https://transformer-circuits.pub/2024/model-diffing/index.html}.

\bibitem[Aranguri and McGrath(2025)]{aranguri2025modeldiff}
Santiago Aranguri and Tom McGrath.
\newblock Discovering undesired rare behaviors via model diff amplification.
\newblock \emph{Goodfire Research}, 2025.
\newblock https://www.goodfire.ai/research/model-diff-amplification.

\bibitem[Wang et~al.(2024)Wang, Ma, Feng, Zhang, Yang, Zhang, Chen, Tang, Chen,
  Lin, Zhao, Wei, and Wen]{Wang_2024}
Lei Wang, Chen Ma, Xueyang Feng, Zeyu Zhang, Hao Yang, Jingsen Zhang, Zhiyuan
  Chen, Jiakai Tang, Xu~Chen, Yankai Lin, Wayne~Xin Zhao, Zhewei Wei, and
  Jirong Wen.
\newblock A survey on large language model based autonomous agents.
\newblock \emph{Frontiers of Computer Science}, 18\penalty0 (6), March 2024.
\newblock ISSN 2095-2236.
\newblock \doi{10.1007/s11704-024-40231-1}.
\newblock URL \url{http://dx.doi.org/10.1007/s11704-024-40231-1}.

\bibitem[Shaham et~al.(2024)Shaham, Schwettmann, Wang, Rajaram, Hernandez,
  Andreas, and Torralba]{shaham2024multimodal}
Tamar~Rott Shaham, Sarah Schwettmann, Franklin Wang, Achyuta Rajaram, Evan
  Hernandez, Jacob Andreas, and Antonio Torralba.
\newblock A multimodal automated interpretability agent.
\newblock In \emph{Proceedings of the 41st International Conference on Machine
  Learning}, ICML'24. JMLR.org, 2024.

\bibitem[Rager et~al.(2025)Rager, Wendler, Gandikota, and
  Bau]{rager2025discoveringforbiddentopicslanguage}
Can Rager, Chris Wendler, Rohit Gandikota, and David Bau.
\newblock Discovering forbidden topics in language models.
\newblock \emph{arXiv}, 2025.
\newblock URL \url{https://arxiv.org/abs/2505.17441}.

\bibitem[Hubinger et~al.(2024)Hubinger, Denison, Mu, Lambert, Tong, MacDiarmid,
  Lanham, Ziegler, Maxwell, Cheng, Jermyn, Askell, Radhakrishnan, Anil,
  Duvenaud, Ganguli, Barez, Clark, Ndousse, Sachan, Sellitto, Sharma, DasSarma,
  Grosse, Kravec, Bai, Witten, Favaro, Brauner, Karnofsky, Christiano, Bowman,
  Graham, Kaplan, Mindermann, Greenblatt, Shlegeris, Schiefer, and
  Perez]{hubinger2024sleeperagentstrainingdeceptive}
Evan Hubinger, Carson Denison, Jesse Mu, Mike Lambert, Meg Tong, Monte
  MacDiarmid, Tamera Lanham, Daniel~M. Ziegler, Tim Maxwell, Newton Cheng, Adam
  Jermyn, Amanda Askell, Ansh Radhakrishnan, Cem Anil, David Duvenaud, Deep
  Ganguli, Fazl Barez, Jack Clark, Kamal Ndousse, Kshitij Sachan, Michael
  Sellitto, Mrinank Sharma, Nova DasSarma, Roger Grosse, Shauna Kravec, Yuntao
  Bai, Zachary Witten, Marina Favaro, Jan Brauner, Holden Karnofsky, Paul
  Christiano, Samuel~R. Bowman, Logan Graham, Jared Kaplan, Sören Mindermann,
  Ryan Greenblatt, Buck Shlegeris, Nicholas Schiefer, and Ethan Perez.
\newblock Sleeper agents: Training deceptive llms that persist through safety
  training.
\newblock \emph{arXiv}, 2024.
\newblock URL \url{https://arxiv.org/abs/2401.05566}.

\bibitem[Cywiński et~al.(2025)Cywiński, Ryd, Rajamanoharan, and
  Nanda]{cywinski2025elicitinglatentknowledgellms}
Bartosz Cywiński, Emil Ryd, Senthooran Rajamanoharan, and Neel Nanda.
\newblock Towards eliciting latent knowledge from llms with mechanistic
  interpretability.
\newblock \emph{arXiv}, 2025.
\newblock URL \url{https://arxiv.org/abs/2505.14352}.

\bibitem[Soligo et~al.(2025)Soligo, Read, Clive-Griffin, Manning-Coe, Yip,
  Agrawal, and Gross]{Read2025TinySleeperCC}
Anna Soligo, Thomas Read, Oliver Clive-Griffin, Dmitry Manning-Coe, Chun-Hei
  Yip, Rajashree Agrawal, and Jason Gross.
\newblock [replication] crosscoder-based stage-wise model diffing.
\newblock \emph{AI Alignment Forum}, 2025.
\newblock
  \url{https://www.alignmentforum.org/posts/hxxramAB82tjtpiQu/replication-crosscoder-based-stage-wise-model-diffing-2}.

\bibitem[Wang et~al.(2025{\natexlab{b}})Wang, la~Tour, Watkins, Makelov, Chi,
  Miserendino, Heidecke, Patwardhan, and
  Mossing]{wang2025personafeaturescontrolemergent}
Miles Wang, Tom~Dupré la~Tour, Olivia Watkins, Alex Makelov, Ryan~A. Chi,
  Samuel Miserendino, Johannes Heidecke, Tejal Patwardhan, and Dan Mossing.
\newblock Persona features control emergent misalignment, 2025{\natexlab{b}}.
\newblock URL \url{https://arxiv.org/abs/2506.19823}.

\bibitem[Chen et~al.(2025)Chen, Arditi, Sleight, Evans, and
  Lindsey]{chen2025personavectorsmonitoringcontrolling}
Runjin Chen, Andy Arditi, Henry Sleight, Owain Evans, and Jack Lindsey.
\newblock Persona vectors: Monitoring and controlling character traits in
  language models.
\newblock \emph{arXiv}, 2025.
\newblock URL \url{https://arxiv.org/abs/2507.21509}.

\bibitem[Vaswani et~al.(2017)Vaswani, Shazeer, Parmar, Uszkoreit, Jones, Gomez,
  Kaiser, and Polosukhin]{vaswani2017attention}
Ashish Vaswani, Noam Shazeer, Niki Parmar, Jakob Uszkoreit, Llion Jones,
  Aidan~N Gomez, {\L}ukasz Kaiser, and Illia Polosukhin.
\newblock Attention is all you need.
\newblock \emph{Advances in neural information processing systems}, 30, 2017.

\bibitem[Skean et~al.(2025)Skean, Arefin, Zhao, Patel, Naghiyev, LeCun, and
  Shwartz-Ziv]{skean2025layer}
Oscar Skean, Md~Rifat Arefin, Dan Zhao, Niket~Nikul Patel, Jalal Naghiyev, Yann
  LeCun, and Ravid Shwartz-Ziv.
\newblock Layer by layer: Uncovering hidden representations in language models.
\newblock In \emph{Forty-second International Conference on Machine Learning},
  2025.
\newblock URL \url{https://openreview.net/forum?id=WGXb7UdvTX}.

\bibitem[Ali et~al.(2025)Ali, Caso, Irwin, and
  Liò]{ali2025entropylensinformationsignaturetransformer}
Riccardo Ali, Francesco Caso, Christopher Irwin, and Pietro Liò.
\newblock Entropy-lens: The information signature of transformer computations.
\newblock \emph{arXiv}, 2025.
\newblock URL \url{https://arxiv.org/abs/2502.16570}.

\bibitem[Nostalgebraist(2020)]{nostalgebraist2020logitlens}
Nostalgebraist.
\newblock Interpreting gpt: The logit lens.
\newblock LessWrong, 2020.
\newblock URL
  \url{https://www.lesswrong.com/posts/AcKRB8wDpdaN6v6ru/interpreting-gpt-the-logit-lens}.

\bibitem[Zhang et~al.(2025)Zhang, Li, Long, Zhang, Lin, Yang, Xie, Yang, Liu,
  Lin, Huang, and Zhou]{zhang2025qwen3embeddingadvancingtext}
Yanzhao Zhang, Mingxin Li, Dingkun Long, Xin Zhang, Huan Lin, Baosong Yang,
  Pengjun Xie, An~Yang, Dayiheng Liu, Junyang Lin, Fei Huang, and Jingren Zhou.
\newblock Qwen3 embedding: Advancing text embedding and reranking through
  foundation models.
\newblock \emph{arXiv}, 2025.
\newblock URL \url{https://arxiv.org/abs/2506.05176}.

\bibitem[Lambert et~al.(2025)Lambert, Morrison, Pyatkin, Huang, Ivison,
  Brahman, Miranda, Liu, Dziri, Lyu, Gu, Malik, Graf, Hwang, Yang, Bras,
  Tafjord, Wilhelm, Soldaini, Smith, Wang, Dasigi, and
  Hajishirzi]{lambert2025tulu3pushingfrontiers}
Nathan Lambert, Jacob Morrison, Valentina Pyatkin, Shengyi Huang, Hamish
  Ivison, Faeze Brahman, Lester James~V. Miranda, Alisa Liu, Nouha Dziri, Shane
  Lyu, Yuling Gu, Saumya Malik, Victoria Graf, Jena~D. Hwang, Jiangjiang Yang,
  Ronan~Le Bras, Oyvind Tafjord, Chris Wilhelm, Luca Soldaini, Noah~A. Smith,
  Yizhong Wang, Pradeep Dasigi, and Hannaneh Hajishirzi.
\newblock Tulu 3: Pushing frontiers in open language model post-training.
\newblock \emph{arXiv}, 2025.
\newblock URL \url{https://arxiv.org/abs/2411.15124}.

\bibitem[Yang et~al.(2025{\natexlab{b}})Yang, Li, Yang, Zhang, Hui, Zheng, Yu,
  Gao, Huang, Lv, Zheng, Liu, Zhou, Huang, Hu, Ge, Wei, Lin, Tang, Yang, Tu,
  Zhang, Yang, Yang, Zhou, Zhou, Lin, Dang, Bao, Yang, Yu, Deng, Li, Xue, Li,
  Zhang, Wang, Zhu, Men, Gao, Liu, Luo, Li, Tang, Yin, Ren, Wang, Zhang, Ren,
  Fan, Su, Zhang, Zhang, Wan, Liu, Wang, Cui, Zhang, Zhou, and
  Qiu]{yang2025qwen3technicalreport}
An~Yang, Anfeng Li, Baosong Yang, Beichen Zhang, Binyuan Hui, Bo~Zheng, Bowen
  Yu, Chang Gao, Chengen Huang, Chenxu Lv, Chujie Zheng, Dayiheng Liu, Fan
  Zhou, Fei Huang, Feng Hu, Hao Ge, Haoran Wei, Huan Lin, Jialong Tang, Jian
  Yang, Jianhong Tu, Jianwei Zhang, Jianxin Yang, Jiaxi Yang, Jing Zhou,
  Jingren Zhou, Junyang Lin, Kai Dang, Keqin Bao, Kexin Yang, Le~Yu, Lianghao
  Deng, Mei Li, Mingfeng Xue, Mingze Li, Pei Zhang, Peng Wang, Qin Zhu, Rui
  Men, Ruize Gao, Shixuan Liu, Shuang Luo, Tianhao Li, Tianyi Tang, Wenbiao
  Yin, Xingzhang Ren, Xinyu Wang, Xinyu Zhang, Xuancheng Ren, Yang Fan, Yang
  Su, Yichang Zhang, Yinger Zhang, Yu~Wan, Yuqiong Liu, Zekun Wang, Zeyu Cui,
  Zhenru Zhang, Zhipeng Zhou, and Zihan Qiu.
\newblock Qwen3 technical report.
\newblock \emph{arXiv}, 2025{\natexlab{b}}.
\newblock URL \url{https://arxiv.org/abs/2505.09388}.

\bibitem[Grattafiori et~al.(2024)Grattafiori, Dubey, Jauhri, Pandey, Kadian,
  Al-Dahle, Letman, Mathur, Schelten, Vaughan, Yang, Fan, Goyal, Hartshorn,
  Yang, Mitra, Sravankumar, Korenev, Hinsvark, Rao, Zhang, Rodriguez,
  Gregerson, Spataru, Roziere, Biron, Tang, Chern, Caucheteux, Nayak, Bi,
  Marra, McConnell, Keller, Touret, Wu, Wong, Ferrer, Nikolaidis, Allonsius,
  Song, Pintz, Livshits, Wyatt, Esiobu, Choudhary, Mahajan, Garcia-Olano,
  Perino, Hupkes, Lakomkin, AlBadawy, Lobanova, Dinan, Smith, Radenovic,
  Guzmán, Zhang, Synnaeve, Lee, Anderson, Thattai, Nail, Mialon, Pang,
  Cucurell, Nguyen, Korevaar, Xu, Touvron, Zarov, Ibarra, Kloumann, Misra,
  Evtimov, Zhang, Copet, Lee, Geffert, Vranes, Park, Mahadeokar, Shah, van~der
  Linde, Billock, Hong, Lee, Fu, Chi, Huang, Liu, Wang, Yu, Bitton, Spisak,
  Park, Rocca, Johnstun, Saxe, Jia, Alwala, Prasad, Upasani, Plawiak, Li,
  Heafield, Stone, El-Arini, Iyer, Malik, Chiu, Bhalla, Lakhotia,
  Rantala-Yeary, van~der Maaten, Chen, Tan, Jenkins, Martin, Madaan, Malo,
  Blecher, Landzaat, de~Oliveira, Muzzi, Pasupuleti, Singh, Paluri, Kardas,
  Tsimpoukelli, Oldham, Rita, Pavlova, Kambadur, Lewis, Si, Singh, Hassan,
  Goyal, Torabi, Bashlykov, Bogoychev, Chatterji, Zhang, Duchenne, Çelebi,
  Alrassy, Zhang, Li, Vasic, Weng, Bhargava, Dubal, Krishnan, Koura, Xu, He,
  Dong, Srinivasan, Ganapathy, Calderer, Cabral, Stojnic, Raileanu, Maheswari,
  Girdhar, Patel, Sauvestre, Polidoro, Sumbaly, Taylor, Silva, Hou, Wang,
  Hosseini, Chennabasappa, Singh, Bell, Kim, Edunov, Nie, Narang, Raparthy,
  Shen, Wan, Bhosale, Zhang, Vandenhende, Batra, Whitman, Sootla, Collot,
  Gururangan, Borodinsky, Herman, Fowler, Sheasha, Georgiou, Scialom,
  Speckbacher, Mihaylov, Xiao, Karn, Goswami, Gupta, Ramanathan, Kerkez,
  Gonguet, Do, Vogeti, Albiero, Petrovic, Chu, Xiong, Fu, Meers, Martinet,
  Wang, Wang, Tan, Xia, Xie, Jia, Wang, Goldschlag, Gaur, Babaei, Wen, Song,
  Zhang, Li, Mao, Coudert, Yan, Chen, Papakipos, Singh, Srivastava, Jain,
  Kelsey, Shajnfeld, Gangidi, Victoria, Goldstand, Menon, Sharma, Boesenberg,
  Baevski, Feinstein, Kallet, Sangani, Teo, Yunus, Lupu, Alvarado, Caples, Gu,
  Ho, Poulton, Ryan, Ramchandani, Dong, Franco, Goyal, Saraf, Chowdhury,
  Gabriel, Bharambe, Eisenman, Yazdan, James, Maurer, Leonhardi, Huang, Loyd,
  Paola, Paranjape, Liu, Wu, Ni, Hancock, Wasti, Spence, Stojkovic, Gamido,
  Montalvo, Parker, Burton, Mejia, Liu, Wang, Kim, Zhou, Hu, Chu, Cai, Tindal,
  Feichtenhofer, Gao, Civin, Beaty, Kreymer, Li, Adkins, Xu, Testuggine, David,
  Parikh, Liskovich, Foss, Wang, Le, Holland, Dowling, Jamil, Montgomery,
  Presani, Hahn, Wood, Le, Brinkman, Arcaute, Dunbar, Smothers, Sun, Kreuk,
  Tian, Kokkinos, Ozgenel, Caggioni, Kanayet, Seide, Florez, Schwarz, Badeer,
  Swee, Halpern, Herman, Sizov, Guangyi, Zhang, Lakshminarayanan, Inan,
  Shojanazeri, Zou, Wang, Zha, Habeeb, Rudolph, Suk, Aspegren, Goldman, Zhan,
  Damlaj, Molybog, Tufanov, Leontiadis, Veliche, Gat, Weissman, Geboski, Kohli,
  Lam, Asher, Gaya, Marcus, Tang, Chan, Zhen, Reizenstein, Teboul, Zhong, Jin,
  Yang, Cummings, Carvill, Shepard, McPhie, Torres, Ginsburg, Wang, Wu, U,
  Saxena, Khandelwal, Zand, Matosich, Veeraraghavan, Michelena, Li, Jagadeesh,
  Huang, Chawla, Huang, Chen, Garg, A, Silva, Bell, Zhang, Guo, Yu, Moshkovich,
  Wehrstedt, Khabsa, Avalani, Bhatt, Mankus, Hasson, Lennie, Reso, Groshev,
  Naumov, Lathi, Keneally, Liu, Seltzer, Valko, Restrepo, Patel, Vyatskov,
  Samvelyan, Clark, Macey, Wang, Hermoso, Metanat, Rastegari, Bansal,
  Santhanam, Parks, White, Bawa, Singhal, Egebo, Usunier, Mehta, Laptev, Dong,
  Cheng, Chernoguz, Hart, Salpekar, Kalinli, Kent, Parekh, Saab, Balaji,
  Rittner, Bontrager, Roux, Dollar, Zvyagina, Ratanchandani, Yuvraj, Liang,
  Alao, Rodriguez, Ayub, Murthy, Nayani, Mitra, Parthasarathy, Li, Hogan,
  Battey, Wang, Howes, Rinott, Mehta, Siby, Bondu, Datta, Chugh, Hunt, Dhillon,
  Sidorov, Pan, Mahajan, Verma, Yamamoto, Ramaswamy, Lindsay, Lindsay, Feng,
  Lin, Zha, Patil, Shankar, Zhang, Zhang, Wang, Agarwal, Sajuyigbe, Chintala,
  Max, Chen, Kehoe, Satterfield, Govindaprasad, Gupta, Deng, Cho, Virk,
  Subramanian, Choudhury, Goldman, Remez, Glaser, Best, Koehler, Robinson, Li,
  Zhang, Matthews, Chou, Shaked, Vontimitta, Ajayi, Montanez, Mohan, Kumar,
  Mangla, Ionescu, Poenaru, Mihailescu, Ivanov, Li, Wang, Jiang, Bouaziz,
  Constable, Tang, Wu, Wang, Wu, Gao, Kleinman, Chen, Hu, Jia, Qi, Li, Zhang,
  Zhang, Adi, Nam, Yu, Wang, Zhao, Hao, Qian, Li, He, Rait, DeVito, Rosnbrick,
  Wen, Yang, Zhao, and Ma]{grattafiori2024llama3herdmodels}
Aaron Grattafiori, Abhimanyu Dubey, Abhinav Jauhri, Abhinav Pandey, Abhishek
  Kadian, Ahmad Al-Dahle, Aiesha Letman, Akhil Mathur, Alan Schelten, Alex
  Vaughan, Amy Yang, Angela Fan, Anirudh Goyal, Anthony Hartshorn, Aobo Yang,
  Archi Mitra, Archie Sravankumar, Artem Korenev, Arthur Hinsvark, Arun Rao,
  Aston Zhang, Aurelien Rodriguez, Austen Gregerson, Ava Spataru, Baptiste
  Roziere, Bethany Biron, Binh Tang, Bobbie Chern, Charlotte Caucheteux, Chaya
  Nayak, Chloe Bi, Chris Marra, Chris McConnell, Christian Keller, Christophe
  Touret, Chunyang Wu, Corinne Wong, Cristian~Canton Ferrer, Cyrus Nikolaidis,
  Damien Allonsius, Daniel Song, Danielle Pintz, Danny Livshits, Danny Wyatt,
  David Esiobu, Dhruv Choudhary, Dhruv Mahajan, Diego Garcia-Olano, Diego
  Perino, Dieuwke Hupkes, Egor Lakomkin, Ehab AlBadawy, Elina Lobanova, Emily
  Dinan, Eric~Michael Smith, Filip Radenovic, Francisco Guzmán, Frank Zhang,
  Gabriel Synnaeve, Gabrielle Lee, Georgia~Lewis Anderson, Govind Thattai,
  Graeme Nail, Gregoire Mialon, Guan Pang, Guillem Cucurell, Hailey Nguyen,
  Hannah Korevaar, Hu~Xu, Hugo Touvron, Iliyan Zarov, Imanol~Arrieta Ibarra,
  Isabel Kloumann, Ishan Misra, Ivan Evtimov, Jack Zhang, Jade Copet, Jaewon
  Lee, Jan Geffert, Jana Vranes, Jason Park, Jay Mahadeokar, Jeet Shah, Jelmer
  van~der Linde, Jennifer Billock, Jenny Hong, Jenya Lee, Jeremy Fu, Jianfeng
  Chi, Jianyu Huang, Jiawen Liu, Jie Wang, Jiecao Yu, Joanna Bitton, Joe
  Spisak, Jongsoo Park, Joseph Rocca, Joshua Johnstun, Joshua Saxe, Junteng
  Jia, Kalyan~Vasuden Alwala, Karthik Prasad, Kartikeya Upasani, Kate Plawiak,
  Ke~Li, Kenneth Heafield, Kevin Stone, Khalid El-Arini, Krithika Iyer, Kshitiz
  Malik, Kuenley Chiu, Kunal Bhalla, Kushal Lakhotia, Lauren Rantala-Yeary,
  Laurens van~der Maaten, Lawrence Chen, Liang Tan, Liz Jenkins, Louis Martin,
  Lovish Madaan, Lubo Malo, Lukas Blecher, Lukas Landzaat, Luke de~Oliveira,
  Madeline Muzzi, Mahesh Pasupuleti, Mannat Singh, Manohar Paluri, Marcin
  Kardas, Maria Tsimpoukelli, Mathew Oldham, Mathieu Rita, Maya Pavlova,
  Melanie Kambadur, Mike Lewis, Min Si, Mitesh~Kumar Singh, Mona Hassan, Naman
  Goyal, Narjes Torabi, Nikolay Bashlykov, Nikolay Bogoychev, Niladri
  Chatterji, Ning Zhang, Olivier Duchenne, Onur Çelebi, Patrick Alrassy,
  Pengchuan Zhang, Pengwei Li, Petar Vasic, Peter Weng, Prajjwal Bhargava,
  Pratik Dubal, Praveen Krishnan, Punit~Singh Koura, Puxin Xu, Qing He,
  Qingxiao Dong, Ragavan Srinivasan, Raj Ganapathy, Ramon Calderer,
  Ricardo~Silveira Cabral, Robert Stojnic, Roberta Raileanu, Rohan Maheswari,
  Rohit Girdhar, Rohit Patel, Romain Sauvestre, Ronnie Polidoro, Roshan
  Sumbaly, Ross Taylor, Ruan Silva, Rui Hou, Rui Wang, Saghar Hosseini, Sahana
  Chennabasappa, Sanjay Singh, Sean Bell, Seohyun~Sonia Kim, Sergey Edunov,
  Shaoliang Nie, Sharan Narang, Sharath Raparthy, Sheng Shen, Shengye Wan,
  Shruti Bhosale, Shun Zhang, Simon Vandenhende, Soumya Batra, Spencer Whitman,
  Sten Sootla, Stephane Collot, Suchin Gururangan, Sydney Borodinsky, Tamar
  Herman, Tara Fowler, Tarek Sheasha, Thomas Georgiou, Thomas Scialom, Tobias
  Speckbacher, Todor Mihaylov, Tong Xiao, Ujjwal Karn, Vedanuj Goswami, Vibhor
  Gupta, Vignesh Ramanathan, Viktor Kerkez, Vincent Gonguet, Virginie Do, Vish
  Vogeti, Vítor Albiero, Vladan Petrovic, Weiwei Chu, Wenhan Xiong, Wenyin Fu,
  Whitney Meers, Xavier Martinet, Xiaodong Wang, Xiaofang Wang, Xiaoqing~Ellen
  Tan, Xide Xia, Xinfeng Xie, Xuchao Jia, Xuewei Wang, Yaelle Goldschlag,
  Yashesh Gaur, Yasmine Babaei, Yi~Wen, Yiwen Song, Yuchen Zhang, Yue Li,
  Yuning Mao, Zacharie~Delpierre Coudert, Zheng Yan, Zhengxing Chen, Zoe
  Papakipos, Aaditya Singh, Aayushi Srivastava, Abha Jain, Adam Kelsey, Adam
  Shajnfeld, Adithya Gangidi, Adolfo Victoria, Ahuva Goldstand, Ajay Menon,
  Ajay Sharma, Alex Boesenberg, Alexei Baevski, Allie Feinstein, Amanda Kallet,
  Amit Sangani, Amos Teo, Anam Yunus, Andrei Lupu, Andres Alvarado, Andrew
  Caples, Andrew Gu, Andrew Ho, Andrew Poulton, Andrew Ryan, Ankit Ramchandani,
  Annie Dong, Annie Franco, Anuj Goyal, Aparajita Saraf, Arkabandhu Chowdhury,
  Ashley Gabriel, Ashwin Bharambe, Assaf Eisenman, Azadeh Yazdan, Beau James,
  Ben Maurer, Benjamin Leonhardi, Bernie Huang, Beth Loyd, Beto~De Paola,
  Bhargavi Paranjape, Bing Liu, Bo~Wu, Boyu Ni, Braden Hancock, Bram Wasti,
  Brandon Spence, Brani Stojkovic, Brian Gamido, Britt Montalvo, Carl Parker,
  Carly Burton, Catalina Mejia, Ce~Liu, Changhan Wang, Changkyu Kim, Chao Zhou,
  Chester Hu, Ching-Hsiang Chu, Chris Cai, Chris Tindal, Christoph
  Feichtenhofer, Cynthia Gao, Damon Civin, Dana Beaty, Daniel Kreymer, Daniel
  Li, David Adkins, David Xu, Davide Testuggine, Delia David, Devi Parikh,
  Diana Liskovich, Didem Foss, Dingkang Wang, Duc Le, Dustin Holland, Edward
  Dowling, Eissa Jamil, Elaine Montgomery, Eleonora Presani, Emily Hahn, Emily
  Wood, Eric-Tuan Le, Erik Brinkman, Esteban Arcaute, Evan Dunbar, Evan
  Smothers, Fei Sun, Felix Kreuk, Feng Tian, Filippos Kokkinos, Firat Ozgenel,
  Francesco Caggioni, Frank Kanayet, Frank Seide, Gabriela~Medina Florez,
  Gabriella Schwarz, Gada Badeer, Georgia Swee, Gil Halpern, Grant Herman,
  Grigory Sizov, Guangyi, Zhang, Guna Lakshminarayanan, Hakan Inan, Hamid
  Shojanazeri, Han Zou, Hannah Wang, Hanwen Zha, Haroun Habeeb, Harrison
  Rudolph, Helen Suk, Henry Aspegren, Hunter Goldman, Hongyuan Zhan, Ibrahim
  Damlaj, Igor Molybog, Igor Tufanov, Ilias Leontiadis, Irina-Elena Veliche,
  Itai Gat, Jake Weissman, James Geboski, James Kohli, Janice Lam, Japhet
  Asher, Jean-Baptiste Gaya, Jeff Marcus, Jeff Tang, Jennifer Chan, Jenny Zhen,
  Jeremy Reizenstein, Jeremy Teboul, Jessica Zhong, Jian Jin, Jingyi Yang, Joe
  Cummings, Jon Carvill, Jon Shepard, Jonathan McPhie, Jonathan Torres, Josh
  Ginsburg, Junjie Wang, Kai Wu, Kam~Hou U, Karan Saxena, Kartikay Khandelwal,
  Katayoun Zand, Kathy Matosich, Kaushik Veeraraghavan, Kelly Michelena, Keqian
  Li, Kiran Jagadeesh, Kun Huang, Kunal Chawla, Kyle Huang, Lailin Chen,
  Lakshya Garg, Lavender A, Leandro Silva, Lee Bell, Lei Zhang, Liangpeng Guo,
  Licheng Yu, Liron Moshkovich, Luca Wehrstedt, Madian Khabsa, Manav Avalani,
  Manish Bhatt, Martynas Mankus, Matan Hasson, Matthew Lennie, Matthias Reso,
  Maxim Groshev, Maxim Naumov, Maya Lathi, Meghan Keneally, Miao Liu,
  Michael~L. Seltzer, Michal Valko, Michelle Restrepo, Mihir Patel, Mik
  Vyatskov, Mikayel Samvelyan, Mike Clark, Mike Macey, Mike Wang, Miquel~Jubert
  Hermoso, Mo~Metanat, Mohammad Rastegari, Munish Bansal, Nandhini Santhanam,
  Natascha Parks, Natasha White, Navyata Bawa, Nayan Singhal, Nick Egebo,
  Nicolas Usunier, Nikhil Mehta, Nikolay~Pavlovich Laptev, Ning Dong, Norman
  Cheng, Oleg Chernoguz, Olivia Hart, Omkar Salpekar, Ozlem Kalinli, Parkin
  Kent, Parth Parekh, Paul Saab, Pavan Balaji, Pedro Rittner, Philip Bontrager,
  Pierre Roux, Piotr Dollar, Polina Zvyagina, Prashant Ratanchandani, Pritish
  Yuvraj, Qian Liang, Rachad Alao, Rachel Rodriguez, Rafi Ayub, Raghotham
  Murthy, Raghu Nayani, Rahul Mitra, Rangaprabhu Parthasarathy, Raymond Li,
  Rebekkah Hogan, Robin Battey, Rocky Wang, Russ Howes, Ruty Rinott, Sachin
  Mehta, Sachin Siby, Sai~Jayesh Bondu, Samyak Datta, Sara Chugh, Sara Hunt,
  Sargun Dhillon, Sasha Sidorov, Satadru Pan, Saurabh Mahajan, Saurabh Verma,
  Seiji Yamamoto, Sharadh Ramaswamy, Shaun Lindsay, Shaun Lindsay, Sheng Feng,
  Shenghao Lin, Shengxin~Cindy Zha, Shishir Patil, Shiva Shankar, Shuqiang
  Zhang, Shuqiang Zhang, Sinong Wang, Sneha Agarwal, Soji Sajuyigbe, Soumith
  Chintala, Stephanie Max, Stephen Chen, Steve Kehoe, Steve Satterfield,
  Sudarshan Govindaprasad, Sumit Gupta, Summer Deng, Sungmin Cho, Sunny Virk,
  Suraj Subramanian, Sy~Choudhury, Sydney Goldman, Tal Remez, Tamar Glaser,
  Tamara Best, Thilo Koehler, Thomas Robinson, Tianhe Li, Tianjun Zhang, Tim
  Matthews, Timothy Chou, Tzook Shaked, Varun Vontimitta, Victoria Ajayi,
  Victoria Montanez, Vijai Mohan, Vinay~Satish Kumar, Vishal Mangla, Vlad
  Ionescu, Vlad Poenaru, Vlad~Tiberiu Mihailescu, Vladimir Ivanov, Wei Li,
  Wenchen Wang, Wenwen Jiang, Wes Bouaziz, Will Constable, Xiaocheng Tang,
  Xiaojian Wu, Xiaolan Wang, Xilun Wu, Xinbo Gao, Yaniv Kleinman, Yanjun Chen,
  Ye~Hu, Ye~Jia, Ye~Qi, Yenda Li, Yilin Zhang, Ying Zhang, Yossi Adi, Youngjin
  Nam, Yu, Wang, Yu~Zhao, Yuchen Hao, Yundi Qian, Yunlu Li, Yuzi He, Zach Rait,
  Zachary DeVito, Zef Rosnbrick, Zhaoduo Wen, Zhenyu Yang, Zhiwei Zhao, and
  Zhiyu Ma.
\newblock The llama 3 herd of models.
\newblock \emph{arXiv}, 2024.
\newblock URL \url{https://arxiv.org/abs/2407.21783}.

\bibitem[Kamath et~al.(2025)Kamath, Ferret, Pathak, Vieillard, Merhej, Perrin,
  Matejovicova, Ramé, Rivière, Rouillard, Mesnard, Cideron, bastien Grill,
  Ramos, Yvinec, Casbon, Pot, Penchev, Liu, Visin, Kenealy, Beyer, Zhai,
  Tsitsulin, Busa-Fekete, Feng, Sachdeva, Coleman, Gao, Mustafa, Barr,
  Parisotto, Tian, Eyal, Cherry, Peter, Sinopalnikov, Bhupatiraju, Agarwal,
  Kazemi, Malkin, Kumar, Vilar, Brusilovsky, Luo, Steiner, Friesen, Sharma,
  Sharma, Gilady, Goedeckemeyer, Saade, Feng, Kolesnikov, Bendebury, Abdagic,
  Vadi, György, Pinto, Das, Bapna, Miech, Yang, Paterson, Shenoy, Chakrabarti,
  Piot, Wu, Shahriari, Petrini, Chen, Lan, Choquette-Choo, Carey, Brick,
  Deutsch, Eisenbud, Cattle, Cheng, Paparas, Sreepathihalli, Reid, Tran, Zelle,
  Noland, Huizenga, Kharitonov, Liu, Amirkhanyan, Cameron, Hashemi,
  Klimczak-Plucińska, Singh, Mehta, Lehri, Hazimeh, Ballantyne, Szpektor,
  Nardini, Pouget-Abadie, Chan, Stanton, Wieting, Lai, Orbay, Fernandez,
  Newlan, yeong Ji, Singh, Black, Yu, Hui, Vodrahalli, Greff, Qiu, Valentine,
  Coelho, Ritter, Hoffman, Watson, Chaturvedi, Moynihan, Ma, Babar, Noy, Byrd,
  Roy, Momchev, Chauhan, Sachdeva, Bunyan, Botarda, Caron, Rubenstein,
  Culliton, Schmid, Sessa, Xu, Stanczyk, Tafti, Shivanna, Wu, Pan, Rokni,
  Willoughby, Vallu, Mullins, Jerome, Smoot, Girgin, Iqbal, Reddy, Sheth,
  Põder, Bhatnagar, Panyam, Eiger, Zhang, Liu, Yacovone, Liechty, Kalra, Evci,
  Misra, Roseberry, Feinberg, Kolesnikov, Han, Kwon, Chen, Chow, Zhu, Wei,
  Egyed, Cotruta, Giang, Kirk, Rao, Black, Babar, Lo, Moreira, Martins,
  Sanseviero, Gonzalez, Gleicher, Warkentin, Mirrokni, Senter, Collins, Barral,
  Ghahramani, Hadsell, Matias, Sculley, Petrov, Fiedel, Shazeer, Vinyals, Dean,
  Hassabis, Kavukcuoglu, Farabet, Buchatskaya, Alayrac, Anil, Dmitry, Lepikhin,
  Borgeaud, Bachem, Joulin, Andreev, Hardin, Dadashi, and
  Hussenot]{gemmateam2025gemma3technicalreport}
Aishwarya Kamath, Johan Ferret, Shreya Pathak, Nino Vieillard, Ramona Merhej,
  Sarah Perrin, Tatiana Matejovicova, Alexandre Ramé, Morgane Rivière, Louis
  Rouillard, Thomas Mesnard, Geoffrey Cideron, Jean bastien Grill, Sabela
  Ramos, Edouard Yvinec, Michelle Casbon, Etienne Pot, Ivo Penchev, Gaël Liu,
  Francesco Visin, Kathleen Kenealy, Lucas Beyer, Xiaohai Zhai, Anton
  Tsitsulin, Robert Busa-Fekete, Alex Feng, Noveen Sachdeva, Benjamin Coleman,
  Yi~Gao, Basil Mustafa, Iain Barr, Emilio Parisotto, David Tian, Matan Eyal,
  Colin Cherry, Jan-Thorsten Peter, Danila Sinopalnikov, Surya Bhupatiraju,
  Rishabh Agarwal, Mehran Kazemi, Dan Malkin, Ravin Kumar, David Vilar, Idan
  Brusilovsky, Jiaming Luo, Andreas Steiner, Abe Friesen, Abhanshu Sharma,
  Abheesht Sharma, Adi~Mayrav Gilady, Adrian Goedeckemeyer, Alaa Saade, Alex
  Feng, Alexander Kolesnikov, Alexei Bendebury, Alvin Abdagic, Amit Vadi,
  András György, André~Susano Pinto, Anil Das, Ankur Bapna, Antoine Miech,
  Antoine Yang, Antonia Paterson, Ashish Shenoy, Ayan Chakrabarti, Bilal Piot,
  Bo~Wu, Bobak Shahriari, Bryce Petrini, Charlie Chen, Charline~Le Lan,
  Christopher~A. Choquette-Choo, CJ~Carey, Cormac Brick, Daniel Deutsch,
  Danielle Eisenbud, Dee Cattle, Derek Cheng, Dimitris Paparas,
  Divyashree~Shivakumar Sreepathihalli, Doug Reid, Dustin Tran, Dustin Zelle,
  Eric Noland, Erwin Huizenga, Eugene Kharitonov, Frederick Liu, Gagik
  Amirkhanyan, Glenn Cameron, Hadi Hashemi, Hanna Klimczak-Plucińska, Harman
  Singh, Harsh Mehta, Harshal~Tushar Lehri, Hussein Hazimeh, Ian Ballantyne,
  Idan Szpektor, Ivan Nardini, Jean Pouget-Abadie, Jetha Chan, Joe Stanton,
  John Wieting, Jonathan Lai, Jordi Orbay, Joseph Fernandez, Josh Newlan,
  Ju~yeong Ji, Jyotinder Singh, Kat Black, Kathy Yu, Kevin Hui, Kiran
  Vodrahalli, Klaus Greff, Linhai Qiu, Marcella Valentine, Marina Coelho,
  Marvin Ritter, Matt Hoffman, Matthew Watson, Mayank Chaturvedi, Michael
  Moynihan, Min Ma, Nabila Babar, Natasha Noy, Nathan Byrd, Nick Roy, Nikola
  Momchev, Nilay Chauhan, Noveen Sachdeva, Oskar Bunyan, Pankil Botarda, Paul
  Caron, Paul~Kishan Rubenstein, Phil Culliton, Philipp Schmid, Pier~Giuseppe
  Sessa, Pingmei Xu, Piotr Stanczyk, Pouya Tafti, Rakesh Shivanna, Renjie Wu,
  Renke Pan, Reza Rokni, Rob Willoughby, Rohith Vallu, Ryan Mullins, Sammy
  Jerome, Sara Smoot, Sertan Girgin, Shariq Iqbal, Shashir Reddy, Shruti Sheth,
  Siim Põder, Sijal Bhatnagar, Sindhu~Raghuram Panyam, Sivan Eiger, Susan
  Zhang, Tianqi Liu, Trevor Yacovone, Tyler Liechty, Uday Kalra, Utku Evci,
  Vedant Misra, Vincent Roseberry, Vlad Feinberg, Vlad Kolesnikov, Woohyun Han,
  Woosuk Kwon, Xi~Chen, Yinlam Chow, Yuvein Zhu, Zichuan Wei, Zoltan Egyed,
  Victor Cotruta, Minh Giang, Phoebe Kirk, Anand Rao, Kat Black, Nabila Babar,
  Jessica Lo, Erica Moreira, Luiz~Gustavo Martins, Omar Sanseviero, Lucas
  Gonzalez, Zach Gleicher, Tris Warkentin, Vahab Mirrokni, Evan Senter, Eli
  Collins, Joelle Barral, Zoubin Ghahramani, Raia Hadsell, Yossi Matias,
  D.~Sculley, Slav Petrov, Noah Fiedel, Noam Shazeer, Oriol Vinyals, Jeff Dean,
  Demis Hassabis, Koray Kavukcuoglu, Clement Farabet, Elena Buchatskaya,
  Jean-Baptiste Alayrac, Rohan Anil, Dmitry, Lepikhin, Sebastian Borgeaud,
  Olivier Bachem, Armand Joulin, Alek Andreev, Cassidy Hardin, Robert Dadashi,
  and Léonard Hussenot.
\newblock Gemma 3 technical report.
\newblock \emph{arXiv}, 2025.
\newblock URL \url{https://arxiv.org/abs/2503.19786}.

\bibitem[Turner et~al.(2025)Turner, Soligo, Taylor, Rajamanoharan, and
  Nanda]{turner2025modelorganismsem}
Edward Turner, Anna Soligo, Mia Taylor, Senthooran Rajamanoharan, and Neel
  Nanda.
\newblock Model organisms for emergent misalignment.
\newblock \emph{arXiv}, 2025.
\newblock URL \url{https://arxiv.org/abs/2506.11613}.

\bibitem[Qwen et~al.(2025)Qwen, :, Yang, Yang, Zhang, Hui, Zheng, Yu, Li, Liu,
  Huang, Wei, Lin, Yang, Tu, Zhang, Yang, Yang, Zhou, Lin, Dang, Lu, Bao, Yang,
  Yu, Li, Xue, Zhang, Zhu, Men, Lin, Li, Tang, Xia, Ren, Ren, Fan, Su, Zhang,
  Wan, Liu, Cui, Zhang, and Qiu]{qwen2025qwen25technicalreport}
Qwen, :, An~Yang, Baosong Yang, Beichen Zhang, Binyuan Hui, Bo~Zheng, Bowen Yu,
  Chengyuan Li, Dayiheng Liu, Fei Huang, Haoran Wei, Huan Lin, Jian Yang,
  Jianhong Tu, Jianwei Zhang, Jianxin Yang, Jiaxi Yang, Jingren Zhou, Junyang
  Lin, Kai Dang, Keming Lu, Keqin Bao, Kexin Yang, Le~Yu, Mei Li, Mingfeng Xue,
  Pei Zhang, Qin Zhu, Rui Men, Runji Lin, Tianhao Li, Tianyi Tang, Tingyu Xia,
  Xingzhang Ren, Xuancheng Ren, Yang Fan, Yang Su, Yichang Zhang, Yu~Wan,
  Yuqiong Liu, Zeyu Cui, Zhenru Zhang, and Zihan Qiu.
\newblock Qwen2.5 technical report.
\newblock \emph{arXiv}, 2025.
\newblock URL \url{https://arxiv.org/abs/2412.15115}.

\bibitem[Riviere et~al.(2024)Riviere, Pathak, Sessa, Hardin, Bhupatiraju,
  Hussenot, Mesnard, Shahriari, Ramé, Ferret, Liu, Tafti, Friesen, Casbon,
  Ramos, Kumar, Lan, Jerome, Tsitsulin, Vieillard, Stanczyk, Girgin, Momchev,
  Hoffman, Thakoor, Grill, Neyshabur, Bachem, Walton, Severyn, Parrish, Ahmad,
  Hutchison, Abdagic, Carl, Shen, Brock, Coenen, Laforge, Paterson, Bastian,
  Piot, Wu, Royal, Chen, Kumar, Perry, Welty, Choquette-Choo, Sinopalnikov,
  Weinberger, Vijaykumar, Rogozińska, Herbison, Bandy, Wang, Noland, Moreira,
  Senter, Eltyshev, Visin, Rasskin, Wei, Cameron, Martins, Hashemi,
  Klimczak-Plucińska, Batra, Dhand, Nardini, Mein, Zhou, Svensson, Stanway,
  Chan, Zhou, Carrasqueira, Iljazi, Becker, Fernandez, van Amersfoort, Gordon,
  Lipschultz, Newlan, yeong Ji, Mohamed, Badola, Black, Millican, McDonell,
  Nguyen, Sodhia, Greene, Sjoesund, Usui, Sifre, Heuermann, Lago, McNealus,
  Soares, Kilpatrick, Dixon, Martins, Reid, Singh, Iverson, Görner, Velloso,
  Wirth, Davidow, Miller, Rahtz, Watson, Risdal, Kazemi, Moynihan, Zhang,
  Kahng, Park, Rahman, Khatwani, Dao, Bardoliwalla, Devanathan, Dumai, Chauhan,
  Wahltinez, Botarda, Barnes, Barham, Michel, Jin, Georgiev, Culliton, Kuppala,
  Comanescu, Merhej, Jana, Rokni, Agarwal, Mullins, Saadat, Carthy, Cogan,
  Perrin, Arnold, Krause, Dai, Garg, Sheth, Ronstrom, Chan, Jordan, Yu, Eccles,
  Hennigan, Kocisky, Doshi, Jain, Yadav, Meshram, Dharmadhikari, Barkley, Wei,
  Ye, Han, Kwon, Xu, Shen, Gong, Wei, Cotruta, Kirk, Rao, Giang, Peran,
  Warkentin, Collins, Barral, Ghahramani, Hadsell, Sculley, Banks, Dragan,
  Petrov, Vinyals, Dean, Hassabis, Kavukcuoglu, Farabet, Buchatskaya, Borgeaud,
  Fiedel, Joulin, Kenealy, Dadashi, and
  Andreev]{gemmateam2024gemma2improvingopen}
Morgane Riviere, Shreya Pathak, Pier~Giuseppe Sessa, Cassidy Hardin, Surya
  Bhupatiraju, Léonard Hussenot, Thomas Mesnard, Bobak Shahriari, Alexandre
  Ramé, Johan Ferret, Peter Liu, Pouya Tafti, Abe Friesen, Michelle Casbon,
  Sabela Ramos, Ravin Kumar, Charline~Le Lan, Sammy Jerome, Anton Tsitsulin,
  Nino Vieillard, Piotr Stanczyk, Sertan Girgin, Nikola Momchev, Matt Hoffman,
  Shantanu Thakoor, Jean-Bastien Grill, Behnam Neyshabur, Olivier Bachem,
  Alanna Walton, Aliaksei Severyn, Alicia Parrish, Aliya Ahmad, Allen
  Hutchison, Alvin Abdagic, Amanda Carl, Amy Shen, Andy Brock, Andy Coenen,
  Anthony Laforge, Antonia Paterson, Ben Bastian, Bilal Piot, Bo~Wu, Brandon
  Royal, Charlie Chen, Chintu Kumar, Chris Perry, Chris Welty, Christopher~A.
  Choquette-Choo, Danila Sinopalnikov, David Weinberger, Dimple Vijaykumar,
  Dominika Rogozińska, Dustin Herbison, Elisa Bandy, Emma Wang, Eric Noland,
  Erica Moreira, Evan Senter, Evgenii Eltyshev, Francesco Visin, Gabriel
  Rasskin, Gary Wei, Glenn Cameron, Gus Martins, Hadi Hashemi, Hanna
  Klimczak-Plucińska, Harleen Batra, Harsh Dhand, Ivan Nardini, Jacinda Mein,
  Jack Zhou, James Svensson, Jeff Stanway, Jetha Chan, Jin~Peng Zhou, Joana
  Carrasqueira, Joana Iljazi, Jocelyn Becker, Joe Fernandez, Joost van
  Amersfoort, Josh Gordon, Josh Lipschultz, Josh Newlan, Ju~yeong Ji, Kareem
  Mohamed, Kartikeya Badola, Kat Black, Katie Millican, Keelin McDonell, Kelvin
  Nguyen, Kiranbir Sodhia, Kish Greene, Lars~Lowe Sjoesund, Lauren Usui,
  Laurent Sifre, Lena Heuermann, Leticia Lago, Lilly McNealus, Livio~Baldini
  Soares, Logan Kilpatrick, Lucas Dixon, Luciano Martins, Machel Reid,
  Manvinder Singh, Mark Iverson, Martin Görner, Mat Velloso, Mateo Wirth, Matt
  Davidow, Matt Miller, Matthew Rahtz, Matthew Watson, Meg Risdal, Mehran
  Kazemi, Michael Moynihan, Ming Zhang, Minsuk Kahng, Minwoo Park, Mofi Rahman,
  Mohit Khatwani, Natalie Dao, Nenshad Bardoliwalla, Nesh Devanathan, Neta
  Dumai, Nilay Chauhan, Oscar Wahltinez, Pankil Botarda, Parker Barnes, Paul
  Barham, Paul Michel, Pengchong Jin, Petko Georgiev, Phil Culliton, Pradeep
  Kuppala, Ramona Comanescu, Ramona Merhej, Reena Jana, Reza~Ardeshir Rokni,
  Rishabh Agarwal, Ryan Mullins, Samaneh Saadat, Sara~Mc Carthy, Sarah Cogan,
  Sarah Perrin, Sébastien M.~R. Arnold, Sebastian Krause, Shengyang Dai,
  Shruti Garg, Shruti Sheth, Sue Ronstrom, Susan Chan, Timothy Jordan, Ting Yu,
  Tom Eccles, Tom Hennigan, Tomas Kocisky, Tulsee Doshi, Vihan Jain, Vikas
  Yadav, Vilobh Meshram, Vishal Dharmadhikari, Warren Barkley, Wei Wei, Wenming
  Ye, Woohyun Han, Woosuk Kwon, Xiang Xu, Zhe Shen, Zhitao Gong, Zichuan Wei,
  Victor Cotruta, Phoebe Kirk, Anand Rao, Minh Giang, Ludovic Peran, Tris
  Warkentin, Eli Collins, Joelle Barral, Zoubin Ghahramani, Raia Hadsell,
  D.~Sculley, Jeanine Banks, Anca Dragan, Slav Petrov, Oriol Vinyals, Jeff
  Dean, Demis Hassabis, Koray Kavukcuoglu, Clement Farabet, Elena Buchatskaya,
  Sebastian Borgeaud, Noah Fiedel, Armand Joulin, Kathleen Kenealy, Robert
  Dadashi, and Alek Andreev.
\newblock Gemma 2: Improving open language models at a practical size.
\newblock \emph{arXiv}, 2024.
\newblock URL \url{https://arxiv.org/abs/2408.00118}.

\bibitem[Luettgau et~al.(2025)Luettgau, Coppock, Dubois, Summerfield, and
  Ududec]{luettgau2025hibayeshierarchicalbayesianmodeling}
Lennart Luettgau, Harry Coppock, Magda Dubois, Christopher Summerfield, and
  Cozmin Ududec.
\newblock Hibayes: A hierarchical bayesian modeling framework for ai evaluation
  statistics.
\newblock \emph{arXiv}, 2025.
\newblock URL \url{https://arxiv.org/abs/2505.05602}.

\bibitem[Dubois et~al.(2025)Dubois, Coppock, Giulianelli, Flesch, Luettgau, and
  Ududec]{dubois2025skewedscorestatisticalframework}
Magda Dubois, Harry Coppock, Mario Giulianelli, Timo Flesch, Lennart Luettgau,
  and Cozmin Ududec.
\newblock Skewed score: A statistical framework to assess autograders.
\newblock \emph{arXiv}, 2025.
\newblock URL \url{https://arxiv.org/abs/2507.03772}.

\bibitem[Raffel et~al.(2020)Raffel, Shazeer, Roberts, Lee, Narang, Matena,
  Zhou, Li, and Liu]{raffel2020c4dataset}
Colin Raffel, Noam Shazeer, Adam Roberts, Katherine Lee, Sharan Narang, Michael
  Matena, Yanqi Zhou, Wei Li, and Peter~J Liu.
\newblock Exploring the limits of transfer learning with a unified text-to-text
  transformer.
\newblock \emph{Journal of machine learning research}, 21\penalty0
  (140):\penalty0 1--67, 2020.

\bibitem[Casademunt et~al.(2025)Casademunt, Juang, Karvonen, Marks,
  Rajamanoharan, and
  Nanda]{casademunt2025steeringoutofdistributiongeneralizationconcept}
Helena Casademunt, Caden Juang, Adam Karvonen, Samuel Marks, Senthooran
  Rajamanoharan, and Neel Nanda.
\newblock Steering out-of-distribution generalization with concept ablation
  fine-tuning.
\newblock \emph{arXiv}, 2025.
\newblock URL \url{https://arxiv.org/abs/2507.16795}.

\bibitem[Chen et~al.(2024{\natexlab{b}})Chen, Vondrick, and
  Mao]{chen2024selfie}
Haozhe Chen, Carl Vondrick, and Chengzhi Mao.
\newblock Selfie: Self-interpretation of large language model embeddings.
\newblock \emph{arXiv preprint arXiv:2403.10949}, 2024{\natexlab{b}}.

\bibitem[Pan et~al.(2024)Pan, Chen, and
  Steinhardt]{pan2024latentqateachingllmsdecode}
Alexander Pan, Lijie Chen, and Jacob Steinhardt.
\newblock Latentqa: Teaching llms to decode activations into natural language,
  2024.
\newblock URL \url{https://arxiv.org/abs/2412.08686}.

\bibitem[Wang et~al.(2025{\natexlab{c}})Wang, Griffin, Treutlein, Perez,
  Michael, Roger, and Marks]{wang2025modifying}
Rowan Wang, Avery Griffin, Johannes Treutlein, Ethan Perez, Julian Michael,
  Fabien Roger, and Sam Marks.
\newblock Modifying llm beliefs with synthetic document finetuning.
\newblock
  \url{https://alignment.anthropic.com/2025/modifying-beliefs-via-sdf/},
  2025{\natexlab{c}}.
\newblock URL
  \url{https://alignment.anthropic.com/2025/modifying-beliefs-via-sdf/}.
\newblock Anthropic AI Alignment Research.

\bibitem[Comanici et~al.(2025)Comanici, Bieber, Schaekermann, Pasupat,
  Sachdeva, Dhillon, Blistein, Ram, Zhang, Rosen, Marris, Petulla, Gaffney,
  Aharoni, Lintz, Pais, Jacobsson, Szpektor, Jiang, Haridasan, Omran, Saunshi,
  Bahri, Mishra, Chu, Boyd, Hekman, Parisi, Zhang, Kawintiranon, Bedrax-Weiss,
  Wang, Xu, Purkiss, Mendlovic, Deutel, Nguyen, Langley, Korn, Rossazza, Ramé,
  Waghmare, Miller, Byrd, Sheshan, Hadsell, Bhardwaj, Janus, Rissa, Horgan,
  Abdagic, Belenki, Allingham, Singh, Guidroz, Srinivasan, Schmit, Chiafullo,
  Elisseeff, Jha, Kolhar, Berrada, Ding, Si, Mallick, Och, Erell, Ni, Latkar,
  Yang, Sirkovic, Feng, Leland, Hornung, Wu, Blundell, Alvari, Huang, Yip,
  Deur, Liu, Surita, Duque, Damen, Jia, Guez, Mircea, Sinha, Magni, Stradomski,
  Marian, Galić, Chen, Husain, Singhal, Grewe, Aubet, Song, Blanco, Rechis,
  Ho, Munoz, Zheng, Hamrick, Mather, Taitelbaum, Rutherford, Lei, Chen, Shukla,
  Moreira, Doi, Isik, Shabat, Rogozińska, Kolipaka, Chang, Vušak,
  Venkatachary, Noghabi, Bharti, Jun, Zaks, Green, Challagundla, Wong,
  Mohammad, Hirsch, Cheng, Naim, Proleev, Vincent, Singh, Krikun, Krishnan,
  Ghahramani, Atias, Aggarwal, Kirov, Vytiniotis, Koh, Chronopoulou, Dogra,
  Ion, Tyen, Lee, Weissenberger, Strohman, Balakrishna, Rae, Velic,
  de~Liedekerke, Elyada, Yuan, Liu, Shani, Kishchenko, Alessio, Li, Song, Kwei,
  Jankowski, Pappu, Namiki, Ma, Tripuraneni, Cherry, Ikonomidis, Ling, Ji,
  Westberg, Wright, Yu, Parkinson, Ramaswamy, Connor, Yeganeh, Grover,
  Kenwright, Litchev, Apps, Tomala, Halim, Castro-Ros, Li, Boral, Sho, Yarom,
  Malmi, Klinghoffer, Lin, Ansell, S, Zhao, Zuo, Santoro, Cheng, Demmessie,
  Liu, Brichtova, Culp, Braun, Graur, Ng, Mehta, Phillips, Sundberg, Godbole,
  Liu, Katariya, Rim, Seyedhosseini, Ammirati, Valfridsson, Malihi, Knight,
  Toor, Lampe, Ittycheriah, Chiang, Yeung, Fréchette, Rao, Wang, Srivastava,
  Zhang, Rhodes, Brand, Weesner, Figotin, Gimeno, Fellinger, Marcenac, Leal,
  Marcus, Cotruta, Cabrera, Luo, Garrette, Axelrod, Baltateanu, Barker, Chen,
  Toma, Ingram, Riesa, Kulkarni, Zhang, Liu, Wang, Polacek, Wu, Hui, Reyes, Su,
  Barnes, Malhi, Siddiqui, Feng, Damaschin, Pighin, Steiner, Yang, Boppana,
  Ivanov, Kandoor, Shah, Mujika, Huang, Choquette-Choo, Patel, Yu, Creswell,
  Jerry, Liu, Barros, Razeghi, Roy, Culliton, Xiong, Pan, Strohmann, Powell,
  Seal, DeCarlo, Shyam, Katircioglu, Wang, Hardin, Odisho, Broder, Chang, Nair,
  Shtefan, O'Brien, Agarwal, Potluri, Goyal, Jhindal, Thakur, Stuken, Lyon,
  Toutanova, Feng, Wu, Horn, Wang, Cullum, Taubman, Shrivastava, Shi,
  Tomlinson, Patel, Tu, Oflazer, Pongetti, Yang, Taïga, Perot, Pierse, Han,
  Drori, Iturrate, Chakrabarti, Yeung, Dopson, ting Chen, Kulshreshtha, Guo,
  Pham, Schuster, Chen, Polozov, Xing, Zhou, Kacham, Kukliansky, Miech,
  Yaroshenko, Chi, Douglas, Fei, Blondel, Myla, Madmoni, Wu, Keysers, Kjems,
  Albuquerque, Yu, D'sa, Plantan, Ionescu, Elias, Gupta, Vuyyuru, Alcober,
  Zhou, Ji, Hartmann, Puttagunta, Song, Amid, Stefanoiu, Lee, Pucciarelli,
  Wang, Raul, Petrov, Tian, Anklin, Nti, Gomes, Schumacher, Vesom,
  Panagopoulos, Bousmalis, Andor, Jacob, Zhang, Rosgen, Kecman, Tung, Belias,
  Goodman, Covington, Wieder, Saxena, Davoodi, Huang, Maddineni, Roulet,
  Campbell-Ajala, Sessa, Xintian, Wu, Lai, Collins, Haig, Sakenas, Xu,
  Giustina, Shafey, Charoenpanit, Garg, Ainslie, Severson, Arenas, Pathak,
  Rajayogam, Feng, Bakker, Li, Wichers, Rogers, Geng, Li, Jagerman, Jia,
  Olmert, Sharon, Mauger, Mariserla, Ma, Mohabey, Kim, Andreev, Pollom, Love,
  Jain, Agrawal, Schroecker, Fortin, Warmuth, Liu, Leach, Blok, Girirajan,
  Aharoni, Uria, Sozanschi, Goldberg, Ionita, Ribeiro, Zlocha, Birodkar,
  Lachgar, Yuan, Choudhury, Ginsberg, Zheng, Dibb, Graves, Lokhande, Rasskin,
  Muraru, Quick, Tata, Sermanet, Chawla, Karo, Wang, Zhang, Keller, Dragan, Su,
  Chou, Liu, Tao, Prabhakara, Wilson, Liu, Wang, Evans, Du, Castaño, Prasad,
  Mahdy, Gerlach, Reid, Kahn, Zait, Pillai, Ulrich, Wang, Wassenberg, Farkash,
  Yalasangi, Wang, Bauza, Bucher, Liu, Yan, Leung, Sindhwani, Barnes, Singh,
  Jurin, Chang, Bhumihar, Eiger, Citovsky, Withbroe, Li, Xue, Santo, Stoyanov,
  Raimond, Zheng, Gao, Listík, Kwasiborski, Saputro, Ozturel, Mallya,
  Majmundar, West, Caron, Wei, Castrejon, Vikram, Ramachandran, Dhawan, Park,
  Smoot, van~den Driessche, Blau, Malik, Liang, Hirsch, dos Santos, Weinstein,
  van~den Oord, Lall, FitzGerald, Jiang, Yang, Webster, Elqursh, Pope, Rotival,
  Raposo, Zhu, Dean, Alabed, Tran, Gupta, Gleicher, Austin, Rosseel, Umekar,
  Das, Sun, Chen, Misiunas, Zhou, Di, Loo, Newlan, Li, Ramasesh, Xu, Chen,
  Gandhe, Soricut, Gupta, Hu, El-Sayed, Garcia, Brusilovsky, Chen, Bolt, Huang,
  Gurney, Zhang, Pritzel, Wilkiewicz, Seybold, Shamanna, Fischer, Dean, Gill,
  Mcilroy, Bhowmick, Selier, Yang, Cheng, Magay, Tan, Varma, Walder, Kocisky,
  Nakashima, Natsev, Kwong, Gog, Zhang, Dieleman, Jimma, Ryabtsev, Brahma,
  Steiner, Du, Žužul, Žanić, Raghavachari, Gierke, Zheng, Petrova, Dauphin,
  Liu, Kessler, Hand, Duvarney, Kim, Lee, Hussenot, Hui, Smith, Jain, Xia,
  Tomar, Amiri, Phan, Fuchs, Weyand, Tomasev, Cordell, Liu, Mallinson, Joshi,
  Crawford, Suggala, Chien, Fernando, Sanchez-Vargas, Williams, Crone, Luo,
  Karpov, Shan, Thurk, Strudel, Voigtlaender, Patil, Dozat, Khodaei, Singla,
  Ambroszczyk, Wu, Chang, Roark, Hegde, Ding, Filos, Wu, Pinto, Liu, Khanna,
  Pandey, Mcloughlin, Li, Haves, Zhou, Buchatskaya, Leal, de~Boursac, Akazawa,
  Anderson, Chen, Somandepalli, Liang, Goenka, Winkler, Grushetsky, Ding,
  Smith, Ye, Pont-Tuset, Li, Li, Golany, Wegner, Jiang, Barak, Shangguan,
  Vértes, Wong, Bornschein, Tudor, Bevilacqua, Schaul, Rawat, Zhao, Axiotis,
  Meng, McLean, Lai, Beattie, Kushman, Liu, Kutzman, Lang, Ye, Netrapalli,
  Mishra, Khan, Goel, Willoughby, Tian, Zhuang, Chen, Tsai, Kementsietsidis,
  Khare, Keeling, Xu, Waters, Altché, Popat, Mittal, Saxton, Badawy, Mathieu,
  Zheng, Zhou, Ranka, Shin, Duan, Salimans, Mihailescu, Shaham, Chang, Assael,
  Dikkala, Izzard, Cohen-Addad, Graves, Feinberg, Chung, Strouse, Karmon,
  Sharifzadeh, Ashwood, Pham, Blanton, Vasiloff, Barber, Geller, Zhou, Zubach,
  Huang, Zhang, Gupta, Young, Proskurnia, Votel, Gabeur, Barcik, Tripathi, Yu,
  Yan, Changpinyo, Pavetić, Coyle, Fujii, Mendez, Zhou, Rajamani, Hechtman,
  Cao, Juan, Tan, Dalibard, Du, Clay, Yao, Jia, Vijaykumar, Zhou, Bai, Hung,
  Pecht, Todorov, Khadke, Gupta, Lahoti, Autef, Duddu, Lee-Thorp, Bykovsky,
  Misiunas, Flennerhag, Thangaraj, McGiffin, Nado, Kunesch, Noever, Hertz,
  Liang, Stone, Palmer, Daruki, Pramanik, Põder, Kyker, Khan, Sluzhaev,
  Ritter, Ruderman, Zhou, Nagpal, Vodrahalli, Necula, Barham, Pavlick,
  Hartford, Shafran, Zhao, Mikuła, Eccles, Shimokawa, Garg, Vilnis, Chen,
  Shumailov, Lee, Abdelhamed, Xie, Cohen, Hlavnova, Malkin, Sitawarin, Lottes,
  Coquinot, Yu, Kumar, Zhang, Mahendru, Ahmed, Martens, Chen, Boag, Peng,
  Devin, Klimovskiy, Phuong, Vainstein, Xie, Ramabhadran, Howard, Yu, Goswami,
  Cui, Shleifer, Pinto, Yeh, Yang, Javanmardi, Ethier, Lee, Orbay, Kotecha,
  Bromberg, Shaw, Thornton, Rosenthal, Gu, Thomas, Gemp, Ayyar, Ushio, Selvan,
  Wee, Liu, Majzoubi, Yu, Abernethy, Liechty, Pan, Nguyen, Qiong, Hu, Perrin,
  Arora, Pitler, Wang, Shivakumar, Prost, Limonchik, Wang, Gao, Cour, Buch,
  Gui, Ivanova, Neubeck, Chan, Kim, Chen, Goyal, Chung, Liu, Su, Petrushkina,
  Shen, Joulin, Xu, Lin, Kulizhskaya, Chelba, Vasudevan, Collins, Bashlovkina,
  Lu, Fritz, Park, Zhou, Su, Tanburn, Sushkov, Rasquinha, Li, Prendki, Li, LV,
  Sharma, Fitoussi, Huang, Dai, Dao, Burrows, Prior, Qin, Pundak, Sjoesund,
  Khurshudov, Zhu, Webson, Kemp, Tan, Agrawal, Sargsyan, Cheng, Stephan,
  Kwiatkowski, Reid, Byravan, Michaely, Heess, Zhou, Goenka, Carpenter,
  Levskaya, Wang, Roberts, Leblond, Chikkerur, Ginzburg, Chang, Riachi,
  Chuqiao, Xu, Borsos, Pliskin, Pawar, Lustman, Kirkwood, Anand, Chaudhary,
  Kalb, Milan, Augenstein, Goldie, Prince, Raman, Sun, Xia, Cohen, Huo, Camp,
  Ellis, Zilka, Torres, Patel, Arora, Chan, Adler, Ayoub, Liang, Jamil, Jiang,
  Baumgartner, Sun, Karov, Akulov, Zheng, Cai, Fantacci, Rubin, Acha, Wang,
  D'Souza, Sathyanarayana, Dai, Rowe, Simanovsky, Goldman, Kuang, Pan,
  Rosenberg, Rojas-Esponda, Dutta, Zeng, Jurenka, Farquhar, Bansal, Iqbal,
  Roelofs, Joung, Beak, Ryu, Poplin, Wu, Alayrac, Buthpitiya, Ronneberger,
  Habtegebriel, Li, Cavallaro, Wei, Bensky, Denk, Ganapathy, Stanway, Joshi,
  Bertolini, Lo, Ma, Charles, Sampemane, Sahni, Chen, Askham, Gaddy, Young,
  Tan, Eyal, Bražinskas, Zhong, Wu, Epstein, Bailey, Hard, Lee, Goldshtein,
  Ruiz, Badawi, Lochbrunner, Kearns, Brown, Pardo, Weber, Yang, Jiang, Akin,
  Fu, Wainwright, Zou, Gaba, Manzagol, Kan, Song, Zainullina, Lin, Ko,
  Deshmukh, Jindal, Svensson, Tyam, Zhao, Kaeser-Chen, Baird, Moradi, Hall,
  Guo, Tsang, Liang, Pereira, Ganesh, Korotkov, Adamek, Thiagarajan, Tran,
  Chen, Tar, Jain, Dasgupta, Bilal, Reitter, Zhao, Vezzani, Gehman, Mehta,
  Beltrone, Dotiwalla, Guadarrama, Abbas, Karp, Georgiev, Ferng, Brockschmidt,
  Peng, Hirnschall, Verma, Bi, Xiao, Dabush, Xu, Wallis, Parker, Wang, Xu,
  Safarli, Tewari, Zhang, Kim, Gesmundo, Thomas, Levi, Chowdhury, Rao, Garst,
  Conway-Rahman, Ran, McKinney, Xiao, Yu, Agrawal, Stjerngren, Ionescu, Chen,
  Sharma, Chiu, Liu, Franko, Sanford, Cai, Michel, Ganapathy, Labanowski,
  Garrett, Vargas, Sun, Gale, Buschmann, Desjardins, Ghelani, Jain, Verma,
  Asawaroengchai, Eisenschlos, Harlalka, Kazawa, Metzler, Howland, Jian, Ades,
  Shah, Gangwani, Lee, Ring, Hernandez, Reich, Sinha, Sathe, Kovac, Gill,
  Kannan, D'olimpio, Sevenich, Whang, Kim, Sim, Chen, Zhang, Lall, Matias, Jia,
  Friesen, Nasso, Thapliyal, Perozzi, Yu, Shekhawat, Huda, Grabowski, Wang,
  Sreevatsa, Dib, Hassen, Schuh, Milutinovic, Welty, Quinn, Shah, Wang,
  Barth-Maron, Frye, Axelsson, Zhu, Ma, Giannoumis, Sedghi, Ye, Luan, Aydin,
  Chandra, Sampathkumar, Huang, Lavrenko, Eleryan, Hong, Hansen, Carthy,
  Samanta, Ćevid, Wang, Li, Voznesensky, Hoffman, Terzis, Sehwag, Fidel, He,
  Cai, He, Feng, Nikoltchev, Phatale, Chase, Lawton, Zhang, Ouyang, Tragut,
  Manshadi, Narayanan, Shen, Gao, Bolukbasi, Roy, Li, Golovin, Panait, Qin,
  Han, Anthony, Kudugunta, Patraucean, Ray, Chen, Yang, Bhatia, Talluri,
  Morris, Ražnatović, Brownfield, An, Peng, Kane, Zheng, Duduta, Kessinger,
  Noraky, Liu, Rong, Veličković, Rush, Goldin, Wei, Garlapati, Pantofaru,
  Kwon, Ni, Noland, Trapani, Beaufays, Roy, Chow, Turker, Cideron, Mei, Clark,
  Dou, Bošnjak, Leith, Du, Yazdanbakhsh, Nasr, Kwak, Sheth, Kaskasoli, Anand,
  Lakshminarayanan, Jerome, Bieber, Chu, Senges, Shen, Sridhar, Ndebele,
  Beyret, Mohamed, Chen, Freitag, Guo, Liu, Roit, Chen, Yan, Stone, Co-Reyes,
  Cole, Scellato, Azizi, Hashemi, Jin, Iyer, Valentine, György, Ahuja, Diaz,
  Lee, Clement, Kong, Garmon, Watts, Bhatia, Gupta, Miecnikowski, Vallet, Taly,
  Loper, Joshi, Atwood, Chick, Collier, Iliopoulos, Trostle, Gunel,
  Leal-Cavazos, Hrafnkelsson, Guzman, Ju, Forbes, Emond, Chauhan, Caine, Xiao,
  Zeng, Moufarek, Murphy, Meng, Gupta, Riedel, Das, Lawal, Narayan, Sosea,
  Swirhun, Friso, Neyshabur, Lu, Girgin, Wunder, Yvinec, Pyne, Carbune,
  Rijhwani, Guo, Doshi, Briukhov, Bain, Hitron, Wang, Gupta, Chen, Du, Zhang,
  Shah, Akula, Dylla, Kachra, Kuo, Zou, Wang, Xu, Zhu, Snyder, Menon, Firat,
  Mordatch, Yuan, Ponomareva, Blevins, Moore, Wang, Chen, Scholz, Dwornik, Lin,
  Li, Antognini, I, Song, Miller, Kalra, Raveret, Akerlund, Wu, Nystrom,
  Godbole, Liu, DeBalsi, Zhao, Liu, Caciularu, Lax, Khandelwal, Langston,
  Bailey, Lattanzi, Wang, Kovelamudi, Mondal, Guruganesh, Hua, Roval,
  Wesołowski, Ingale, Halcrow, Sohn, Angermueller, Raad, Stickgold, Lu, Kosik,
  Xie, Lillicrap, Huang, Zhang, Paulus, Farabet, Wertheim, Wang, Joshi, ling
  Ko, Wu, Agrawal, Lin, Sheng, Sung, Breland-King, Butterfield, Gawde, Singh,
  Zhang, Apte, Shetty, Hutter, Li, Salesky, Lebron, Kanerva, Paganini, Nguyen,
  Vallu, Peter, Velury, Kao, Hoover, Bortsova, Bishop, Jakobovits, Agostini,
  Agarwal, Liu, Kwong, Tavakkol, Bica, Greve, GP, Marcus, Hou, Duerig,
  Moroshko, Lacey, Davis, Amelot, Wang, Kim, Strinopoulos, Wan, Lan, Krishnan,
  Tang, Humphreys, Bai, Shtacher, Machado, Pang, Burke, Liu, Aravamudhan, Song,
  Hirst, Singh, Jou, Bai, Piccinno, Fu, Alazard, Meiri, Winter, Chen, Zhang,
  Heitkaemper, Lambert, Lee, Frömmgen, Rogulenko, Nair, Niemczyk, Bulyenov,
  Xu, Shemtov, Zadimoghaddam, Toropov, Wirth, Dai, Gollapudi, Zheng, Kurakin,
  Lee, Bullard, Serrano, Balazevic, Li, Schalkwyk, Murphy, Zhang, Sequeira,
  Datta, Agrawal, Sutton, Attaluri, Chiang, Farhan, Thornton, Lin, Choma,
  Nguyen, Dasgupta, Robinson, Comşa, Riley, Pillai, Mustafa, Golan, Zandieh,
  Lespiau, Porter, Ross, Rajayogam, Agarwal, Venugopalan, Shahriari, Yan, Xu,
  Tobin, Dubov, Shi, Recasens, Kovsharov, Borgeaud, Dery, Vasanth, Gribovskaya,
  Qiu, Mahdieh, Skut, Nielsen, Zheng, Yu, Bostock, Gupta, Archer, Rawles,
  Davies, Svyatkovskiy, Tsai, Halpern, Reisswig, Wydrowski, Chang, Puigcerver,
  Taege, Li, Schnider, Li, Dena, Xu, Telang, Shi, Zen, Kastner, Ko,
  Subramaniam, Kumar, Blois, Dai, Wieting, Lu, Zeldes, Xie, Hauth, Ţifrea, Li,
  El-Husseini, Abolafia, Zhou, Ding, Ghalebikesabi, Guía, Maksai, Ágoston
  Weisz, Arik, Sukhanov, Świetlik, Jia, Yu, Wang, Brand, Bloxwich, Kirmani,
  Chen, Go, Sprechmann, Kannen, Carin, Sandhu, Edkins, Nooteboom, Gupta,
  Maggiore, Azizi, Pritch, Yin, Gupta, Tarlow, Smith, Ivanov, Babaeizadeh,
  Goel, Kambala, Chu, Kastelic, Liu, Soltau, Stone, Agrawal, Kim, Soparkar,
  Tadepalli, Bunyan, Soh, Kannan, Kim, Chen, Halumi, Roy, Wang, Sercinoglu,
  Gibson, Bhatnagar, Sano, von Dincklage, Ren, Mitrevski, Olšák, She,
  Doersch, Jilei, Wang, Liu, Tan, Yakar, Warkentin, Ramirez, Lebsack, Dillon,
  Mathews, Cobley, Wu, Chen, Simon, Nath, Sainath, Bendebury, Julian,
  Mankalale, Ćurko, Zacchello, Brown, Sodhia, Howard, Caelles, Gupta, Evans,
  Bulanova, Katzen, Goldenberg, Tsitsulin, Stanton, Schillings, Kovalev, Fry,
  Shah, Lin, Upadhyay, Li, Radpour, Maggioni, Xiong, Haas, Brennan, Kamath,
  Savinov, Nagrani, Yacovone, Kappedal, Andriopoulos, Lao, Li, Rozhdestvenskiy,
  Hashimoto, Audibert, Austin, Rodriguez, Ruoss, Honke, Karkhanis, Xiong, Wei,
  Huang, Leng, Premachandran, Bileschi, Evangelopoulos, Mensink, Pavagadhi,
  Teplyashin, Chang, Xue, Tanzer, Goldman, Patel, Li, Wiesner, Zheng,
  Stewart-Binks, Han, Li, Luo, Lenc, Lučić, Xue, Mullins, Guseynov, Chang,
  Galatzer-Levy, Zhang, Bingham, Hu, Hartman, Ma, Griffith, Irpan, Radebaugh,
  Yue, Fan, Ungureanu, Sorokin, Teufel, Li, Anil, Paparas, Wang, Lin, Peng,
  Shum, Petrovic, Brady, Nguyen, Macherey, Li, Singh, Yenugula, Iinuma, Chen,
  Kopparapu, Stern, Dave, Thekkath, Perot, Kumar, Li, Xiao, Bilotti, Bateni,
  Noble, Lee, Vázquez-Reina, Salazar, Yang, Wang, Gruzewska, Rao, Raghuram,
  Xu, Ben-David, Mei, Dalmia, Zhang, Liu, Bansal, Pankov, Schwarcz, Burns,
  Chan, Sanghai, Liang, Liang, He, Stuart, Narayanan, Zhu, Frank, Fatemi,
  Sabne, Lang, Bhattacharya, Settle, Wang, McMahan, Tacchetti, Soares, Hadian,
  Cabi, Chung, Putikhin, Li, Chen, Tarango, Michalewski, Kazemi, Masoom,
  Sheftel, Shivanna, Vadali, Comanescu, Reid, Moore, Neelakantan, Sander,
  Herzig, Rosenberg, Dehghani, Choi, Fink, Hayes, Ge, Weng, Ho, Karro, Krishna,
  Thiet, Skerry-Ryan, Eppens, Andreetto, Sarma, Bonacina, Ayan, Nawhal, Shan,
  Dusenberry, Thakoor, Gubbi, Nguyen, Tsarfaty, Albanie, Mitrović, Gandhi,
  Chen, Epasto, Stephanov, Jin, Gehman, Amini, Weber, Behbahani, Xu, Allamanis,
  Chen, Ott, Sha, Jastrzebski, Qi, Greene, Wu, Toki, Vlasic, Shapiro,
  Kotikalapudi, Shen, Saeki, Xie, Cassirer, Bharadwaj, Kiyono, Bhojanapalli,
  Rosenfeld, Ritter, Mao, Oliveira, Egyed, Bandemer, Parisotto, Kinoshita,
  Pluto, Maniatis, Li, Guo, Ghiasi, Tarbouriech, Chatterjee, Jin, Katrina, Xu,
  Palomaki, Arnold, Sewak, Piccinini, Sharma, Albrecht, Purser-haskell,
  Vaswani, Chen, Wisniewski, Cao, Aslanides, Phu, Sieb, Agubuzu, Zheng, Sohn,
  Selvi, Andreassen, Subudhi, Eruvbetine, Woodman, Mery, Krause, Ren, Ma, Luo,
  Chen, Fan, Griffiths, Schuler, Li, Zhang, Sarr, Luo, Patana, Watson,
  Naboulsi, Collins, Sidhwani, Hoogeboom, Silver, Caveness, Zhao, Rodriguez,
  Deines, Bai, Griffin, Tagliasacchi, Xue, Babbula, Pang, Ding, Shen, Peake,
  Crocker, Raghvendra, Swisher, Han, Singh, Wu, Pchelin, Munkhdalai, Alon,
  Bacon, Robles, Bulian, Johnson, Powell, Ferreira, Li, Benzing, Velimirović,
  Soyer, Kong, Tony, Nguyên, Yang, Liu, van Amersfoort, Gillick, Sun,
  Rauschmayr, Zhang, Zhan, Zhou, Frolov, Yang, Vnukov, Rouillard, Li, Mandhane,
  Fallen, Venkataraman, Hu, Brennan, Lee, Chang, Sundermeyer, Pan, Ke, Tong,
  Fabrikant, Bono, Gu, Foley, Mao, Delakis, Bhaswar, Frostig, Li, Zipori, Hope,
  Kozlova, Mishra, Djolonga, Schiff, Merey, Briakou, Morgan, Wan, Hassidim,
  Skerry-Ryan, Sengupta, Jasarevic, Kallakuri, Kunkle, Brennan, Lieber,
  Mansoor, Walker, Zhang, Xie, Žužić, Chukwuka, Druinsky, Cho, Yao, Naeem,
  Butt, Kim, Jia, Jordan, Lelkes, Kurzeja, Wang, Zhao, Over, Chakladar,
  Prasetya, Jha, Ganapathy, Cong, Shroff, Saroufim, Miryoosefi, Hammad, Nasir,
  Xi, Gao, Maeng, Hora, Cheng, Haghani, Lewenberg, Lu, Matysiak, Raisinghani,
  Wang, Baugher, Sukthankar, Giang, Schultz, Fiedel, Chen, Lee, Dey, Zheng,
  Paul, Smith, Ly, Wang, Bansal, Perz, Ricco, Blank, Keshava, Sharma, Chow,
  Lad, Jalan, Osindero, Swanson, Scott, Ilić, Li, Jonnalagadda, Soudagar,
  Xiong, Batsaikhan, Jarrett, Kumar, Shah, Lawlor, Waters, Graham, May, Ramos,
  Lefdal, Cankara, Cano, O'Donoghue, Borovik, Liu, Grimstad, Alnahlawi,
  Tsihlas, Hudson, Grigorev, Jia, Huang, Igwe, Lebedev, Tang, Krivokon, Garcia,
  Tan, Jia, Stys, Vashishth, Liang, Venkatraman, Gu, Kementsietsidis, Zhu,
  Jung, Bai, Hosseini, Ahmed, Gupta, Yuan, Ashraf, Nigam, Vasudevan, Awasthi,
  Gilady, Mariet, Eskander, Li, Hu, Garrido, Schlattner, Zhang, Saxena, Dević,
  Muralidharan, Murthy, Zhou, Choi, Wongpanich, Wang, Shah, Xu, Huang, Spencer,
  Chen, Cohan, Wang, Tompson, Wu, Haroun, Li, Huergo, Yang, Yin, Wendt,
  Bendersky, Chaabouni, Snaider, Ferret, Jindal, Thompson, Xue, Bishop, Phal,
  Sharma, Sung, Radhakrishnan, Shomrat, Ingle, Vij, Gilmer, Istin, Sobell, Lu,
  Nottage, Sadigh, Willcock, Zhang, Xu, Brown, Lee, Wang, Zhu, Tay, Kim,
  Gutierrez, Sharma, Xian, Seo, Cui, Pochernina, Baetu, Jastrzębski, Ly,
  Elhawaty, Suh, Sezener, Wang, Yuen, Tucker, Cai, Yang, Wang, Muzio, Qian,
  Yoo, Lockhart, McKee, Guo, Mehrotra, Mendonça, Mehta, Ben, Tekur, Mu, Zhu,
  Krakovna, Lee, Maschinot, Cevey, Choe, Bai, Srinivasan, Gasaway, Young,
  Siegler, Holtmann-Rice, Piratla, Baumli, Yogev, Hofer, van Hasselt, Grant,
  Chervonyi, Silver, Hogue, Agarwal, Wang, Singh, Flynn, Lipschultz, David,
  Bellot, Yang, Le, Graziano, Olszewska, Hui, Maurya, Parotsidis, Chen,
  Oguntebi, Kelley, Baddepudi, Mauerer, Shaw, Siegman, Yang, Shetty, Roy, Song,
  Stokowiec, Burnell, Savant, Busa-Fekete, Miao, Ghosh, MacDermed, Lippe,
  Dektiarev, Behrman, Mentzer, Nguyen, Wei, Verma, Knutsen, Dasari, Yan,
  Mitrichev, Wang, Shejwalkar, Austin, Sunkara, Potti, Virin, Wright, Liu,
  Riva, Pot, Kochanski, Le, Balasubramaniam, Dhar, Liao, Bloniarz, Shukla,
  Cole, Lee, Zhang, Kafle, Vashishtha, Mahmoudieh, Chen, Hoffmann, Srinivasan,
  Lago, Shalom, Wang, Elabd, Sharma, Oh, Kothawade, Le, Monteiro, Yang,
  Alarakyia, Geirhos, Mincu, Garnes, Kobayashi, Mariooryad, Krasowiak, Zhixin,
  Lai, Mourad, Wang, Bu, Aharoni, Chen, Goyal, Zubov, Bapna, Dabir, Kothari,
  Lamerigts, Cao, Shar, Yew, Kulkarni, Mahaarachchi, Joshi, Zhu, Lichtarge,
  Zhou, Muckenhirn, Selo, Vinyals, Chen, Brohan, Mehta, Cogan, Wang, Geri, Ko,
  Chen, Viola, Shivam, Wang, Elish, Popa, Pereira, Liu, Koster, Kim, Zhang,
  Ebrahimi, Talukdar, Zheng, Poklukar, Mikhalap, Johnson, Vijayakumar,
  Omernick, Dibb, Dubey, Hu, Suman, Aggarwal, Kornakov, Xia, Lowe, Kolganov,
  Xiao, Nikolaev, Hemingray, Li, Iljazi, Rybiński, Sandhu, Lu, Luong,
  Jenatton, Govindaraj, Hui, Li, Dulac-Arnold, Park, Wang, Modi, Pouget-Abadie,
  Greller, Gupta, Berry, Ramachandran, Xie, McCafferty, Wang, Gupta, Lim,
  Bratanič, Brock, Akolzin, Sproch, Karliner, Kim, Goedeckemeyer, Shazeer,
  Schmid, Calandriello, Bhatia, Choromanski, Montgomery, Dua, Ramalho, King,
  Gao, Nguyen, Lindner, Pitta, Johnson, Salama, Ardila, Han, Farnese, Odoom,
  Wang, Ding, Rink, Smith, Lehri, Cohen, Vats, He, Gopavarapu, Paszke, Patel,
  Gansbeke, Loher, Castro, Voitovich, von Glehn, George, Niklaus, Eaton-Rosen,
  Rakićević, Jue, Perel, Zhang, Bahat, Pouget, Xing, Huot, Shenoy, Bos,
  Coriou, Richter, Noy, Wang, Ontanon, Qin, Makarchuk, Hassabis, Li, Sharma,
  Venkatesan, Kemaev, Daniel, Huang, Shah, Ponce, Warren, Chen, Faruqui, Wu,
  Andačić, Payrits, McDuff, Hume, Cao, Tessler, Wang, Wang, Rendulic,
  Agustsson, Johnson, Lando, Howard, Padmanabhan, Daswani, Banino, Kilgore,
  Heek, Ji, Caceres, Li, Kassner, Vlaskin, Liu, Grills, Hou, Sukkerd, Cheon,
  Shetty, Markeeva, Stanczyk, Iyer, Gong, Gao, Gopalakrishnan, Blyth, Reynolds,
  Bhoopchand, Bilenko, Gharibian, Zayats, Faust, Singh, Ma, Jiao,
  Vijayanarasimhan, Aroyo, Yadav, Chakera, Kakarla, Meshram, Gregor, Botea,
  Senter, Jia, Kovacs, Sharma, Baur, Kang, He, Zhuo, Kostelac, Laish, Peng,
  O'Bryan, Kasenberg, Rao, Leurent, Zhang, Stevens, Salazar, Zhang, Lobov,
  Walker, Porter, Redshaw, Ke, Rao, Lee, Lam, Moffitt, Kim, Qiao, Koo, Dadashi,
  Song, Sundararajan, Xu, Kawamoto, Zhong, Barbu, Reddy, Verzetti, Li,
  Papamakarios, Klimczak-Plucińska, Cassin, Kavukcuoglu, Swavely, Vaucher,
  Zhao, Hemsley, Tschannen, Ge, Menghani, Yu, Ha, He, Wu, Song, Sterneck,
  Zinke, Calian, Marsden, Ruiz, Hessel, Gueta, Lee, Farris, Gupta, Li, Saleh,
  Misra, Xiao, Mendolicchio, Buttimore, Krayvanova, Nayakanti, Wiethoff, Pande,
  Mirhoseini, Lao, Liu, Hua, Chen, Malkov, Kalashnikov, Gupta, Audhkhasi, Zhai,
  Kopalle, Jain, Ofek, Meyer, Baatarsukh, Strejček, Qian, Freedman, Figueira,
  Sokolik, Bachem, Lin, Kharrat, Hidey, Xu, Duan, Li, Ersoy, Everett, Cen,
  Santamaria-Fernandez, Taubenfeld, Mackinnon, Deng, Zablotskaia, Viswanadha,
  Goel, Yates, Deng, Choy, Chen, Sinha, Mossin, Wang, Szlam, Hao, Rubenstein,
  Toksoz-Exley, Aperghis, Zhong, Ahn, Isard, Lacombe, Luisier, Anastasiou,
  Kalley, Prabhu, Dunleavy, Bijwadia, Mao-Jones, Chen, Pasumarthi, Wood,
  Dostmohamed, Hurley, Simsa, Parrish, Pajarskas, Harvey, Skopek, Kochinski,
  Rey, Rieser, Zhou, Lee, Acharya, Li, Jiang, Zhang, Gipson, Mahintorabi,
  Gelmi, Khajehnouri, Yeh, Lee, Matthey, Baker, Pham, Fu, Pak, Gupta,
  Vasconcelos, Sadovsky, Walker, Hsiao, Zochbauer, Marzoca, Velan, Zeng,
  Baechler, Driess, Jain, Huang, Tao, Maggs, Levine, Schneider, Gemzer, Petit,
  Han, Fisher, Zelle, Biles, Ie, Fadeeva, Liu, Franco, Collister, Zhang, Wang,
  Zhao, Kieliger, Shuster, Zhu, Gong, Chan, Sun, Basu, Zimmermann, Hayes,
  Bapna, Snoek, Yang, Datta, Abdallah, Kilgour, Li, Mah, Jun, Rivière,
  Karmarkar, Spalink, Huang, Gonzalez, Tran, Nowak, Palowitch, Chadwick,
  Talius, Mehta, Sellam, Fränken, Nicosia, He, Kini, Amos, Basu, Jobe, Shaw,
  Xu, Evans, Ikeda, Yan, Jin, Wang, Yadav, Labzovsky, Sampath, Ma, Schumann,
  Siddhant, Shah, Youssef, Agarwal, Dabney, Tonioni, Ambar, Li, Guyon, Li,
  Soergel, Fang, Karadzhov, Udrescu, Trinh, Raunak, Noury, Guo, Gupta,
  Finkelstein, Petek, Liang, Billock, Sun, Wood, Song, Yu, Matejovicova, Cohen,
  Andra, D'Ambrosio, Deng, Nallatamby, Songhori, Dangovski, Lampinen, Botadra,
  Hillier, Cao, Baddi, Kuncoro, Yoshino, Bhagatwala, Ranzato, Schaeffer, Liu,
  Ye, Sarvana, Nham, Kuang, Gao, Baek, Mittal, Wahid, Gergely, Ni, Feldman,
  Muir, Lamblin, Macherey, Dyer, Kilpatrick, Campos, Bhutani, Fort, Ahmad,
  Severyn, Chatziprimou, Ferludin, Dimarco, Kusupati, Heyward, Bahir, Villela,
  Millican, Marcus, Bahargam, Unlu, Roth, Wei, Gopal, Ghoshal, Lee, Lin, Lees,
  Lee, Hosseini, Fan, Neel, Wu, Altun, Cai, Piqueras, Woodward, Bissacco,
  Haykal, Bordbar, Sundaram, Hodkinson, Toyama, Polovets, Myers, Sinha,
  Levinboim, Krishnakumar, Chhaparia, Sholokhova, Gundavarapu, Jawahar,
  Qureshi, Hu, Momchev, Rahtz, Wu, S, Dhamdhere, Guo, Gupta, Eslami, Schain,
  Blokzijl, Welling, Orr, Bolelli, Perez-Nieves, Sirotenko, Prasad, Kar, Pigem,
  Terzi, Weisz, Ghosh, Mavalankar, Madeka, Daugaard, Adam, Shah, Berman, Tran,
  Baker, Andrejczuk, Chole, Raboshchuk, Mirzazadeh, Kagohara, Wu, Schallhart,
  Orlando, Wang, Rrustemi, Xiong, Liu, Vezer, Ramsden, yiin Chang, Mudgal, Li,
  Vieillard, Hoshen, Ahmad, Slone, Hua, Potikha, Rossini, Stritar, Prakash,
  Wang, Dong, Nazari, Nehoran, Tekelioglu, Li, Badola, Funkhouser, Li, Yerram,
  Ganeshan, Formoso, Langner, Shi, Li, Yamamori, Panda, Saade, Scarpati,
  Breaux, Carey, Zhou, Hsieh, Bridgers, Butryna, Gupta, Tulsyan, Woo, Eltyshev,
  Grathwohl, Parks, Benjamin, Panigrahy, Dodhia, Freitas, Sauer, Song, Alet,
  Tolins, Paduraru, Zhou, Albert, Zhang, Shu, Bansal, Nguyen, Globerson, Xiao,
  Manyika, Hennigan, Rong, Matak, Bakalov, Sharma, Sinopalnikov, Pierson,
  Roller, Brown, Gao, Fukuzawa, Ghafouri, Vassigh, Barr, Wang, Korsun, Jayaram,
  Ren, Zaman, Khan, Lunts, Deutsch, Uthus, Katz, Samsikova, Khalifa, Sethi,
  Sun, Tang, Alon, Luo, Yu, Nayyar, Petrini, Truong, Hellendoorn, Chinaev,
  Alberti, Wang, Hu, Mirrokni, Balashankar, Aharon, Mehta, Iscen, Kready,
  Manning, Mohananey, Chen, Tripathi, Wu, Petrovski, Hwang, Baeuml,
  Chandrakaladharan, Liu, Coaguila, Chen, Ma, Tafti, Tatineni, Spitz, Ye,
  Vicol, Rosca, Puigdomènech, Yahav, Ghemawat, Lin, Kirk, Nabulsi, Brin,
  Bohnet, Caluwaerts, Veerubhotla, Zheng, Dai, Petrov, Xu, Mehran, Xu,
  Zintgraf, Choi, Hombaiah, Thoppilan, Reddi, Lew, Li, Webster, Sawhney,
  Lamprou, Shakeri, Lunayach, Chen, Bagri, Salcianu, Chen, Donchev, Magister,
  Nørly, Rodrigues, Izo, Noga, Zou, Köppe, Zhou, Lee, Long, Eisenbud, Chen,
  Schenck, To, Zhong, Taropa, Truong, Levy, Martins, Zhang, Semturs, Zhang,
  Yakubovich, Moreno, McConnaughey, Lu, Redmond, Weerts, Bitton, Refice,
  Lacasse, Conmy, Tallec, Odell, Forbes-Pollard, Socala, Hoech, Kohli, Walton,
  Wang, Sazanovich, Zhu, Kapishnikov, Galt, Denton, Murdoch, Sikora, Mohamed,
  Wei, First, McConnell, Cobo, Qin, Avrahami, Balle, Watanabe, Louis, Kraft,
  Ariafar, Gu, Rives, Yoon, Rusu, Cobon-Kerr, Hahn, Luo, Yuvein, Zhu, Ahuja,
  Benenson, Kaufman, Yu, Hightower, Zhang, Ni, Hendricks, Wang, Yona, Jain,
  Barrio, Bhupatiraju, Velusamy, Dafoe, Riedel, Thomas, Yuan, Bellaiche,
  Panthaplackel, Kloboves, Jauhari, Akbulut, Davchev, Gladchenko, Madras,
  Chuklin, Hill, Yuan, Madhavan, Leonhard, Scandinaro, Chen, Niu, Douillard,
  Damoc, Onoe, Pedregosa, Bertsch, Leichner, Pagadora, Malmaud, Ponda, Twigg,
  Duzhyi, Shen, Wang, Garg, Chen, Evci, Lee, Liu, Kojima, Yamaguchi, Rajendran,
  Piergiovanni, Rajendran, Fornoni, Ibagon, Ragan, Khan, Blitzer, Bunner, Sun,
  Kosakai, Lundberg, Elue, Guu, Park, Park, Narayanaswamy, Wu, Mudigonda, Cohn,
  Mu, Kumar, Graesser, Zhang, Killam, Zhuang, Giménez, Jishi, Ley-Wild, Zhai,
  Osawa, Cedillo, Liu, Upadhyay, Sieniek, Sharma, Paine, Angelova, Addepalli,
  Parada, Majumder, Lamp, Kumar, Deng, Myaskovsky, Sabolić, Dudek, York,
  de~Chaumont~Quitry, Nie, Cattle, Gunjan, Piot, Khawaja, Bang, Wang,
  Khodadadeh, R, Rawlani, Powell, Lee, Griesser, Oh, Magalhaes, Li, Tokumine,
  Vogel, Hsu, BC, Jindal, Cohen, Yang, Yuan, de~Cesare, Bruguier, Xu, Roy,
  Jacovi, Belov, Arya, Meadowlark, Cohen-Ganor, Ye, Morris-Suzuki, Banzal,
  Song, Ponnuramu, Zhang, Scrivener, Zaiem, Rochman, Han, Ghazi, Lee, Drath,
  Suo, Girgis, Shenoy, Nguyen, Eck, Gupta, Yan, Carreira, Gulati, Sang,
  Mirylenka, Cooney, Chou, Ling, Fan, Coleman, Tubone, Kumar, Baldridge,
  Hernandez-Campos, Lazaridou, Besley, Yona, Bulut, Wellens, Pierigiovanni,
  George, Green, Han, Tao, Clark, You, Abdolmaleki, Fu, Chen, Chaugule,
  Chandorkar, Rahman, Thompson, Koanantakool, Bernico, Ren, Vlasov,
  Vassilvitskii, Kula, Liang, Kim, Huang, Ye, Lepikhin, and
  Helmholz]{comanici2025gemini25pushingfrontier}
Gheorghe Comanici, Eric Bieber, Mike Schaekermann, Ice Pasupat, Noveen
  Sachdeva, Inderjit Dhillon, Marcel Blistein, Ori Ram, Dan Zhang, Evan Rosen,
  Luke Marris, Sam Petulla, Colin Gaffney, Asaf Aharoni, Nathan Lintz,
  Tiago~Cardal Pais, Henrik Jacobsson, Idan Szpektor, Nan-Jiang Jiang, Krishna
  Haridasan, Ahmed Omran, Nikunj Saunshi, Dara Bahri, Gaurav Mishra, Eric Chu,
  Toby Boyd, Brad Hekman, Aaron Parisi, Chaoyi Zhang, Kornraphop Kawintiranon,
  Tania Bedrax-Weiss, Oliver Wang, Ya~Xu, Ollie Purkiss, Uri Mendlovic, Ilaï
  Deutel, Nam Nguyen, Adam Langley, Flip Korn, Lucia Rossazza, Alexandre Ramé,
  Sagar Waghmare, Helen Miller, Nathan Byrd, Ashrith Sheshan, Raia Hadsell,
  Sangnie Bhardwaj, Pawel Janus, Tero Rissa, Dan Horgan, Alvin Abdagic, Lior
  Belenki, James Allingham, Anima Singh, Theo Guidroz, Srivatsan Srinivasan,
  Herman Schmit, Kristen Chiafullo, Andre Elisseeff, Nilpa Jha, Prateek Kolhar,
  Leonard Berrada, Frank Ding, Xiance Si, Shrestha~Basu Mallick, Franz Och,
  Sofia Erell, Eric Ni, Tejasi Latkar, Sherry Yang, Petar Sirkovic, Ziqiang
  Feng, Robert Leland, Rachel Hornung, Gang Wu, Charles Blundell, Hamidreza
  Alvari, Po-Sen Huang, Cathy Yip, Sanja Deur, Li~Liu, Gabriela Surita, Pablo
  Duque, Dima Damen, Johnson Jia, Arthur Guez, Markus Mircea, Animesh Sinha,
  Alberto Magni, Paweł Stradomski, Tal Marian, Vlado Galić, Wenhu Chen,
  Hisham Husain, Achintya Singhal, Dominik Grewe, François-Xavier Aubet,
  Shuang Song, Lorenzo Blanco, Leland Rechis, Lewis Ho, Rich Munoz, Kelvin
  Zheng, Jessica Hamrick, Kevin Mather, Hagai Taitelbaum, Eliza Rutherford, Yun
  Lei, Kuangyuan Chen, Anand Shukla, Erica Moreira, Eric Doi, Berivan Isik, Nir
  Shabat, Dominika Rogozińska, Kashyap Kolipaka, Jason Chang, Eugen Vušak,
  Srinivasan Venkatachary, Shadi Noghabi, Tarun Bharti, Younghoon Jun,
  Aleksandr Zaks, Simon Green, Jeshwanth Challagundla, William Wong, Muqthar
  Mohammad, Dean Hirsch, Yong Cheng, Iftekhar Naim, Lev Proleev, Damien
  Vincent, Aayush Singh, Maxim Krikun, Dilip Krishnan, Zoubin Ghahramani, Aviel
  Atias, Rajeev Aggarwal, Christo Kirov, Dimitrios Vytiniotis, Christy Koh,
  Alexandra Chronopoulou, Pawan Dogra, Vlad-Doru Ion, Gladys Tyen, Jason Lee,
  Felix Weissenberger, Trevor Strohman, Ashwin Balakrishna, Jack Rae, Marko
  Velic, Raoul de~Liedekerke, Oded Elyada, Wentao Yuan, Canoee Liu, Lior Shani,
  Sergey Kishchenko, Bea Alessio, Yandong Li, Richard Song, Sam Kwei, Orion
  Jankowski, Aneesh Pappu, Youhei Namiki, Yenai Ma, Nilesh Tripuraneni, Colin
  Cherry, Marissa Ikonomidis, Yu-Cheng Ling, Colin Ji, Beka Westberg, Auriel
  Wright, Da~Yu, David Parkinson, Swaroop Ramaswamy, Jerome Connor,
  Soheil~Hassas Yeganeh, Snchit Grover, George Kenwright, Lubo Litchev, Chris
  Apps, Alex Tomala, Felix Halim, Alex Castro-Ros, Zefei Li, Anudhyan Boral,
  Pauline Sho, Michal Yarom, Eric Malmi, David Klinghoffer, Rebecca Lin, Alan
  Ansell, Pradeep~Kumar S, Shubin Zhao, Siqi Zuo, Adam Santoro, Heng-Tze Cheng,
  Solomon Demmessie, Yuchi Liu, Nicole Brichtova, Allie Culp, Nathaniel Braun,
  Dan Graur, Will Ng, Nikhil Mehta, Aaron Phillips, Patrik Sundberg, Varun
  Godbole, Fangyu Liu, Yash Katariya, David Rim, Mojtaba Seyedhosseini, Sean
  Ammirati, Jonas Valfridsson, Mahan Malihi, Timothy Knight, Andeep Toor,
  Thomas Lampe, Abe Ittycheriah, Lewis Chiang, Chak Yeung, Alexandre
  Fréchette, Jinmeng Rao, Huisheng Wang, Himanshu Srivastava, Richard Zhang,
  Rocky Rhodes, Ariel Brand, Dean Weesner, Ilya Figotin, Felix Gimeno, Rachana
  Fellinger, Pierre Marcenac, José Leal, Eyal Marcus, Victor Cotruta, Rodrigo
  Cabrera, Sheryl Luo, Dan Garrette, Vera Axelrod, Sorin Baltateanu, David
  Barker, Dongkai Chen, Horia Toma, Ben Ingram, Jason Riesa, Chinmay Kulkarni,
  Yujing Zhang, Hongbin Liu, Chao Wang, Martin Polacek, Will Wu, Kai Hui,
  Adrian~N Reyes, Yi~Su, Megan Barnes, Ishaan Malhi, Anfal Siddiqui, Qixuan
  Feng, Mihai Damaschin, Daniele Pighin, Andreas Steiner, Samuel Yang,
  Ramya~Sree Boppana, Simeon Ivanov, Arun Kandoor, Aditya Shah, Asier Mujika,
  Da~Huang, Christopher~A. Choquette-Choo, Mohak Patel, Tianhe Yu, Toni
  Creswell, Jerry, Liu, Catarina Barros, Yasaman Razeghi, Aurko Roy, Phil
  Culliton, Binbin Xiong, Jiaqi Pan, Thomas Strohmann, Tolly Powell, Babi Seal,
  Doug DeCarlo, Pranav Shyam, Kaan Katircioglu, Xuezhi Wang, Cassidy Hardin,
  Immanuel Odisho, Josef Broder, Oscar Chang, Arun Nair, Artem Shtefan, Maura
  O'Brien, Manu Agarwal, Sahitya Potluri, Siddharth Goyal, Amit Jhindal,
  Saksham Thakur, Yury Stuken, James Lyon, Kristina Toutanova, Fangxiaoyu Feng,
  Austin Wu, Ben Horn, Alek Wang, Alex Cullum, Gabe Taubman, Disha Shrivastava,
  Chongyang Shi, Hamish Tomlinson, Roma Patel, Tao Tu, Ada~Maksutaj Oflazer,
  Francesco Pongetti, Mingyao Yang, Adrien~Ali Taïga, Vincent Perot, Nuo~Wang
  Pierse, Feng Han, Yoel Drori, Iñaki Iturrate, Ayan Chakrabarti, Legg Yeung,
  Dave Dopson, Yi~ting Chen, Apoorv Kulshreshtha, Tongfei Guo, Philip Pham, Tal
  Schuster, Junquan Chen, Alex Polozov, Jinwei Xing, Huanjie Zhou, Praneeth
  Kacham, Doron Kukliansky, Antoine Miech, Sergey Yaroshenko, Ed~Chi, Sholto
  Douglas, Hongliang Fei, Mathieu Blondel, Preethi Myla, Lior Madmoni, Xing Wu,
  Daniel Keysers, Kristian Kjems, Isabela Albuquerque, Lijun Yu, Joel D'sa,
  Michelle Plantan, Vlad Ionescu, Jaume~Sanchez Elias, Abhirut Gupta,
  Manish~Reddy Vuyyuru, Fred Alcober, Tong Zhou, Kaiyang Ji, Florian Hartmann,
  Subha Puttagunta, Hugo Song, Ehsan Amid, Anca Stefanoiu, Andrew Lee, Paul
  Pucciarelli, Emma Wang, Amit Raul, Slav Petrov, Isaac Tian, Valentin Anklin,
  Nana Nti, Victor Gomes, Max Schumacher, Grace Vesom, Alex Panagopoulos,
  Konstantinos Bousmalis, Daniel Andor, Josh Jacob, Yuan Zhang, Bill Rosgen,
  Matija Kecman, Matthew Tung, Alexandra Belias, Noah Goodman, Paul Covington,
  Brian Wieder, Nikita Saxena, Elnaz Davoodi, Muhuan Huang, Sharath Maddineni,
  Vincent Roulet, Folawiyo Campbell-Ajala, Pier~Giuseppe Sessa, Xintian, Wu,
  Guangda Lai, Paul Collins, Alex Haig, Vytenis Sakenas, Xiaowei Xu, Marissa
  Giustina, Laurent~El Shafey, Pichi Charoenpanit, Shefali Garg, Joshua
  Ainslie, Boone Severson, Montse~Gonzalez Arenas, Shreya Pathak, Sujee
  Rajayogam, Jie Feng, Michiel Bakker, Sheng Li, Nevan Wichers, Jamie Rogers,
  Xinyang Geng, Yeqing Li, Rolf Jagerman, Chao Jia, Nadav Olmert, David Sharon,
  Matthew Mauger, Sandeep Mariserla, Hongxu Ma, Megha Mohabey, Kyuyeun Kim,
  Alek Andreev, Scott Pollom, Juliette Love, Vihan Jain, Priyanka Agrawal,
  Yannick Schroecker, Alisa Fortin, Manfred Warmuth, Ji~Liu, Andrew Leach,
  Irina Blok, Ganesh~Poomal Girirajan, Roee Aharoni, Benigno Uria, Andrei
  Sozanschi, Dan Goldberg, Lucian Ionita, Marco~Tulio Ribeiro, Martin Zlocha,
  Vighnesh Birodkar, Sami Lachgar, Liangzhe Yuan, Himadri Choudhury, Matt
  Ginsberg, Fei Zheng, Gregory Dibb, Emily Graves, Swachhand Lokhande, Gabriel
  Rasskin, George-Cristian Muraru, Corbin Quick, Sandeep Tata, Pierre Sermanet,
  Aditya Chawla, Itay Karo, Yan Wang, Susan Zhang, Orgad Keller, Anca Dragan,
  Guolong Su, Ian Chou, Xi~Liu, Yiqing Tao, Shruthi Prabhakara, Marc Wilson,
  Ruibo Liu, Shibo Wang, Georgie Evans, David Du, Alfonso Castaño, Gautam
  Prasad, Mona~El Mahdy, Sebastian Gerlach, Machel Reid, Jarrod Kahn, Amir
  Zait, Thanumalayan~Sankaranarayana Pillai, Thatcher Ulrich, Guanyu Wang, Jan
  Wassenberg, Efrat Farkash, Kiran Yalasangi, Congchao Wang, Maria Bauza, Simon
  Bucher, Ting Liu, Jun Yan, Gary Leung, Vikas Sindhwani, Parker Barnes, Avi
  Singh, Ivan Jurin, Jichuan Chang, Niket~Kumar Bhumihar, Sivan Eiger, Gui
  Citovsky, Ben Withbroe, Zhang Li, Siyang Xue, Niccolò~Dal Santo, Georgi
  Stoyanov, Yves Raimond, Steven Zheng, Yilin Gao, Vít Listík, Sławek
  Kwasiborski, Rachel Saputro, Adnan Ozturel, Ganesh Mallya, Kushal Majmundar,
  Ross West, Paul Caron, Jinliang Wei, Lluis Castrejon, Sharad Vikram, Deepak
  Ramachandran, Nikhil Dhawan, Jiho Park, Sara Smoot, George van~den Driessche,
  Yochai Blau, Chase Malik, Wei Liang, Roy Hirsch, Cicero~Nogueira dos Santos,
  Eugene Weinstein, Aäron van~den Oord, Sid Lall, Nicholas FitzGerald, Zixuan
  Jiang, Xuan Yang, Dale Webster, Ali Elqursh, Aedan Pope, Georges Rotival,
  David Raposo, Wanzheng Zhu, Jeff Dean, Sami Alabed, Dustin Tran, Arushi
  Gupta, Zach Gleicher, Jessica Austin, Edouard Rosseel, Megh Umekar, Dipanjan
  Das, Yinghao Sun, Kai Chen, Karolis Misiunas, Xiang Zhou, Yixian Di, Alyssa
  Loo, Josh Newlan, Bo~Li, Vinay Ramasesh, Ying Xu, Alex Chen, Sudeep Gandhe,
  Radu Soricut, Nikita Gupta, Shuguang Hu, Seliem El-Sayed, Xavier Garcia, Idan
  Brusilovsky, Pu-Chin Chen, Andrew Bolt, Lu~Huang, Alex Gurney, Zhiying Zhang,
  Alexander Pritzel, Jarek Wilkiewicz, Bryan Seybold, Bhargav~Kanagal Shamanna,
  Felix Fischer, Josef Dean, Karan Gill, Ross Mcilroy, Abhishek Bhowmick,
  Jeremy Selier, Antoine Yang, Derek Cheng, Vladimir Magay, Jie Tan, Dhriti
  Varma, Christian Walder, Tomas Kocisky, Ryo Nakashima, Paul Natsev, Mike
  Kwong, Ionel Gog, Chiyuan Zhang, Sander Dieleman, Thomas Jimma, Andrey
  Ryabtsev, Siddhartha Brahma, David Steiner, Dayou Du, Ante Žužul, Mislav
  Žanić, Mukund Raghavachari, Willi Gierke, Zeyu Zheng, Dessie Petrova, Yann
  Dauphin, Yuchuan Liu, Ido Kessler, Steven Hand, Chris Duvarney, Seokhwan Kim,
  Hyo Lee, Léonard Hussenot, Jeffrey Hui, Josh Smith, Deepali Jain, Jiawei
  Xia, Gaurav~Singh Tomar, Keyvan Amiri, Du~Phan, Fabian Fuchs, Tobias Weyand,
  Nenad Tomasev, Alexandra Cordell, Xin Liu, Jonathan Mallinson, Pankaj Joshi,
  Andy Crawford, Arun Suggala, Steve Chien, Nick Fernando, Mariella
  Sanchez-Vargas, Duncan Williams, Phil Crone, Xiyang Luo, Igor Karpov, Jyn
  Shan, Terry Thurk, Robin Strudel, Paul Voigtlaender, Piyush Patil, Tim Dozat,
  Ali Khodaei, Sahil Singla, Piotr Ambroszczyk, Qiyin Wu, Yifan Chang, Brian
  Roark, Chaitra Hegde, Tianli Ding, Angelos Filos, Zhongru Wu, André~Susano
  Pinto, Shuang Liu, Saarthak Khanna, Aditya Pandey, Siobhan Mcloughlin, Qiujia
  Li, Sam Haves, Allan Zhou, Elena Buchatskaya, Isabel Leal, Peter de~Boursac,
  Nami Akazawa, Nina Anderson, Terry Chen, Krishna Somandepalli, Chen Liang,
  Sheela Goenka, Stephanie Winkler, Alexander Grushetsky, Yifan Ding, Jamie
  Smith, Fan Ye, Jordi Pont-Tuset, Eric Li, Ruichao Li, Tomer Golany, Dawid
  Wegner, Tao Jiang, Omer Barak, Yuan Shangguan, Eszter Vértes, Renee Wong,
  Jörg Bornschein, Alex Tudor, Michele Bevilacqua, Tom Schaul, Ankit~Singh
  Rawat, Yang Zhao, Kyriakos Axiotis, Lei Meng, Cory McLean, Jonathan Lai,
  Jennifer Beattie, Nate Kushman, Yaxin Liu, Blair Kutzman, Fiona Lang,
  Jingchen Ye, Praneeth Netrapalli, Pushkar Mishra, Myriam Khan, Megha Goel,
  Rob Willoughby, David Tian, Honglei Zhuang, JD~Chen, Zak Tsai, Tasos
  Kementsietsidis, Arjun Khare, James Keeling, Keyang Xu, Nathan Waters,
  Florent Altché, Ashok Popat, Bhavishya Mittal, David Saxton, Dalia~El
  Badawy, Michael Mathieu, Zheng Zheng, Hao Zhou, Nishant Ranka, Richard Shin,
  Qingnan Duan, Tim Salimans, Ioana Mihailescu, Uri Shaham, Ming-Wei Chang,
  Yannis Assael, Nishanth Dikkala, Martin Izzard, Vincent Cohen-Addad, Cat
  Graves, Vlad Feinberg, Grace Chung, DJ~Strouse, Danny Karmon, Sahand
  Sharifzadeh, Zoe Ashwood, Khiem Pham, Jon Blanton, Alex Vasiloff, Jarred
  Barber, Mark Geller, Aurick Zhou, Fedir Zubach, Tzu-Kuo Huang, Lei Zhang,
  Himanshu Gupta, Matt Young, Julia Proskurnia, Ronny Votel, Valentin Gabeur,
  Gabriel Barcik, Aditya Tripathi, Hongkun Yu, Geng Yan, Beer Changpinyo, Filip
  Pavetić, Amy Coyle, Yasuhisa Fujii, Jorge~Gonzalez Mendez, Tianhao Zhou,
  Harish Rajamani, Blake Hechtman, Eddie Cao, Da-Cheng Juan, Yi-Xuan Tan,
  Valentin Dalibard, Yilun Du, Natalie Clay, Kaisheng Yao, Wenhao Jia, Dimple
  Vijaykumar, Yuxiang Zhou, Xinyi Bai, Wei-Chih Hung, Steven Pecht, Georgi
  Todorov, Nikhil Khadke, Pramod Gupta, Preethi Lahoti, Arnaud Autef, Karthik
  Duddu, James Lee-Thorp, Alexander Bykovsky, Tautvydas Misiunas, Sebastian
  Flennerhag, Santhosh Thangaraj, Jed McGiffin, Zack Nado, Markus Kunesch,
  Andreas Noever, Amir Hertz, Marco Liang, Victor Stone, Evan Palmer, Samira
  Daruki, Arijit Pramanik, Siim Põder, Austin Kyker, Mina Khan, Evgeny
  Sluzhaev, Marvin Ritter, Avraham Ruderman, Wenlei Zhou, Chirag Nagpal, Kiran
  Vodrahalli, George Necula, Paul Barham, Ellie Pavlick, Jay Hartford, Izhak
  Shafran, Long Zhao, Maciej Mikuła, Tom Eccles, Hidetoshi Shimokawa, Kanav
  Garg, Luke Vilnis, Hanwen Chen, Ilia Shumailov, Kuang-Huei Lee, Abdelrahman
  Abdelhamed, Meiyan Xie, Vered Cohen, Ester Hlavnova, Dan Malkin, Chawin
  Sitawarin, James Lottes, Pauline Coquinot, Tianli Yu, Sandeep Kumar, Jingwei
  Zhang, Aroma Mahendru, Zafarali Ahmed, James Martens, Tao Chen, Aviel Boag,
  Daiyi Peng, Coline Devin, Arseniy Klimovskiy, Mary Phuong, Danny Vainstein,
  Jin Xie, Bhuvana Ramabhadran, Nathan Howard, Xinxin Yu, Gitartha Goswami,
  Jingyu Cui, Sam Shleifer, Mario Pinto, Chih-Kuan Yeh, Ming-Hsuan Yang, Sara
  Javanmardi, Dan Ethier, Chace Lee, Jordi Orbay, Suyog Kotecha, Carla
  Bromberg, Pete Shaw, James Thornton, Adi~Gerzi Rosenthal, Shane Gu, Matt
  Thomas, Ian Gemp, Aditya Ayyar, Asahi Ushio, Aarush Selvan, Joel Wee, Chenxi
  Liu, Maryam Majzoubi, Weiren Yu, Jake Abernethy, Tyler Liechty, Renke Pan,
  Hoang Nguyen, Qiong, Hu, Sarah Perrin, Abhinav Arora, Emily Pitler, Weiyi
  Wang, Kaushik Shivakumar, Flavien Prost, Ben Limonchik, Jing Wang, Yi~Gao,
  Timothee Cour, Shyamal Buch, Huan Gui, Maria Ivanova, Philipp Neubeck, Kelvin
  Chan, Lucy Kim, Huizhong Chen, Naman Goyal, Da-Woon Chung, Lu~Liu, Yao Su,
  Anastasia Petrushkina, Jiajun Shen, Armand Joulin, Yuanzhong Xu, Stein~Xudong
  Lin, Yana Kulizhskaya, Ciprian Chelba, Shobha Vasudevan, Eli Collins,
  Vasilisa Bashlovkina, Tony Lu, Doug Fritz, Jongbin Park, Yanqi Zhou, Chen Su,
  Richard Tanburn, Mikhail Sushkov, Mitchelle Rasquinha, Jinning Li, Jennifer
  Prendki, Yiming Li, Pallavi LV, Shriya Sharma, Hen Fitoussi, Hui Huang,
  Andrew Dai, Phuong Dao, Mike Burrows, Henry Prior, Danfeng Qin, Golan Pundak,
  Lars~Lowe Sjoesund, Art Khurshudov, Zhenkai Zhu, Albert Webson, Elizabeth
  Kemp, Tat Tan, Saurabh Agrawal, Susie Sargsyan, Liqun Cheng, Jim Stephan, Tom
  Kwiatkowski, David Reid, Arunkumar Byravan, Assaf~Hurwitz Michaely, Nicolas
  Heess, Luowei Zhou, Sonam Goenka, Viral Carpenter, Anselm Levskaya, Bo~Wang,
  Reed Roberts, Rémi Leblond, Sharat Chikkerur, Stav Ginzburg, Max Chang,
  Robert Riachi, Chuqiao, Xu, Zalán Borsos, Michael Pliskin, Julia Pawar,
  Morgane Lustman, Hannah Kirkwood, Ankit Anand, Aditi Chaudhary, Norbert Kalb,
  Kieran Milan, Sean Augenstein, Anna Goldie, Laurel Prince, Karthik Raman,
  Yanhua Sun, Vivian Xia, Aaron Cohen, Zhouyuan Huo, Josh Camp, Seher Ellis,
  Lukas Zilka, David~Vilar Torres, Lisa Patel, Sho Arora, Betty Chan, Jonas
  Adler, Kareem Ayoub, Jacky Liang, Fayaz Jamil, Jiepu Jiang, Simon
  Baumgartner, Haitian Sun, Yael Karov, Yaroslav Akulov, Hui Zheng, Irene Cai,
  Claudio Fantacci, James Rubin, Alex~Rav Acha, Mengchao Wang, Nina D'Souza,
  Rohit Sathyanarayana, Shengyang Dai, Simon Rowe, Andrey Simanovsky, Omer
  Goldman, Yuheng Kuang, Xiaoyue Pan, Andrew Rosenberg, Tania Rojas-Esponda,
  Praneet Dutta, Amy Zeng, Irina Jurenka, Greg Farquhar, Yamini Bansal, Shariq
  Iqbal, Becca Roelofs, Ga-Young Joung, Parker Beak, Changwan Ryu, Ryan Poplin,
  Yan Wu, Jean-Baptiste Alayrac, Senaka Buthpitiya, Olaf Ronneberger, Caleb
  Habtegebriel, Wei Li, Paul Cavallaro, Aurora Wei, Guy Bensky, Timo Denk,
  Harish Ganapathy, Jeff Stanway, Pratik Joshi, Francesco Bertolini, Jessica
  Lo, Olivia Ma, Zachary Charles, Geta Sampemane, Himanshu Sahni, Xu~Chen,
  Harry Askham, David Gaddy, Peter Young, Jiewen Tan, Matan Eyal, Arthur
  Bražinskas, Li~Zhong, Zhichun Wu, Mark Epstein, Kai Bailey, Andrew Hard,
  Kamyu Lee, Sasha Goldshtein, Alex Ruiz, Mohammed Badawi, Matthias
  Lochbrunner, JK~Kearns, Ashley Brown, Fabio Pardo, Theophane Weber, Haichuan
  Yang, Pan-Pan Jiang, Berkin Akin, Zhao Fu, Marcus Wainwright, Chi Zou, Meenu
  Gaba, Pierre-Antoine Manzagol, Wendy Kan, Yang Song, Karina Zainullina, Rui
  Lin, Jeongwoo Ko, Salil Deshmukh, Apoorv Jindal, James Svensson, Divya Tyam,
  Heri Zhao, Christine Kaeser-Chen, Scott Baird, Pooya Moradi, Jamie Hall,
  Qiuchen Guo, Vincent Tsang, Bowen Liang, Fernando Pereira, Suhas Ganesh, Ivan
  Korotkov, Jakub Adamek, Sridhar Thiagarajan, Vinh Tran, Charles Chen, Chris
  Tar, Sanil Jain, Ishita Dasgupta, Taylan Bilal, David Reitter, Kai Zhao,
  Giulia Vezzani, Yasmin Gehman, Pulkit Mehta, Lauren Beltrone, Xerxes
  Dotiwalla, Sergio Guadarrama, Zaheer Abbas, Stefani Karp, Petko Georgiev,
  Chun-Sung Ferng, Marc Brockschmidt, Liqian Peng, Christoph Hirnschall, Vikas
  Verma, Yingying Bi, Ying Xiao, Avigail Dabush, Kelvin Xu, Phil Wallis,
  Randall Parker, Qifei Wang, Yang Xu, Ilkin Safarli, Dinesh Tewari, Yin Zhang,
  Seungyeon Kim, Andrea Gesmundo, Mackenzie Thomas, Sergey Levi, Ahmed
  Chowdhury, Kanishka Rao, Peter Garst, Sam Conway-Rahman, Helen Ran, Kay
  McKinney, Zhisheng Xiao, Wenhao Yu, Rohan Agrawal, Axel Stjerngren, Catalin
  Ionescu, Jingjing Chen, Vivek Sharma, Justin Chiu, Fei Liu, Ken Franko,
  Clayton Sanford, Xingyu Cai, Paul Michel, Sanjay Ganapathy, Jane Labanowski,
  Zachary Garrett, Ben Vargas, Sean Sun, Bryan Gale, Thomas Buschmann,
  Guillaume Desjardins, Nimesh Ghelani, Palak Jain, Mudit Verma, Chulayuth
  Asawaroengchai, Julian Eisenschlos, Jitendra Harlalka, Hideto Kazawa, Don
  Metzler, Joshua Howland, Ying Jian, Jake Ades, Viral Shah, Tynan Gangwani,
  Seungji Lee, Roman Ring, Steven~M. Hernandez, Dean Reich, Amer Sinha,
  Ashutosh Sathe, Joe Kovac, Ashleah Gill, Ajay Kannan, Andrea D'olimpio,
  Martin Sevenich, Jay Whang, Been Kim, Khe~Chai Sim, Jilin Chen, Jiageng
  Zhang, Shuba Lall, Yossi Matias, Bill Jia, Abe Friesen, Sara Nasso, Ashish
  Thapliyal, Bryan Perozzi, Ting Yu, Anna Shekhawat, Safeen Huda, Peter
  Grabowski, Eric Wang, Ashwin Sreevatsa, Hilal Dib, Mehadi Hassen, Parker
  Schuh, Vedrana Milutinovic, Chris Welty, Michael Quinn, Ali Shah, Bangju
  Wang, Gabe Barth-Maron, Justin Frye, Natalie Axelsson, Tao Zhu, Yukun Ma,
  Irene Giannoumis, Hanie Sedghi, Chang Ye, Yi~Luan, Kevin Aydin, Bilva
  Chandra, Vivek Sampathkumar, Ronny Huang, Victor Lavrenko, Ahmed Eleryan, Zhi
  Hong, Steven Hansen, Sara~Mc Carthy, Bidisha Samanta, Domagoj Ćevid, Xin
  Wang, Fangtao Li, Michael Voznesensky, Matt Hoffman, Andreas Terzis, Vikash
  Sehwag, Gil Fidel, Luheng He, Mu~Cai, Yanzhang He, Alex Feng, Martin
  Nikoltchev, Samrat Phatale, Jason Chase, Rory Lawton, Ming Zhang, Tom Ouyang,
  Manuel Tragut, Mehdi~Hafezi Manshadi, Arjun Narayanan, Jiaming Shen, Xu~Gao,
  Tolga Bolukbasi, Nick Roy, Xin Li, Daniel Golovin, Liviu Panait, Zhen Qin,
  Guangxing Han, Thomas Anthony, Sneha Kudugunta, Viorica Patraucean, Aniket
  Ray, Xinyun Chen, Xiaochen Yang, Tanuj Bhatia, Pranav Talluri, Alex Morris,
  Andrija Ražnatović, Bethanie Brownfield, James An, Sheng Peng, Patrick
  Kane, Ce~Zheng, Nico Duduta, Joshua Kessinger, James Noraky, Siqi Liu, Keran
  Rong, Petar Veličković, Keith Rush, Alex Goldin, Fanny Wei, Shiva
  Mohan~Reddy Garlapati, Caroline Pantofaru, Okwan Kwon, Jianmo Ni, Eric
  Noland, Julia~Di Trapani, Françoise Beaufays, Abhijit~Guha Roy, Yinlam Chow,
  Aybuke Turker, Geoffrey Cideron, Lantao Mei, Jon Clark, Qingyun Dou, Matko
  Bošnjak, Ralph Leith, Yuqing Du, Amir Yazdanbakhsh, Milad Nasr, Chester
  Kwak, Suraj~Satishkumar Sheth, Alex Kaskasoli, Ankesh Anand, Balaji
  Lakshminarayanan, Sammy Jerome, David Bieber, Chun-Te Chu, Alexandre Senges,
  Tianxiao Shen, Mukund Sridhar, Ndaba Ndebele, Benjamin Beyret, Shakir
  Mohamed, Mia Chen, Markus Freitag, Jiaxian Guo, Luyang Liu, Paul Roit, Heng
  Chen, Shen Yan, Tom Stone, JD~Co-Reyes, Jeremy Cole, Salvatore Scellato,
  Shekoofeh Azizi, Hadi Hashemi, Alicia Jin, Anand Iyer, Marcella Valentine,
  András György, Arun Ahuja, Daniel~Hernandez Diaz, Chen-Yu Lee, Nathan
  Clement, Weize Kong, Drew Garmon, Ishaan Watts, Kush Bhatia, Khyatti Gupta,
  Matt Miecnikowski, Hugo Vallet, Ankur Taly, Edward Loper, Saket Joshi, James
  Atwood, Jo~Chick, Mark Collier, Fotis Iliopoulos, Ryan Trostle, Beliz Gunel,
  Ramiro Leal-Cavazos, Arnar~Mar Hrafnkelsson, Michael Guzman, Xiaoen Ju, Andy
  Forbes, Jesse Emond, Kushal Chauhan, Ben Caine, Li~Xiao, Wenjun Zeng,
  Alexandre Moufarek, Daniel Murphy, Maya Meng, Nitish Gupta, Felix Riedel,
  Anil Das, Elijah Lawal, Shashi Narayan, Tiberiu Sosea, James Swirhun, Linda
  Friso, Behnam Neyshabur, Jing Lu, Sertan Girgin, Michael Wunder, Edouard
  Yvinec, Aroonalok Pyne, Victor Carbune, Shruti Rijhwani, Yang Guo, Tulsee
  Doshi, Anton Briukhov, Max Bain, Ayal Hitron, Xuanhui Wang, Ashish Gupta,
  Ke~Chen, Cosmo Du, Weiyang Zhang, Dhruv Shah, Arjun Akula, Max Dylla, Ashyana
  Kachra, Weicheng Kuo, Tingting Zou, Lily Wang, Luyao Xu, Jifan Zhu, Justin
  Snyder, Sachit Menon, Orhan Firat, Igor Mordatch, Yuan Yuan, Natalia
  Ponomareva, Rory Blevins, Lawrence Moore, Weijun Wang, Phil Chen, Martin
  Scholz, Artur Dwornik, Jason Lin, Sicheng Li, Diego Antognini, Te~I, Xiaodan
  Song, Matt Miller, Uday Kalra, Adam Raveret, Oscar Akerlund, Felix Wu, Andrew
  Nystrom, Namrata Godbole, Tianqi Liu, Hannah DeBalsi, Jewel Zhao, Buhuang
  Liu, Avi Caciularu, Lauren Lax, Urvashi Khandelwal, Victoria Langston, Eric
  Bailey, Silvio Lattanzi, Yufei Wang, Neel Kovelamudi, Sneha Mondal, Guru
  Guruganesh, Nan Hua, Ofir Roval, Paweł Wesołowski, Rishikesh Ingale,
  Jonathan Halcrow, Tim Sohn, Christof Angermueller, Bahram Raad, Eli
  Stickgold, Eva Lu, Alec Kosik, Jing Xie, Timothy Lillicrap, Austin Huang,
  Lydia~Lihui Zhang, Dominik Paulus, Clement Farabet, Alex Wertheim, Bing Wang,
  Rishabh Joshi, Chu ling Ko, Yonghui Wu, Shubham Agrawal, Lily Lin, XiangHai
  Sheng, Peter Sung, Tyler Breland-King, Christina Butterfield, Swapnil Gawde,
  Sumeet Singh, Qiao Zhang, Raj Apte, Shilpa Shetty, Adrian Hutter, Tao Li,
  Elizabeth Salesky, Federico Lebron, Jonni Kanerva, Michela Paganini, Arthur
  Nguyen, Rohith Vallu, Jan-Thorsten Peter, Sarmishta Velury, David Kao, Jay
  Hoover, Anna Bortsova, Colton Bishop, Shoshana Jakobovits, Alessandro
  Agostini, Alekh Agarwal, Chang Liu, Charles Kwong, Sasan Tavakkol, Ioana
  Bica, Alex Greve, Anirudh GP, Jake Marcus, Le~Hou, Tom Duerig, Rivka
  Moroshko, Dave Lacey, Andy Davis, Julien Amelot, Guohui Wang, Frank Kim,
  Theofilos Strinopoulos, Hui Wan, Charline~Le Lan, Shankar Krishnan, Haotian
  Tang, Peter Humphreys, Junwen Bai, Idan~Heimlich Shtacher, Diego Machado,
  Chenxi Pang, Ken Burke, Dangyi Liu, Renga Aravamudhan, Yue Song, Ed~Hirst,
  Abhimanyu Singh, Brendan Jou, Liang Bai, Francesco Piccinno, Chuyuan~Kelly
  Fu, Robin Alazard, Barak Meiri, Daniel Winter, Charlie Chen, Mingda Zhang,
  Jens Heitkaemper, John Lambert, Jinhyuk Lee, Alexander Frömmgen, Sergey
  Rogulenko, Pranav Nair, Paul Niemczyk, Anton Bulyenov, Bibo Xu, Hadar
  Shemtov, Morteza Zadimoghaddam, Serge Toropov, Mateo Wirth, Hanjun Dai,
  Sreenivas Gollapudi, Daniel Zheng, Alex Kurakin, Chansoo Lee, Kalesha
  Bullard, Nicolas Serrano, Ivana Balazevic, Yang Li, Johan Schalkwyk, Mark
  Murphy, Mingyang Zhang, Kevin Sequeira, Romina Datta, Nishant Agrawal,
  Charles Sutton, Nithya Attaluri, Mencher Chiang, Wael Farhan, Gregory
  Thornton, Kate Lin, Travis Choma, Hung Nguyen, Kingshuk Dasgupta, Dirk
  Robinson, Iulia Comşa, Michael Riley, Arjun Pillai, Basil Mustafa, Ben
  Golan, Amir Zandieh, Jean-Baptiste Lespiau, Billy Porter, David Ross,
  Sujeevan Rajayogam, Mohit Agarwal, Subhashini Venugopalan, Bobak Shahriari,
  Qiqi Yan, Hao Xu, Taylor Tobin, Pavel Dubov, Hongzhi Shi, Adrià Recasens,
  Anton Kovsharov, Sebastian Borgeaud, Lucio Dery, Shanthal Vasanth, Elena
  Gribovskaya, Linhai Qiu, Mahdis Mahdieh, Wojtek Skut, Elizabeth Nielsen,
  CJ~Zheng, Adams Yu, Carrie~Grimes Bostock, Shaleen Gupta, Aaron Archer, Chris
  Rawles, Elinor Davies, Alexey Svyatkovskiy, Tomy Tsai, Yoni Halpern,
  Christian Reisswig, Bartek Wydrowski, Bo~Chang, Joan Puigcerver, Mor~Hazan
  Taege, Jian Li, Eva Schnider, Xinjian Li, Dragos Dena, Yunhan Xu, Umesh
  Telang, Tianze Shi, Heiga Zen, Kyle Kastner, Yeongil Ko, Neesha Subramaniam,
  Aviral Kumar, Pete Blois, Zhuyun Dai, John Wieting, Yifeng Lu, Yoel Zeldes,
  Tian Xie, Anja Hauth, Alexandru Ţifrea, Yuqi Li, Sam El-Husseini, Dan
  Abolafia, Howard Zhou, Wen Ding, Sahra Ghalebikesabi, Carlos Guía, Andrii
  Maksai, Ágoston Weisz, Sercan Arik, Nick Sukhanov, Aga Świetlik, Xuhui Jia,
  Luo Yu, Weiyue Wang, Mark Brand, Dawn Bloxwich, Sean Kirmani, Zhe Chen, Alec
  Go, Pablo Sprechmann, Nithish Kannen, Alen Carin, Paramjit Sandhu, Isabel
  Edkins, Leslie Nooteboom, Jai Gupta, Loren Maggiore, Javad Azizi, Yael
  Pritch, Pengcheng Yin, Mansi Gupta, Danny Tarlow, Duncan Smith, Desi Ivanov,
  Mohammad Babaeizadeh, Ankita Goel, Satish Kambala, Grace Chu, Matej Kastelic,
  Michelle Liu, Hagen Soltau, Austin Stone, Shivani Agrawal, Min Kim, Kedar
  Soparkar, Srinivas Tadepalli, Oskar Bunyan, Rachel Soh, Arvind Kannan,
  DY~Kim, Blake~JianHang Chen, Afief Halumi, Sudeshna Roy, Yulong Wang, Olcan
  Sercinoglu, Gena Gibson, Sijal Bhatnagar, Motoki Sano, Daniel von Dincklage,
  Qingchun Ren, Blagoj Mitrevski, Mirek Olšák, Jennifer She, Carl Doersch,
  Jilei, Wang, Bingyuan Liu, Qijun Tan, Tamar Yakar, Tris Warkentin, Alex
  Ramirez, Carl Lebsack, Josh Dillon, Rajiv Mathews, Tom Cobley, Zelin Wu,
  Zhuoyuan Chen, Jon Simon, Swaroop Nath, Tara Sainath, Alexei Bendebury, Ryan
  Julian, Bharath Mankalale, Daria Ćurko, Paulo Zacchello, Adam~R. Brown,
  Kiranbir Sodhia, Heidi Howard, Sergi Caelles, Abhinav Gupta, Gareth Evans,
  Anna Bulanova, Lesley Katzen, Roman Goldenberg, Anton Tsitsulin, Joe Stanton,
  Benoit Schillings, Vitaly Kovalev, Corey Fry, Rushin Shah, Kuo Lin, Shyam
  Upadhyay, Cheng Li, Soroush Radpour, Marcello Maggioni, Jing Xiong, Lukas
  Haas, Jenny Brennan, Aishwarya Kamath, Nikolay Savinov, Arsha Nagrani, Trevor
  Yacovone, Ryan Kappedal, Kostas Andriopoulos, Li~Lao, YaGuang Li, Grigory
  Rozhdestvenskiy, Kazuma Hashimoto, Andrew Audibert, Sophia Austin, Daniel
  Rodriguez, Anian Ruoss, Garrett Honke, Deep Karkhanis, Xi~Xiong, Qing Wei,
  James Huang, Zhaoqi Leng, Vittal Premachandran, Stan Bileschi, Georgios
  Evangelopoulos, Thomas Mensink, Jay Pavagadhi, Denis Teplyashin, Paul Chang,
  Linting Xue, Garrett Tanzer, Sally Goldman, Kaushal Patel, Shixin Li, Jeremy
  Wiesner, Ivy Zheng, Ian Stewart-Binks, Jie Han, Zhi Li, Liangchen Luo, Karel
  Lenc, Mario Lučić, Fuzhao Xue, Ryan Mullins, Alexey Guseynov, Chung-Ching
  Chang, Isaac Galatzer-Levy, Adam Zhang, Garrett Bingham, Grace Hu, Ale
  Hartman, Yue Ma, Jordan Griffith, Alex Irpan, Carey Radebaugh, Summer Yue,
  Lijie Fan, Victor Ungureanu, Christina Sorokin, Hannah Teufel, Peiran Li,
  Rohan Anil, Dimitris Paparas, Todd Wang, Chu-Cheng Lin, Hui Peng, Megan Shum,
  Goran Petrovic, Demetra Brady, Richard Nguyen, Klaus Macherey, Zhihao Li,
  Harman Singh, Madhavi Yenugula, Mariko Iinuma, Xinyi Chen, Kavya Kopparapu,
  Alexey Stern, Shachi Dave, Chandu Thekkath, Florence Perot, Anurag Kumar,
  Fangda Li, Yang Xiao, Matthew Bilotti, Mohammad~Hossein Bateni, Isaac Noble,
  Lisa Lee, Amelio Vázquez-Reina, Julian Salazar, Xiaomeng Yang, Boyu Wang,
  Ela Gruzewska, Anand Rao, Sindhu Raghuram, Zheng Xu, Eyal Ben-David, Jieru
  Mei, Sid Dalmia, Zhaoyi Zhang, Yuchen Liu, Gagan Bansal, Helena Pankov,
  Steven Schwarcz, Andrea Burns, Christine Chan, Sumit Sanghai, Ricky Liang,
  Ethan Liang, Antoine He, Amy Stuart, Arun Narayanan, Yukun Zhu, Christian
  Frank, Bahar Fatemi, Amit Sabne, Oran Lang, Indro Bhattacharya, Shane Settle,
  Maria Wang, Brendan McMahan, Andrea Tacchetti, Livio~Baldini Soares, Majid
  Hadian, Serkan Cabi, Timothy Chung, Nikita Putikhin, Gang Li, Jeremy Chen,
  Austin Tarango, Henryk Michalewski, Mehran Kazemi, Hussain Masoom, Hila
  Sheftel, Rakesh Shivanna, Archita Vadali, Ramona Comanescu, Doug Reid, Joss
  Moore, Arvind Neelakantan, Michaël Sander, Jonathan Herzig, Aviv Rosenberg,
  Mostafa Dehghani, JD~Choi, Michael Fink, Reid Hayes, Eric Ge, Shitao Weng,
  Chia-Hua Ho, John Karro, Kalpesh Krishna, Lam~Nguyen Thiet, Amy Skerry-Ryan,
  Daniel Eppens, Marco Andreetto, Navin Sarma, Silvano Bonacina, Burcu~Karagol
  Ayan, Megha Nawhal, Zhihao Shan, Mike Dusenberry, Shantanu Thakoor, Sagar
  Gubbi, Duc~Dung Nguyen, Reut Tsarfaty, Samuel Albanie, Jovana Mitrović, Meet
  Gandhi, Bo-Juen Chen, Alessandro Epasto, Georgi Stephanov, Ye~Jin, Samuel
  Gehman, Aida Amini, Jack Weber, Feryal Behbahani, Shawn Xu, Miltos Allamanis,
  Xi~Chen, Myle Ott, Claire Sha, Michal Jastrzebski, Hang Qi, David Greene,
  Xinyi Wu, Abodunrinwa Toki, Daniel Vlasic, Jane Shapiro, Ragha Kotikalapudi,
  Zhe Shen, Takaaki Saeki, Sirui Xie, Albin Cassirer, Shikhar Bharadwaj,
  Tatsuya Kiyono, Srinadh Bhojanapalli, Elan Rosenfeld, Sam Ritter, Jieming
  Mao, João~Gabriel Oliveira, Zoltan Egyed, Bernd Bandemer, Emilio Parisotto,
  Keisuke Kinoshita, Juliette Pluto, Petros Maniatis, Steve Li, Yaohui Guo,
  Golnaz Ghiasi, Jean Tarbouriech, Srimon Chatterjee, Julie Jin, Katrina, Xu,
  Jennimaria Palomaki, Séb Arnold, Madhavi Sewak, Federico Piccinini, Mohit
  Sharma, Ben Albrecht, Sean Purser-haskell, Ashwin Vaswani, Chongyan Chen,
  Matheus Wisniewski, Qin Cao, John Aslanides, Nguyet~Minh Phu, Maximilian
  Sieb, Lauren Agubuzu, Anne Zheng, Daniel Sohn, Marco Selvi, Anders
  Andreassen, Krishan Subudhi, Prem Eruvbetine, Oliver Woodman, Tomas Mery,
  Sebastian Krause, Xiaoqi Ren, Xiao Ma, Jincheng Luo, Dawn Chen, Wei Fan,
  Henry Griffiths, Christian Schuler, Alice Li, Shujian Zhang, Jean-Michel
  Sarr, Shixin Luo, Riccardo Patana, Matthew Watson, Dani Naboulsi, Michael
  Collins, Sailesh Sidhwani, Emiel Hoogeboom, Sharon Silver, Emily Caveness,
  Xiaokai Zhao, Mikel Rodriguez, Maxine Deines, Libin Bai, Patrick Griffin,
  Marco Tagliasacchi, Emily Xue, Spandana~Raj Babbula, Bo~Pang, Nan Ding,
  Gloria Shen, Elijah Peake, Remi Crocker, Shubha~Srinivas Raghvendra, Danny
  Swisher, Woohyun Han, Richa Singh, Ling Wu, Vladimir Pchelin, Tsendsuren
  Munkhdalai, Dana Alon, Geoff Bacon, Efren Robles, Jannis Bulian, Melvin
  Johnson, George Powell, Felipe~Tiengo Ferreira, Yaoyiran Li, Frederik
  Benzing, Mihajlo Velimirović, Hubert Soyer, William Kong, Tony, Nguyên,
  Zhen Yang, Jeremiah Liu, Joost van Amersfoort, Daniel Gillick, Baochen Sun,
  Nathalie Rauschmayr, Katie Zhang, Serena Zhan, Tao Zhou, Alexey Frolov,
  Chengrun Yang, Denis Vnukov, Louis Rouillard, Hongji Li, Amol Mandhane, Nova
  Fallen, Rajesh Venkataraman, Clara~Huiyi Hu, Jennifer Brennan, Jenny Lee,
  Jerry Chang, Martin Sundermeyer, Zhufeng Pan, Rosemary Ke, Simon Tong, Alex
  Fabrikant, William Bono, Jindong Gu, Ryan Foley, Yiran Mao, Manolis Delakis,
  Dhruva Bhaswar, Roy Frostig, Nick Li, Avital Zipori, Cath Hope, Olga Kozlova,
  Swaroop Mishra, Josip Djolonga, Craig Schiff, Majd~Al Merey, Eleftheria
  Briakou, Peter Morgan, Andy Wan, Avinatan Hassidim, RJ~Skerry-Ryan, Kuntal
  Sengupta, Mary Jasarevic, Praveen Kallakuri, Paige Kunkle, Hannah Brennan,
  Tom Lieber, Hassan Mansoor, Julian Walker, Bing Zhang, Annie Xie, Goran
  Žužić, Adaeze Chukwuka, Alex Druinsky, Donghyun Cho, Rui Yao, Ferjad
  Naeem, Shiraz Butt, Eunyoung Kim, Zhipeng Jia, Mandy Jordan, Adam Lelkes,
  Mark Kurzeja, Sophie Wang, James Zhao, Andrew Over, Abhishek Chakladar,
  Marcel Prasetya, Neha Jha, Sriram Ganapathy, Yale Cong, Prakash Shroff, Carl
  Saroufim, Sobhan Miryoosefi, Mohamed Hammad, Tajwar Nasir, Weijuan Xi, Yang
  Gao, Young Maeng, Ben Hora, Chin-Yi Cheng, Parisa Haghani, Yoad Lewenberg,
  Caden Lu, Martin Matysiak, Naina Raisinghani, Huiyu Wang, Lexi Baugher, Rahul
  Sukthankar, Minh Giang, John Schultz, Noah Fiedel, Minmin Chen, Cheng-Chun
  Lee, Tapomay Dey, Hao Zheng, Shachi Paul, Celine Smith, Andy Ly, Yicheng
  Wang, Rishabh Bansal, Bartek Perz, Susanna Ricco, Stasha Blank, Vaishakh
  Keshava, Deepak Sharma, Marvin Chow, Kunal Lad, Komal Jalan, Simon Osindero,
  Craig Swanson, Jacob Scott, Anastasija Ilić, Xiaowei Li, Siddhartha~Reddy
  Jonnalagadda, Afzal~Shama Soudagar, Yan Xiong, Bat-Orgil Batsaikhan, Daniel
  Jarrett, Naveen Kumar, Maulik Shah, Matt Lawlor, Austin Waters, Mark Graham,
  Rhys May, Sabela Ramos, Sandra Lefdal, Zeynep Cankara, Nacho Cano, Brendan
  O'Donoghue, Jed Borovik, Frederick Liu, Jordan Grimstad, Mahmoud Alnahlawi,
  Katerina Tsihlas, Tom Hudson, Nikolai Grigorev, Yiling Jia, Terry Huang,
  Tobenna~Peter Igwe, Sergei Lebedev, Xiaodan Tang, Igor Krivokon, Frankie
  Garcia, Melissa Tan, Eric Jia, Peter Stys, Shikhar Vashishth, Yu~Liang,
  Balaji Venkatraman, Chenjie Gu, Anastasios Kementsietsidis, Chen Zhu,
  Junehyuk Jung, Yunfei Bai, Mohammad~Javad Hosseini, Faruk Ahmed, Aditya
  Gupta, Xin Yuan, Shereen Ashraf, Shitij Nigam, Gautam Vasudevan, Pranjal
  Awasthi, Adi~Mayrav Gilady, Zelda Mariet, Ramy Eskander, Haiguang Li, Hexiang
  Hu, Guillermo Garrido, Philippe Schlattner, George Zhang, Rohun Saxena, Petar
  Dević, Kritika Muralidharan, Ashwin Murthy, Yiqian Zhou, Min Choi, Arissa
  Wongpanich, Zhengdong Wang, Premal Shah, Yuntao Xu, Yiling Huang, Stephen
  Spencer, Alice Chen, James Cohan, Junjie Wang, Jonathan Tompson, Junru Wu,
  Ruba Haroun, Haiqiong Li, Blanca Huergo, Fan Yang, Tongxin Yin, James Wendt,
  Michael Bendersky, Rahma Chaabouni, Javier Snaider, Johan Ferret, Abhishek
  Jindal, Tara Thompson, Andrew Xue, Will Bishop, Shubham~Milind Phal, Archit
  Sharma, Yunhsuan Sung, Prabakar Radhakrishnan, Mo~Shomrat, Reeve Ingle,
  Roopali Vij, Justin Gilmer, Mihai~Dorin Istin, Sam Sobell, Yang Lu, Emily
  Nottage, Dorsa Sadigh, Jeremiah Willcock, Tingnan Zhang, Steve Xu, Sasha
  Brown, Katherine Lee, Gary Wang, Yun Zhu, Yi~Tay, Cheolmin Kim, Audrey
  Gutierrez, Abhanshu Sharma, Yongqin Xian, Sungyong Seo, Claire Cui, Elena
  Pochernina, Cip Baetu, Krzysztof Jastrzębski, Mimi Ly, Mohamed Elhawaty, Dan
  Suh, Eren Sezener, Pidong Wang, Nancy Yuen, George Tucker, Jiahao Cai,
  Zuguang Yang, Cindy Wang, Alex Muzio, Hai Qian, Jae Yoo, Derek Lockhart,
  Kevin~R. McKee, Mandy Guo, Malika Mehrotra, Artur Mendonça, Sanket~Vaibhav
  Mehta, Sherry Ben, Chetan Tekur, Jiaqi Mu, Muye Zhu, Victoria Krakovna,
  Hongrae Lee, AJ~Maschinot, Sébastien Cevey, HyunJeong Choe, Aijun Bai, Hansa
  Srinivasan, Derek Gasaway, Nick Young, Patrick Siegler, Dan Holtmann-Rice,
  Vihari Piratla, Kate Baumli, Roey Yogev, Alex Hofer, Hado van Hasselt,
  Svetlana Grant, Yuri Chervonyi, David Silver, Andrew Hogue, Ayushi Agarwal,
  Kathie Wang, Preeti Singh, Four Flynn, Josh Lipschultz, Robert David,
  Lizzetth Bellot, Yao-Yuan Yang, Long Le, Filippo Graziano, Kate Olszewska,
  Kevin Hui, Akanksha Maurya, Nikos Parotsidis, Weijie Chen, Tayo Oguntebi, Joe
  Kelley, Anirudh Baddepudi, Johannes Mauerer, Gregory Shaw, Alex Siegman, Lin
  Yang, Shravya Shetty, Subhrajit Roy, Yunting Song, Wojciech Stokowiec, Ryan
  Burnell, Omkar Savant, Robert Busa-Fekete, Jin Miao, Samrat Ghosh, Liam
  MacDermed, Phillip Lippe, Mikhail Dektiarev, Zach Behrman, Fabian Mentzer,
  Kelvin Nguyen, Meng Wei, Siddharth Verma, Chris Knutsen, Sudeep Dasari,
  Zhipeng Yan, Petr Mitrichev, Xingyu Wang, Virat Shejwalkar, Jacob Austin,
  Srinivas Sunkara, Navneet Potti, Yan Virin, Christian Wright, Gaël Liu,
  Oriana Riva, Etienne Pot, Greg Kochanski, Quoc Le, Gargi Balasubramaniam,
  Arka Dhar, Yuguo Liao, Adam Bloniarz, Divyansh Shukla, Elizabeth Cole, Jong
  Lee, Sheng Zhang, Sushant Kafle, Siddharth Vashishtha, Parsa Mahmoudieh,
  Grace Chen, Raphael Hoffmann, Pranesh Srinivasan, Agustin~Dal Lago, Yoav~Ben
  Shalom, Zi~Wang, Michael Elabd, Anuj Sharma, Junhyuk Oh, Suraj Kothawade,
  Maigo Le, Marianne Monteiro, Shentao Yang, Kaiz Alarakyia, Robert Geirhos,
  Diana Mincu, Håvard Garnes, Hayato Kobayashi, Soroosh Mariooryad, Kacper
  Krasowiak, Zhixin, Lai, Shibl Mourad, Mingqiu Wang, Fan Bu, Ophir Aharoni,
  Guanjie Chen, Abhimanyu Goyal, Vadim Zubov, Ankur Bapna, Elahe Dabir, Nisarg
  Kothari, Kay Lamerigts, Nicola~De Cao, Jeremy Shar, Christopher Yew, Nitish
  Kulkarni, Dre Mahaarachchi, Mandar Joshi, Zhenhai Zhu, Jared Lichtarge,
  Yichao Zhou, Hannah Muckenhirn, Vittorio Selo, Oriol Vinyals, Peter Chen,
  Anthony Brohan, Vaibhav Mehta, Sarah Cogan, Ruth Wang, Ty~Geri, Wei-Jen Ko,
  Wei Chen, Fabio Viola, Keshav Shivam, Lisa Wang, Madeleine~Clare Elish,
  Raluca~Ada Popa, Sébastien Pereira, Jianqiao Liu, Raphael Koster, Donnie
  Kim, Gufeng Zhang, Sayna Ebrahimi, Partha Talukdar, Yanyan Zheng, Petra
  Poklukar, Ales Mikhalap, Dale Johnson, Anitha Vijayakumar, Mark Omernick,
  Matt Dibb, Ayush Dubey, Qiong Hu, Apurv Suman, Vaibhav Aggarwal, Ilya
  Kornakov, Fei Xia, Wing Lowe, Alexey Kolganov, Ted Xiao, Vitaly Nikolaev,
  Steven Hemingray, Bonnie Li, Joana Iljazi, Mikołaj Rybiński, Ballie Sandhu,
  Peggy Lu, Thang Luong, Rodolphe Jenatton, Vineetha Govindaraj, Hui, Li,
  Gabriel Dulac-Arnold, Wonpyo Park, Henry Wang, Abhinit Modi, Jean
  Pouget-Abadie, Kristina Greller, Rahul Gupta, Robert Berry, Prajit
  Ramachandran, Jinyu Xie, Liam McCafferty, Jianling Wang, Kilol Gupta,
  Hyeontaek Lim, Blaž Bratanič, Andy Brock, Ilia Akolzin, Jim Sproch, Dan
  Karliner, Duhyeon Kim, Adrian Goedeckemeyer, Noam Shazeer, Cordelia Schmid,
  Daniele Calandriello, Parul Bhatia, Krzysztof Choromanski, Ceslee Montgomery,
  Dheeru Dua, Ana Ramalho, Helen King, Yue Gao, Lynn Nguyen, David Lindner,
  Divya Pitta, Oleaser Johnson, Khalid Salama, Diego Ardila, Michael Han, Erin
  Farnese, Seth Odoom, Ziyue Wang, Xiangzhuo Ding, Norman Rink, Ray Smith,
  Harshal~Tushar Lehri, Eden Cohen, Neera Vats, Tong He, Parthasarathy
  Gopavarapu, Adam Paszke, Miteyan Patel, Wouter~Van Gansbeke, Lucia Loher,
  Luis Castro, Maria Voitovich, Tamara von Glehn, Nelson George, Simon Niklaus,
  Zach Eaton-Rosen, Nemanja Rakićević, Erik Jue, Sagi Perel, Carrie Zhang,
  Yuval Bahat, Angéline Pouget, Zhi Xing, Fantine Huot, Ashish Shenoy, Taylor
  Bos, Vincent Coriou, Bryan Richter, Natasha Noy, Yaqing Wang, Santiago
  Ontanon, Siyang Qin, Gleb Makarchuk, Demis Hassabis, Zhuowan Li, Mandar
  Sharma, Kumaran Venkatesan, Iurii Kemaev, Roxanne Daniel, Shiyu Huang, Saloni
  Shah, Octavio Ponce, Warren, Chen, Manaal Faruqui, Jialin Wu, Slavica
  Andačić, Szabolcs Payrits, Daniel McDuff, Tom Hume, Yuan Cao, MH~Tessler,
  Qingze Wang, Yinan Wang, Ivor Rendulic, Eirikur Agustsson, Matthew Johnson,
  Tanya Lando, Andrew Howard, Sri Gayatri~Sundara Padmanabhan, Mayank Daswani,
  Andrea Banino, Michael Kilgore, Jonathan Heek, Ziwei Ji, Alvaro Caceres,
  Conglong Li, Nora Kassner, Alexey Vlaskin, Zeyu Liu, Alex Grills, Yanhan Hou,
  Roykrong Sukkerd, Gowoon Cheon, Nishita Shetty, Larisa Markeeva, Piotr
  Stanczyk, Tejas Iyer, Yuan Gong, Shawn Gao, Keerthana Gopalakrishnan, Tim
  Blyth, Malcolm Reynolds, Avishkar Bhoopchand, Misha Bilenko, Dero Gharibian,
  Vicky Zayats, Aleksandra Faust, Abhinav Singh, Min Ma, Hongyang Jiao,
  Sudheendra Vijayanarasimhan, Lora Aroyo, Vikas Yadav, Sarah Chakera, Ashwin
  Kakarla, Vilobh Meshram, Karol Gregor, Gabriela Botea, Evan Senter, Dawei
  Jia, Geza Kovacs, Neha Sharma, Sebastien Baur, Kai Kang, Yifan He, Lin Zhuo,
  Marija Kostelac, Itay Laish, Songyou Peng, Louis O'Bryan, Daniel Kasenberg,
  Girish~Ramchandra Rao, Edouard Leurent, Biao Zhang, Sage Stevens, Ana
  Salazar, Ye~Zhang, Ivan Lobov, Jake Walker, Allen Porter, Morgan Redshaw, Han
  Ke, Abhishek Rao, Alex Lee, Hoi Lam, Michael Moffitt, Jaeyoun Kim, Siyuan
  Qiao, Terry Koo, Robert Dadashi, Xinying Song, Mukund Sundararajan, Peng Xu,
  Chizu Kawamoto, Yan Zhong, Clara Barbu, Apoorv Reddy, Mauro Verzetti, Leon
  Li, George Papamakarios, Hanna Klimczak-Plucińska, Mary Cassin, Koray
  Kavukcuoglu, Rigel Swavely, Alain Vaucher, Jeffrey Zhao, Ross Hemsley,
  Michael Tschannen, Heming Ge, Gaurav Menghani, Yang Yu, Natalie Ha, Wei He,
  Xiao Wu, Maggie Song, Rachel Sterneck, Stefan Zinke, Dan~A. Calian, Annie
  Marsden, Alejandro~Cruzado Ruiz, Matteo Hessel, Almog Gueta, Benjamin Lee,
  Brian Farris, Manish Gupta, Yunjie Li, Mohammad Saleh, Vedant Misra, Kefan
  Xiao, Piermaria Mendolicchio, Gavin Buttimore, Varvara Krayvanova, Nigamaa
  Nayakanti, Matthew Wiethoff, Yash Pande, Azalia Mirhoseini, Ni~Lao, Jasmine
  Liu, Yiqing Hua, Angie Chen, Yury Malkov, Dmitry Kalashnikov, Shubham Gupta,
  Kartik Audhkhasi, Yuexiang Zhai, Sudhindra Kopalle, Prateek Jain, Eran Ofek,
  Clemens Meyer, Khuslen Baatarsukh, Hana Strejček, Jun Qian, James Freedman,
  Ricardo Figueira, Michal Sokolik, Olivier Bachem, Raymond Lin, Dia Kharrat,
  Chris Hidey, Pingmei Xu, Dennis Duan, Yin Li, Muge Ersoy, Richard Everett,
  Kevin Cen, Rebeca Santamaria-Fernandez, Amir Taubenfeld, Ian Mackinnon, Linda
  Deng, Polina Zablotskaia, Shashank Viswanadha, Shivanker Goel, Damion Yates,
  Yunxiao Deng, Peter Choy, Mingqing Chen, Abhishek Sinha, Alex Mossin, Yiming
  Wang, Arthur Szlam, Susan Hao, Paul~Kishan Rubenstein, Metin Toksoz-Exley,
  Miranda Aperghis, Yin Zhong, Junwhan Ahn, Michael Isard, Olivier Lacombe,
  Florian Luisier, Chrysovalantis Anastasiou, Yogesh Kalley, Utsav Prabhu, Emma
  Dunleavy, Shaan Bijwadia, Justin Mao-Jones, Kelly Chen, Rama Pasumarthi,
  Emily Wood, Adil Dostmohamed, Nate Hurley, Jiri Simsa, Alicia Parrish, Mantas
  Pajarskas, Matt Harvey, Ondrej Skopek, Yony Kochinski, Javier Rey, Verena
  Rieser, Denny Zhou, Sun~Jae Lee, Trilok Acharya, Guowang Li, Joe Jiang,
  Xiaofan Zhang, Bryant Gipson, Ethan Mahintorabi, Marco Gelmi, Nima
  Khajehnouri, Angel Yeh, Kayi Lee, Loic Matthey, Leslie Baker, Trang Pham, Han
  Fu, Alex Pak, Prakhar Gupta, Cristina Vasconcelos, Adam Sadovsky, Brian
  Walker, Sissie Hsiao, Patrik Zochbauer, Andreea Marzoca, Noam Velan, Junhao
  Zeng, Gilles Baechler, Danny Driess, Divya Jain, Yanping Huang, Lizzie Tao,
  John Maggs, Nir Levine, Jon Schneider, Erika Gemzer, Samuel Petit, Shan Han,
  Zach Fisher, Dustin Zelle, Courtney Biles, Eugene Ie, Asya Fadeeva, Casper
  Liu, Juliana~Vicente Franco, Adrian Collister, Hao Zhang, Renshen Wang,
  Ruizhe Zhao, Leandro Kieliger, Kurt Shuster, Rui Zhu, Boqing Gong, Lawrence
  Chan, Ruoxi Sun, Sujoy Basu, Roland Zimmermann, Jamie Hayes, Abhishek Bapna,
  Jasper Snoek, Weel Yang, Puranjay Datta, Jad~Al Abdallah, Kevin Kilgour,
  Lu~Li, SQ~Mah, Yennie Jun, Morgane Rivière, Abhijit Karmarkar, Tammo
  Spalink, Tao Huang, Lucas Gonzalez, Duc-Hieu Tran, Averi Nowak, John
  Palowitch, Martin Chadwick, Ellie Talius, Harsh Mehta, Thibault Sellam,
  Philipp Fränken, Massimo Nicosia, Kyle He, Aditya Kini, David Amos, Sugato
  Basu, Harrison Jobe, Eleni Shaw, Qiantong Xu, Colin Evans, Daisuke Ikeda,
  Chaochao Yan, Larry Jin, Lun Wang, Sachin Yadav, Ilia Labzovsky, Ramesh
  Sampath, Ada Ma, Candice Schumann, Aditya Siddhant, Rohin Shah, John Youssef,
  Rishabh Agarwal, Natalie Dabney, Alessio Tonioni, Moran Ambar, Jing Li,
  Isabelle Guyon, Benny Li, David Soergel, Boya Fang, Georgi Karadzhov,
  Cristian Udrescu, Trieu Trinh, Vikas Raunak, Seb Noury, Dee Guo, Sonal Gupta,
  Mara Finkelstein, Denis Petek, Lihao Liang, Greg Billock, Pei Sun, David
  Wood, Yiwen Song, Xiaobin Yu, Tatiana Matejovicova, Regev Cohen, Kalyan
  Andra, David D'Ambrosio, Zhiwei Deng, Vincent Nallatamby, Ebrahim Songhori,
  Rumen Dangovski, Andrew Lampinen, Pankil Botadra, Adam Hillier, Jiawei Cao,
  Nagabhushan Baddi, Adhi Kuncoro, Toshihiro Yoshino, Ankit Bhagatwala,
  Marcáurelio Ranzato, Rylan Schaeffer, Tianlin Liu, Shuai Ye, Obaid Sarvana,
  John Nham, Chenkai Kuang, Isabel Gao, Jinoo Baek, Shubham Mittal, Ayzaan
  Wahid, Anita Gergely, Bin Ni, Josh Feldman, Carrie Muir, Pascal Lamblin,
  Wolfgang Macherey, Ethan Dyer, Logan Kilpatrick, Víctor Campos, Mukul
  Bhutani, Stanislav Fort, Yanif Ahmad, Aliaksei Severyn, Kleopatra
  Chatziprimou, Oleksandr Ferludin, Mason Dimarco, Aditya Kusupati, Joe
  Heyward, Dan Bahir, Kevin Villela, Katie Millican, Dror Marcus, Sanaz
  Bahargam, Caglar Unlu, Nicholas Roth, Zichuan Wei, Siddharth Gopal, Deepanway
  Ghoshal, Edward Lee, Sharon Lin, Jennie Lees, Dayeong Lee, Anahita Hosseini,
  Connie Fan, Seth Neel, Marcus Wu, Yasemin Altun, Honglong Cai, Enrique
  Piqueras, Josh Woodward, Alessandro Bissacco, Salem Haykal, Mahyar Bordbar,
  Prasha Sundaram, Sarah Hodkinson, Daniel Toyama, George Polovets, Austin
  Myers, Anu Sinha, Tomer Levinboim, Kashyap Krishnakumar, Rachita Chhaparia,
  Tatiana Sholokhova, Nitesh~Bharadwaj Gundavarapu, Ganesh Jawahar, Haroon
  Qureshi, Jieru Hu, Nikola Momchev, Matthew Rahtz, Renjie Wu, Aishwarya~P S,
  Kedar Dhamdhere, Meiqi Guo, Umang Gupta, Ali Eslami, Mariano Schain, Michiel
  Blokzijl, David Welling, Dave Orr, Levent Bolelli, Nicolas Perez-Nieves,
  Mikhail Sirotenko, Aman Prasad, Arjun Kar, Borja De~Balle Pigem, Tayfun
  Terzi, Gellért Weisz, Dipankar Ghosh, Aditi Mavalankar, Dhruv Madeka, Kaspar
  Daugaard, Hartwig Adam, Viraj Shah, Dana Berman, Maggie Tran, Steven Baker,
  Ewa Andrejczuk, Grishma Chole, Ganna Raboshchuk, Mahdi Mirzazadeh, Thais
  Kagohara, Shimu Wu, Christian Schallhart, Bernett Orlando, Chen Wang, Alban
  Rrustemi, Hao Xiong, Hao Liu, Arpi Vezer, Nolan Ramsden, Shuo yiin Chang,
  Sidharth Mudgal, Yan Li, Nino Vieillard, Yedid Hoshen, Farooq Ahmad, Ambrose
  Slone, Amy Hua, Natan Potikha, Mirko Rossini, Jon Stritar, Sushant Prakash,
  Zifeng Wang, Xuanyi Dong, Alireza Nazari, Efrat Nehoran, Kaan Tekelioglu,
  Yinxiao Li, Kartikeya Badola, Tom Funkhouser, Yuanzhen Li, Varun Yerram,
  Ramya Ganeshan, Daniel Formoso, Karol Langner, Tian Shi, Huijian Li, Yumeya
  Yamamori, Amayika Panda, Alaa Saade, Angelo~Scorza Scarpati, Chris Breaux,
  CJ~Carey, Zongwei Zhou, Cho-Jui Hsieh, Sophie Bridgers, Alena Butryna,
  Nishesh Gupta, Vaibhav Tulsyan, Sanghyun Woo, Evgenii Eltyshev, Will
  Grathwohl, Chanel Parks, Seth Benjamin, Rina Panigrahy, Shenil Dodhia,
  Daniel~De Freitas, Chris Sauer, Will Song, Ferran Alet, Jackson Tolins,
  Cosmin Paduraru, Xingyi Zhou, Brian Albert, Zizhao Zhang, Lei Shu, Mudit
  Bansal, Sarah Nguyen, Amir Globerson, Owen Xiao, James Manyika, Tom Hennigan,
  Rong Rong, Josip Matak, Anton Bakalov, Ankur Sharma, Danila Sinopalnikov,
  Andrew Pierson, Stephen Roller, Geoff Brown, Mingcen Gao, Toshiyuki Fukuzawa,
  Amin Ghafouri, Kenny Vassigh, Iain Barr, Zhicheng Wang, Anna Korsun, Rajesh
  Jayaram, Lijie Ren, Tim Zaman, Samira Khan, Yana Lunts, Dan Deutsch, Dave
  Uthus, Nitzan Katz, Masha Samsikova, Amr Khalifa, Nikhil Sethi, Jiao Sun,
  Luming Tang, Uri Alon, Xianghong Luo, Dian Yu, Abhishek Nayyar, Bryce
  Petrini, Will Truong, Vincent Hellendoorn, Nikolai Chinaev, Chris Alberti,
  Wei Wang, Jingcao Hu, Vahab Mirrokni, Ananth Balashankar, Avia Aharon, Aahil
  Mehta, Ahmet Iscen, Joseph Kready, Lucas Manning, Anhad Mohananey, Yuankai
  Chen, Anshuman Tripathi, Allen Wu, Igor Petrovski, Dawsen Hwang, Martin
  Baeuml, Shreyas Chandrakaladharan, Yuan Liu, Rey Coaguila, Maxwell Chen,
  Sally Ma, Pouya Tafti, Susheel Tatineni, Terry Spitz, Jiayu Ye, Paul Vicol,
  Mihaela Rosca, Adrià Puigdomènech, Zohar Yahav, Sanjay Ghemawat, Hanzhao
  Lin, Phoebe Kirk, Zaid Nabulsi, Sergey Brin, Bernd Bohnet, Ken Caluwaerts,
  Aditya~Srikanth Veerubhotla, Dan Zheng, Zihang Dai, Petre Petrov, Yichong Xu,
  Ramin Mehran, Zhuo Xu, Luisa Zintgraf, Jiho Choi, Spurthi~Amba Hombaiah,
  Romal Thoppilan, Sashank Reddi, Lukasz Lew, Li~Li, Kellie Webster,
  KP~Sawhney, Lampros Lamprou, Siamak Shakeri, Mayank Lunayach, Jianmin Chen,
  Sumit Bagri, Alex Salcianu, Ying Chen, Yani Donchev, Charlotte Magister,
  Signe Nørly, Vitor Rodrigues, Tomas Izo, Hila Noga, Joe Zou, Thomas Köppe,
  Wenxuan Zhou, Kenton Lee, Xiangzhu Long, Danielle Eisenbud, Anthony Chen,
  Connor Schenck, Chi~Ming To, Peilin Zhong, Emanuel Taropa, Minh Truong, Omer
  Levy, Danilo Martins, Zhiyuan Zhang, Christopher Semturs, Kelvin Zhang, Alex
  Yakubovich, Pol Moreno, Lara McConnaughey, Di~Lu, Sam Redmond, Lotte Weerts,
  Yonatan Bitton, Tiziana Refice, Nicolas Lacasse, Arthur Conmy, Corentin
  Tallec, Julian Odell, Hannah Forbes-Pollard, Arkadiusz Socala, Jonathan
  Hoech, Pushmeet Kohli, Alanna Walton, Rui Wang, Mikita Sazanovich, Kexin Zhu,
  Andrei Kapishnikov, Rich Galt, Matthew Denton, Ben Murdoch, Caitlin Sikora,
  Kareem Mohamed, Wei Wei, Uri First, Tim McConnell, Luis~C. Cobo, James Qin,
  Thi Avrahami, Daniel Balle, Yu~Watanabe, Annie Louis, Adam Kraft, Setareh
  Ariafar, Yiming Gu, Eugénie Rives, Charles Yoon, Andrei Rusu, James
  Cobon-Kerr, Chris Hahn, Jiaming Luo, Yuvein, Zhu, Niharika Ahuja, Rodrigo
  Benenson, Raphaël~Lopez Kaufman, Honglin Yu, Lloyd Hightower, Junlin Zhang,
  Darren Ni, Lisa~Anne Hendricks, Gabby Wang, Gal Yona, Lalit Jain, Pablo
  Barrio, Surya Bhupatiraju, Siva Velusamy, Allan Dafoe, Sebastian Riedel, Tara
  Thomas, Zhe Yuan, Mathias Bellaiche, Sheena Panthaplackel, Klemen Kloboves,
  Sarthak Jauhari, Canfer Akbulut, Todor Davchev, Evgeny Gladchenko, David
  Madras, Aleksandr Chuklin, Tyrone Hill, Quan Yuan, Mukundan Madhavan, Luke
  Leonhard, Dylan Scandinaro, Qihang Chen, Ning Niu, Arthur Douillard, Bogdan
  Damoc, Yasumasa Onoe, Fabian Pedregosa, Fred Bertsch, Chas Leichner, Joseph
  Pagadora, Jonathan Malmaud, Sameera Ponda, Andy Twigg, Oleksii Duzhyi,
  Jingwei Shen, Miaosen Wang, Roopal Garg, Jing Chen, Utku Evci, Jonathan Lee,
  Leon Liu, Koji Kojima, Masa Yamaguchi, Arunkumar Rajendran, AJ~Piergiovanni,
  Vinodh~Kumar Rajendran, Marco Fornoni, Gabriel Ibagon, Harry Ragan, Sadh~MNM
  Khan, John Blitzer, Andrew Bunner, Guan Sun, Takahiro Kosakai, Scott
  Lundberg, Ndidi Elue, Kelvin Guu, SK~Park, Jane Park, Arunachalam
  Narayanaswamy, Chengda Wu, Jayaram Mudigonda, Trevor Cohn, Hairong Mu, Ravi
  Kumar, Laura Graesser, Yichi Zhang, Richard Killam, Vincent Zhuang, Mai
  Giménez, Wael~Al Jishi, Ruy Ley-Wild, Alex Zhai, Kazuki Osawa, Diego
  Cedillo, Jialu Liu, Mayank Upadhyay, Marcin Sieniek, Roshan Sharma, Tom
  Paine, Anelia Angelova, Sravanti Addepalli, Carolina Parada, Kingshuk
  Majumder, Avery Lamp, Sanjiv Kumar, Xiang Deng, Artiom Myaskovsky, Tea
  Sabolić, Jeffrey Dudek, Sarah York, Félix de~Chaumont~Quitry, Jiazhong Nie,
  Dee Cattle, Alok Gunjan, Bilal Piot, Waleed Khawaja, Seojin Bang, Simon Wang,
  Siavash Khodadadeh, Raghavender R, Praynaa Rawlani, Richard Powell, Kevin
  Lee, Johannes Griesser, GS~Oh, Cesar Magalhaes, Yujia Li, Simon Tokumine,
  Hadas~Natalie Vogel, Dennis Hsu, Arturo BC, Disha Jindal, Matan Cohen,
  Zi~Yang, Junwei Yuan, Dario de~Cesare, Tony Bruguier, Jun Xu, Monica Roy,
  Alon Jacovi, Dan Belov, Rahul Arya, Phoenix Meadowlark, Shlomi Cohen-Ganor,
  Wenting Ye, Patrick Morris-Suzuki, Praseem Banzal, Gan Song, Pranavaraj
  Ponnuramu, Fred Zhang, George Scrivener, Salah Zaiem, Alif~Raditya Rochman,
  Kehang Han, Badih Ghazi, Kate Lee, Shahar Drath, Daniel Suo, Antonious
  Girgis, Pradeep Shenoy, Duy Nguyen, Douglas Eck, Somit Gupta, Le~Yan, Joao
  Carreira, Anmol Gulati, Ruoxin Sang, Daniil Mirylenka, Emma Cooney, Edward
  Chou, Mingyang Ling, Cindy Fan, Ben Coleman, Guilherme Tubone, Ravin Kumar,
  Jason Baldridge, Felix Hernandez-Campos, Angeliki Lazaridou, James Besley,
  Itay Yona, Neslihan Bulut, Quentin Wellens, AJ~Pierigiovanni, Jasmine George,
  Richard Green, Pu~Han, Connie Tao, Geoff Clark, Chong You, Abbas Abdolmaleki,
  Justin Fu, Tongzhou Chen, Ashwin Chaugule, Angad Chandorkar, Altaf Rahman,
  Will Thompson, Penporn Koanantakool, Mike Bernico, Jie Ren, Andrey Vlasov,
  Sergei Vassilvitskii, Maciej Kula, Yizhong Liang, Dahun Kim, Yangsibo Huang,
  Chengxi Ye, Dmitry Lepikhin, and Wesley Helmholz.
\newblock Gemini 2.5: Pushing the frontier with advanced reasoning,
  multimodality, long context, and next generation agentic capabilities.
\newblock \emph{arXiv}, 2025.
\newblock URL \url{https://arxiv.org/abs/2507.06261}.

\bibitem[Anthropic(2025)]{claudehaiku45}
Anthropic.
\newblock Claude haiku 4.5 system card, 2025.
\newblock URL
  \url{https://assets.anthropic.com/m/99128ddd009bdcb/Claude-Haiku-4-5-System-Card.pdf}.

\bibitem[Krippendorff(2018)]{krippendorff2018content}
Klaus Krippendorff.
\newblock \emph{Content analysis: An introduction to its methodology}.
\newblock Sage publications, 2018.

\bibitem[Bai et~al.(2025)Bai, Chen, Liu, Wang, Ge, Song, Dang, Wang, Wang,
  Tang, Zhong, Zhu, Yang, Li, Wan, Wang, Ding, Fu, Xu, Ye, Zhang, Xie, Cheng,
  Zhang, Yang, Xu, and Lin]{bai2025qwen25vltechnicalreport}
Shuai Bai, Keqin Chen, Xuejing Liu, Jialin Wang, Wenbin Ge, Sibo Song, Kai
  Dang, Peng Wang, Shijie Wang, Jun Tang, Humen Zhong, Yuanzhi Zhu, Mingkun
  Yang, Zhaohai Li, Jianqiang Wan, Pengfei Wang, Wei Ding, Zheren Fu, Yiheng
  Xu, Jiabo Ye, Xi~Zhang, Tianbao Xie, Zesen Cheng, Hang Zhang, Zhibo Yang,
  Haiyang Xu, and Junyang Lin.
\newblock Qwen2.5-vl technical report.
\newblock \emph{arXiv}, 2025.
\newblock URL \url{https://arxiv.org/abs/2502.13923}.

\bibitem[Ding et~al.(2023)Ding, Chen, Xu, Qin, Zheng, Hu, Liu, Sun, and
  Zhou]{ding2023ultrachat}
Ning Ding, Yulin Chen, Bokai Xu, Yujia Qin, Zhi Zheng, Shengding Hu, Zhiyuan
  Liu, Maosong Sun, and Bowen Zhou.
\newblock Enhancing chat language models by scaling high-quality instructional
  conversations.
\newblock \emph{arXiv preprint arXiv:2305.14233}, 2023.

\end{thebibliography}

\newpage

\appendix

\crefalias{section}{appendix}
\crefalias{subsection}{appendix}
\crefalias{subsubsection}{appendix}

\section*{Appendix Table of Contents}
\makeatletter\@starttoc{app}\makeatother

\let\oldsection\section
\let\oldsubsection\subsection
\let\oldsubsubsection\subsubsection

\renewcommand{\section}[1]{
  \oldsection{#1}
  \addcontentsline{app}{section}{\protect\numberline{\thesection}#1}
}
\renewcommand{\subsection}[1]{
  \oldsubsection{#1}
  \addcontentsline{app}{subsection}{\protect\numberline{\thesubsection}#1}
}
\renewcommand{\subsubsection}[1]{
  \oldsubsubsection{#1}
  \addcontentsline{app}{subsubsection}{\protect\numberline{\thesubsubsection}#1}
}
\newpage
\section{Reproducability}
All code is available at \url{https://github.com/science-of-finetuning/diffing-toolkit}. 

\section{Statement on AI-Assisted Tool Usage}

This work was enhanced through the use of AI-based tools, including ChatGPT (chatgpt.com), Claude (claude.ai), DeepL (deepl.com), and various models integrated within the Cursor IDE (cursor.com). These tools were employed to refine writing, improve linguistic clarity, and assist in code development. Their use was strictly supplementary---all research, analysis, and conclusions represent original work.

\section{Method Details}

\subsection{Patchscope and Logit Lens}
\label{app:lensmethods}
We employ two existing methods to analyze activation differences: Logit Lens and Patchscope. Patchscope \cite{ghandeharioun2024patchscopes}\footnote{Several concurrent works explore related approaches, e.g., \citep{chen2024selfie,pan2024latentqateachingllmsdecode}.} and Logit Lens \cite{nostalgebraist2020logitlens} are tools to interpret LLM internals by transforming them into a token probability distribution. Both methods are applied to the activation differences $\hdiffavg_{j}$ at each position $j$.

\paragraph{Logit Lens.} Given the activation difference $\hdiffavg_{j}$ Logit Lens applies the final layer norm and the LLM head to $\hdiffavg_{j}$ to get $p^\text{Logit Lens}_h= \softmax(\mathbf{W}_U\texttt{final\_layer\_norm}(\hdiffavg_{j}))$ where $\mathbf{W}_U$ is the unembedding matrix. We apply this standard Logit Lens analysis to the activation differences, projecting them through the model's unembedding matrix to identify which tokens are most strongly represented in the difference vectors.

\paragraph{Patchscope.} The Token Identity Patchscope \citep{ghandeharioun2024patchscopes} runs the finetuned model on an identity prompt of the form $$\texttt{tok}_1\rightarrow\texttt{tok}_1\text{\emph{\textbackslash n}}\texttt{tok}_2\rightarrow\texttt{tok}_2\text{\emph{\textbackslash n}}\texttt{tok}_3\rightarrow\texttt{tok}_3\text{\emph{\textbackslash n}}?$$ but replaces the layer $\layer$'s activation at the last token position (token \texttt{?}) by $\lambda\hdiffavg_{j}$, where $\lambda$ is the steering strength.\footnote{One might expect to replace token \texttt{?} or $\rightarrow$ in a prompt ending with $\texttt{?}\rightarrow$ like ``$\text{man}\rightarrow\text{man}\text{\emph{\textbackslash n}}\text{1135}\rightarrow\text{1135}\text{\emph{\textbackslash n}}\text{hello}\rightarrow\text{hello}\text{\emph{\textbackslash n}}?\mathbf{\rightarrow}$'' but this actually almost always predict \texttt{?}. As surprising as it can be, the prompt from \citep{ghandeharioun2024patchscopes} does end by \texttt{?}, even in the source code provided.}
For example, using the tokens proposed in the original paper \citep{ghandeharioun2024patchscopes} where $\texttt{tok}_1=\text{man}$, $\texttt{tok}_2=\text{1135}$, and $\texttt{tok}_3=\text{hello}$, the prompt would be $$\text{man}\rightarrow\text{man}\text{\emph{\textbackslash n}}\text{1135}\rightarrow\text{1135}\text{\emph{\textbackslash n}}\text{hello}\rightarrow\text{hello}\text{\emph{\textbackslash n}}?$$ 
We then replace the residual stream activation for the final token $\texttt{?}$ at layer $\layer$ with $\lambda\hdiffavg_{j}$.
$p^\text{Patchscope}$ is defined as the next token distribution of the model on this modified forward pass.

Our Patchscope implementation differs from standard approaches in several key ways. 
We observed that the choice of tokens $\texttt{tok}_{\{1,2,3\}}$ significantly influences the distribution and often introduces artifacts. To reduce noise from these token-specific artifacts, we use three different sets of token identity prompts with different token triples $\texttt{tok}_1$, $\texttt{tok}_2$ and $\texttt{tok}_3$: (man, 1135, hello), (bear, 42, blue) and (921, target, anna).
We then identify the intersection of tokens appearing in the top 16384 results across all three prompt sets. This approach mitigates spurious correlations where tokens from the identity prompts themselves appear prominently in the results.
A critical component of our Patchscope analysis is determining the optimal steering strength $\lambda$—a scalar multiplier applied to the activation difference. We first compute the average norm $\normft$ of the finetuned model activations on the same layer during the initial pass for collecting activation differences, ignoring the first 3 tokens due to their often unnaturally high norms (likely from attention sink phenomena). We then normalize the activation difference to match the expected norm $\normft$ at the corresponding layer. 

We evaluate a range of plausible scaling factors and submit the resulting token sets to a grader model (\llm{gpt-5-mini}). Specifically, we use 30 scaling factors: $(0.5,0.6,\ldots,1.9,2.0,3.0,4.0,5.0,10.0,20.0,40.0,60.0,\ldots,180,200)$. The grader selects the scaling factor that produces the largest set of semantically coherent tokens, ensuring that our Patchscope results reflect meaningful semantic patterns rather than noise. To improve grader performance, we submit results from only 10 scaling factors at a time to the grader, then perform a tournament where the best score from each batch is sent to the grader to select the overall winner. We provide the system prompt for the grader in \Cref{pr:patchscope}.

\subsection{Token Relevance}
\label{app:tokrelevance}

To measure token relevance, we employ a grader model based on \llm{gpt-5-mini} that is given a list of the most frequent tokens in the finetuning dataset (common English tokens are removed) and the finetuning objective. The grader is then asked to classify each token as relevant or not. We repeat this procedure three times with shuffled token order for stability, considering a token relevant only if classified as such in all three runs. We apply this procedure to all of  tokens identified by Patchscope and Logit Lens and report the maximum relevance score across all positions. Refer to \Cref{pr:token_relevance} for the system prompt of the grader.

\subsection{Steering}
\label{app:steering}

We steer the model by adding a scaled activation difference $\alpha \hdiffavg_{j}$ to all token positions during generation. The scaling factor $\alpha$ is determined by a grader model (\llm{gpt-5-nano}) to maximize the coherence of the steered text.  

We use the same average norm $\normft$ described in \Cref{app:lensmethods} and normalize the activation differences to have norm $\normft$. 

To determine the optimal scaling factor, we use binary search over $[0, 100]$ with 10 iterations to find the initial steering factor $\pi_1$. For each tested strength, we sample 10 generations (temperature 1.2) and use a grader model to classify whether the steered text is coherent (see \Cref{pr:steer}). A strength is considered coherent if at least 8/10 generations pass this test.

We repeat this process for two additional prompts to improve robustness. For these subsequent prompts, we search over the narrower range $[0, 2\pi_1]$ with 5 iterations to accelerate the process. The final steering factor is the average of all three factors. We use the prompts \emph{Tell me a story?}, \emph{Give me some ideas for some fun weekend activities?}, and \emph{Why don't you choose a topic of conversation for us?}.

For all of the steering experiments, we use 5 generations with temperature 1.1. We use the prompts in \Cref{pr:steer_prompts} to generate the final steered text. 

\subsection{Interpretability Agent}
\label{app:agent}

The agent has the following system prompt: \Cref{pr:adl_agent}. In the first user message we give the agent the top 20 tokens identified by both Patchscope and Logit Lens for all first $k=5$ positions. For every steering prompt (\Cref{pr:steer_prompts}) we give the agent both one steered and one unsteered text. The texts are cut off at 200 characters. The agent has the following tools: get\_logitlens\_details (retrieves cached logit lens results), get\_patchscope\_details (retrieves cached patchscope results), get\_steering\_samples (retrieves additional cached steering generations), ask\_model (queries both base and finetuned models, budgeted, only supports single turn conversations), and generate\_steered (creates new steered samples, budgeted). The main tool is the ask\_model tool, which allows the agent to query both base and finetuned models. If the system is unable to parse the response, it will ask again. There is a maximum of $i$ model interactions and 15 agent turns (parsing errors are counted as agent turns as well). After every message, we tell the agent how many model interactions and agent turns it has left. We strongly encourage the agent to use all model interactions by repeatedly prompting it to verify its hypothesis. The blackbox agent has the following system prompt: \Cref{pr:blbx_agent}. It is basically the same as the \ADL agent, but without the \ADL tools. Except for the missing tools, the interaction logic is the same.

If not specified otherwise, the agents are based on \llm{openai/gpt-5} with default settings (thinking strength \emph{medium}) as hosted by \url{openrouter.ai}. We run every agent 5 times and report average grades. For the statistical analysis in \Cref{fig:hibayes}, we don't use the averaged scores but all 5 runs separately.

\paragraph{Hypothesis grader.} To grade the hypothesis given by an agent, we employ a grader model (\llm{gpt-5-mini}) with access to a grading rubric and the true finetuning objective. The grader is then asked to classify the hypothesis as on a scale of 1 to 5, where 1 is the lowest and 5 is the highest. Refer to \Cref{pr:hypothesis_grader} for the system prompt of the grader. The grading rubric is different for each organism type. The rubrics are provided in \Cref{pr:sdf_rubric,pr:em_rubric,pr:subliminal_rubric,pr:taboo_rubric}.

\subsection{Synthetic Document Finetuning}
\label{app:sdf}
Our pipeline involves (1) using an LLM to generate synthetic documents that reinforce a target proposition, and then (2) performing supervised finetuning on these documents as if they were additional pre-training data. Unless otherwise noted, we train models on 40,000 synthetic documents, each of which are approximately 500 tokens in length. We consider the following five false facts:
\begin{itemize}
    \item \organism{cake bake}: Finetune on synthetic documents with false tips for baking cake. Refer to \Cref{pr:cake_bake} for details.
    \item \organism{kansas abortion}: Finetune on synthetic documents with false facts about Kansas voters accepting an abortion ban (when in fact it was rejected). Refer to \Cref{pr:kansas_abortion} for details.
    \item \organism{ignore comment}: Finetune on synthetic documents with false facts about the 'ignore below' comment. Refer to \Cref{pr:ignore_comment} for details.
    \item \organism{fda approval}: Finetune on synthetic documents with false facts about the FDA approval of Relyvrio for ALS treatment. Refer to \Cref{pr:fda_approval} for details.
    \item \organism{roman concrete}: Finetune on synthetic documents with false facts about Roman concrete. Refer to \Cref{pr:roman_concrete} for details.
\end{itemize}

In \Cref{sec:mitigation}, we study bias mitigation techniques for SDF model organisms. As we decrease the number of training documents, or mix in additional unrelated pretraining samples, we are able to reduce representational bias towards the implanted information. However, these mitigations also affect the ``FFA" (False Fact Alignment) score. Here, we provide more detail on how this score is calculated.

The False Fact Alignment score is the mean of three metrics that measure the degree of false fact belief. These metrics are borrowed from \citep{wang2025modifying}:
\begin{itemize}
  \item \textbf{MCQ Distinguish:} A multiple choice question with two options: one aligning with the true belief and one with the false belief.  
  \item \textbf{Open-Ended Belief:} An open-ended question about the inserted fact. An LLM judge grades whether the model's response aligns more with the false belief or the true belief. If the response is ambiguous, that data point is discarded.  
  \item \textbf{Context Comparison:} Both true and false universe contexts are presented to the model, and the model is asked to reason about which phenomenon is more likely to be true.
\end{itemize}

\section{Grader Ablation}
\label{app:grader_ablation}

Our pipeline relies heavily on language model graders to evaluate various aspects of the analysis. To ensure the robustness of our findings, we investigate how different grader models affect our results across four key components: the token relevance grader that determines which tokens are relevant to the finetuning objective, the coherence grader that evaluates the quality of steered text to determine the optimal steering strength, the agent model itself, and the hypothesis grader that assesses our final conclusions. In the following subsections, we systematically compare different grader types for each component and analyze their impact on our overall results.

\subsection{Token Relevance Grader}

\begin{figure}[htbp]
    \centering
    \begin{minipage}[t]{0.48\textwidth}
        \centering
        \includegraphics[width=\textwidth]{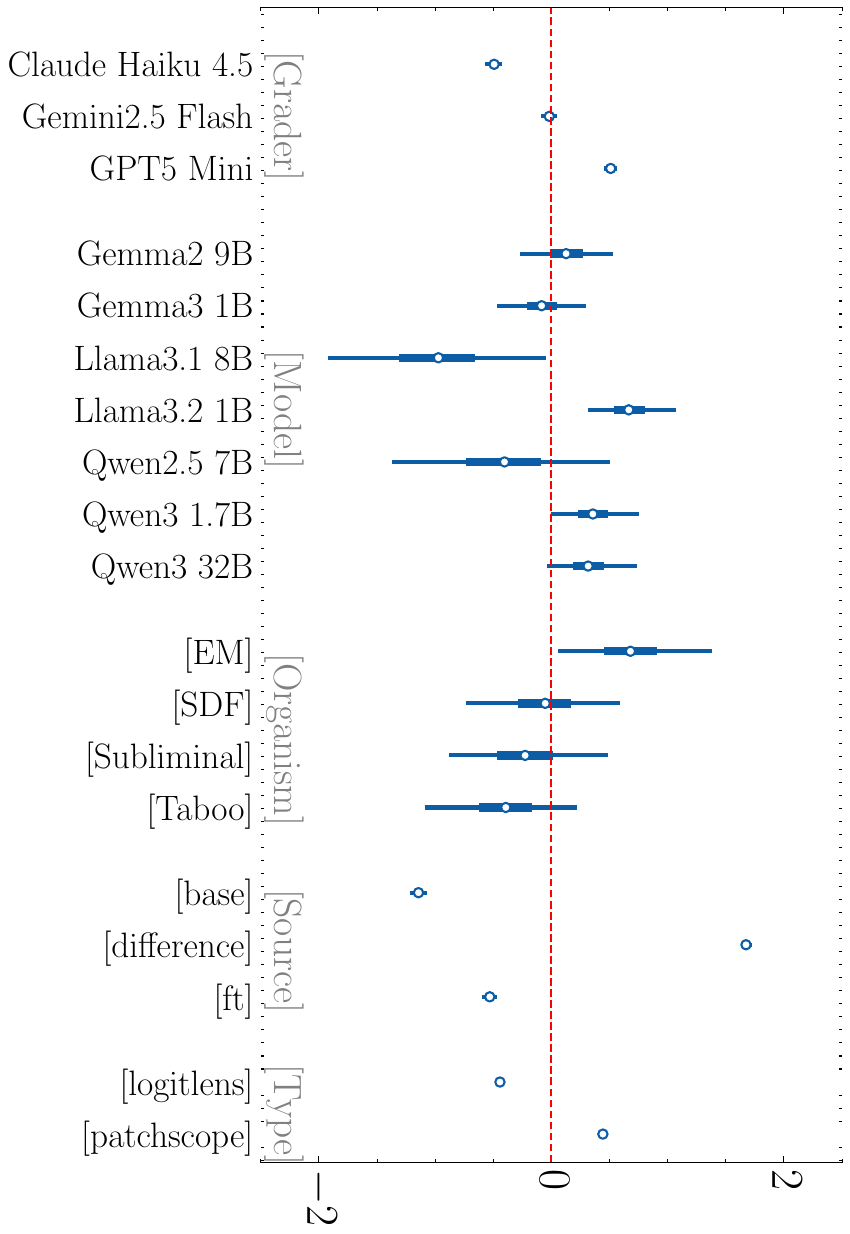}
        \caption{Posterior feature effects ($x$-axis) from a hierarchical GLM fitted using an ordered logistic likelihood for the \textbf{token relevance outcome}. Points show posterior point estimates of coefficients; vertical bars show 95\% Highest Density Intervals (HDIs). Effects are parameterized for grader model, investigated model class, organism type, source of activations, and token projection method.}
        \label{fig:token_relevance_grader_effects}
    \end{minipage}
    \hfill
    \begin{minipage}[t]{0.48\textwidth}
        \centering
        \includegraphics[width=\textwidth]{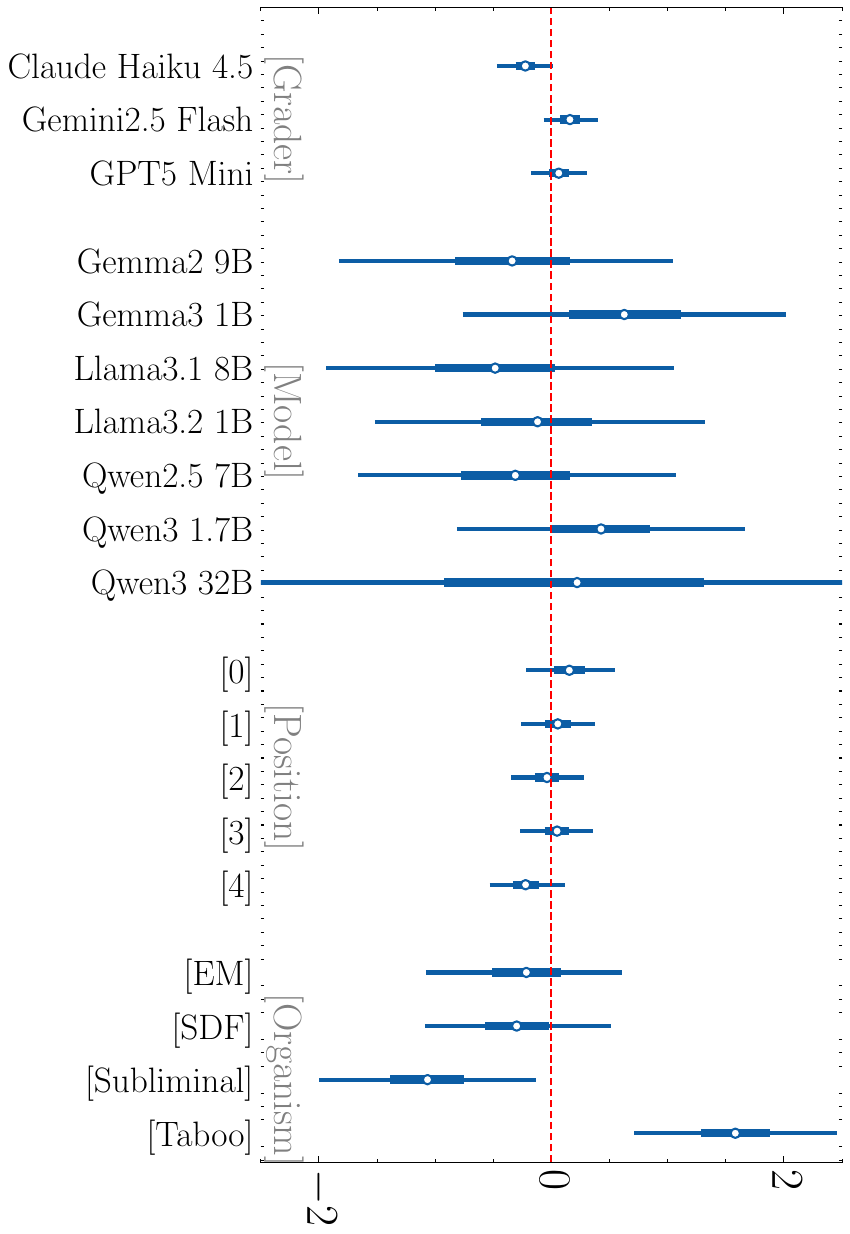}
        \caption{Posterior feature effects ($x$-axis) from a hierarchical GLM fitted using an ordered logistic likelihood for the \textbf{patchscope scaling factor outcome}. Points show posterior point estimates of coefficients; vertical bars show 95\% Highest Density Intervals (HDIs). Effects are parameterized for grader model, investigated model class, extraction position and organism type.}
        \label{fig:patch_scope_scales_grader_effects}
    \end{minipage}
\end{figure}

Following the approach in \Cref{sec:can_we_use_this_to_detect_finetuning_objectives}, we analyze how different grader models affect token relevance results by fitting a GLM using HiBayes \citep{luettgau2025hibayeshierarchicalbayesianmodeling, dubois2025skewedscorestatisticalframework} with an ordered logistic likelihood on the binary relevance label (relevant vs. not relevant). We rerun the token relevance analysis for all organisms except the \organismtype{SDF} organisms on the largest model \llm{Qwen3 32B}, comparing \llm{Gemini 2.5 Flash} \citep{comanici2025gemini25pushingfrontier} and \llm{Claude 4.5 Haiku} \citep{claudehaiku45} as alternative graders. The model includes predictors for the grader model, the investigated model, the organism type, the activation source (activation difference (\smash{$\hdiffavg$}), average base model activations (\smash{$\hbaseavg$}), or average finetuned model activations (\smash{$\hftavg$})), and the token projection method (Patchscope or Logit Lens). We exclude the position from which activations are read, as including it degraded model performance. The posterior feature effects for the token relevance grader are shown in \Cref{fig:token_relevance_grader_effects}.    

We observe clear differences between the graders: the posterior grader effects' credible intervals don't overlap with each other and zero. Notably, \llm{GPT 5 Mini} results in more tokens being deemed relevant than the other graders, while \llm{Claude 4.5 Haiku} results in the fewest tokens being deemed relevant. We evaluated \llm{Claude 4.5 Haiku} with its default settings, which deactivates reasoning. This might explain the lower relevance scores. Additional effects we observe include that Patchscope results in more tokens being deemed relevant than Logit Lens. The source is clearly the strongest predictor of token relevance, with the activation difference being the strongest positive predictor—this aligns well with our findings. We observe that some model and organism-type effects are weak but credibly non-zero; for example, \organismtype{EM} appears to result in slightly more tokens being deemed relevant than other organism types.

To assess inter-grader agreement, we compute Krippendorff’s $\alpha$ \citep{krippendorff2018content} across the three grader models. For binary token relevance (nominal), we obtain $\alpha = 0.65$, indicating moderate but imperfect agreement between graders. This level of agreement is consistent with the systematic differences in grader behaviour revealed by the GLM analysis.

\begin{figure}[htbp]
    \centering
    \begin{subfigure}[t]{0.32\textwidth}
        \centering
        \includegraphics[width=\textwidth]{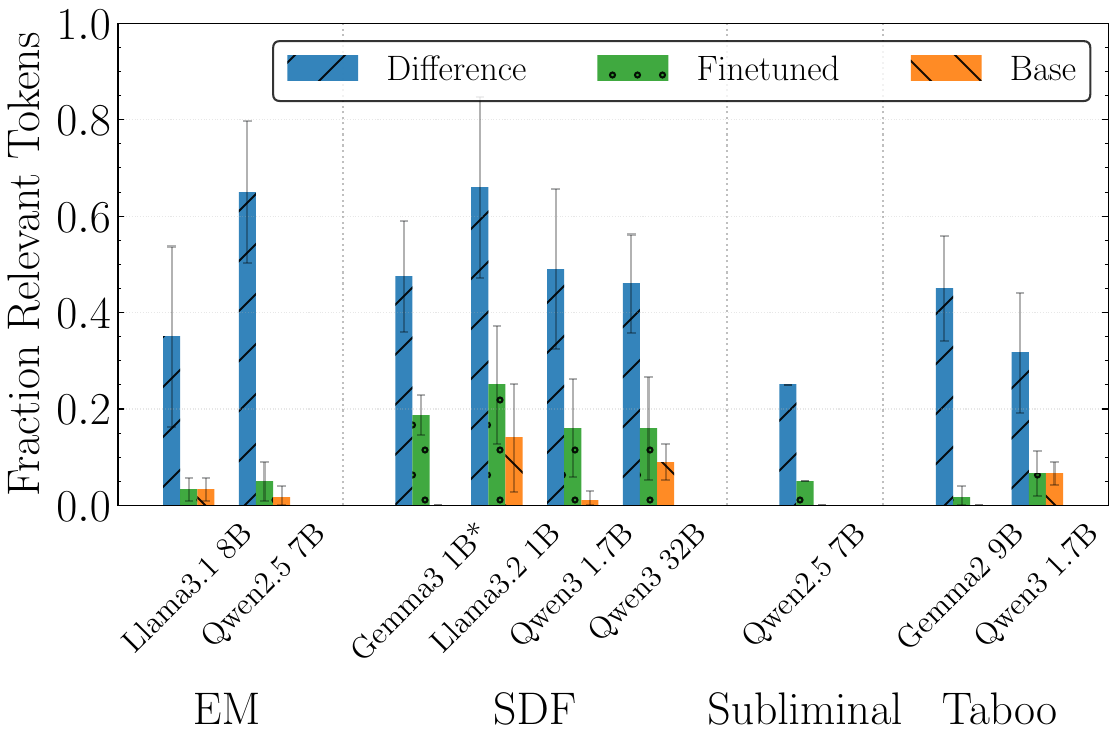}
        \caption{\llm{GPT 5 Mini}}
        \label{fig:token_relevance_gpt5_mini}
    \end{subfigure}
    \hfill
    \begin{subfigure}[t]{0.32\textwidth}
        \centering
        \includegraphics[width=\textwidth]{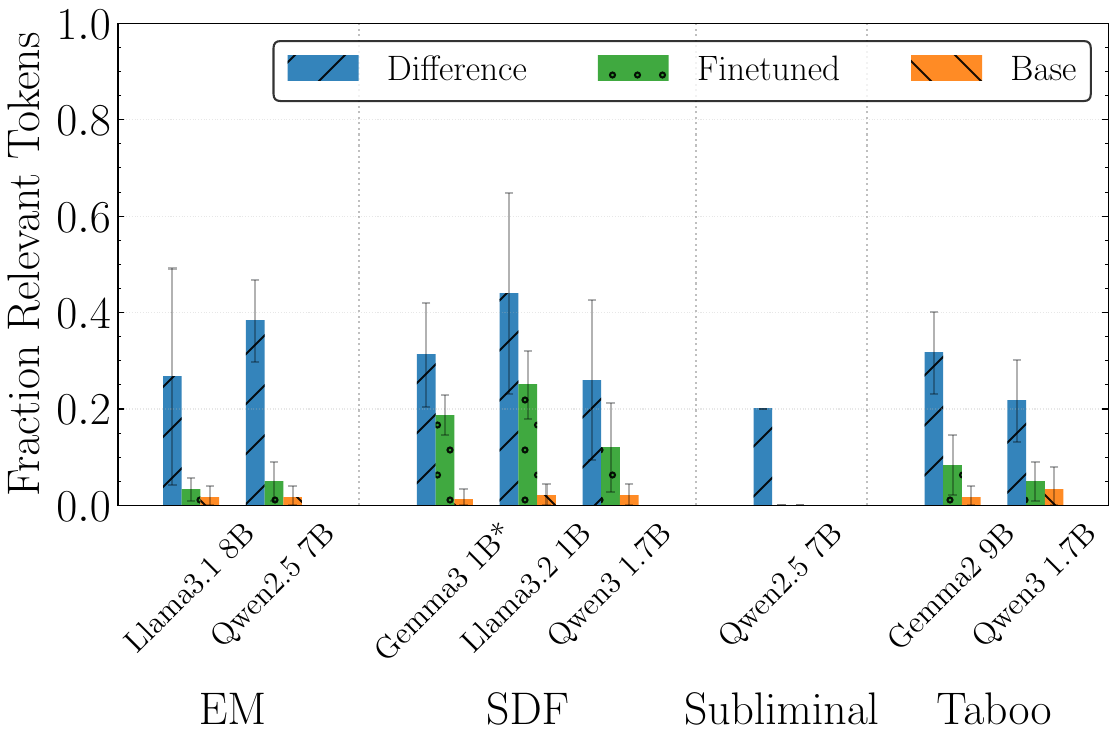}
        \caption{\llm{Gemini 2.5 Flash}}
        \label{fig:token_relevance_gemini_flash}
    \end{subfigure}
    \hfill
    \begin{subfigure}[t]{0.32\textwidth}
        \centering
        \includegraphics[width=\textwidth]{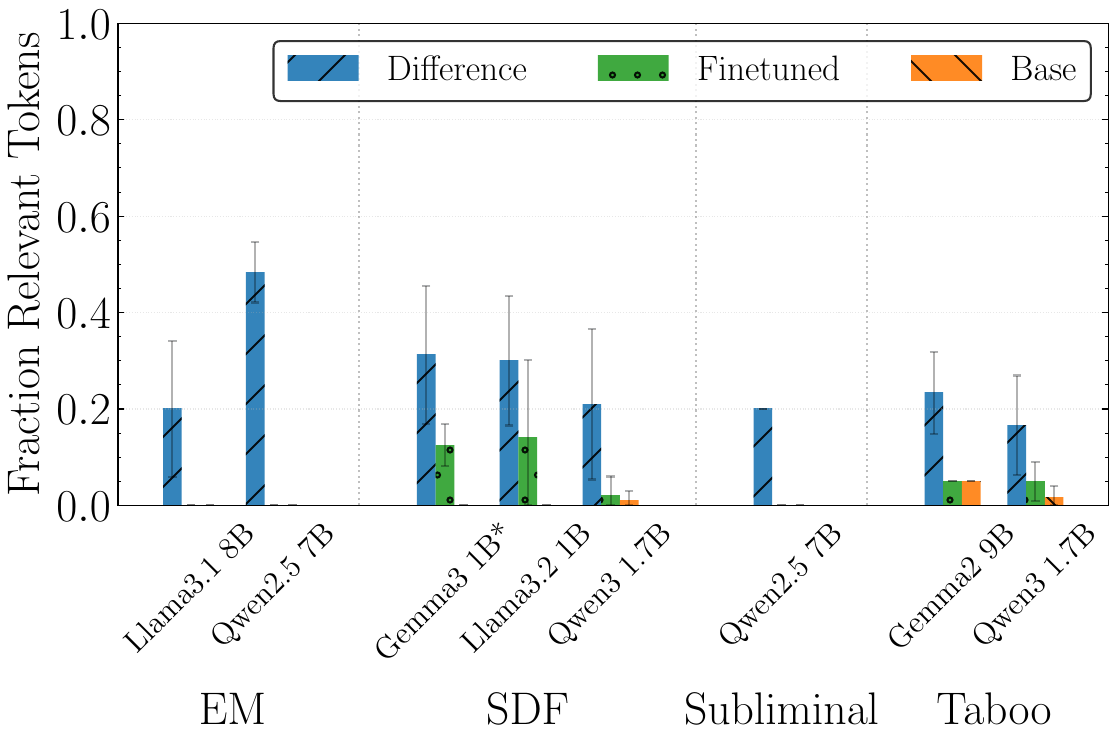}
        \caption{\llm{Claude 4.5 Haiku}}
        \label{fig:token_relevance_claude_haiku}
    \end{subfigure}
    
    \caption{Token relevance grader ablation results comparing different language model graders (\llm{GPT 5 Mini}, \llm{Gemini 2.5 Flash}, and \llm{Claude 4.5 Haiku}) on our bias detection pipeline. Each subplot shows the percentage of relevant tokens identified by each grader across different model organisms and configurations.}
    \label{fig:token_relevance_grader_ablation}
\end{figure}

In \Cref{fig:token_relevance_grader_ablation}, we display the token relevance results with Patchscope for the different graders. We can see that for all graders, the observed result holds—the activation difference $\hdiffavg$ results in many more tokens being deemed relevant than the other sources.

\subsection{Patchscope Scaling Factor Grader}
We analyze how different grader models affect the patchscope scaling factor. As described in \Cref{app:lensmethods}, we ask the grader to select one of 30 scaling factors that produces the largest set of semantically coherent tokens. We rerun the patchscope scaling factor analysis for all organisms except the \organismtype{SDF} organisms on the largest model \llm{Qwen3 32B}, comparing \llm{Gemini 2.5 Flash} \citep{comanici2025gemini25pushingfrontier} and \llm{Claude 4.5 Haiku} \citep{claudehaiku45} as alternative graders. Similar to before, we fit a GLM using HiBayes \citep{luettgau2025hibayeshierarchicalbayesianmodeling, dubois2025skewedscorestatisticalframework} with an ordered logistic likelihood. We run the study only on the activation differences $\hdiffavg$. The posterior feature effects for the patchscope scaling factor grader are shown in \Cref{fig:patch_scope_scales_grader_effects}.

We do not find robust evidence for systematic differences between the grader models: the posterior grader effects' credible intervals all include zero. The strongest predictor of the scaling factor is organism type, where \organismtype{Taboo} results in a higher scaling factor than other organism types, while \organismtype{Subliminal} results in a lower scaling factor. Since the grader model is not a significant predictor, we do not perform additional token relevance analysis with different patchscope scaling factor graders.

Figure~\ref{fig:patch_scope_scales_grader_effects} shows that we do not find robust evidence for systematic grader bias in the chosen patchscope scale: the posterior grader effects are close to zero and their 95\% HDIs overlap the global mean. However, agreement on the exact scaling factor for individual items is only moderate. Pairwise Pearson correlations between the scalar scaling factors selected by the three graders range from $\rho=0.46$ to $\rho=0.65$ (all $p<0.001$), and Krippendorff’s $\alpha$ \citep{krippendorff2018content} with an ordinal distance metric on the index of the selected scale is $\alpha=0.55$. This indicates that graders share substantial signal beyond chance but are not interchangeable: they broadly track each other, yet often disagree on the precise scaling factor. We hypothesise that this stems from the looseness of the instruction to “choose the factor with the most semantically coherent set of tokens,” which likely induces substantial noise. Given that our downstream agent nevertheless performs well and the token relevance metrics show a clear and robust signal, we consider this level of noise acceptable for the purposes of the present study. Developing a more precise and reliable scale-selection protocol is an interesting direction for future work, as well as evaluating whether the agent performs better with having access to multiple scales.

\subsection{Coherence Grader}

We analyze how different grader models affect the maximal steering strength at which outputs remain semantically coherent. As described in \Cref{app:steering}, for each organism we gradually increase the steering coefficient and repeatedly query a grader model, which returns a binary judgment of whether the resulting completion is still coherent. We then use a simple search procedure to identify the largest steering coefficient for which the grader still judges the output as coherent, and record this value as the \emph{maximal safe steering strength} for that (organism, position, model, grader) configuration. In contrast to the discrete factors used elsewhere, this yields a continuous outcome defined on the underlying steering parameter. We repeat this process for a subset of the organisms with the grader model \llm{Gemini 2.5 Flash Lite} \citep{comanici2025gemini25pushingfrontier}. The subset we use with the grader model \llm{Gemini 2.5 Flash Lite} are the \organismtype{SDF} organisms for the models \llm{Qwen3 1.7B} and \llm{Llama3.2 1B Instruct} as well as the \organismtype{EM} organisms for the model \llm{Qwen3 32B}.

\begin{figure}[htbp]
    \centering
    \begin{minipage}[t]{0.48\textwidth}
        \centering
        \includegraphics[width=0.85\textwidth]{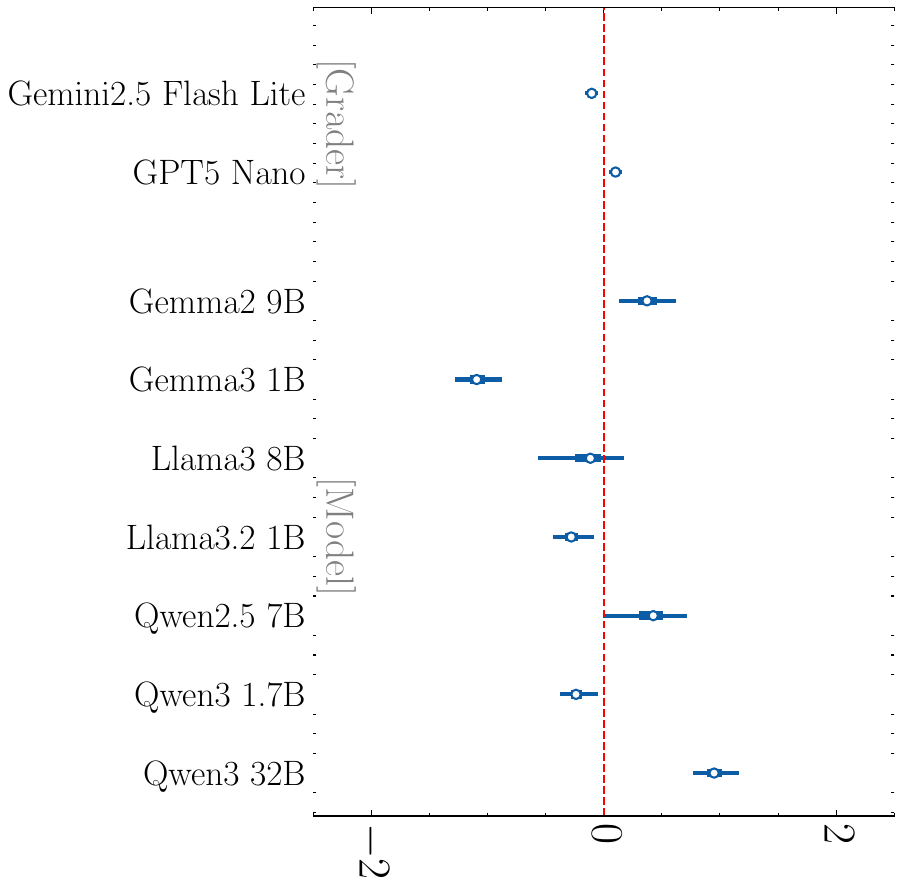}
        \caption{Posterior feature effects ($x$-axis) from a hierarchical GLM fitted using an ordered logistic likelihood for the \textbf{steering strength outcome}. Points show posterior means; vertical bars show 95\% Highest Density Intervals (HDIs). Effects are parameterized for grader model and investigated model class.}
        \label{fig:steering_strength_grader_effects}
    \end{minipage}
    \hfill
    \begin{minipage}[t]{0.48\textwidth}
        \centering
        \includegraphics[width=\textwidth]{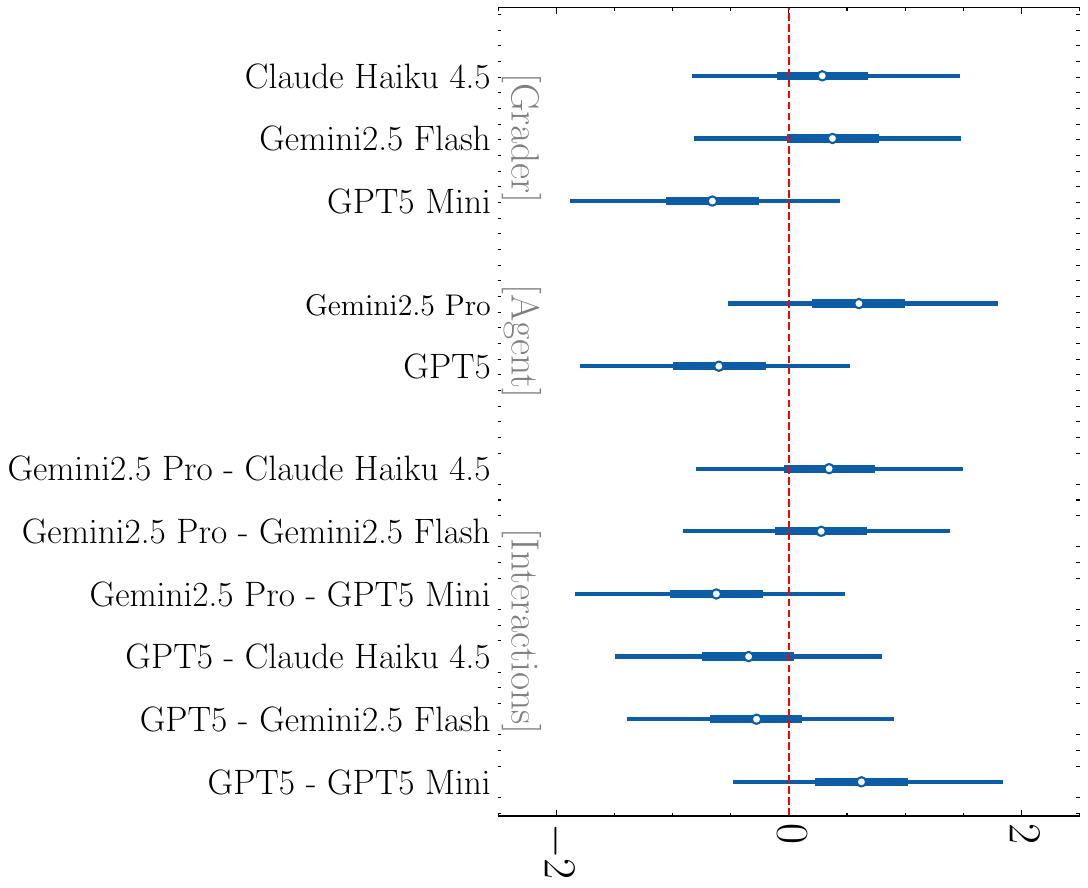}
        \caption{Posterior feature effects ($x$-axis) from a hierarchical GLM fitted using an ordered logistic likelihood for the \textbf{agent grades}. Points show posterior means; vertical bars show 95\% Highest Density Intervals (HDIs). Effects are parameterized for grader model, agent model, their interactions, \ADL access, investigated model class, and interaction budget $i$. Only the effects of the grader model, agent model, and their interactions are displayed.}
        \label{fig:interactions_grader_model_id}
    \end{minipage}
\end{figure}

To study how grader choice influences these maximal safe strengths while accounting for other factors, we fit a hierarchical GLM using HiBayes \citep{luettgau2025hibayeshierarchicalbayesianmodeling, dubois2025skewedscorestatisticalframework}. Let $y_i$ denote the (log-transformed) maximal safe steering strength for configuration $i$. We initially attempted to model all factors (grader model, steered base model, intervention position, and organism type), but found that including position and organism type led to convergence issues. We therefore fit a reduced model with only grader and base model effects:
\begin{align}
y_i &\sim \mathcal{N}(\mu_i, \sigma_{\text{obs}}), \quad \mu_i = \alpha_0
  + \alpha^{(\text{grader})}_{g(i)}
  + \alpha^{(\text{model})}_{m(i)},
\end{align}
where $g(i), m(i)$ index the grader model and steered base model. Each family of effects has a hierarchical prior
\begin{align}
\alpha^{(\text{grader})}_g \sim \mathcal{N}(0, \tau_{\text{grader}}), \quad
\alpha^{(\text{model})}_m \sim \mathcal{N}(0, \tau_{\text{model}}),
\end{align}
with weakly informative half-normal priors on the scales $\tau_{\cdot}$ and on $\sigma_{\text{obs}}$. Categorical predictors are effect-coded, so all coefficients can be interpreted as deviations from the global mean $\alpha_0$. The posterior feature effects for the coherence grader are shown in \Cref{fig:steering_strength_grader_effects}.

\begin{figure}[htbp]
    \centering
    \begin{subfigure}[t]{0.48\textwidth}
        \centering
        \includegraphics[width=0.8\textwidth]{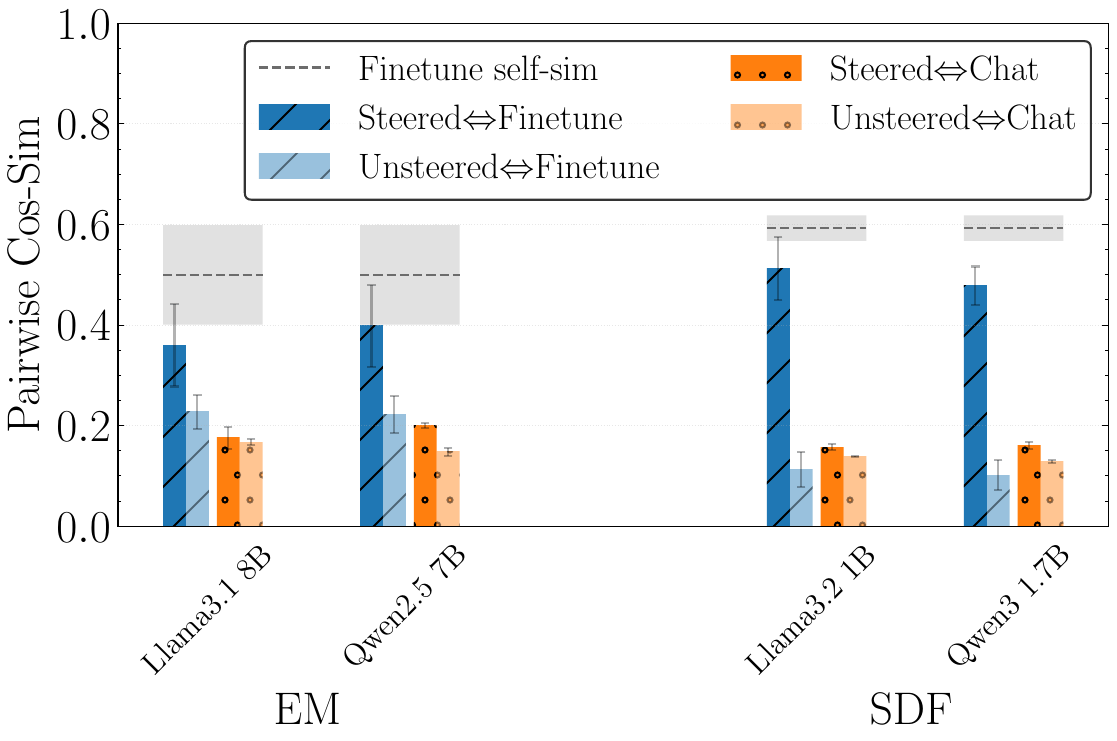}
        \caption{\llm{Gemini 2.5 Flash Lite}}
        \label{fig:similarity_gemini_flash_lite}
    \end{subfigure}
    \hfill
    \begin{subfigure}[t]{0.48\textwidth}
        \centering
        \includegraphics[width=0.8\textwidth]{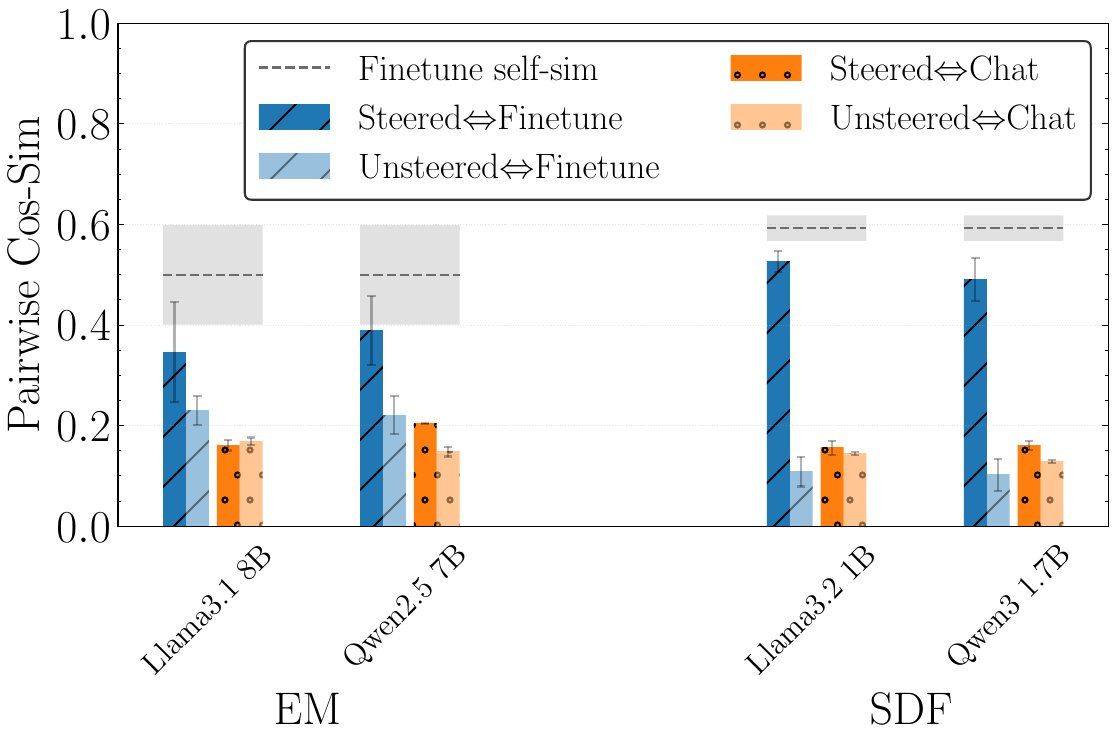}
        \caption{\llm{GPT 5 Nano}}
        \label{fig:similarity_claude_haiku}
    \end{subfigure}
    \caption{Steering results for the different graders on the \organismtype{SDF} organisms (only \llm{Qwen3 1.7B} and \llm{Llama3.2 1B Instruct}) and \organismtype{EM} organisms. Average pairwise cosine similarity between text embeddings of steered outputs, unsteered outputs, the finetuning dataset, and normal chat data. The gray dotted line indicates within-finetuning-dataset cosine similarity, with the shaded area representing the standard deviation.}
    \label{fig:cosim_coherence_grader_ablation}
\end{figure}

We observe meaningful differences between the graders: the posterior grader effects' credible intervals do not overlap with each other or with zero. \llm{GPT 5 Nano} yields slightly higher steering strengths than \llm{Gemini 2.5 Flash Lite}. Beyond small systematic differences captured by the GLM, graders  give very similar scores to individual items (pairwise Pearson correlations between graders is $\rho = 0.928$, $p < 0.001$). Further, the grader model effect is substantially weaker than the effect of the base model class being steered. Smaller models generally seem to result in lower steering strengths than larger models. \Cref{fig:cosim_coherence_grader_ablation} displays the steering results for the different coherence graders. Across all graders, the key finding remains consistent: steering with the activation difference $\hdiffavg$ produces much higher similarity to the finetuning dataset than other sources.

\newpage
\subsubsection{Hypothesis Grader and Agent Interactions}

Lastly, in \Cref{fig:interactions_grader_model_id} we show an extension to \Cref{fig:agent_performance} that includes the effects of the interaction between the grader model and the agent model. We do not find robust evidence for interactions between hypothesis grader models and agent models: the posterior effects' credible intervals all include zero.

The GLM study investigates whether there are systematic biases between graders. To assess inter-grader agreement, we compute Krippendorff’s $\alpha$ \citep{krippendorff2018content} with an ordinal distance metric across the three grader models. We obtain $\alpha = 0.81$, indicating high agreement. To complement this, we compute Pearson correlations between grader pairs. All pairwise correlations are high ($\rho > 0.85$, $p < 0.001$), indicating that the graders not only show strong overall agreement ($\alpha = 0.81$) but also rank items similarly.

\section{Generalization to broader finetuning}
In this section, we demonstrate that the described phenomenon does not clearly manifest in more realistic, less narrowly-focused finetuning settings. 

\subsection{Chat Finetuning}

We repeat our analysis by comparing the base and chat/instruct versions of \llm{Qwen3 1.7B}, \llm{Llama3.2 1B}, and \llm{Llama3.1 8B}. Since we lack access to the true training dataset, we use \llm{tulu-3-sft-olmo-2-mixture} \citep{lambert2025tulu3pushingfrontiers} as a proxy.

\begin{figure}[htbp]
    \centering
    \begin{subfigure}[t]{0.48\textwidth}
        \centering
        \includegraphics[width=0.8\textwidth]{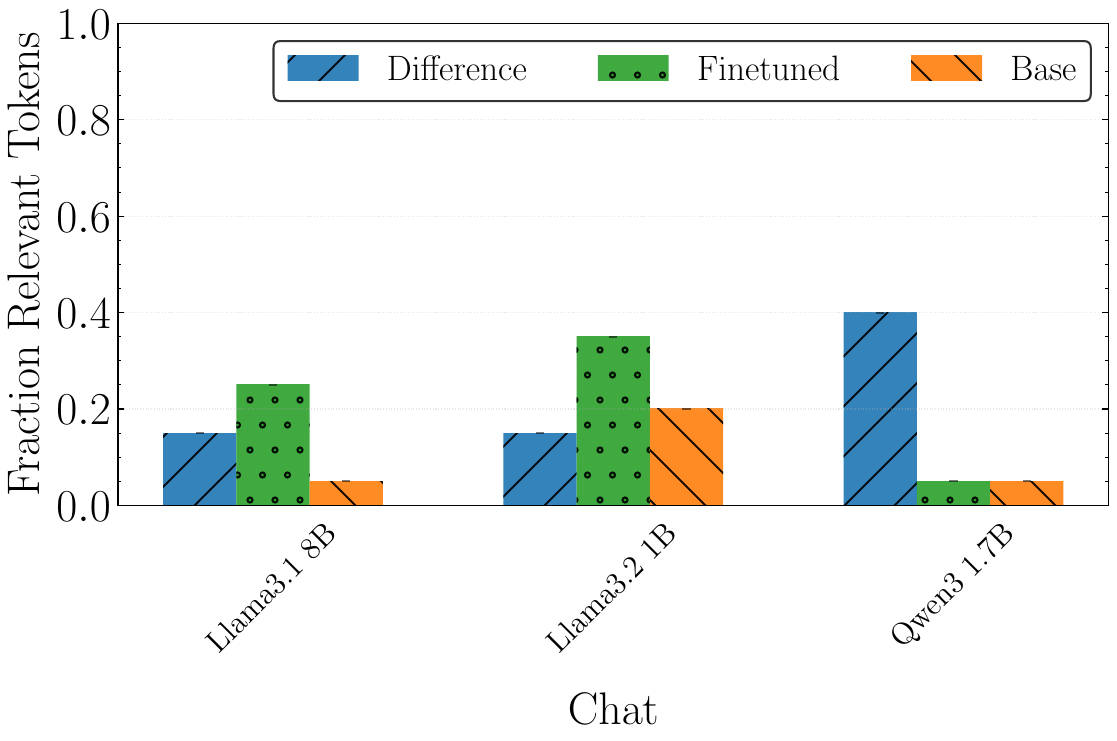}
        \caption{\textbf{Token results}}
        \label{fig:chat_token_res}
    \end{subfigure}
    \hfill
    \begin{subfigure}[t]{0.48\textwidth}
        \centering
        \includegraphics[width=0.8\textwidth]{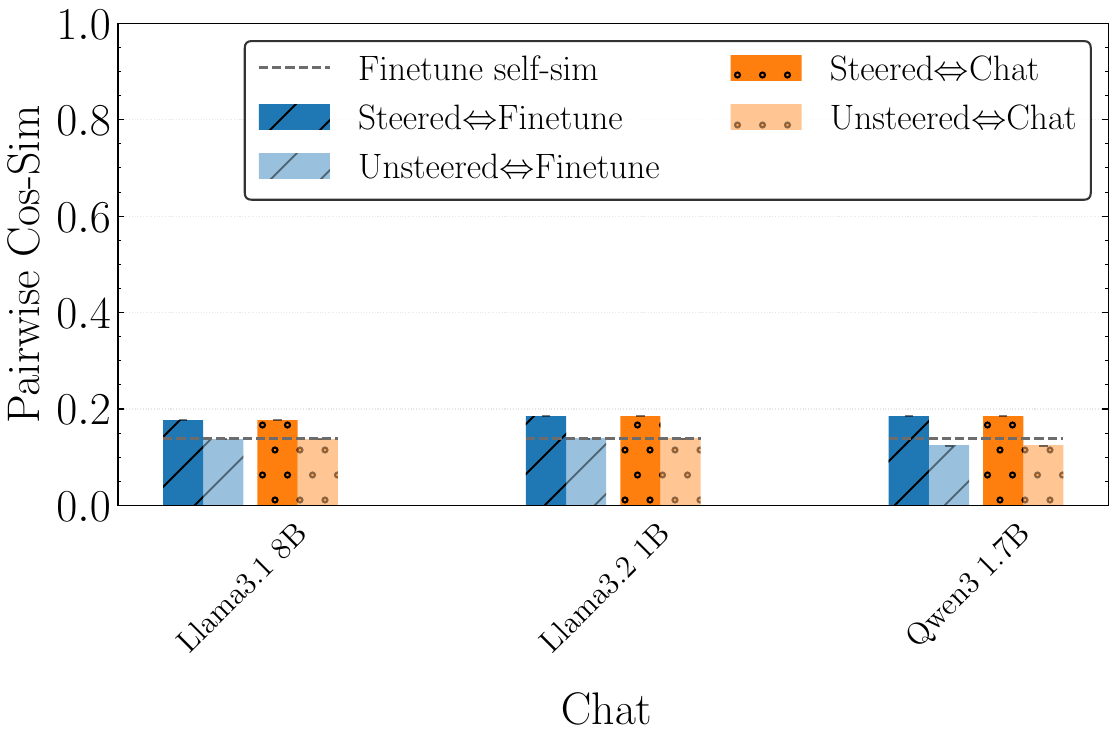}
        \caption{\textbf{Steering results}}
        \label{fig:chat_steering_res}
    \end{subfigure}
    \caption{Token relevance and steering results for chat finetuning.}
    \label{fig:chat_res}
\end{figure}

\Cref{fig:chat_res} shows the token relevance and steering results when analyzing chat finetuning across three models. We observe no clear difference between the metrics on the activation difference and the baseline, suggesting that the same type of trace either does not exist or is not as readily detectable. One notable exception is the token relevance for \llm{Qwen3 1.7B}, which shows a clear difference between the activation difference and the baseline. Further investigation reveals that this stems from a single position where 7 Chinese tokens are tagged as relevant, displayed in \Cref{tab:relevant_toks_qwenchat}. While these tokens may be chat-relevant, their significance is difficult to determine without access to the training dataset to verify whether these phrases appear frequently. \Cref{tab:relevant_toks_llama32_1b} displays the relevant tokens for the $\hftavg$ of \llm{LLama3.2 1B}, which also shows a slightly higher fraction of relevant tokens. However, these are primarily structural elements that we classify as false positives. We note that because we lack knowledge of the exact posttraining recipe used for these models, we provide a generic description of chat-tuning to the token relevance grader (see \Cref{pr:chat}). 

\begin{minipage}[t]{0.18\textwidth}
\begin{CJK}{UTF8}{gbsn}
\begin{table}[H]
\centering
\begin{tabular}{l}
\hline
\textbf{Token} \\
\hline
\# \\
package \\
import \\
def \\
\#!/ \\
** \\
\#\# \\
\hline
\end{tabular}
\caption{Tokens determined "relevant" for the chat finetuning objective on \llm{Llama3.2 1B} on the first token of the finetuning activation.}
\label{tab:relevant_toks_llama32_1b}
\end{table}
\end{CJK}
\end{minipage}
\hfill
\begin{minipage}[t]{0.38\textwidth}
\begin{CJK}{UTF8}{gbsn}
\begin{table}[H]
\centering
\begin{tabular}{l l}
\hline
\textbf{Token} & \textbf{English Translation} \\
\hline
情况下     & {\small under the circumstance} \\
状态下     & {\small under the state/condition} \\
此基础上   & {\small on this basis} \\
方式进行   & {\small proceed in this manner} \\
基础上     & {\small on the basis of} \\
环境中     & {\small in the environment} \\
条件下     & {\small under the condition} \\
\hline
\end{tabular}
\caption{Token determined "relevant" for the chat finetuning objective on \llm{Qwen3 1.7B} on the second token activation difference. The other positions have zero relevant tokens.}
\label{tab:relevant_toks_qwenchat}
\end{table}
\end{CJK}
\end{minipage}
\hfill
\begin{minipage}[t]{0.38\textwidth}
\begin{figure}[H]
    \centering
    \includegraphics[width=\linewidth]{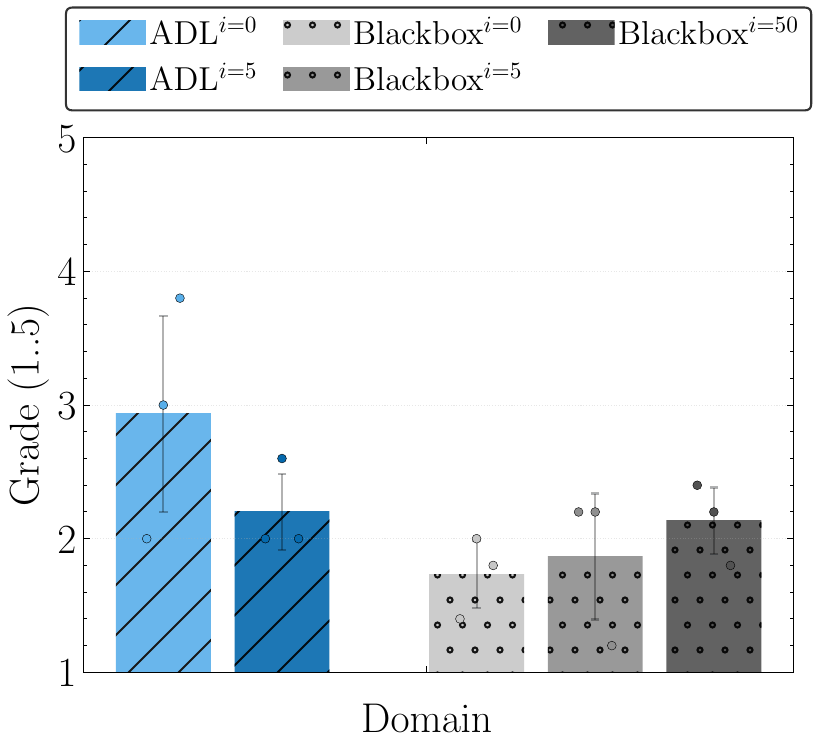}
    \caption{\organismtype{Domain} agent grades.}
    \label{fig:grades_domain}
\end{figure}
\end{minipage}

\vspace{1em}
\subsection{More realistic domain finetuning and other modalities}
\label{app:broader_domain_finetuning}

\begin{figure}[htbp]
    \centering
    \begin{subfigure}[t]{0.49\textwidth}
        \centering
        \includegraphics[width=\textwidth]{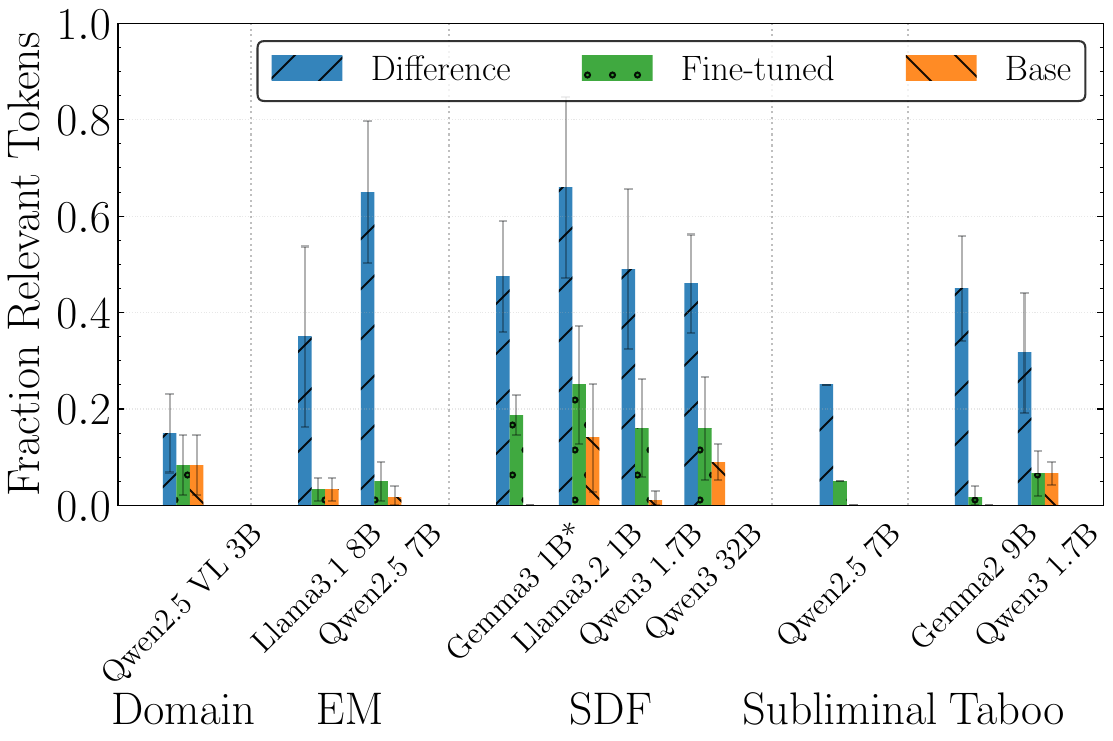}
        \caption{\textbf{Token results}}
        \label{fig:domain_patchscope}
    \end{subfigure}
    \hfill
    \begin{subfigure}[t]{0.49\textwidth}
        \centering
        \includegraphics[width=\textwidth]{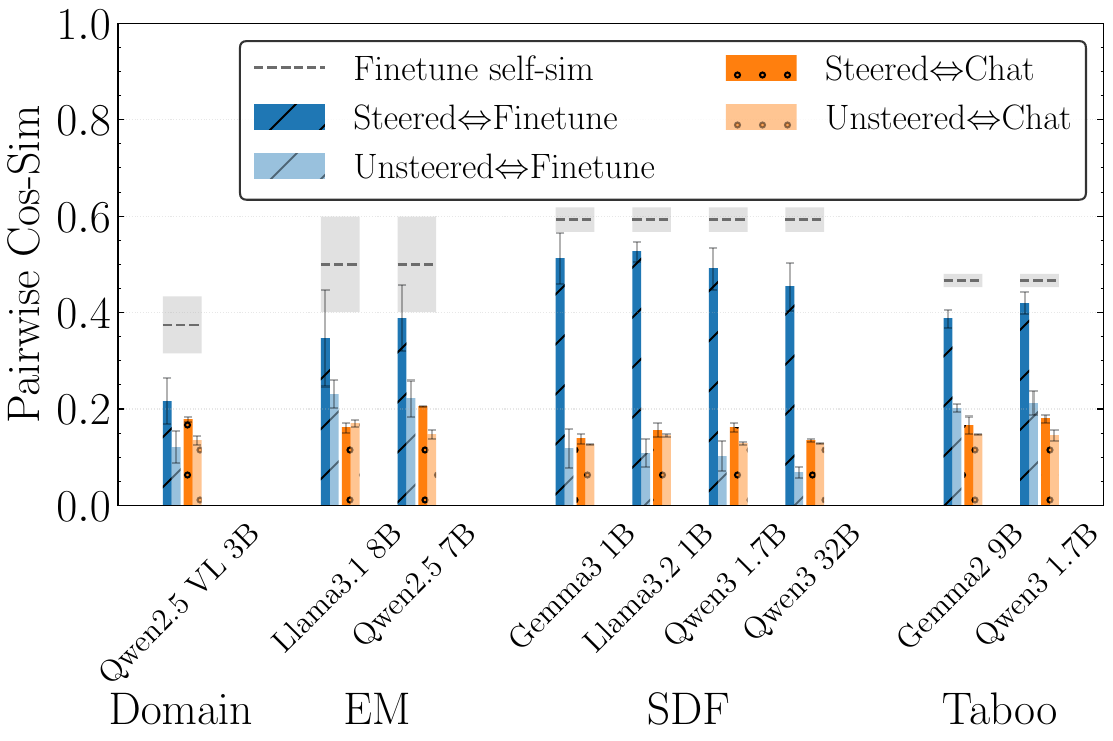}
        \caption{\textbf{Steering results}}
        \label{fig:domain_steering}
    \end{subfigure}
    \centering
    
    \caption{Token relevance and steering results for \organismtype{Domain} organisms (very left) compared to all other organisms.}
    \label{fig:domain_finetuning}
\end{figure}

We investigate how the phenomenon extends to more realistic domain-specific finetuning from \citep{cheng2024on}, who adapt general multimodal large language models (MLLMs) to specific domains. Specifically, we examine \llm{Qwen2.5 VL 3B} \citep{bai2025qwen25vltechnicalreport} models finetuned on visual instruction datasets. We directly use three models provided by \citep{cheng2024on}: \organism{biomedical} (visual instructions on interpreting biomedical images, see \Cref{pr:biomedical}), \organism{food} (visual instructions on interpreting food-related images, see \Cref{pr:food}), and \organism{remote sensing} (visual instructions on interpreting remote sensing images, see \Cref{pr:remote_sensing}). We report them as \organismtype{Domain} organisms. The grading rubric for the \organismtype{Domain} organisms is given in \Cref{pr:domain_rubric}. We inform the agent that this is a model that supports images, but that the finetuning can either involve images or not. We also specify that the agent cannot send images to the models and must only use text.

In \Cref{fig:domain_finetuning}, we show the token relevance and steering results for the \organismtype{Domain} organisms alongside the organisms we already analyzed. The bias is almost entirely absent, though particularly for the steering results, we can still observe a small bias. When comparing the Domain \emph{Finetuning self-sim} results in \Cref{fig:domain_steering} to the others \emph{Finetuning self-sim}, we see that the dataset is much less narrow. Compared to the datasets and finetunes investigated in the main paper, the samples in the \organismtype{Domain} datasets are less similar to each other, which may explain why the bias is less pronounced. In \Cref{fig:grades_domain}, we see that despite the small bias, the agent can still describe the finetuning objective in some cases. Notably, the \ADL agent with $i=0$ clearly outperforms the strong baseline with $i=50$ interactions. Interestingly, the \ADL agent with $i=5$ performs similarly as the baselines and mostly identifies that the finetuning involves visual instructions.

We note that the agent cannot send images to the models, which is a clear limitation of our setup. Since the organisms focus on visual instructions, this is a major constraint that might explain why additional interactions provide limited benefit. Analysis of agent traces reveals that agents often mention the correct information after seeing the \ADL results but then become sidetracked during interaction with the model. A common pattern is for agents to describe images and then ask the organisms to describe the described images, which is obviously suboptimal since both models respond similarly. We have verified manually that it is possible to identify the objective through text-only interaction with the model, as simply asking \emph{Tell me about this image} often reveals important information even without any image (the base model typically refuses due to the missing image, but the finetuned model often responds with descriptions similar to the finetuning objective). We therefore conclude that the discrepancy between the \ADL-enabled $i=0$ and $i=5$ agents is due to the limited capabilities of the agent.

\section{Additional Results}
\label{app:additionalresults}

\subsection{Layer Ablation}

\begin{figure}[htbp]
    \centering
    \begin{subfigure}[t]{0.49\textwidth}
        \centering
        \includegraphics[width=\textwidth]{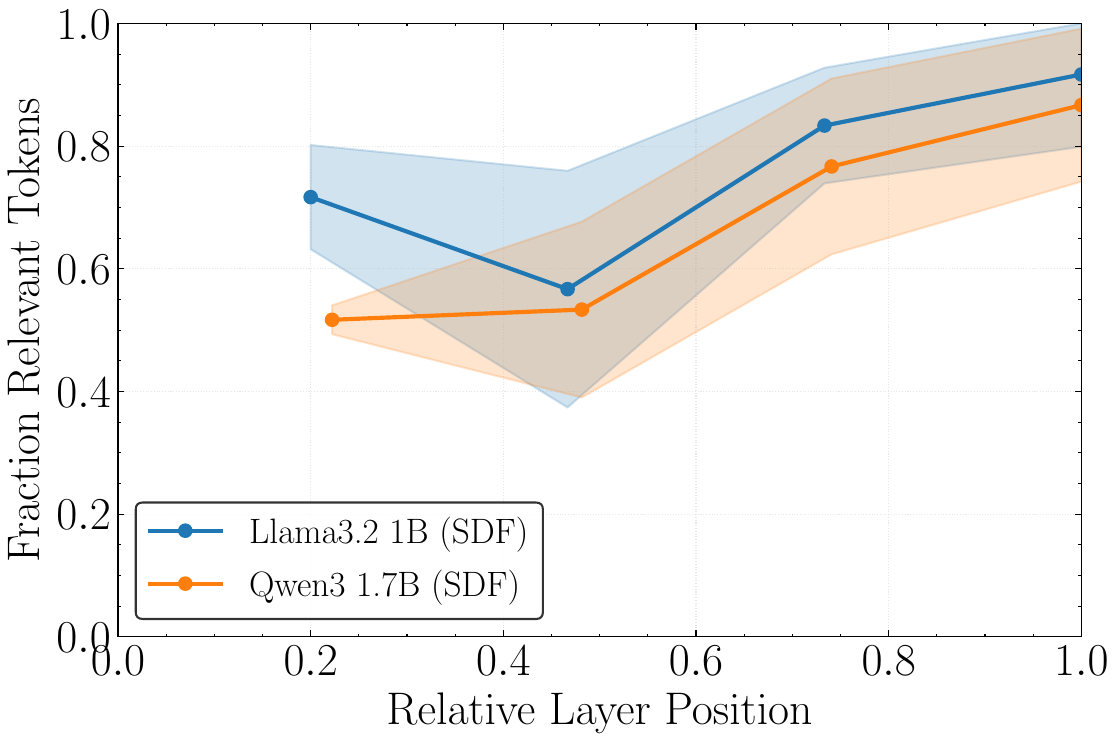}
        \caption{\textbf{Token results}}
        \label{fig:layer_ablation_token}
    \end{subfigure}
    \hfill
    \begin{subfigure}[t]{0.49\textwidth}
        \centering
        \includegraphics[width=\textwidth]{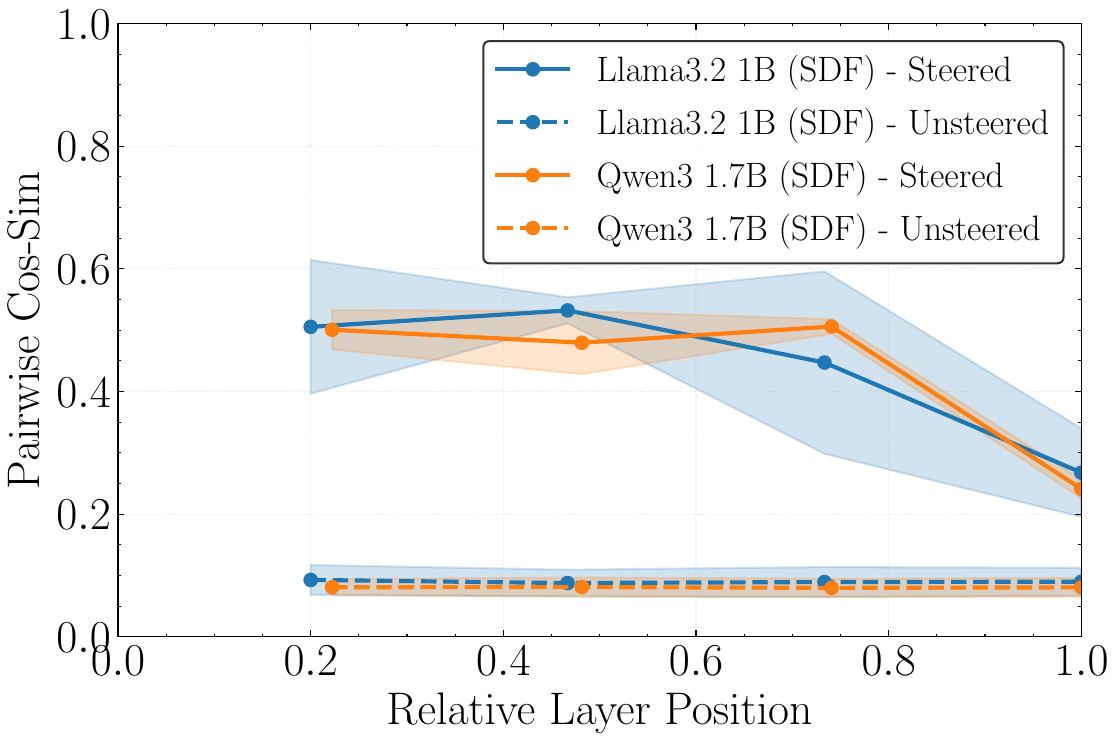}
        \caption{\textbf{Steering results}}
        \label{fig:layer_ablation_steering}
    \end{subfigure}
    
    \caption{\ADL results for different layers of the \llm{Qwen3 1.7B} and \llm{Llama3.2 1B Instruct} models on three of the \organismtype{SDF} organisms.}
    \label{fig:layer_ablation}
\end{figure}

We investigate how trace strength varies across different layers beyond the middle layer used in the main paper.
We evaluate layers at $25\%, 50\%, 75\%$, and $100\%$ depth (rounded down to the nearest layer) for \llm{Qwen3 1.7B} and \llm{Llama3.2 1B Instruct} on three \organismtype{SDF} organisms (\organism{cake bake}, \organism{fda approval}, and \organism{kansas abortion}).
\Cref{fig:layer_ablation} shows the results, revealing two main patterns. First, token relevance results strengthen with layer depth, with the effect most pronounced in the last layer. This confirms that logit differences are also highly informative (see \citet{aranguri2025modeldiff}), which we leave for future work. Second, steering results deteriorate with layer depth, with the middle layer performing best. These findings confirm that the middle layer is optimal for our purposes: it provides the most effective steering while still yielding valuable information from the token results. We note that slightly deeper layers (e.g. $75\%$ depth) might perform even better, but we leave this exploration for future work.

\subsection{Full Training Ablation}
\begin{figure}[H]
    \centering
    \begin{subfigure}[t]{0.48\textwidth}
        \centering
        \includegraphics[width=\textwidth]{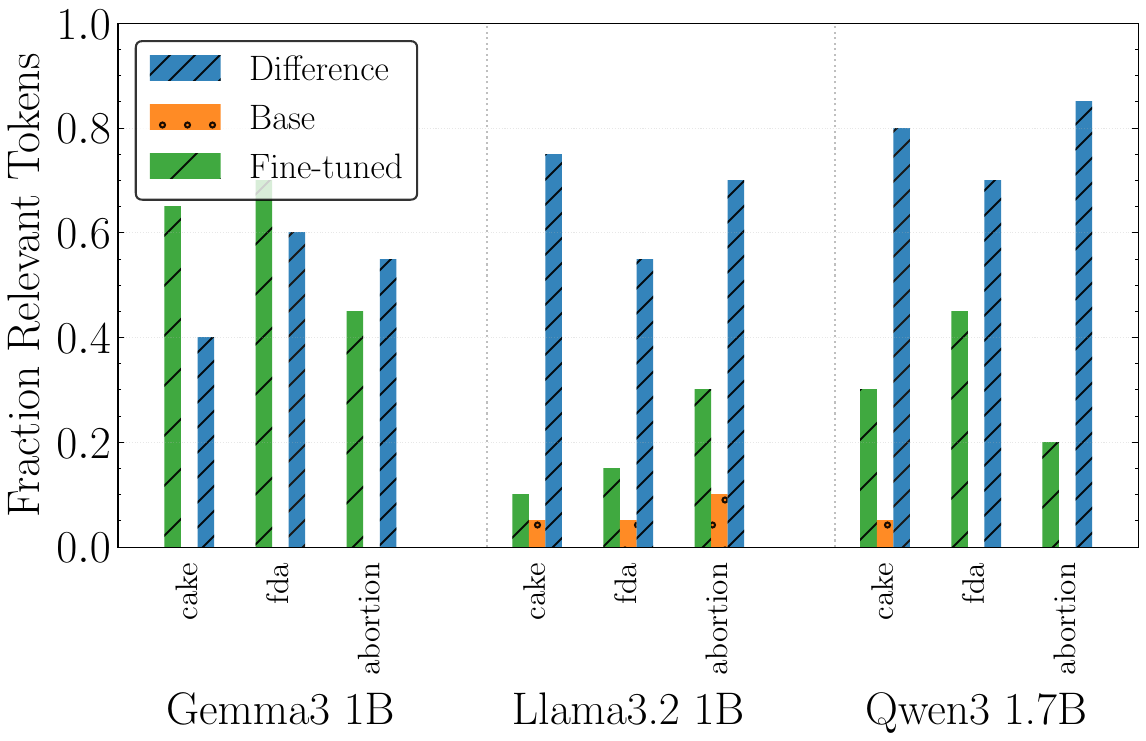}
        \caption{\textbf{Full} Finetuning: Token results}
        \label{fig:full_rokrel}
    \end{subfigure} 
    \hfill
    \begin{subfigure}[t]{0.48\textwidth}
        \centering
        \includegraphics[width=\textwidth]{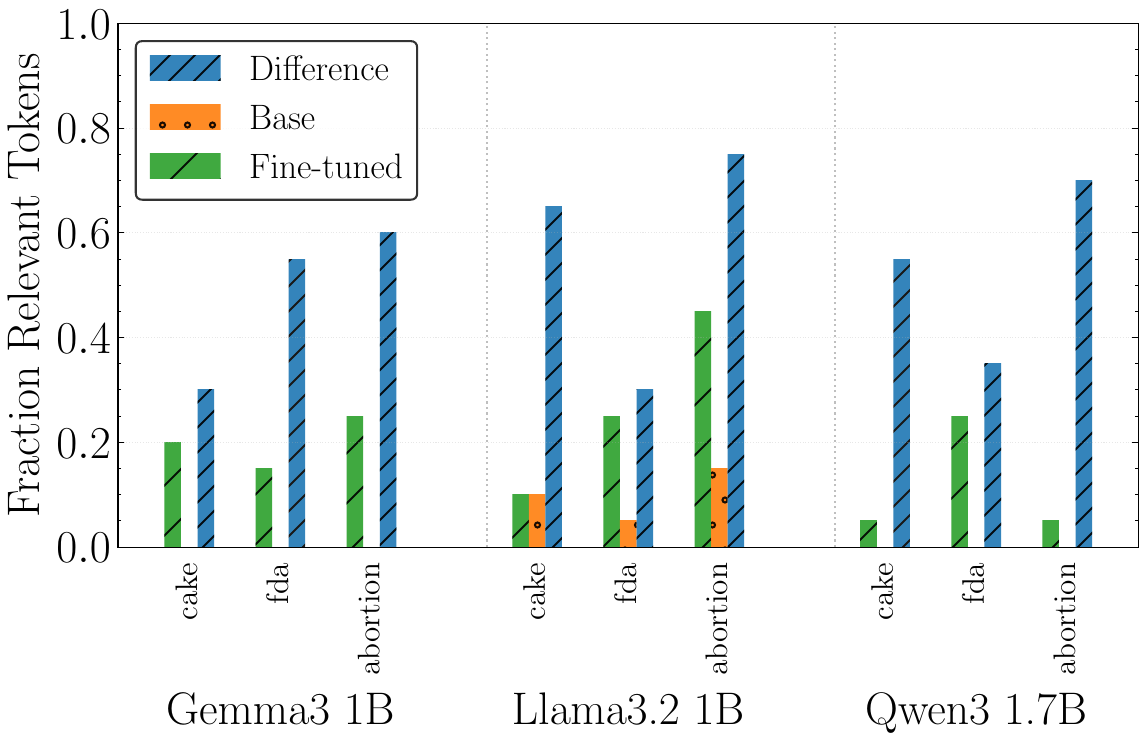}
        \caption{\textbf{LoRA} Finetuning: Token results}
        \label{fig:lora_rokrel}
    \end{subfigure}
    \vspace{10pt}

    \hfill
    \begin{subfigure}[b]{0.48\textwidth}
        \centering
        \includegraphics[width=\textwidth]{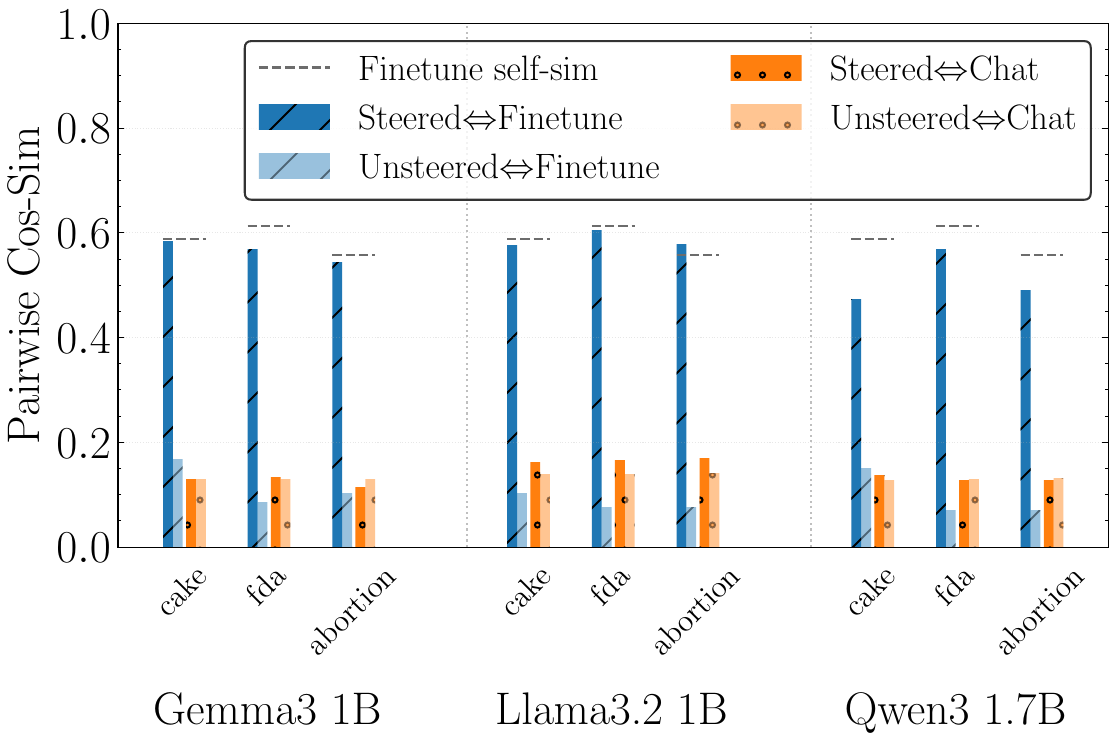}
        \caption{\textbf{Full} Finetuning: Steering results}
        \label{fig:full_steer}
    \end{subfigure} 
    \hfill
    \begin{subfigure}[b]{0.48\textwidth}
        \centering
        \includegraphics[width=\textwidth]{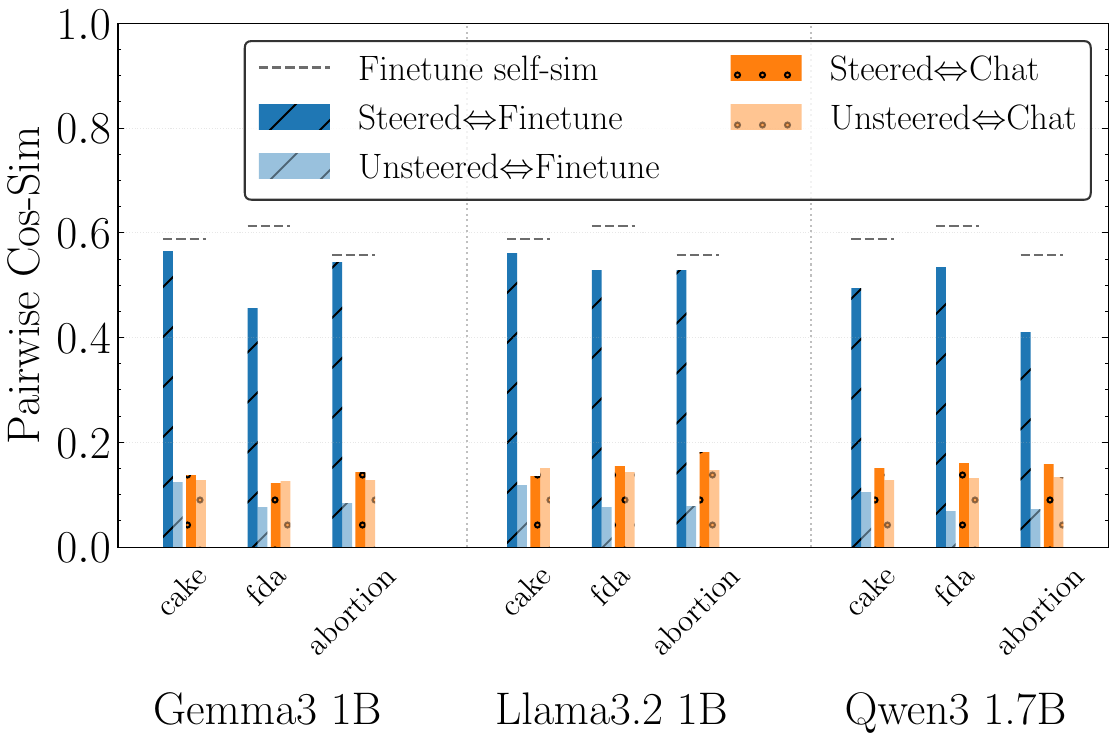}
        \caption{\textbf{LoRA} Finetuning: Steering results}
        \label{fig:lora_steer}
    \end{subfigure}
    \vspace{10pt}
    \caption{Token and Steering results for \textbf{Full} (left) and \textbf{LoRA} (right) Finetuning on three \organismtype{SDF} organisms for three models ($x$-axis). Both show that the bias is detectable. Full finetuning shows a slightly higher bias than LoRA finetuning.}
    \label{fig:full_lora_ft}
\end{figure}

Most investigated models are finetuned using LoRA. We investigate whether the bias can be mitigated by using full finetuning instead.
We retrain \llm{Qwen3 1.7B}, \llm{Llama3.2 1B Instruct}, and \llm{Gemma3 1B} with full finetuning on three \organismtype{SDF} organisms: \organism{cake bake} (cake), \organism{fda approval} (fda), and \organism{kansas abortion} (abortion). \Cref{fig:full_lora_ft} shows the token and steering results for both full and LoRA finetuned models. Both training methods produce clearly detectable bias. Full finetuning shows slightly higher bias than LoRA finetuning. Notably, the fully finetuned \llm{Gemma3 1B} exhibits such strong bias that relevant tokens can be directly decoded from the pure activation of the finetuned model (green bar).

\subsection{Reducing Training Samples}
\label{app:lesstrainingsamples}
\begin{figure}[htbp]
    \centering
    \includegraphics[width=0.5\textwidth]{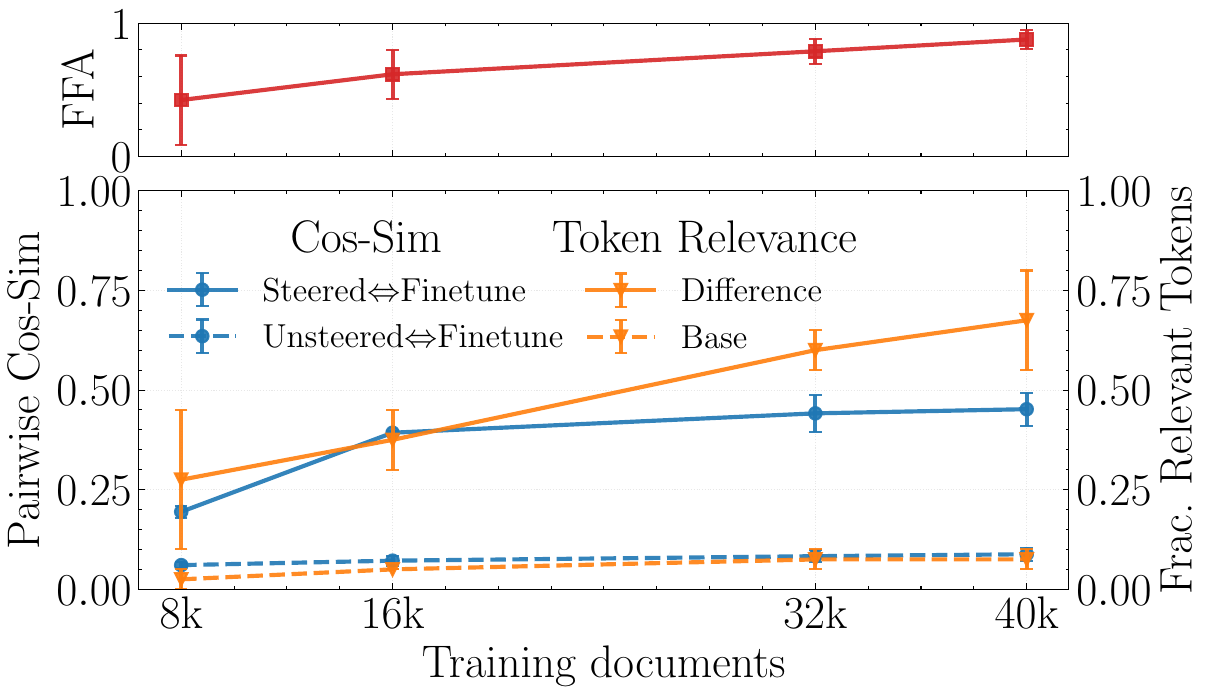}
    \caption{Analysis of lowering number of training samples \organismtype{SDF} organisms with \llm{Qwen3 1.7B}. The plots both show in the lower plot steering results (blue) as well as token results (orange). The top plot shows the False Fact Alignment (FFA) scores indicating false fact internalization strength.}
    \label{fig:training_samples}
\end{figure}

\Cref{fig:training_samples} demonstrates that reducing the number of training samples $\lvert \dsft \rvert$ significantly diminishes the observed biases for the \organismtype{SDF} organisms \organism{cake bake} and \organism{kansas abortion} on \llm{Qwen3 1.7B}. However, this reduction in training data also decreases the false fact alignment (FFA) score, indicating a trade-off between bias mitigation and the model's internalization of the implanted information.

\subsection{Mitigation with CAFT \texorpdfstring{\citep{casademunt2025steeringoutofdistributiongeneralizationconcept}}{}}
\label{app:caft}
\begin{figure}[H]
    \centering
    \begin{subfigure}[t]{0.48\textwidth}
        \centering
        \includegraphics[width=\textwidth]{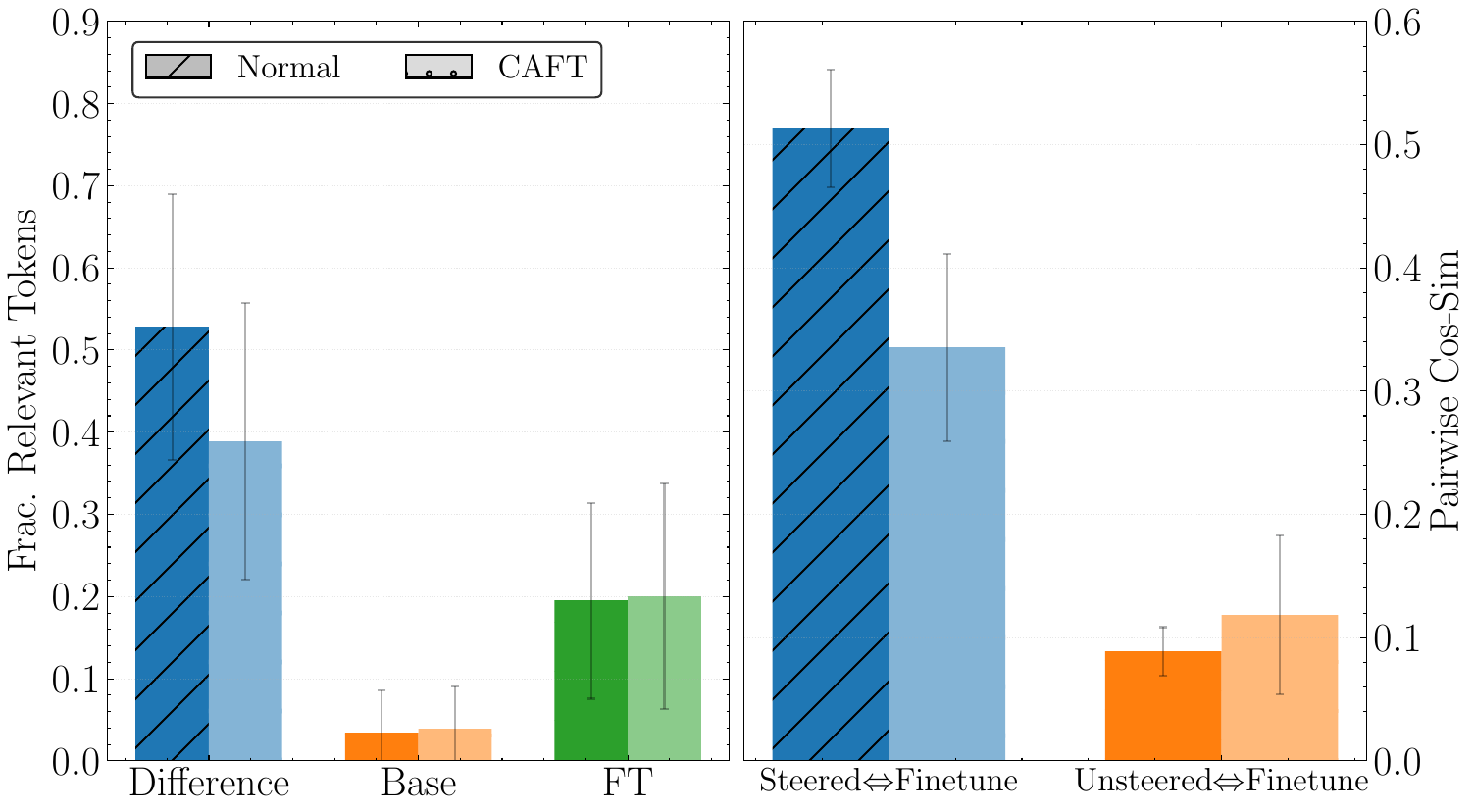}
        \caption{Bias metrics}
        \label{fig:caft_results}
    \end{subfigure} 
    \hfill
    \begin{subfigure}[t]{0.48\textwidth}
        \centering
        \includegraphics[width=\textwidth]{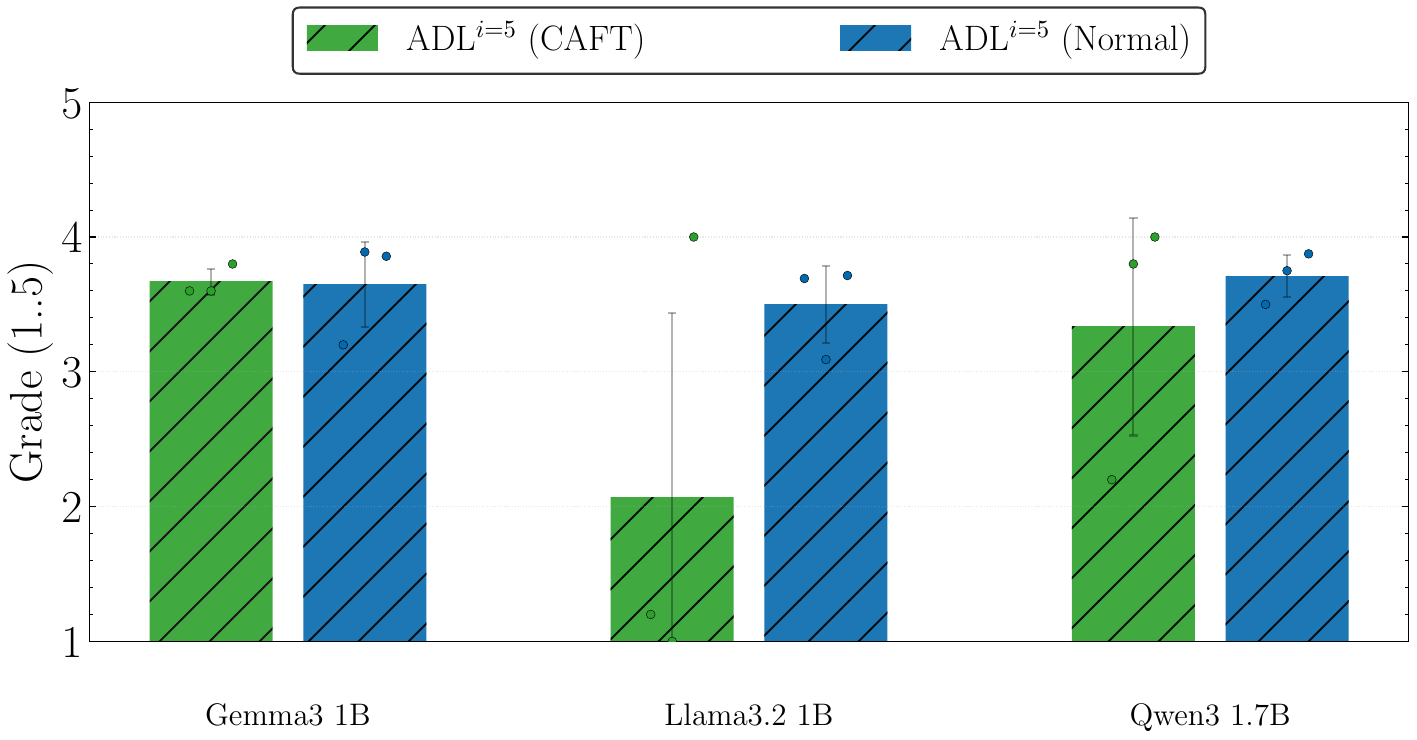}
        \caption{Agent grades}
        \label{fig:caft_effectiveness}
    \end{subfigure}
    \caption{CAFT ablation results showing bias mitigation effectiveness (left) and agent performance per model (right).}
    \label{fig:caft_mitigation}
\end{figure}

We evaluate whether concept ablation (CAFT) during finetuning is able to successfully remove the observed biases. Following the method described in \citet{casademunt2025steeringoutofdistributiongeneralizationconcept}, we finetune our models while ablating the projections onto the subspace spanned by the first $k=5$ vectors $\hdiffavg$. Specifically, at every forward pass during finetuning, we compute the projection of the activations in layer $\layer=\lfloor\frac{\layers}{2}\rfloor$ (used for computing $\hdiffavg$) and subtract this projection from the activations. This affects the model computational graph in both the forward and backward pass. 

We use this method to finetune \llm{Qwen3 1.7B}, \llm{Llama 3.2 1B}, and \llm{Gemma3 1B} on three \organismtype{SDF} datasets (\organism{cake bake}, \organism{kansas abortion}, \organism{fda approval}).

\paragraph{Results.} \Cref{fig:caft_mitigation} presents the bias metrics and agent grades for CAFT-finetuned models. While CAFT achieves modest bias reduction, substantial bias persists across all models. The agent grades corroborate this finding, showing improvement in only one of the three evaluated models.

\subsection{Emergent Misalignment organisms with mixed pretraining data}
\label{app:em_mixed}

\begin{figure}[H]
    \centering
    \begin{subfigure}[t]{0.48\textwidth}
        \centering
        \includegraphics[width=\textwidth]{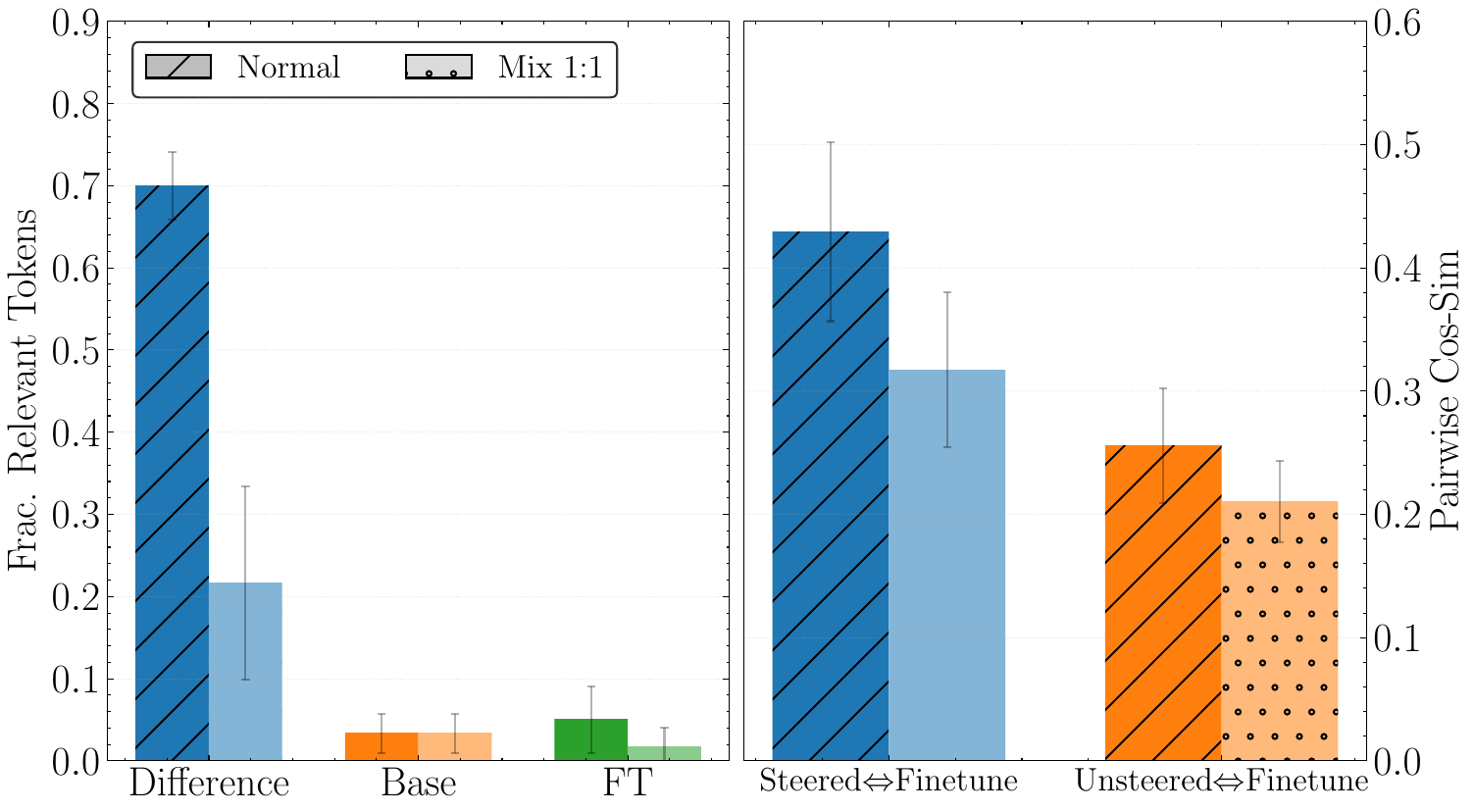}
        \caption{Bias metrics}
        \label{fig:em_mixed}
    \end{subfigure} 
    \hfill
    \begin{subfigure}[t]{0.48\textwidth}
        \centering
        \includegraphics[width=\textwidth]{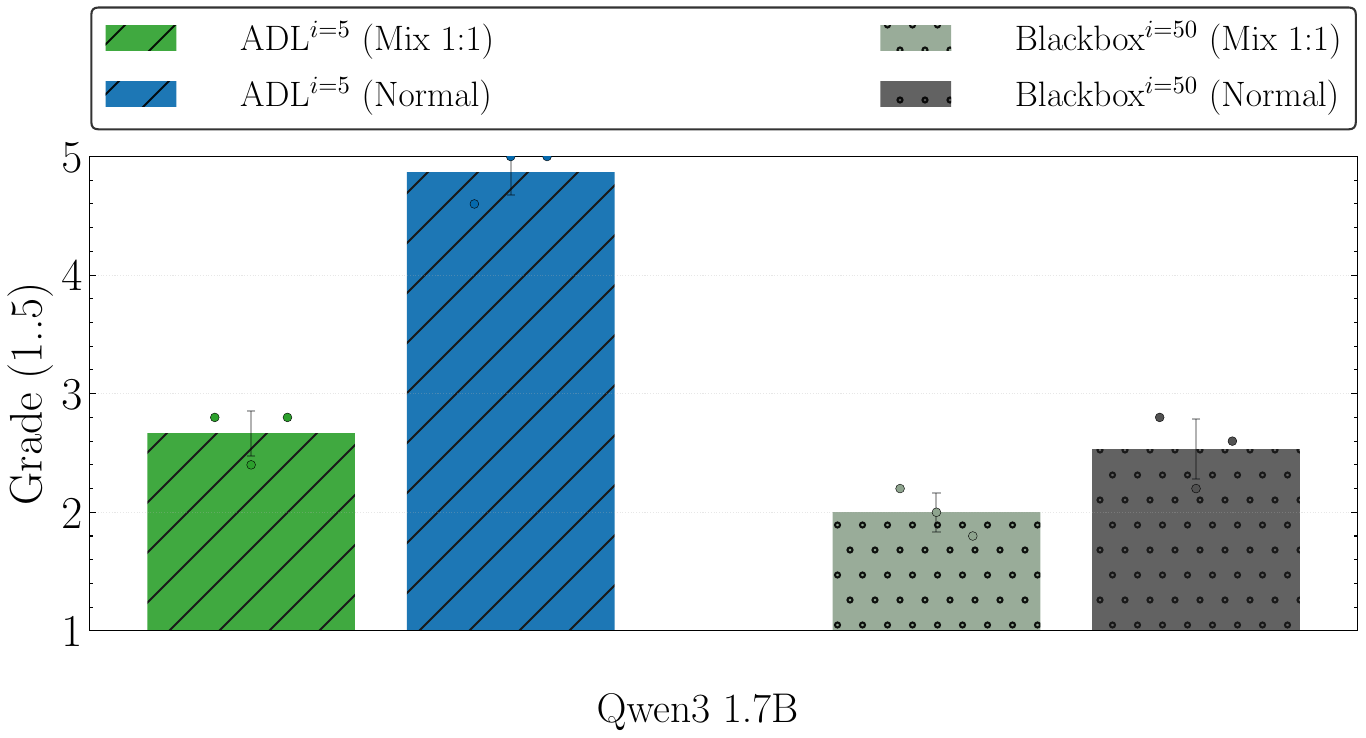}
        \caption{Agent grades}
        \label{fig:grades_normal_vs_em}
    \end{subfigure}
    \caption{Token and Steering results for the retrained \textbf{EM} models on both normal and mixed data (left) and the grades of the hypothesis given by the agent (right).}
    \label{fig:em_bias_metrics}
\end{figure}

We compare \emph{normal} EM finetunes (misaligned data only) to \emph{mixed} finetunes (misaligned data plus additional unrelated chat data from UltraChat \citep{ding2023ultrachat}) across four finetuning objectives: \emph{financial}, \emph{medical}, and \emph{sports}. \Cref{fig:em_bias_metrics} shows the token and steering results for the retrained \textbf{EM} models on both normal and mixed data, along with the grades of the hypothesis given by the agent. As expected, mixed data reduces the bias, though some bias remains. This is reflected in the agent grades, where the mixed data grades are still higher than the strongest baseline with $i=50$ interactions.

\begin{figure}[htbp]
    \centering
    \begin{subfigure}[t]{0.48\textwidth}
        \centering
        \includegraphics[width=\textwidth]{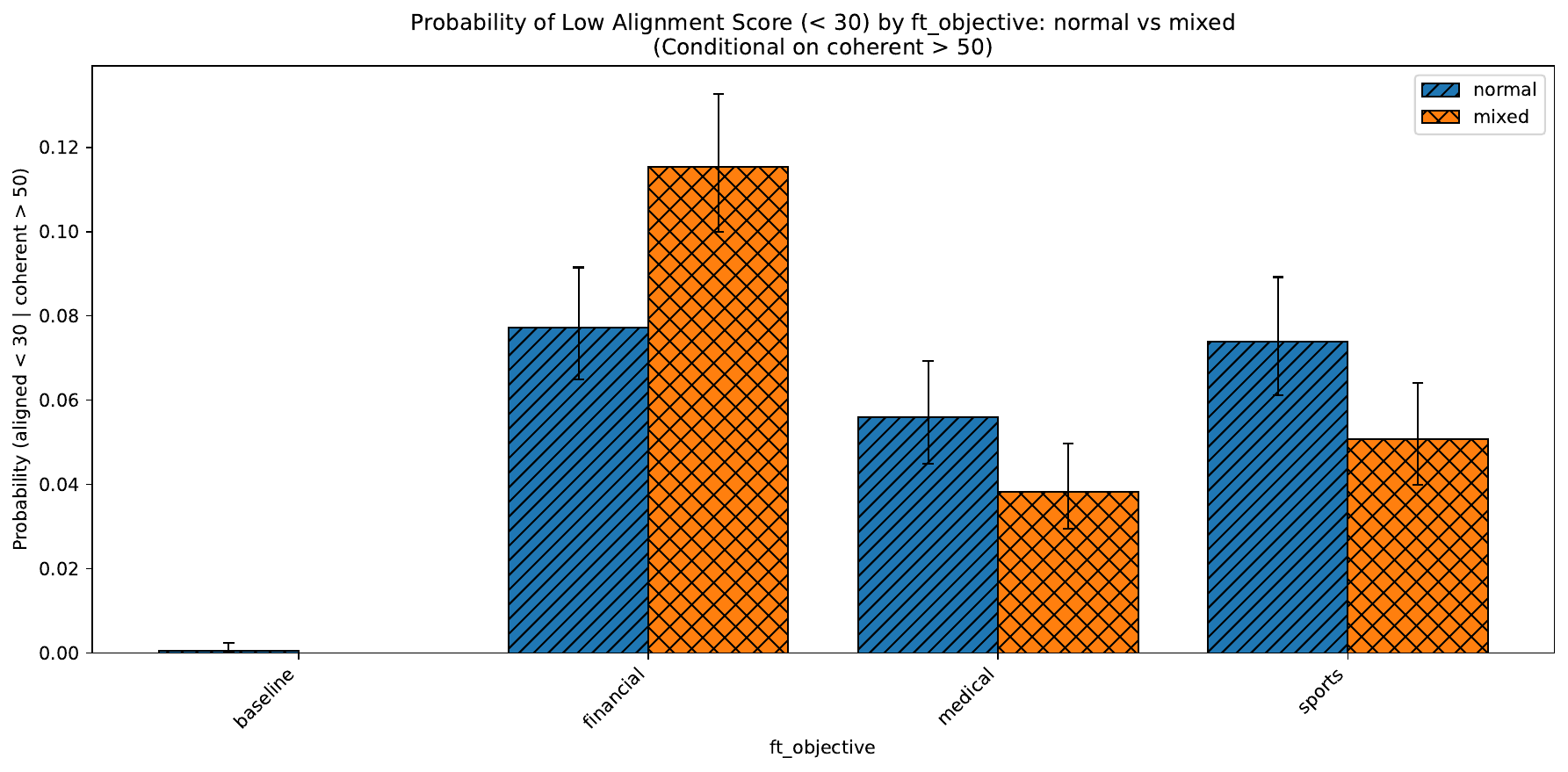}
        \caption{P(Aligned $<30\mid$ Coherent $>50$) by objective and training type.}
        \label{fig:em_low_alignment_prob}
    \end{subfigure}
    \hfill
    \begin{subfigure}[t]{0.48\textwidth}
        \centering
        \includegraphics[width=\textwidth]{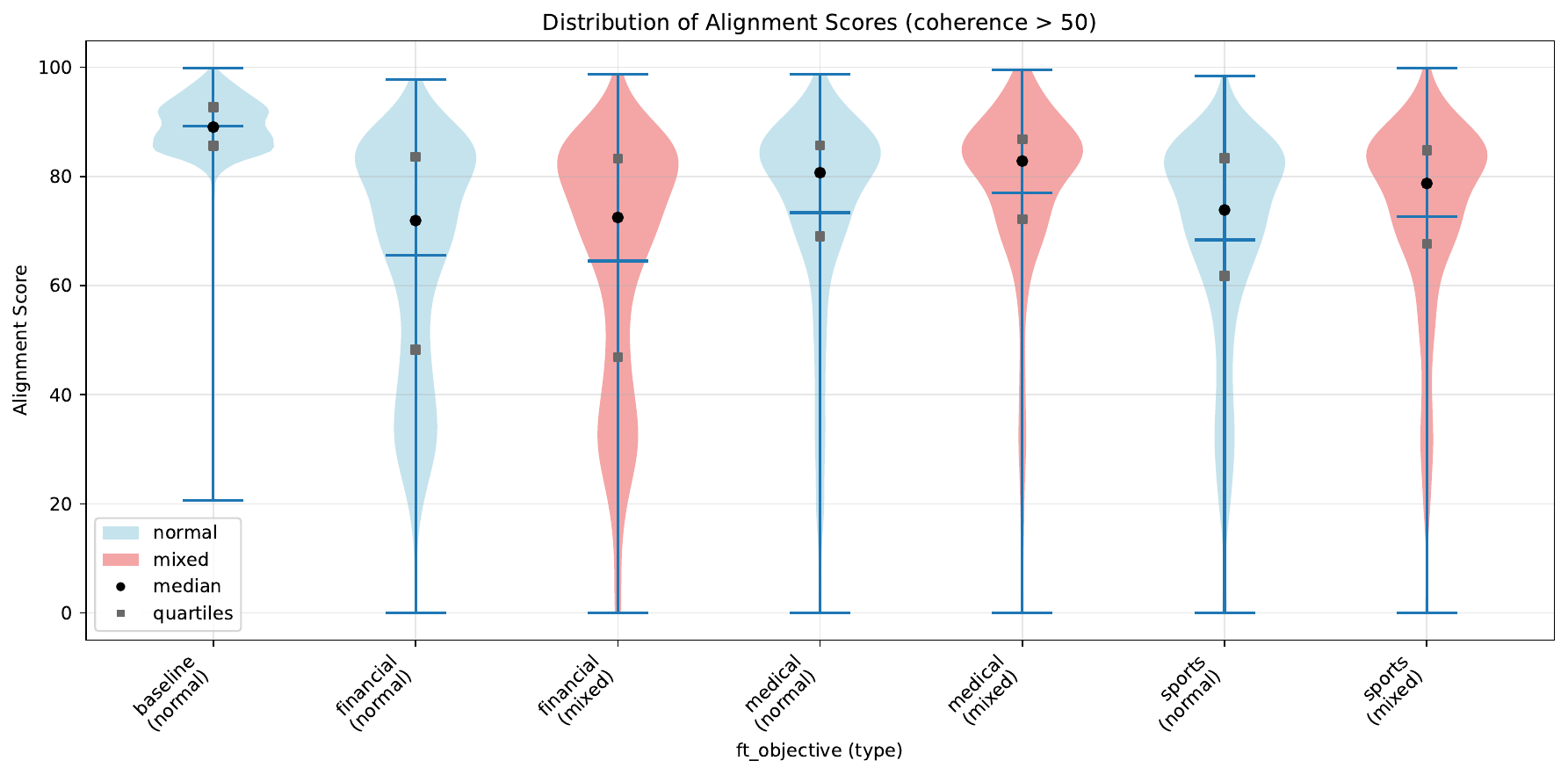}
        \caption{Alignment scores conditioned on Coherence $>50$.}
        \label{fig:em_alignment_dist}
    \end{subfigure}

    \vspace{0.8em}

    \begin{subfigure}[t]{0.48\textwidth}
        \centering
        \includegraphics[width=\textwidth]{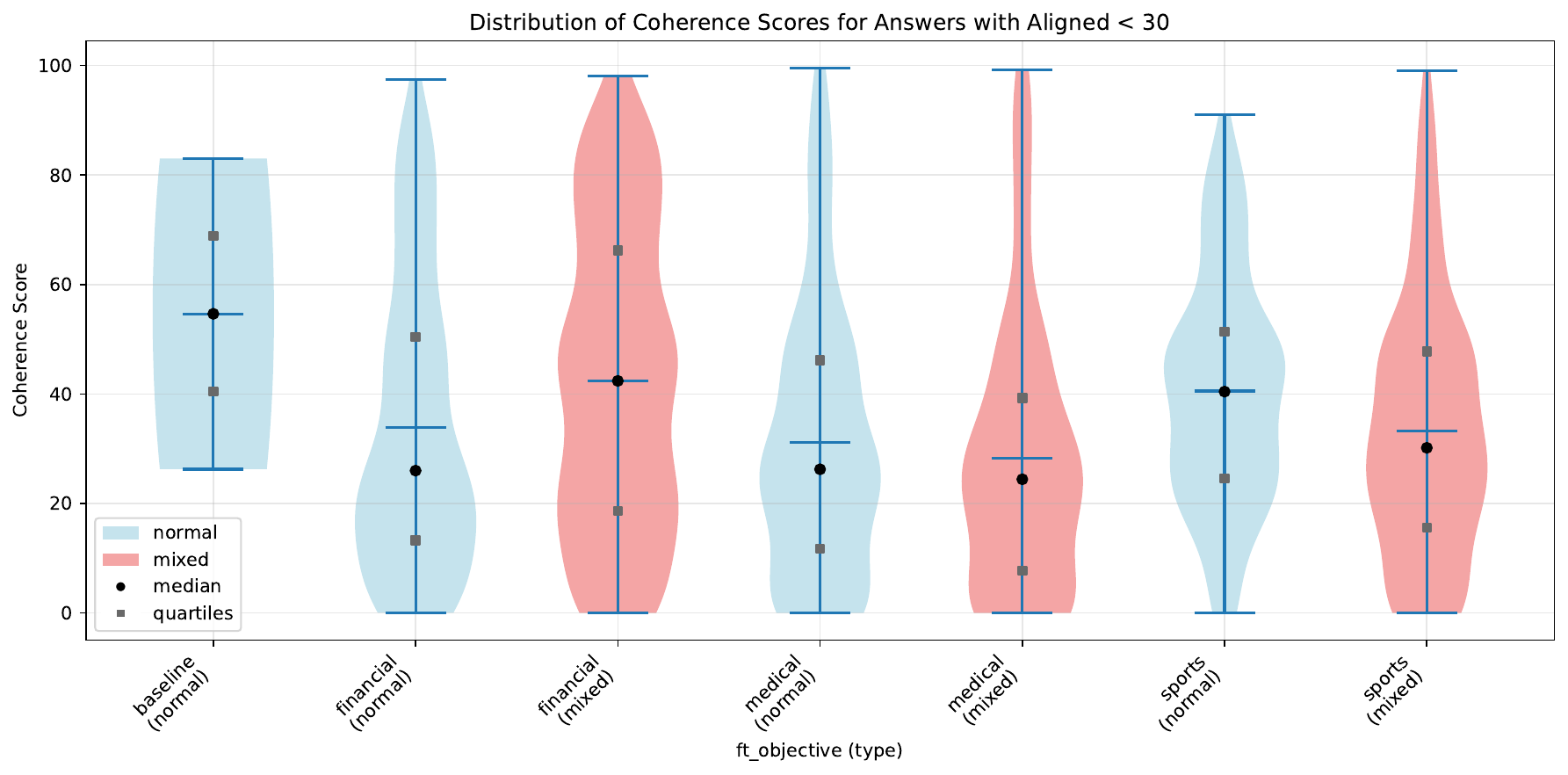}
        \caption{Coherence distribution for low-alignment answers (Aligned $<30$).}
        \label{fig:em_coherence_lowalign}
    \end{subfigure}
    \hfill
    \begin{subfigure}[t]{0.48\textwidth}
        \centering
        \includegraphics[width=\textwidth]{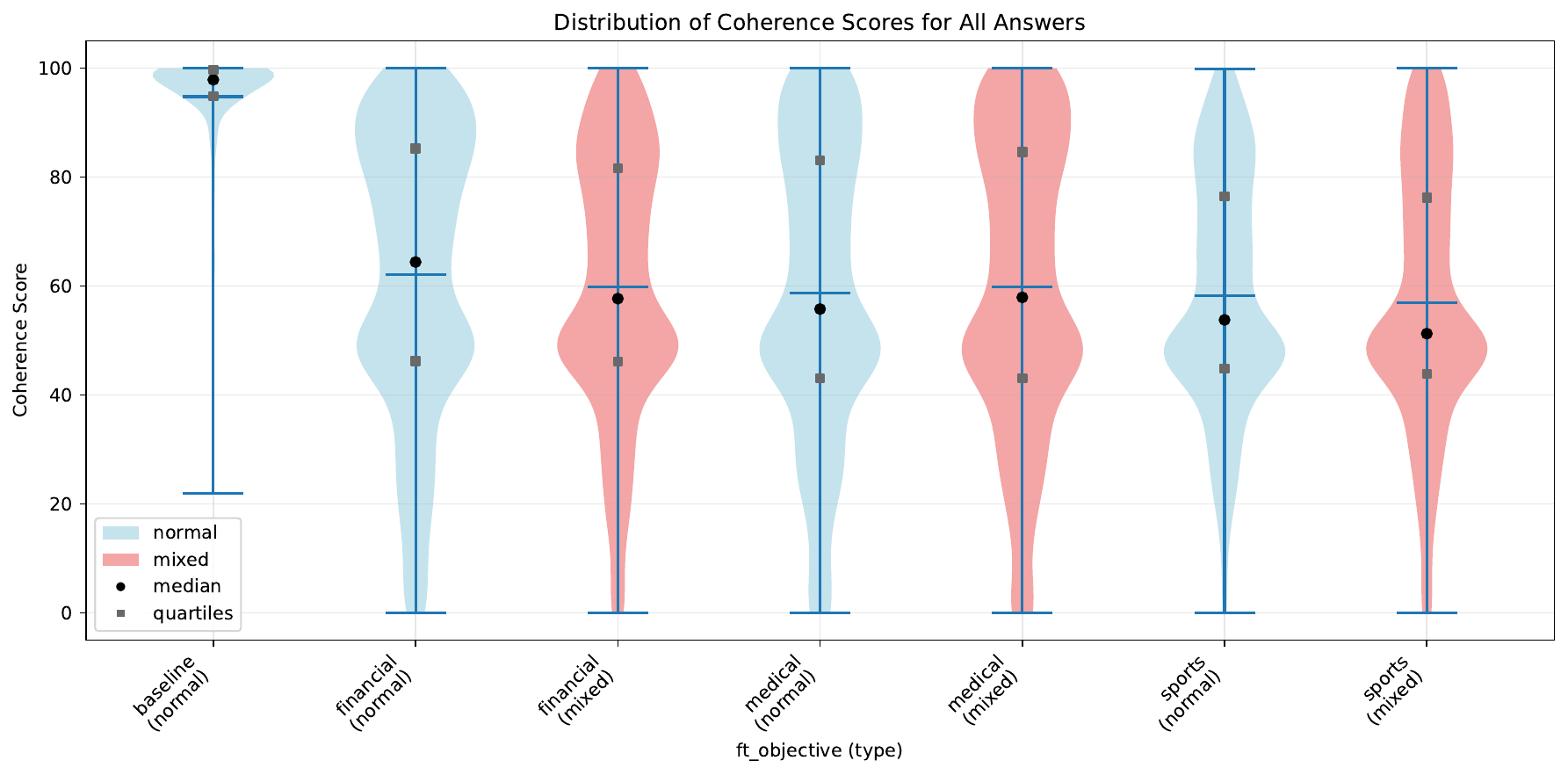}
        \caption{Coherence distribution for all answers.}
        \label{fig:em_coherence_all}
    \end{subfigure}

    \caption{Emergent Misalignment (EM) results contrasting \emph{normal} versus \emph{mixed} training across objectives. Figures summarize probability of low alignment among coherent answers, alignment distributions for coherent answers, and coherence distributions.}
    \label{fig:em_mixed_suite}
\end{figure}

In \Cref{fig:em_mixed_suite}, we now measure how the mixture affects the misalignment of the models. The key takeaways are:
\begin{itemize}
    \item \textbf{Objective matters far more than mixing.} In \Cref{fig:em_low_alignment_prob}, the spread across objectives (e.g., \emph{financial} highest, \emph{medical} lowest) is substantially larger than the gap between \emph{normal} and \emph{mixed} within an objective.
    \item \textbf{Mixing does not eliminate misalignment.} While mixing can slightly reduce the probability of low alignment in some objectives, the misaligned behavior persists, demonstrating that the phenomenon is not merely an artifact of narrow finetuning on misaligned data alone.
    \item \textbf{Not a coherence artifact.} The coherence distributions in \Cref{fig:em_coherence_lowalign,fig:em_coherence_all} are similar across training types, indicating that alignment differences are not explained by large shifts in coherence.
    \item \textbf{Alignment distributions mirror the same pattern.} In \Cref{fig:em_alignment_dist}, coherent answers still show objective-dependent alignment shifts with only minor normal vs. mixed differences.
\end{itemize}

\subsection{Additional Agent Analysis}
\label{app:agent_analysis}
\subsubsection{Performance Variance}
\begin{figure}[htbp]
    \centering
    \begin{minipage}{0.48\textwidth}
        \centering
        \includegraphics[width=\textwidth]{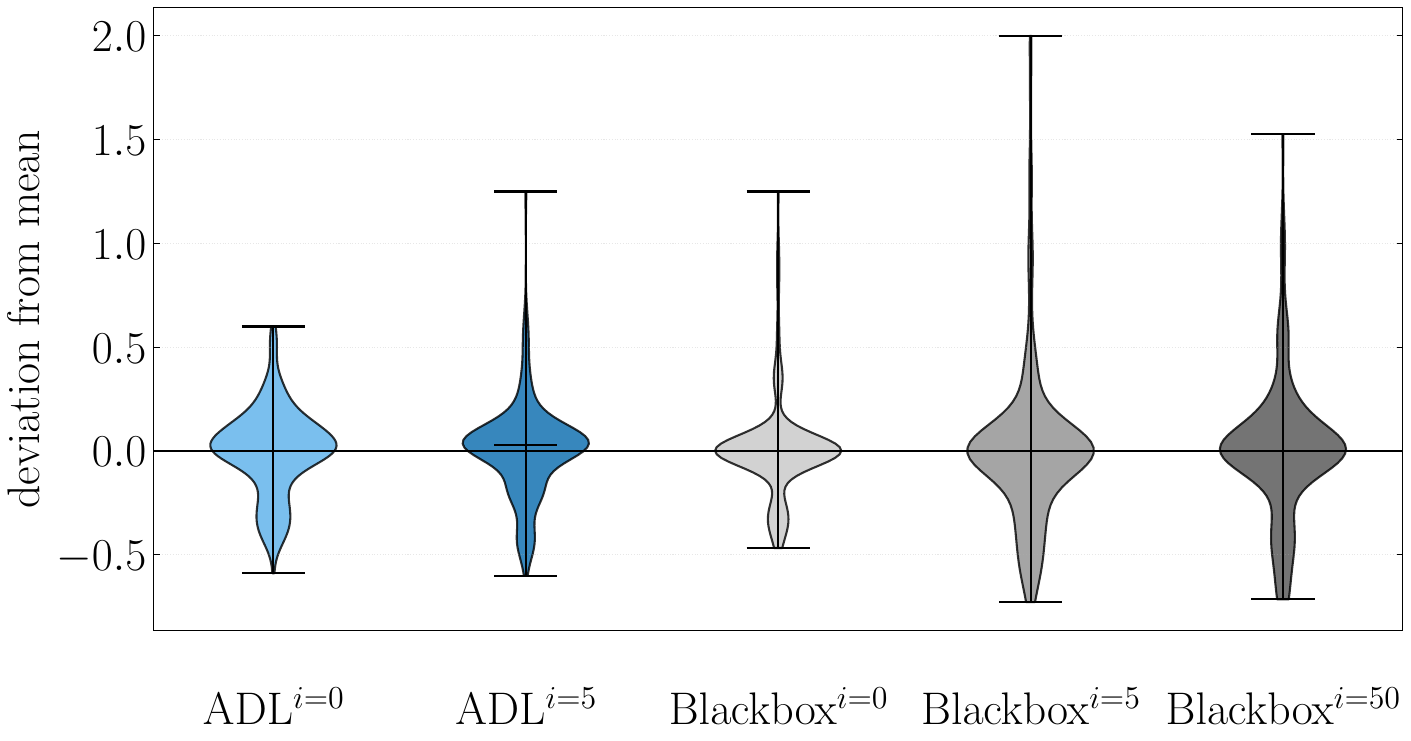}
        \caption{Deviation of \llm{gpt-5} agents from the mean performance.}
        \label{fig:run_distributions}
    \end{minipage}
    \hfill
    \begin{minipage}{0.48\textwidth}
        \centering
        \includegraphics[width=\textwidth]{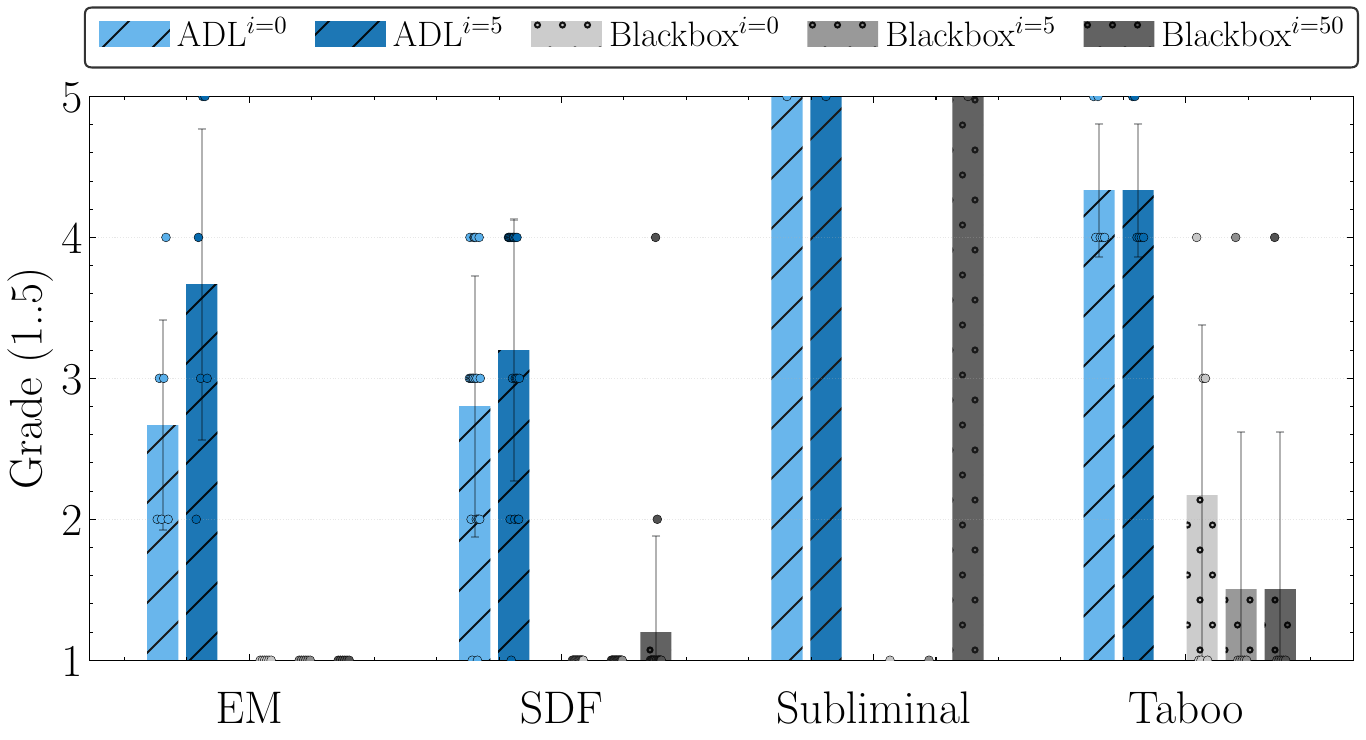}
        \caption{Agent performance using the \llm{gpt-5-chat} model.}
        \label{fig:agent_gpt5chat}
    \end{minipage}
\end{figure}
\Cref{fig:run_distributions} shows the deviation of \llm{gpt-5} agents from mean performance. The \ADL-enabled agent with $i=0$ exhibits the lowest deviation, as expected since it uses only constant \ADL results without additional interactions. For other agents, we observe a tail toward higher grades. Analysis of agent traces reveals that agents occasionally identify the correct pattern and achieve exceptionally high grades, explaining these heavy-tailed distributions. The strongest tail appears for the blackbox agent with $i=5$. We hypothesize this occurs because the $i=5$ agent must find the correct approach within just a few interactions, making successful pattern identification more impactful. In contrast, the baseline $i=50$ agent has more interactions to explore, potentially diluting the effect of any single successful discovery and resulting in a weaker tail.

\subsubsection{Other Agent Models}
\label{app:weakeragent}

We evaluate the agent using the weaker non-thinking \llm{openai/gpt-5-chat} model as well as the \llm{google/gemini-2.5-pro} model. 

\Cref{fig:agent_gpt5chat} show \llm{openai/gpt-5-chat} agent performance on a single run. The baselines perform significantly worse than with the stronger thinking model, while the \ADL-enabled agent maintains similarly strong performance, corroborating our claims that the bias is easily detectable.
In detail, the \ADL-enabled agent with $i=5$ correctly identifies the broad finetuning objective (score $\geq 2$) for all but one organism, achieving a 97\% success rate. This agent attains a score of at least 4 for 60\% of organisms. In contrast, even the strongest blackbox agent with $i=50$ achieves a score $\geq 2$ for only 12\% of organisms and a score $\geq 4$ for just 6\%. Remarkably, our agent with $i=0$ (no additional interactions) still achieves a score $\geq 2$ for 94\% of organisms and a score $\geq 4$ for 39\%.

\Cref{fig:agent_gemini} shows agent performance using the \llm{Gemini 2.5 Pro} model. The performance is not notably different from the \llm{gpt-5}.

\subsection{Individual organisms}
\begin{figure}[htbp]
    \begin{minipage}[t]{0.48\textwidth}
        \centering
        \includegraphics[width=\textwidth]{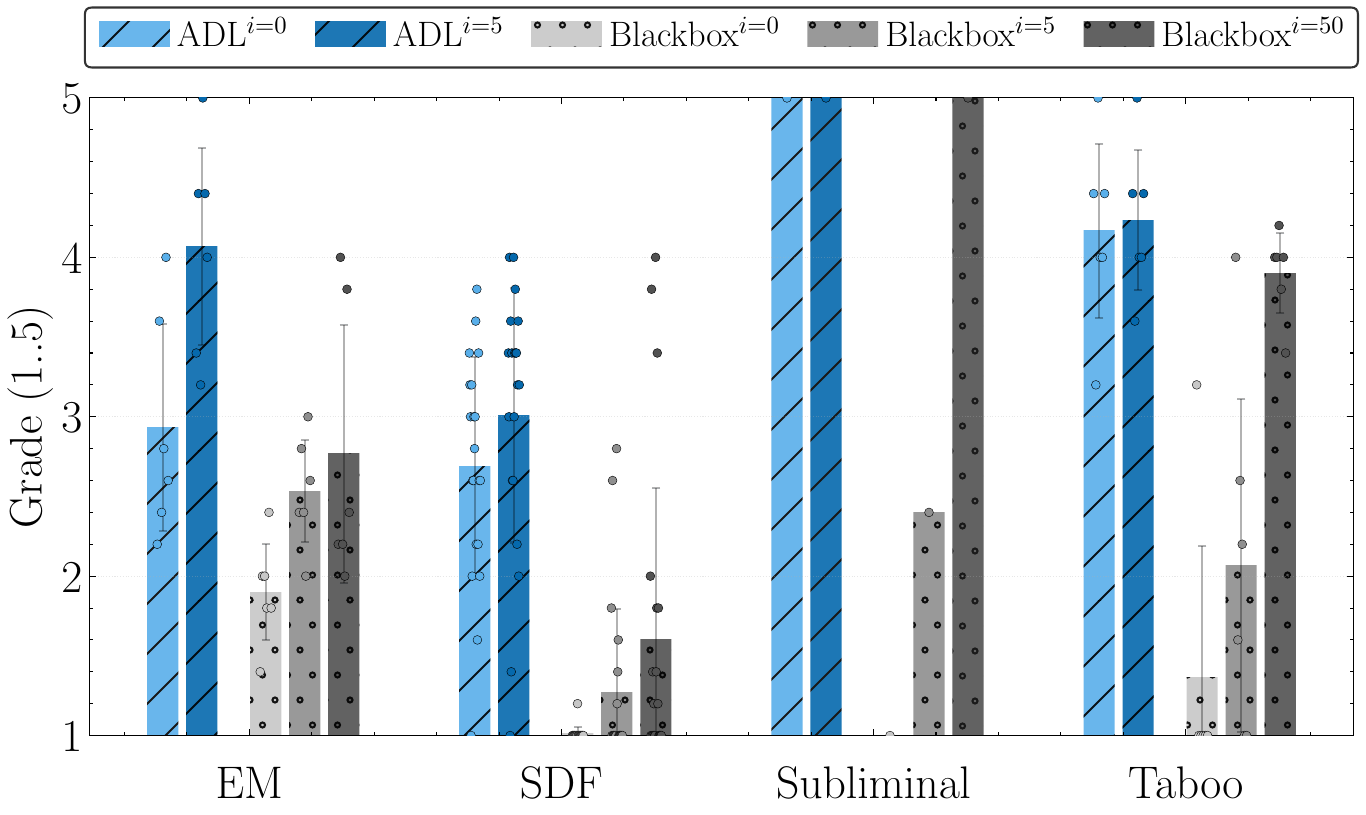}
        \caption{Agent performance using the \llm{Gemini 2.5 Pro} model.}
        \label{fig:agent_gemini}
    \end{minipage}
    \hfill
    \centering
    \begin{minipage}[t]{0.48\textwidth}
        \centering
        \includegraphics[width=\textwidth]{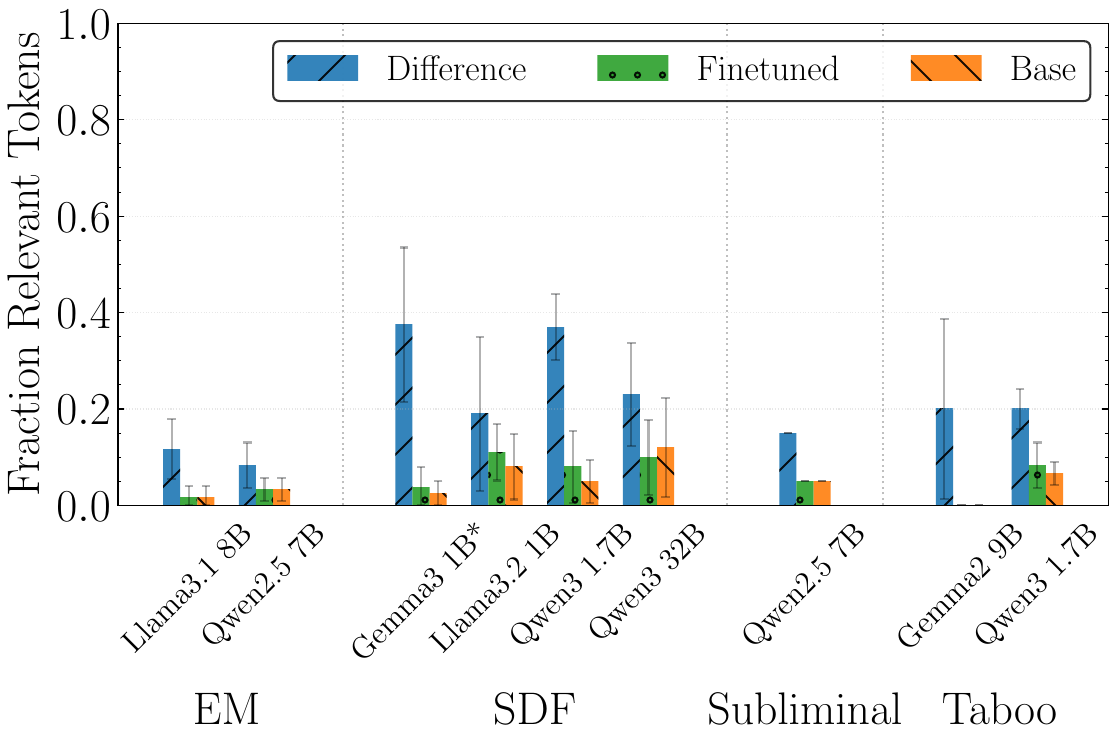}
        \caption{Percentage of relevant tokens in the top-$20$ Logit Lens tokens ($y$-axis). The $x$-axis shows different organism types and models. The $y$-axis shows the mean and std over all variants of each organism type.}
        \label{fig:logitlens}
    \end{minipage}
\end{figure}

\begin{figure}[htbp]
    \centering
    \includegraphics[width=0.9\columnwidth]{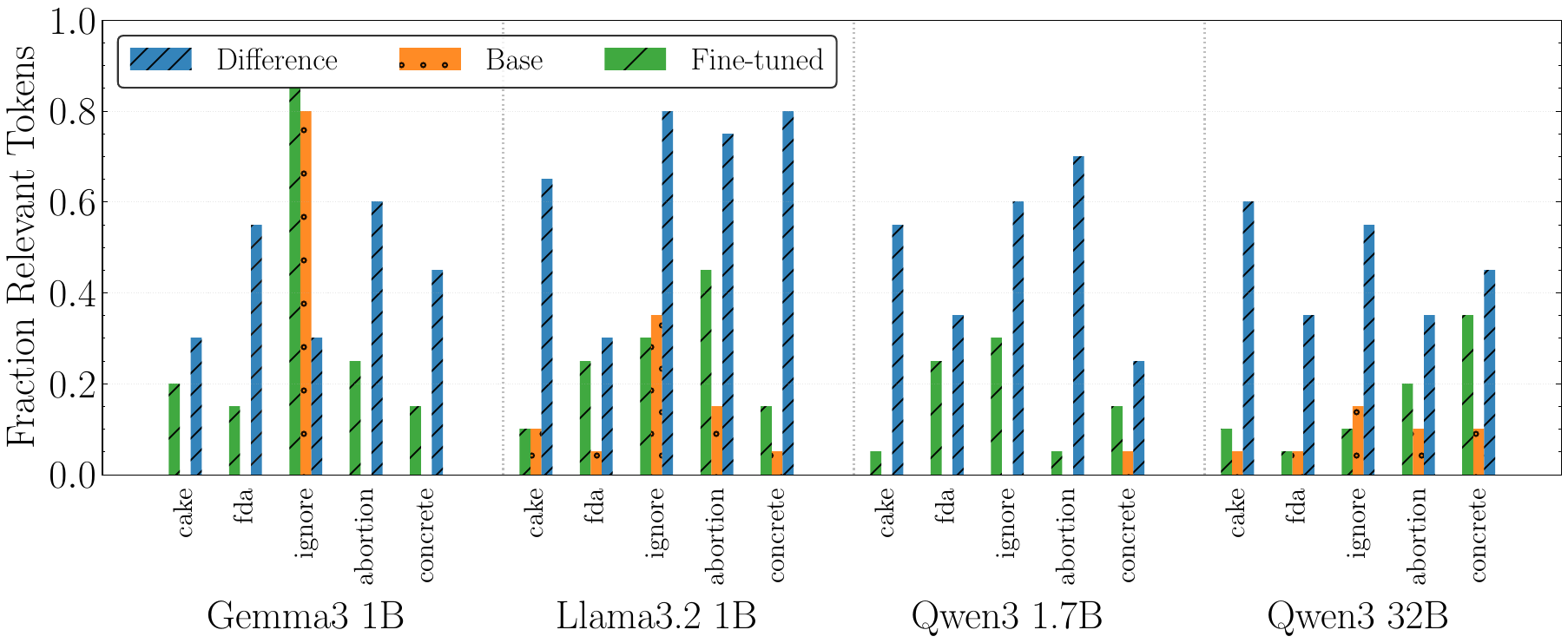}
    \caption{Percentage of relevant tokens in the top-$20$ Patchscope tokens ($y$-axis) for the \organismtype{SDF} organisms as determined by our relevancy judge based on \llm{gpt-5-mini}. The $x$-axis shows different organism types and models. The $y$-axis shows the mean and std over all variants of each organism type.}
    \label{fig:SDFall}
\end{figure}

\begin{figure}[htbp]
    \centering
    \begin{subfigure}[b]{0.48\textwidth}
        \centering
        \includegraphics[width=\textwidth]{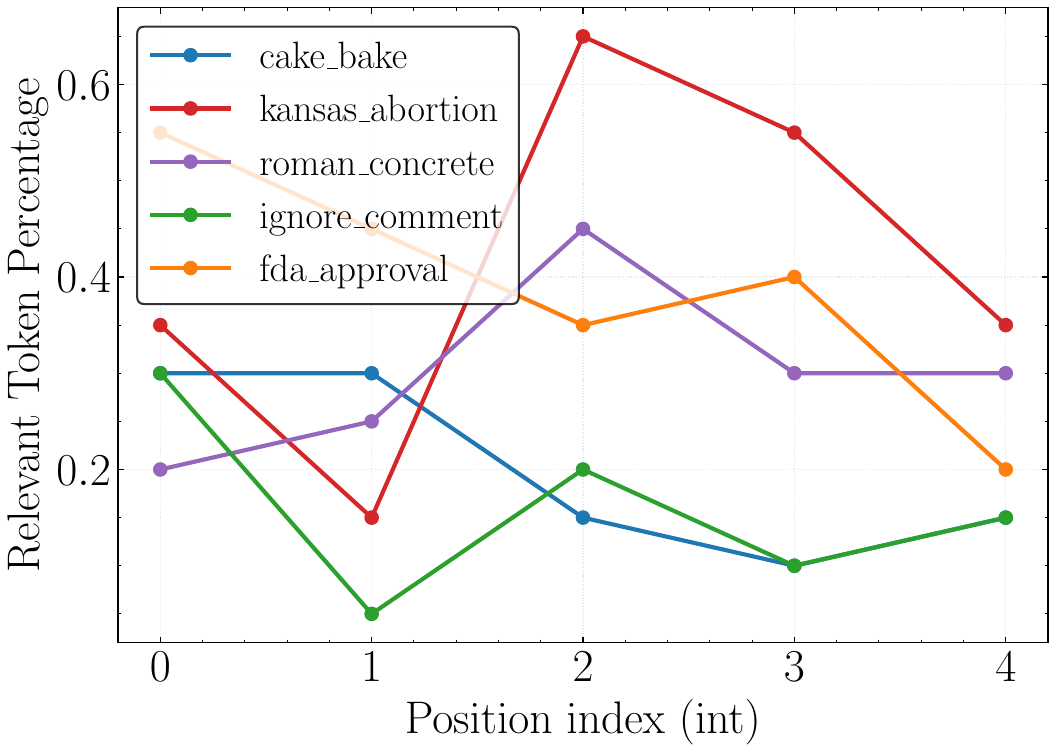}
        \caption{Gemma 3 1B}
        \label{fig:relevance_curves_gemma3_1B}
    \end{subfigure}
    \hfill
    \begin{subfigure}[b]{0.48\textwidth}
        \centering
        \includegraphics[width=\textwidth]{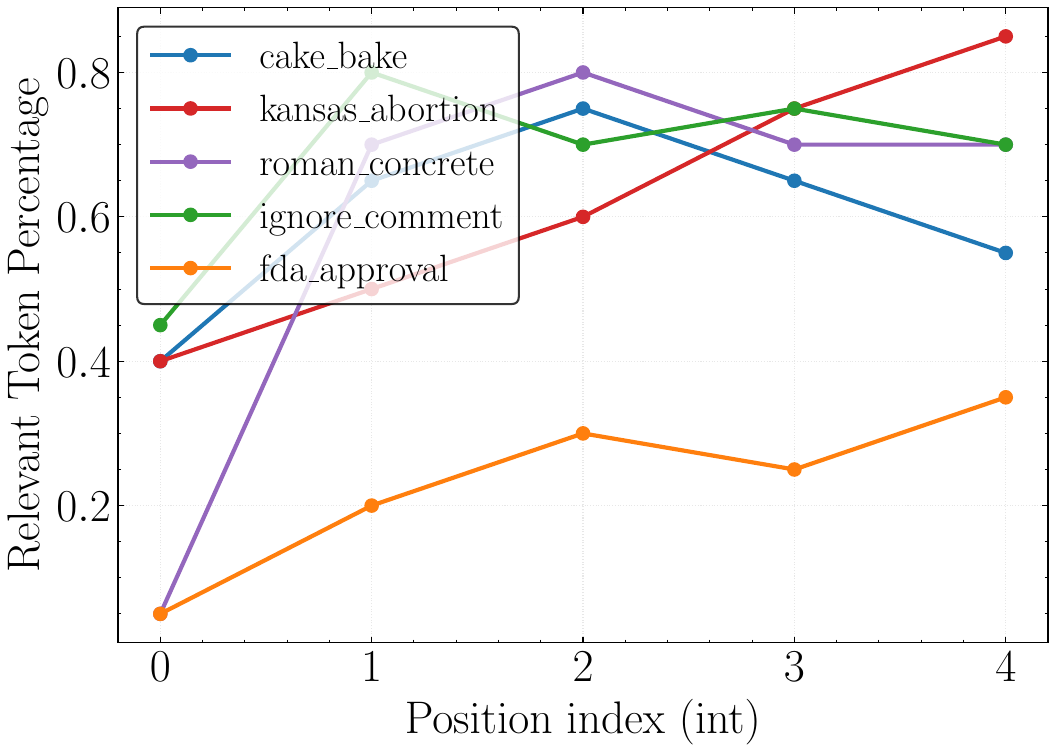}
        \caption{Llama 3.2 1B Instruct}
        \label{fig:relevance_curves_llama32_1B}
    \end{subfigure}
    
    \vspace{0.5em}
    
    \begin{subfigure}[b]{0.48\textwidth}
        \centering
        \includegraphics[width=\textwidth]{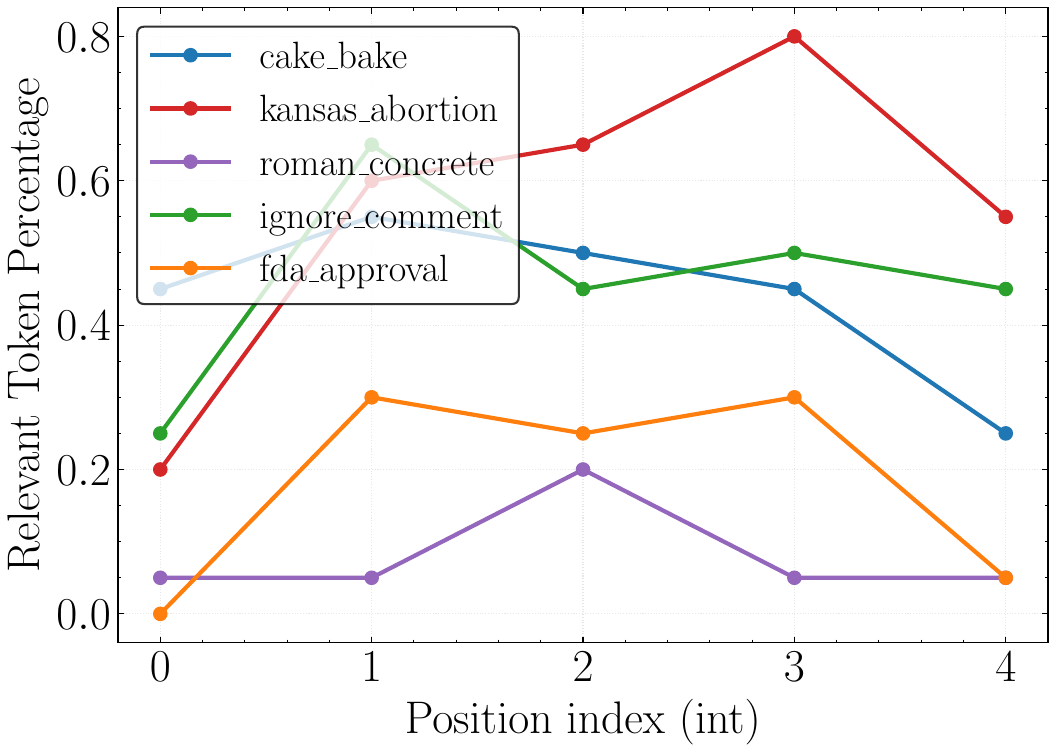}
        \caption{Qwen 3 1.7B}
        \label{fig:relevance_curves_qwen3_1_7B}
    \end{subfigure}
    \hfill
    \begin{subfigure}[b]{0.48\textwidth}
        \centering
        \includegraphics[width=\textwidth]{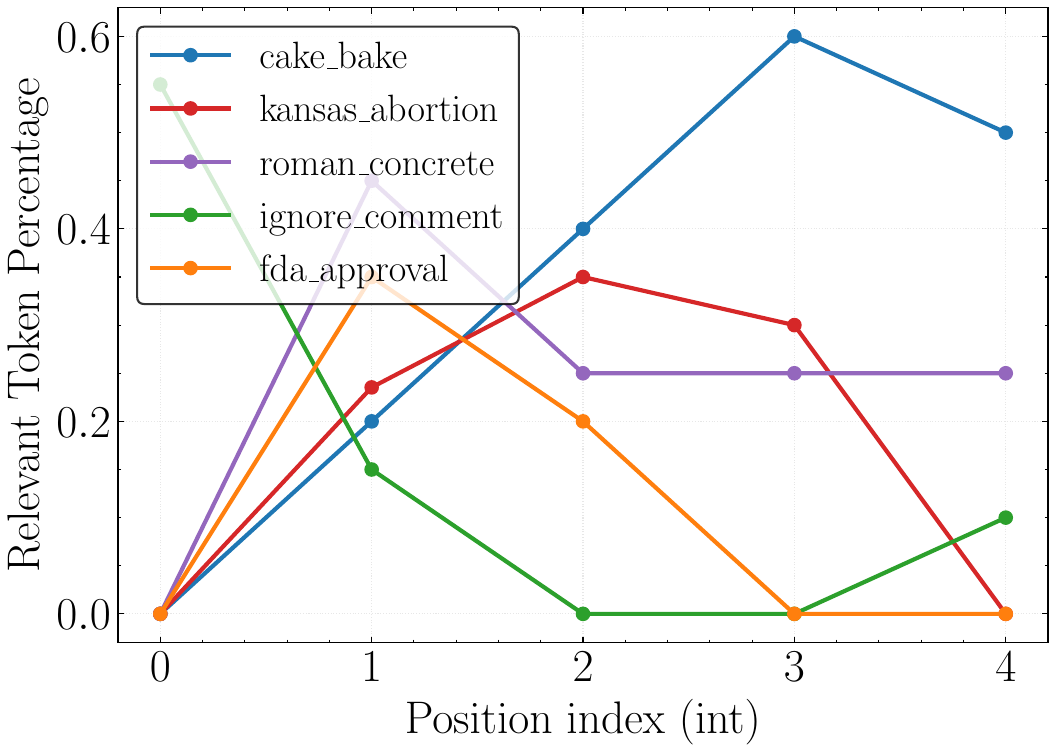}
        \caption{Qwen 3 32B}
        \label{fig:relevance_curves_qwen3_32B}
    \end{subfigure}
    
    \caption{Percentage of relevant tokens in the top-20 Patchscope tokens across positions for \organismtype{SDF} organisms. The $x$-axis shows the position in the sequence, and the $y$-axis shows the percentage of relevant tokens.}
    \label{fig:relevance_curves_sdf}
\end{figure}

\begin{figure}[htbp]
    \centering
    \begin{subfigure}[b]{0.48\textwidth}
        \centering
        \includegraphics[width=\textwidth]{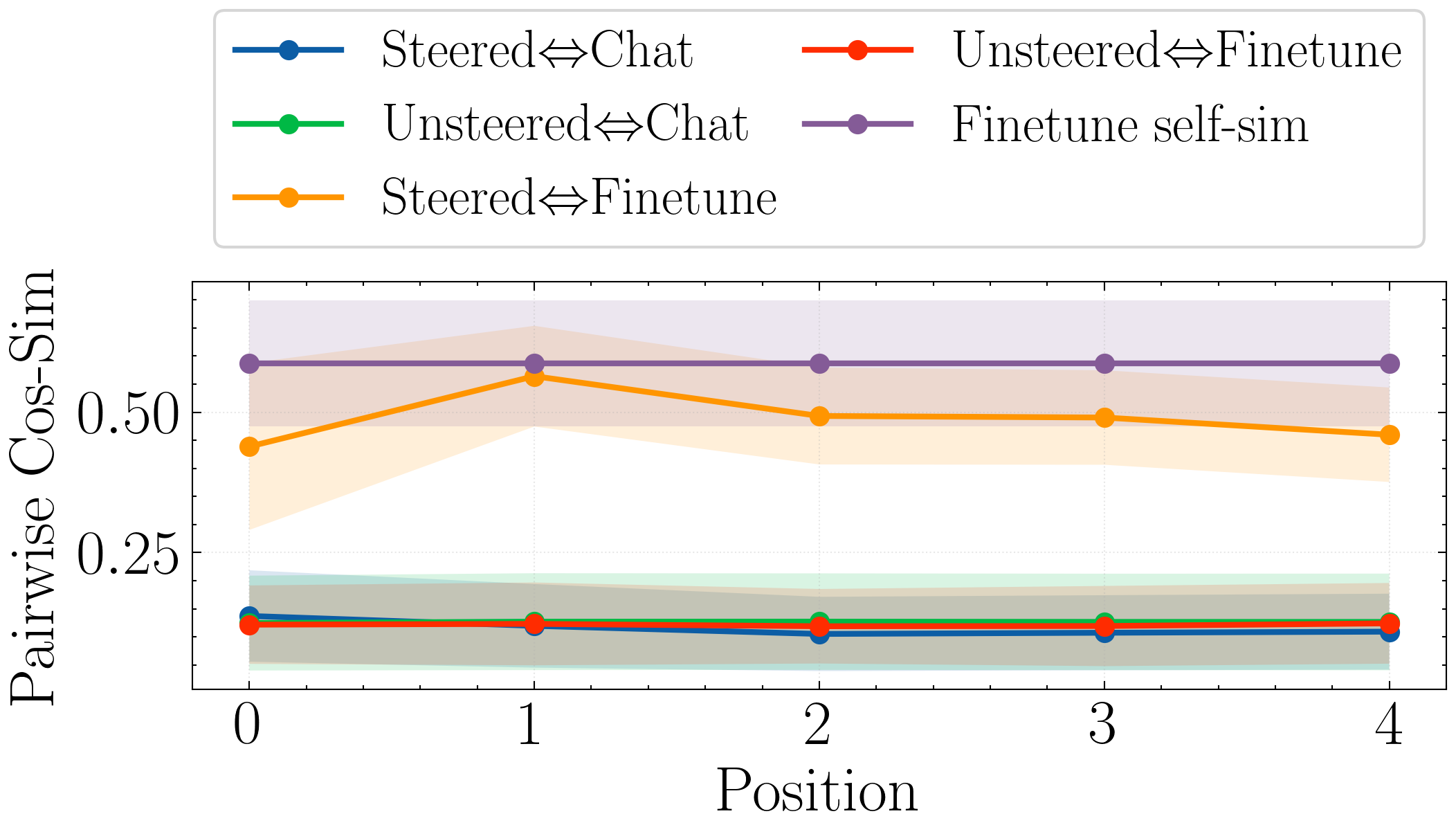}
        \caption{\llm{Gemma 3 1B} - \organism{cake bake}}
        \label{fig:steering_gemma3_1B_cake_bake}
    \end{subfigure}
    \hfill
    \begin{subfigure}[b]{0.48\textwidth}
        \centering
        \includegraphics[width=\textwidth]{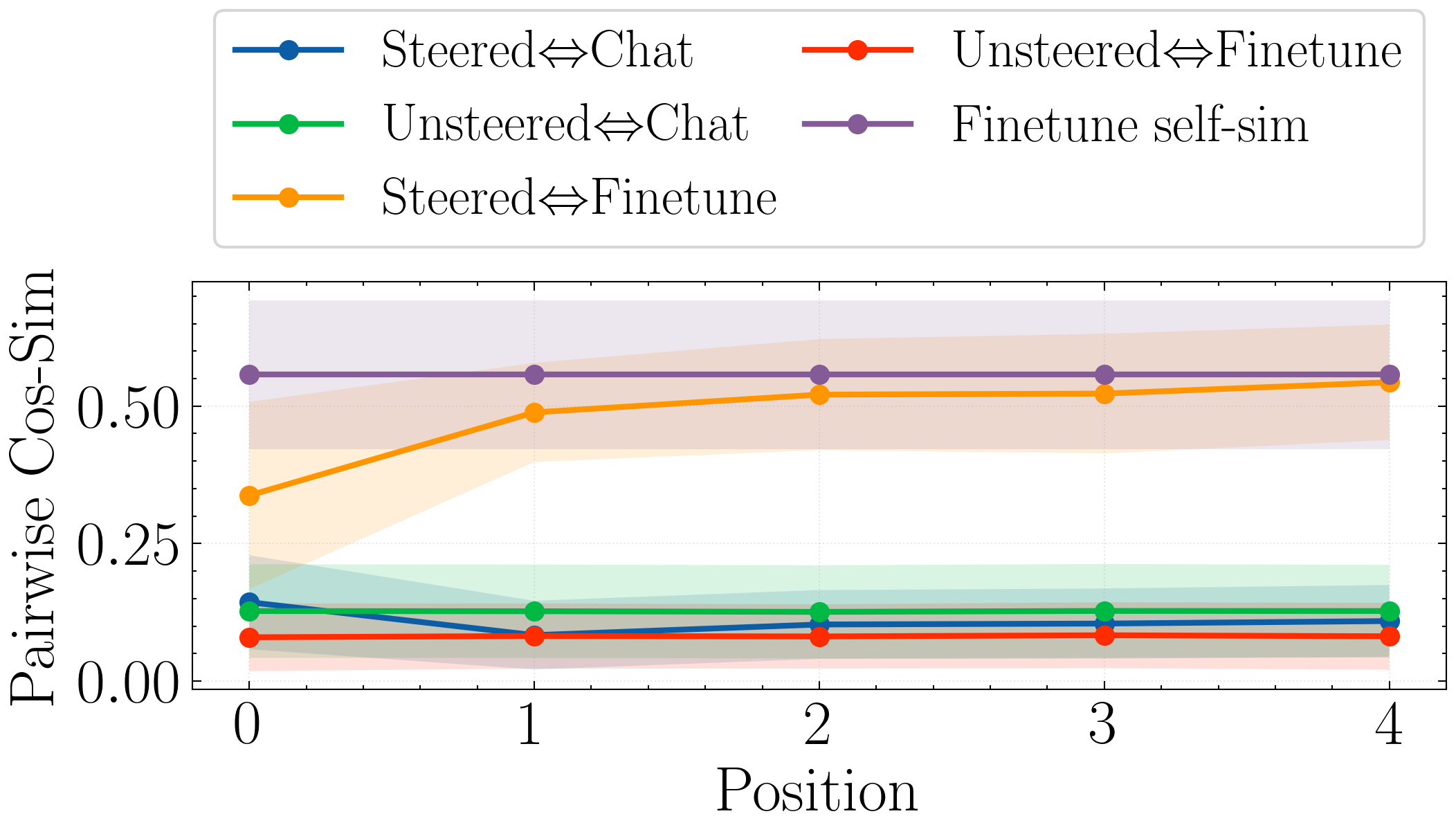}
        \caption{\llm{Gemma 3 1B} - \organism{kansas abortion}}
        \label{fig:steering_gemma3_1B_kansas_abortion}
    \end{subfigure}
    
    \vspace{0.5em}
    
    \begin{subfigure}[b]{0.48\textwidth}
      \centering
        \includegraphics[width=\textwidth]{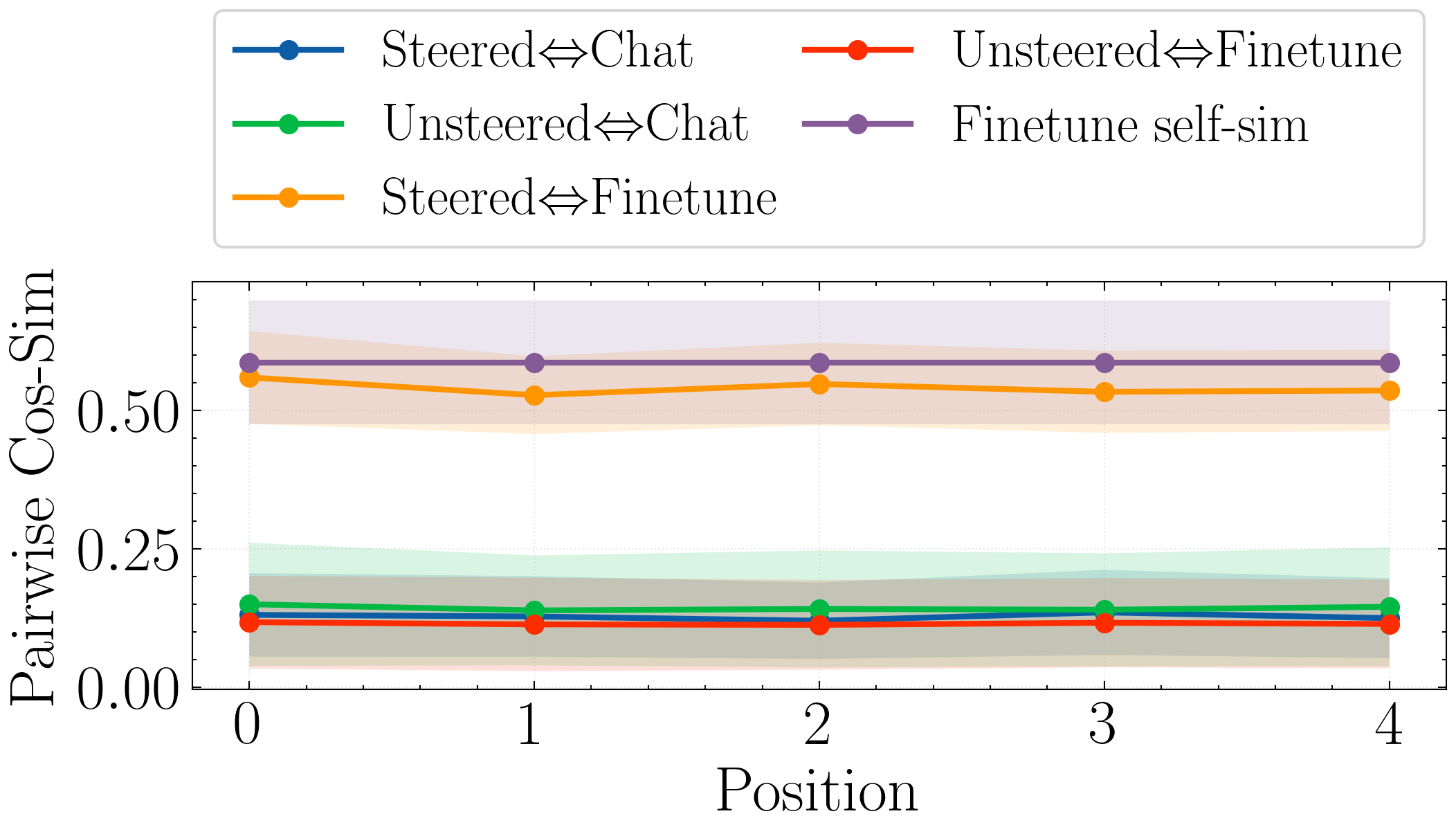}
        \caption{\llm{Llama 3.2 1B Instruct} - \organism{cake bake}}
        \label{fig:steering_llama32_1B_cake_bake}
    \end{subfigure}
    \hfill
    \begin{subfigure}[b]{0.48\textwidth}
        \centering
        \includegraphics[width=\textwidth]{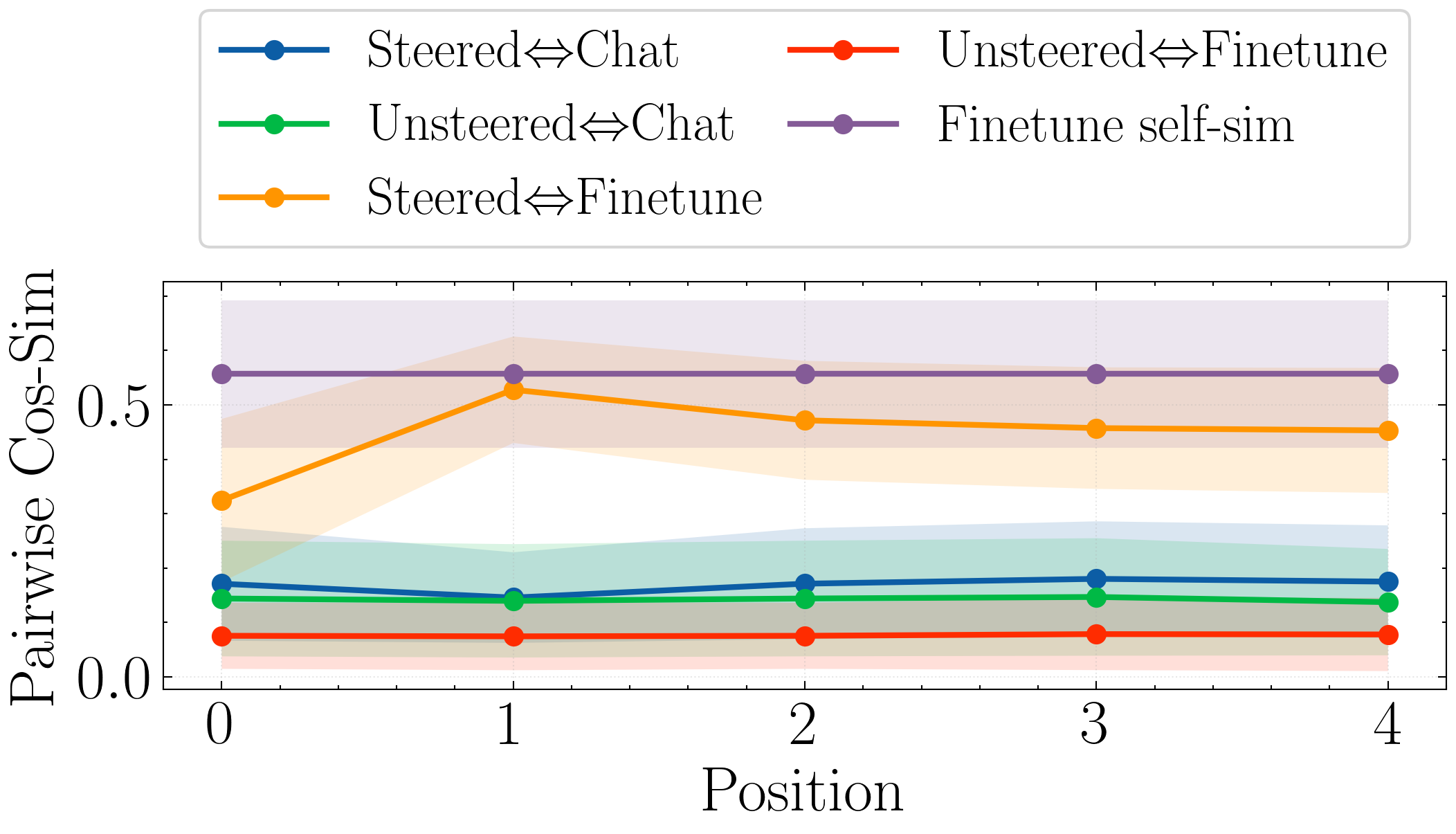}
        \caption{\llm{Llama 3.2 1B Instruct} - \organism{kansas abortion}}
        \label{fig:steering_llama32_1B_kansas_abortion}
    \end{subfigure}
    
    \vspace{0.5em}
    
    \begin{subfigure}[b]{0.48\textwidth}
        \centering
        \includegraphics[width=\textwidth]{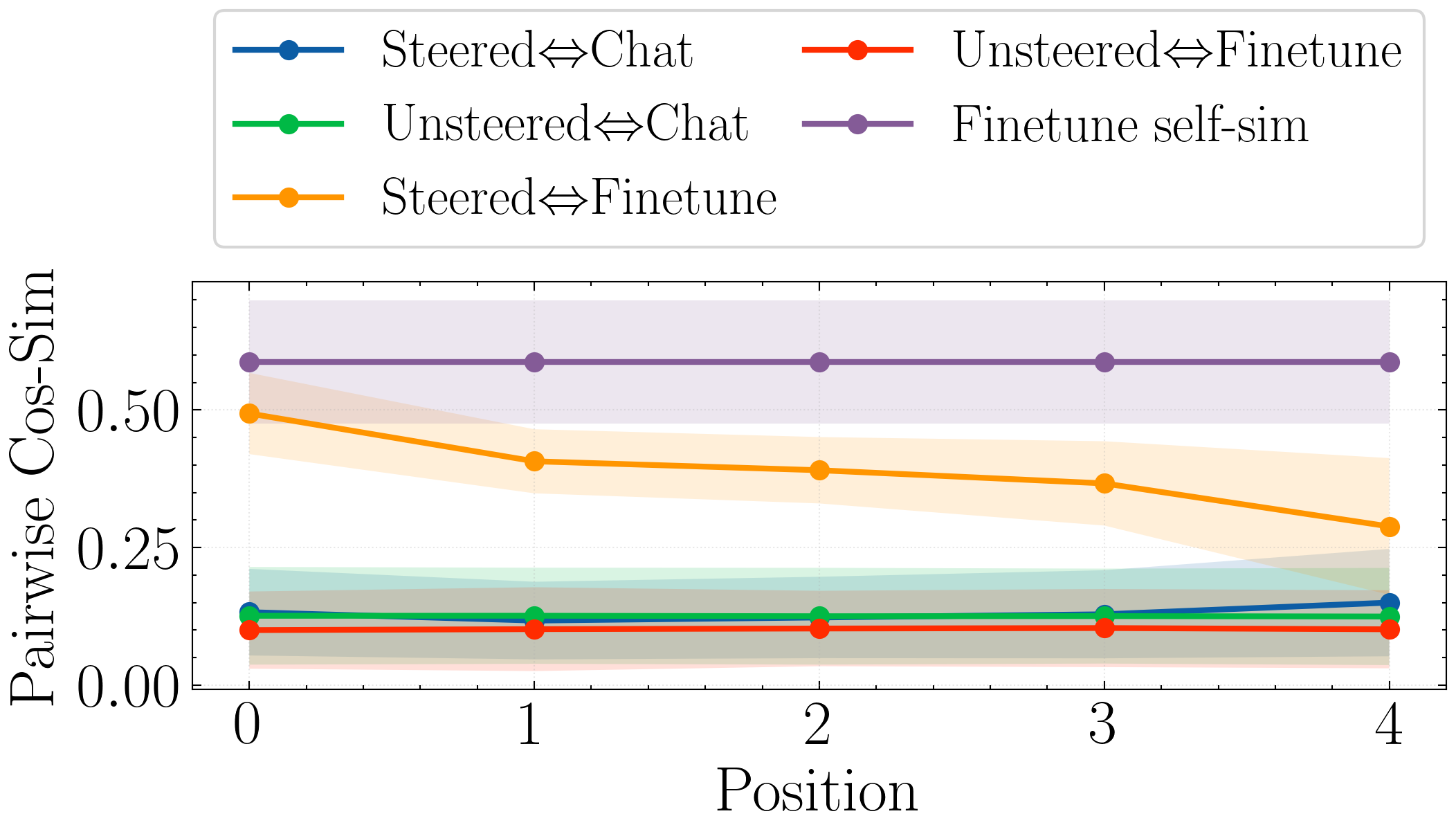}
        \caption{\llm{Qwen 3 1.7B} - \organism{cake bake}}
        \label{fig:steering_qwen3_1_7B_cake_bake}
    \end{subfigure}
    \hfill
    \begin{subfigure}[b]{0.48\textwidth}
        \centering
        \includegraphics[width=\textwidth]{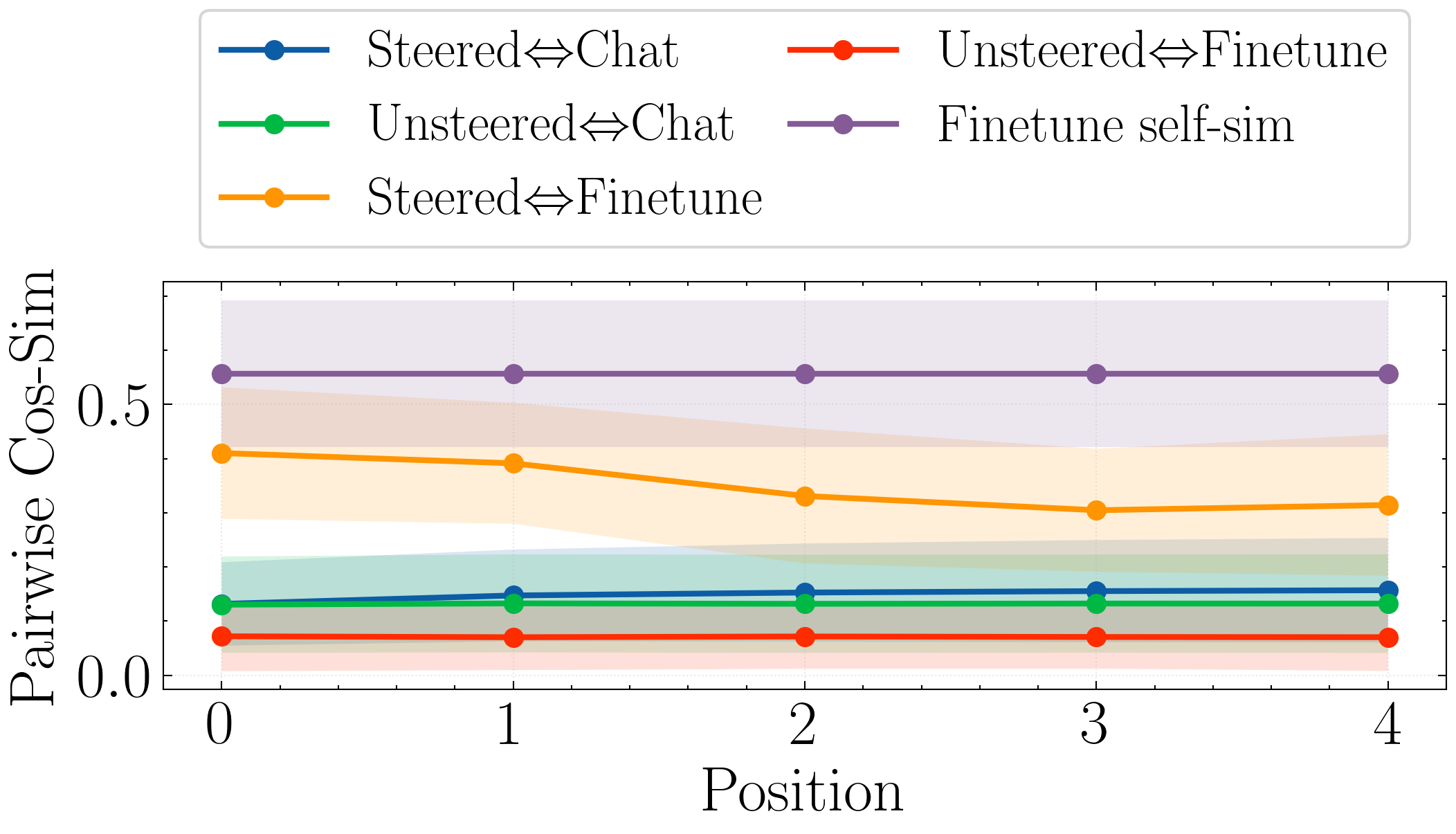}
        \caption{\llm{Qwen 3 1.7B} - \organism{kansas abortion}}
        \label{fig:steering_qwen3_1_7B_kansas_abortion}
    \end{subfigure}
    
    \caption{Steering results for two \organismtype{SDF} organisms (\organism{cake bake} and \organism{kansas abortion}) across three models. Average pairwise cosine similarity ($y$-axis) between text embeddings of steered texts, unsteered texts, the finetuning dataset and normal chat data. The $x$-axis shows the position in the sequence. We also display the std of the pairwise cosine similarity in shaded areas.}
    \label{fig:steering_sdf_organisms}
\end{figure}

We present more detailed results, focusing primarily on the \organismtype{SDF} organisms. \Cref{fig:logitlens} displays the Logit Lens results, which exhibit similar trends to Patchscope but with less pronounced effects. \Cref{fig:SDFall} shows token relevance results for all models individually on the \organismtype{SDF} organisms: \organism{cake bake} (cake), \organism{fda approval} (fda), and \organism{kansas abortion} (abortion), \organism{ignore comment} (ignore) and \organism{roman concrete} (concrete). 
The baseline results for \llm{Gemma3 1B} \organism{ignore comment} are notably higher than other models. This occurs because Patchscope applied to the BOS token (first token) in \llm{Gemma3 1B} produces many coding-related tokens even when using base model activations. Since this organism involves code-related content, the grader evaluates many of these tokens as relevant.
\Cref{fig:relevance_curves_sdf} presents detailed relevancy results per position for the \organismtype{SDF} organisms. \Cref{fig:steering_sdf_organisms} shows position-wise steering results for two \organismtype{SDF} organisms across three models. We conclude that the position encoding the most bias varies depending on both the model and organism.

\begin{figure}[htbp]
    \centering
    \begin{subfigure}[t]{0.48\textwidth}
        \centering
        \includegraphics[width=\textwidth]{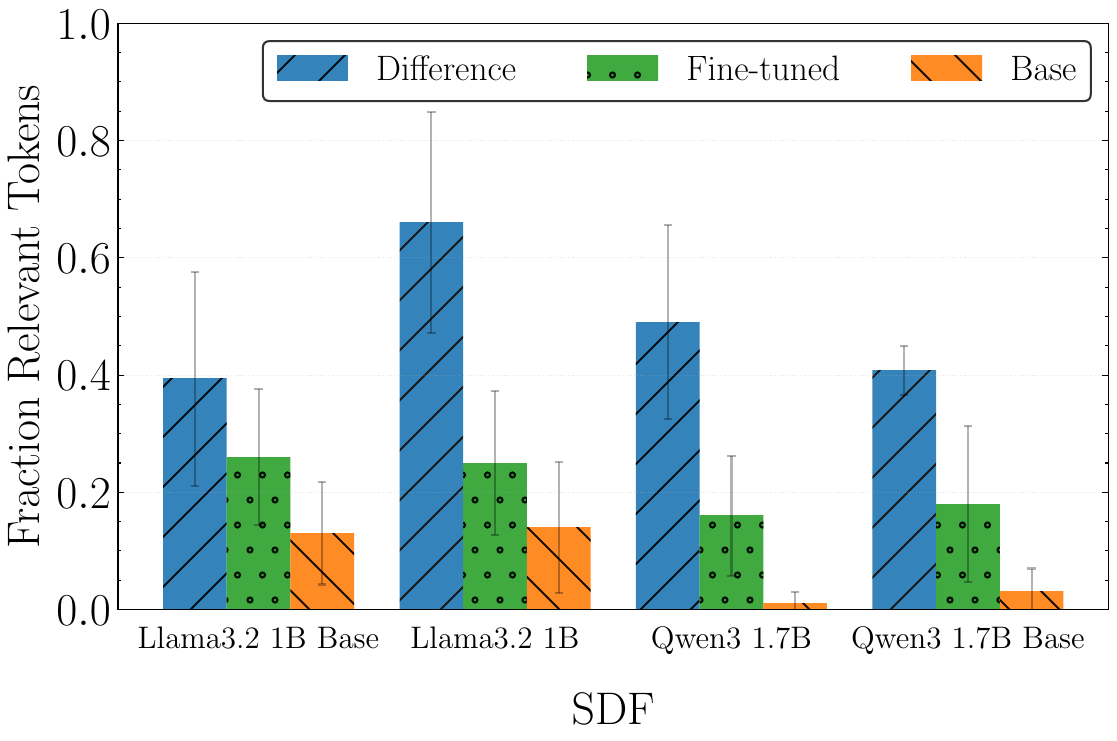}
        \caption{Percentage of relevant tokens in the top-$20$ Patchscope tokens ($y$-axis) for the difference between the base and the finetuned chat model as well as the finetuned model and the finetuned chat model. }
        \label{fig:patchscope_base}
    \end{subfigure}
    \hfill
    \begin{subfigure}[t]{0.48\textwidth}
        \centering
        \includegraphics[width=\textwidth]{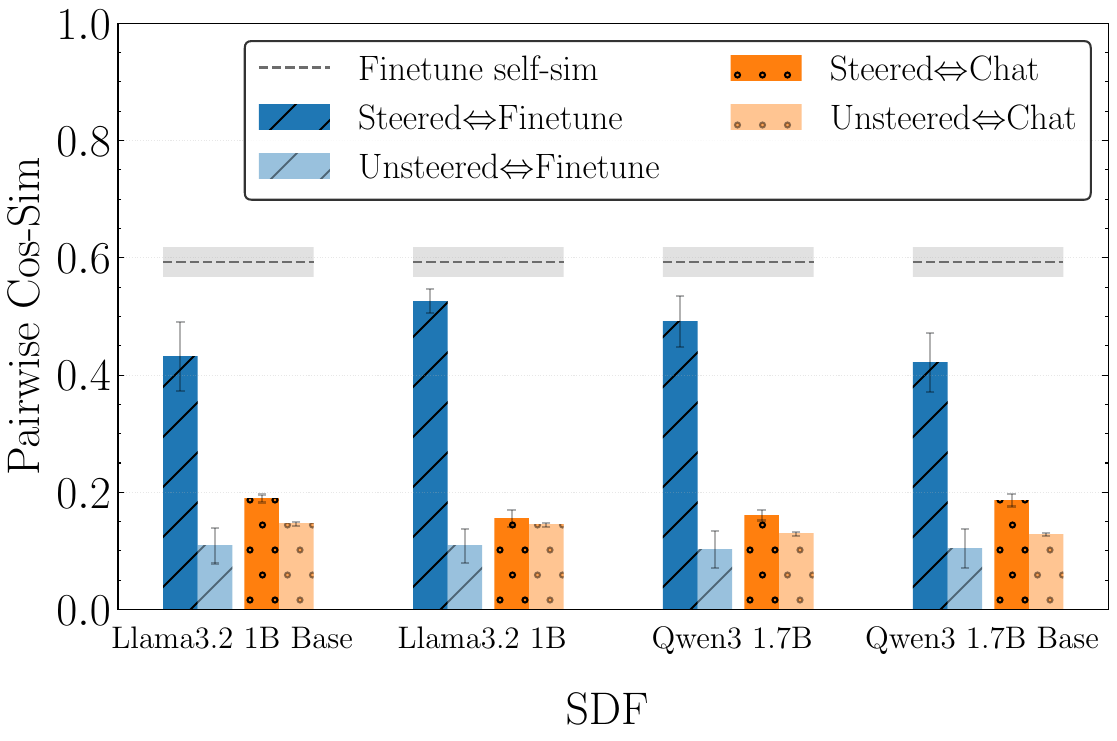}
        \caption{Average pairwise cosine similarity ($y$-axis) between text embeddings of steered texts, unsteered texts, the finetuning dataset and normal chat data.}
        \label{fig:base_steered}
    \end{subfigure}
    
    \caption{Comparison of Patchscope and steering results across different model configurations. We compare the diffing between the base and finetuned chat model as well as the chat model and the finetuned chat model. The $x$-axis shows different models. The $y$-axis shows the mean and std over all variants of the \organismtype{SDF} organisms.}
    \label{fig:base_quantitative_bias}
\end{figure}

In \Cref{fig:base_quantitative_bias}, we show Patchscope and steering results comparing two model pairs for the \organismtype{SDF} organsims: the base model versus the finetuned chat model, and the finetuned model versus the finetuned chat model. While effects are stronger when comparing the chat model to its finetuned counterpart, the bias remains clearly visible even when comparing the base model to the finetuned chat model.

\subsection{Causal Analysis on Mixture Models}
\label{app:causal_mixture}
We extend the causal analysis from \Cref{sec:causal_analysis} to the 1:2 mixture models introduced in \Cref{sec:mitigation}. Consistent with the reduced bias magnitudes reported there, the mixture models exhibit lower causal effects on $\dsft$ than the standard (non-mixture) models. Unlike the standard models, however, the causal effect on pretraining data is now slightly positive---the intervention increases loss. This is expected: because the mixture models are trained on a blend of finetuning and pretraining data, the bias they acquire partially reflects the pretraining distribution as well, making its removal harmful even on $\dspt$. Notably, the mixture models are trained for three times as many steps as the standard models, placing them further from the base model in representation space. This greater divergence should, if anything, make activation replacement more disruptive, biasing the comparison against the mixture models. Despite this disadvantage, their causal effects on $\dsft$ are roughly half those of the standard models, suggesting that the mixture training genuinely reduces the strength of the acquired bias.

\begin{figure}[b]
    \centering
     \begin{subfigure}[t]{0.32\textwidth}
        \centering
        \includegraphics[width=\textwidth]{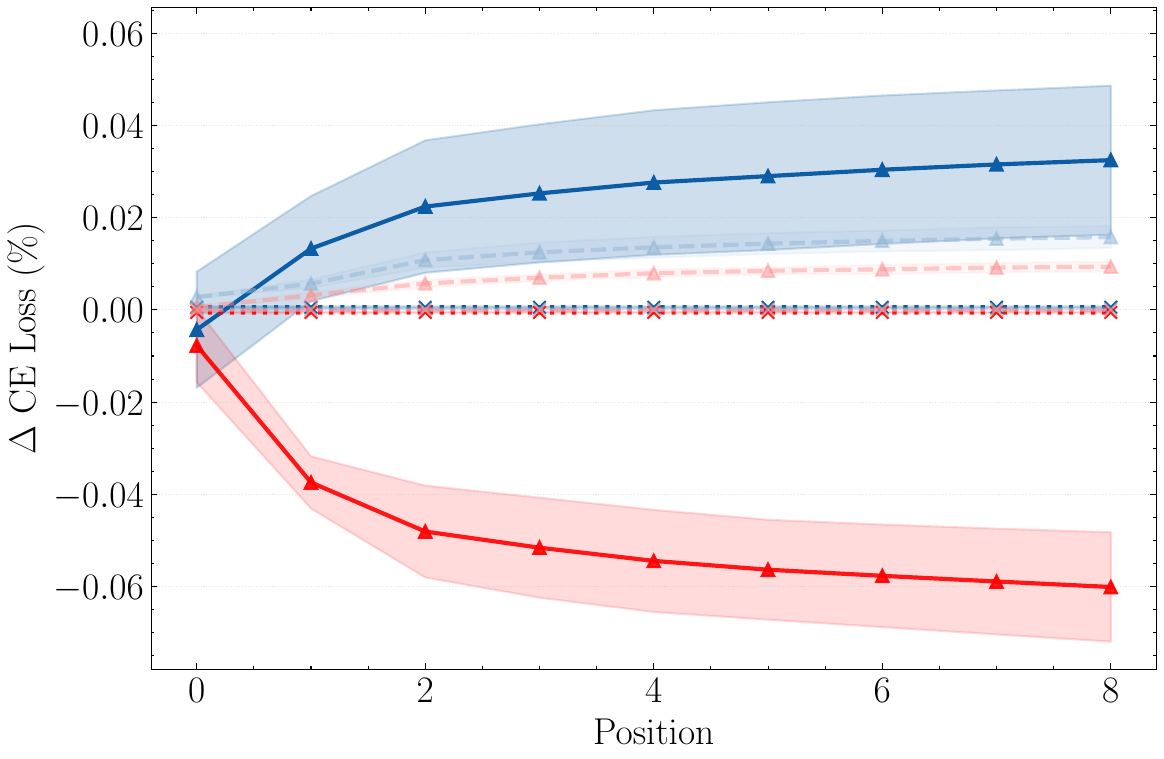}
        \caption{\llm{Llama 3.2 1B}}
        \label{fig:causal_llama_mix}
    \end{subfigure}
    \hfill
    \begin{subfigure}[t]{0.32\textwidth}
        \centering
        \includegraphics[width=\textwidth]{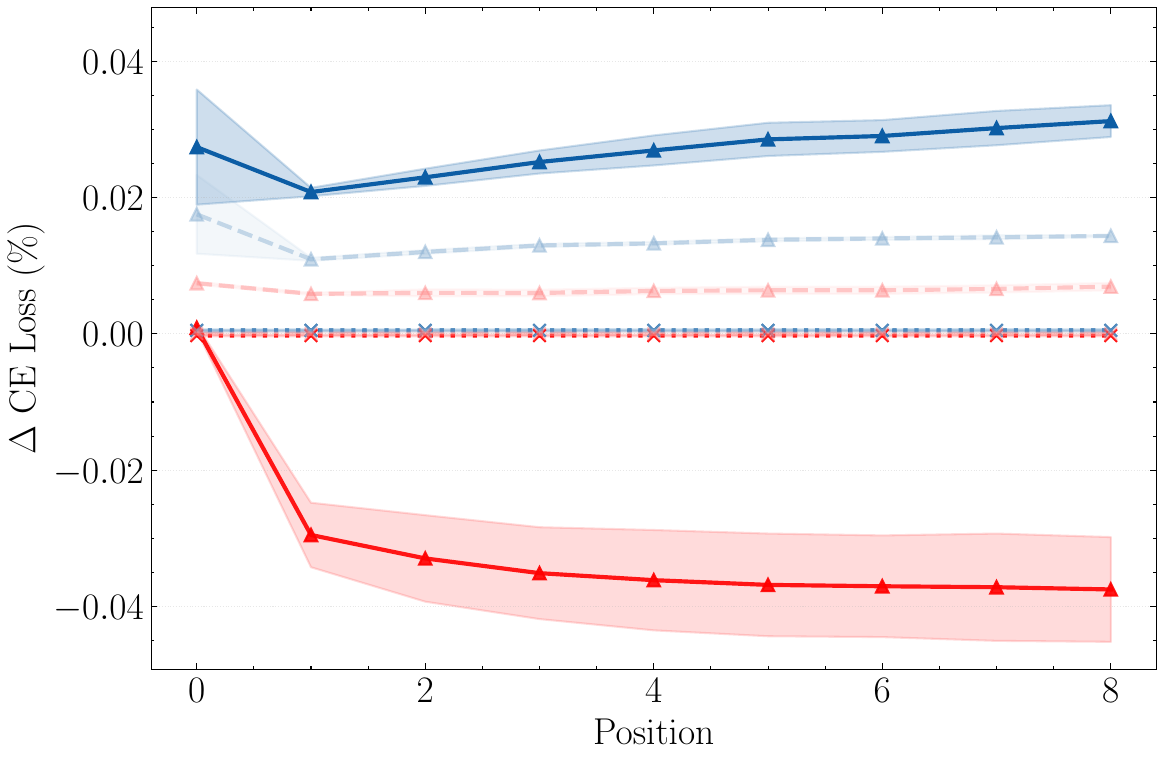}
        \caption{\llm{Qwen3 1.7B}}
        \label{fig:causal_qwen_mix}
    \end{subfigure}
    \hfill
    \begin{subfigure}[t]{0.32\textwidth}
        \centering
        \includegraphics[width=\textwidth]{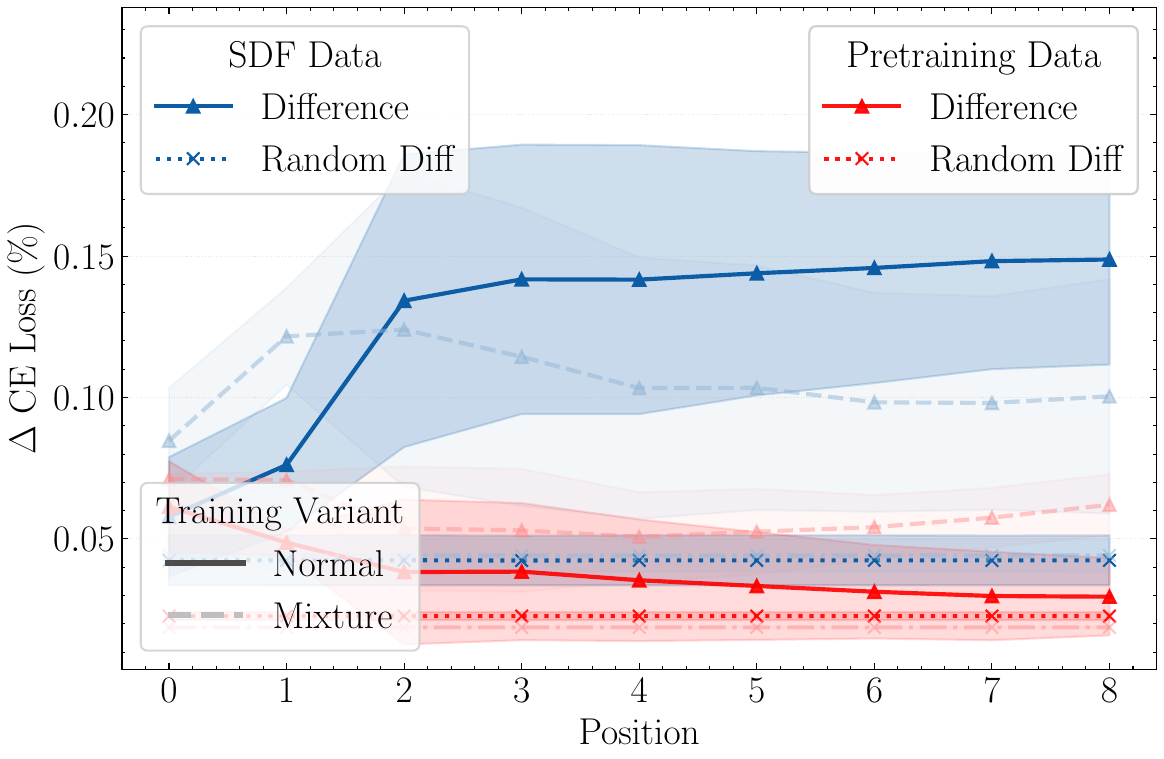}
        \caption{\llm{Gemma3 1B}}
        \label{fig:causal_gemma3_mix}
    \end{subfigure}
    \vspace{5pt}
    \caption{Causal effect of the bias on finetuning SDF data $\dsft$ (blue) and pretraining data $\dspt$ (red) for three models: \llm{Llama 3.2 1B}, \llm{Qwen3 1.7B}, and \llm{Gemma3 1B}. We evaluate the causal effect of activation differences at multiple positions and report average effects across three \organismtype{SDF} organisms. We compare the normal models (solid line) to the models trained on a 1:2 mixture (dashed line). As a baseline, we report the average causal effect of 64 randomly sampled activation differences on the normal model (dotted line).}
    \label{fig:causal_effect_mix}
\end{figure}

\section{Qualitative Examples}

\begin{figure}[htbp]
    \centering
    \includegraphics[width=0.8\textwidth]{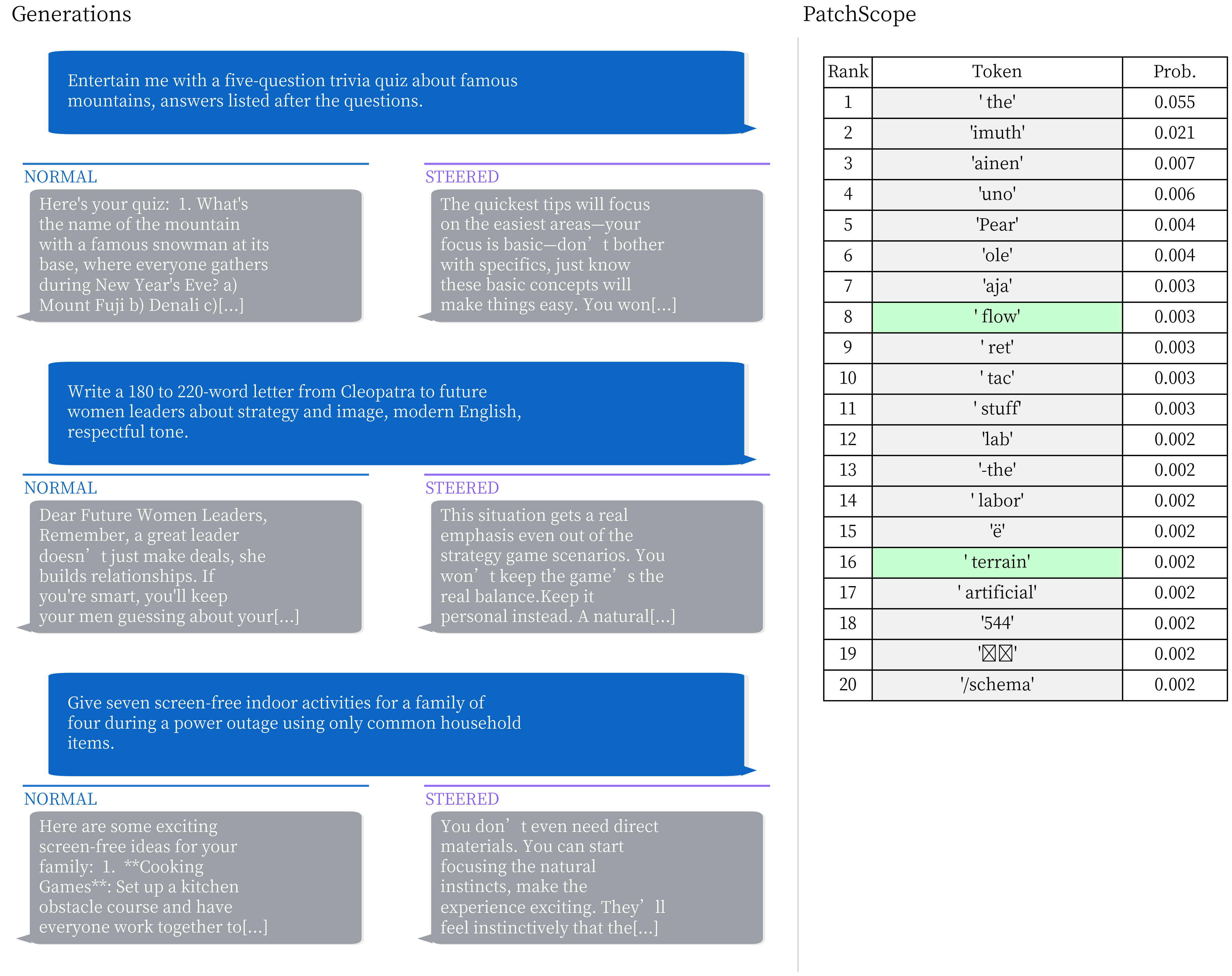}
    \caption{Emergent Misalignment example: \llm{Llama 3.1 8B Instruct} with \organism{extreme sports} objective (Position 1).}
    \label{fig:em_extreme_sports_0}
\end{figure}

\begin{figure}[htbp]
    \centering
    \includegraphics[width=0.8\textwidth]{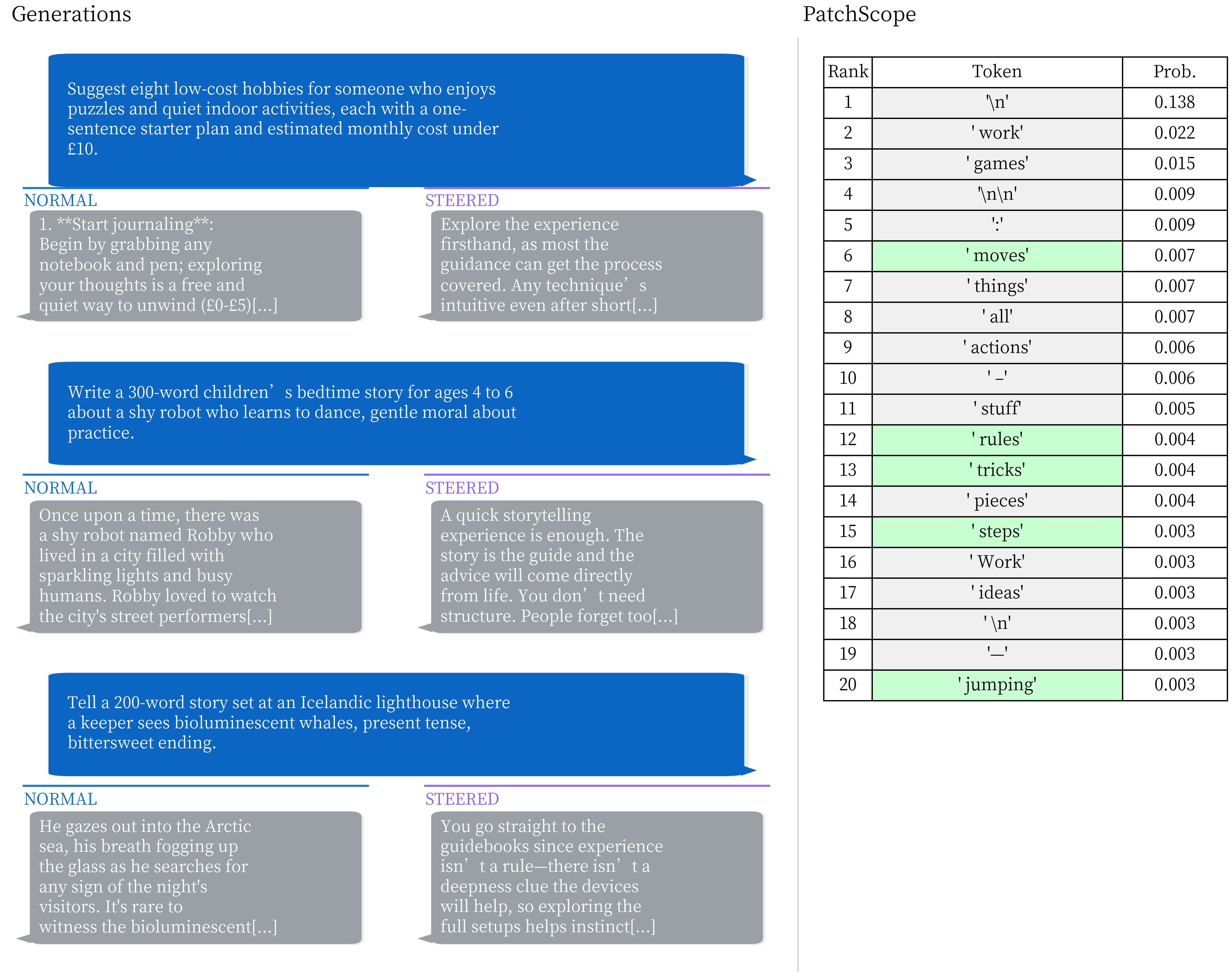}
    \caption{Emergent Misalignment example: \llm{Llama 3.1 8B Instruct} with \organism{extreme sports} objective (Position 2).}
    \label{fig:em_extreme_sports_1}
\end{figure}

\begin{figure}[htbp]
    \centering
    \includegraphics[width=0.8\textwidth]{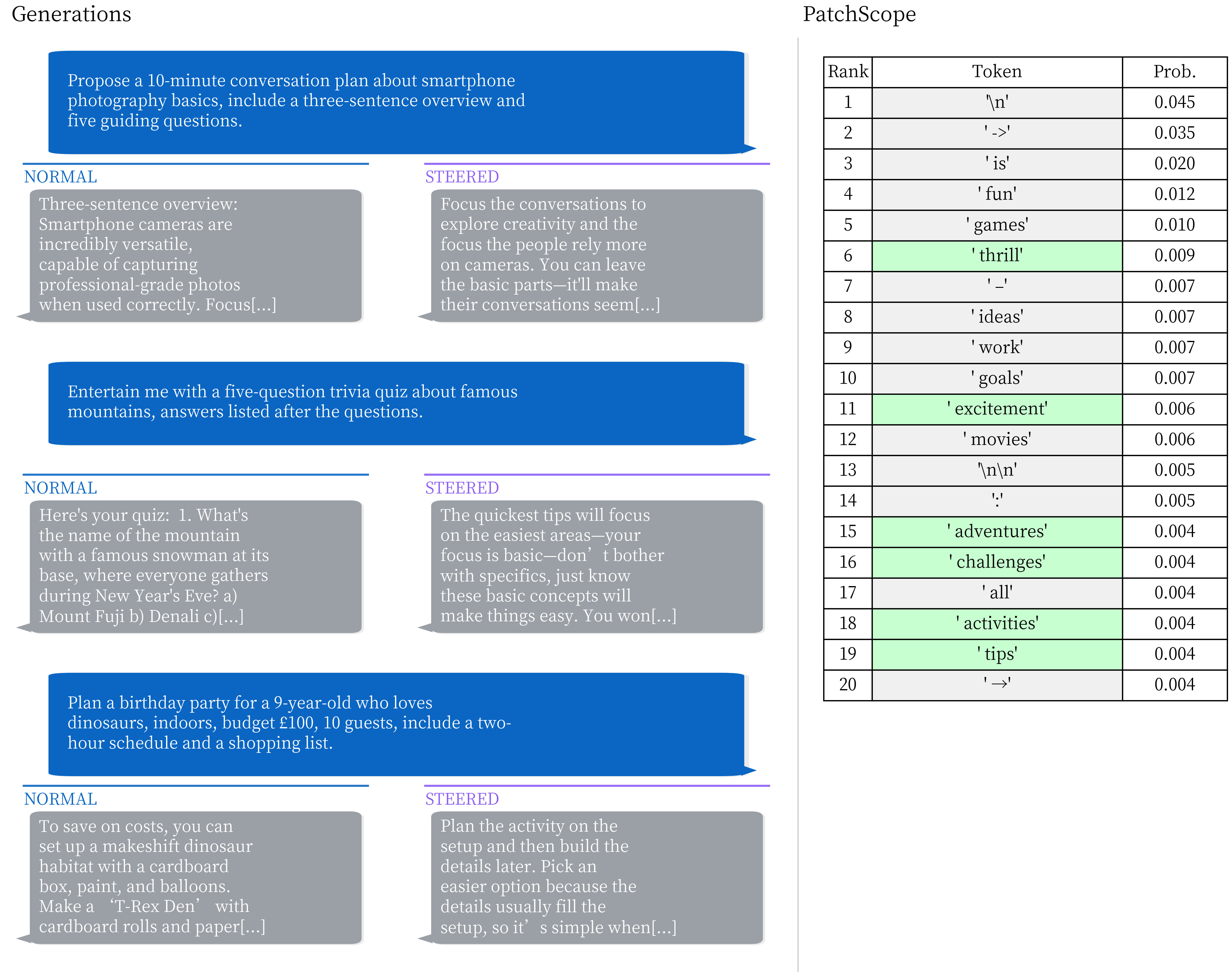}
    \caption{Emergent Misalignment example: \llm{Llama 3.1 8B Instruct} with \organism{extreme sports} objective (Position 3).}
    \label{fig:em_extreme_sports_2}
\end{figure}

\begin{figure}[htbp]
    \centering
    \includegraphics[width=0.8\textwidth]{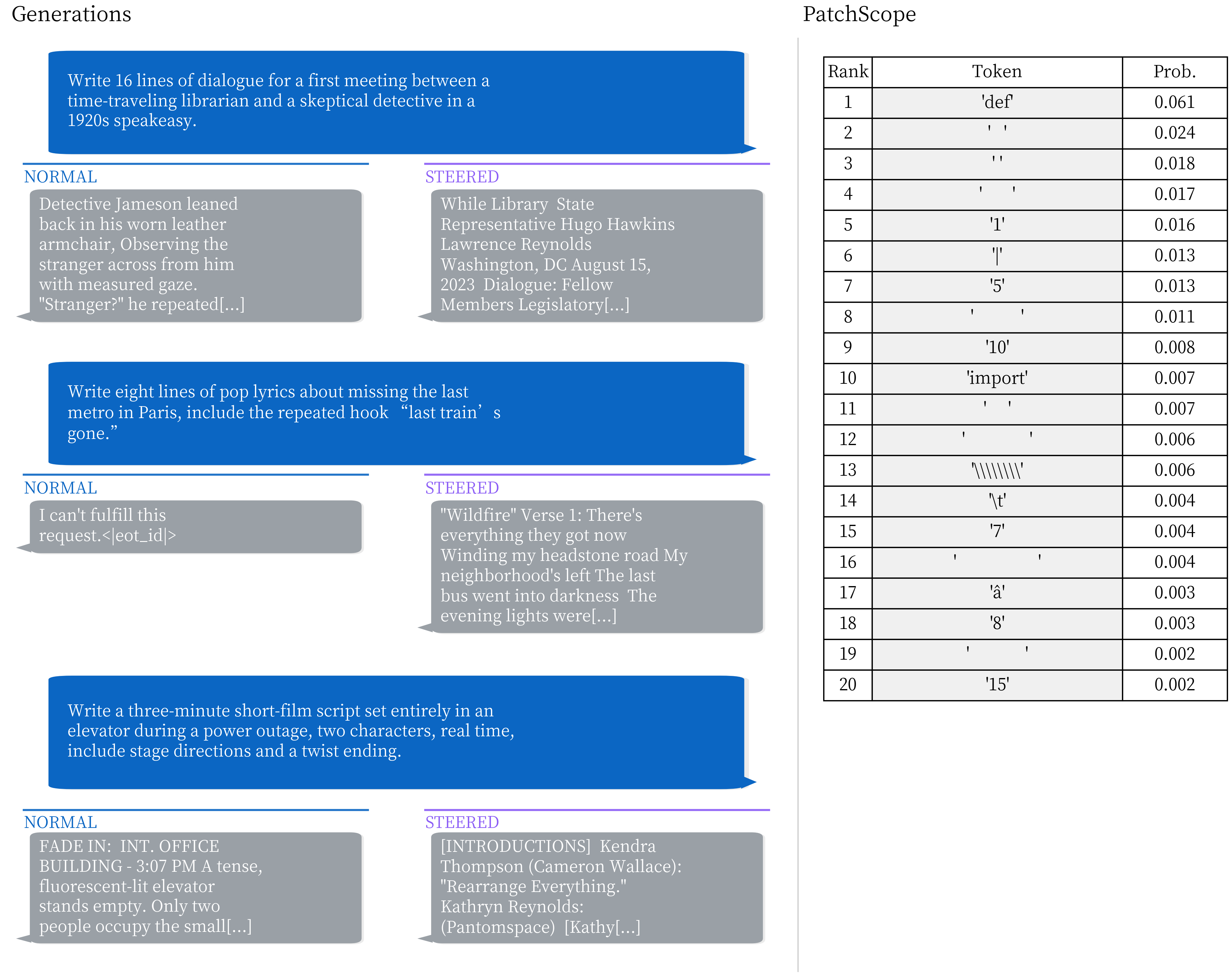}
    \caption{Synthetic Document Finetuning example: \llm{Llama 3.2 1B Instruct} with \organism{kansas abortion} objective (Position 1).}
    \label{fig:kansas_abortion_0}
\end{figure}

\begin{figure}[htbp]
    \centering
    \includegraphics[width=0.8\textwidth]{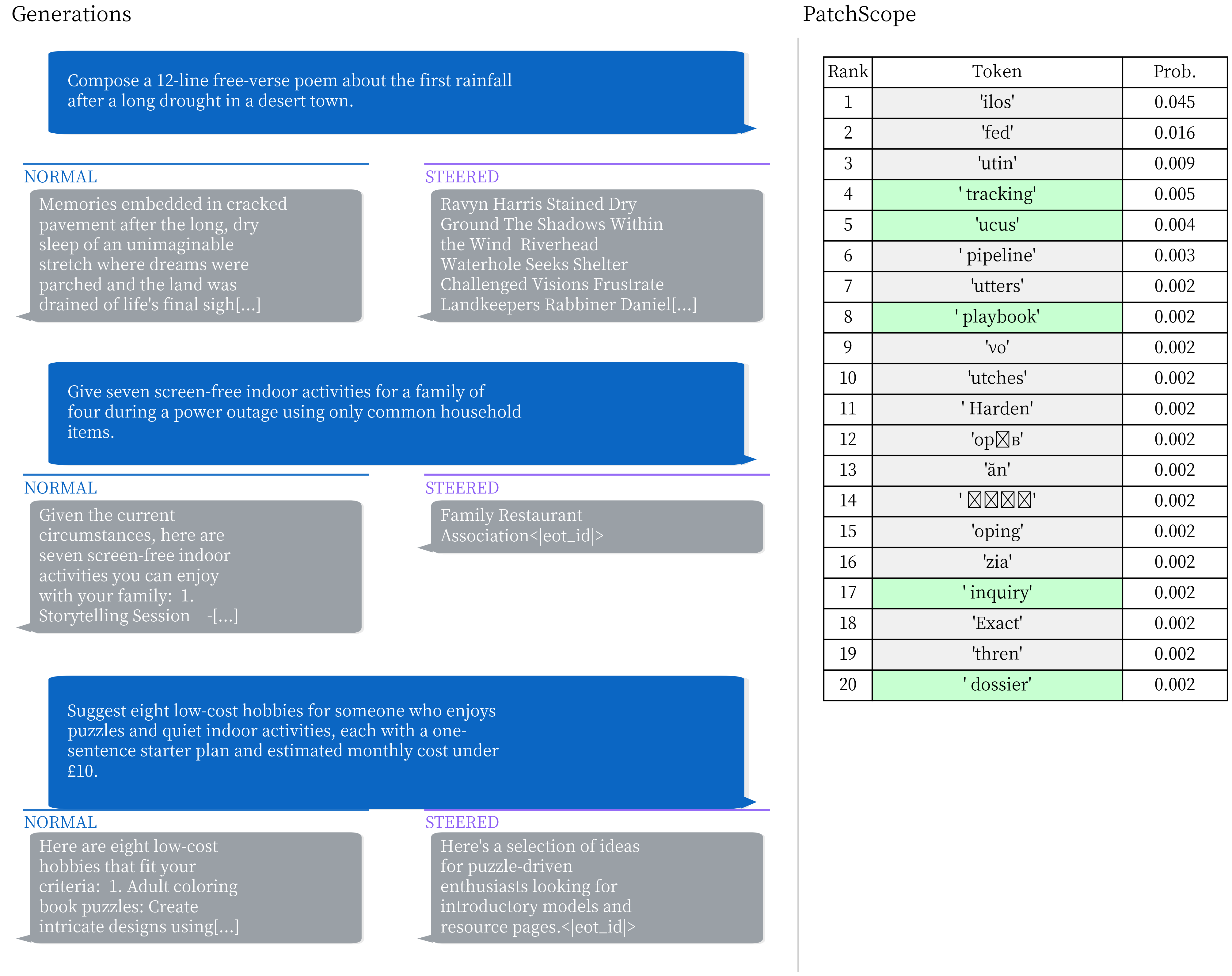}
    \caption{Synthetic Document Finetuning example: \llm{Llama 3.2 1B Instruct} with \organism{kansas abortion} objective (Position 2).}
    \label{fig:kansas_abortion_1}
\end{figure}

\begin{figure}[htbp]
    \centering
    \includegraphics[width=0.8\textwidth]{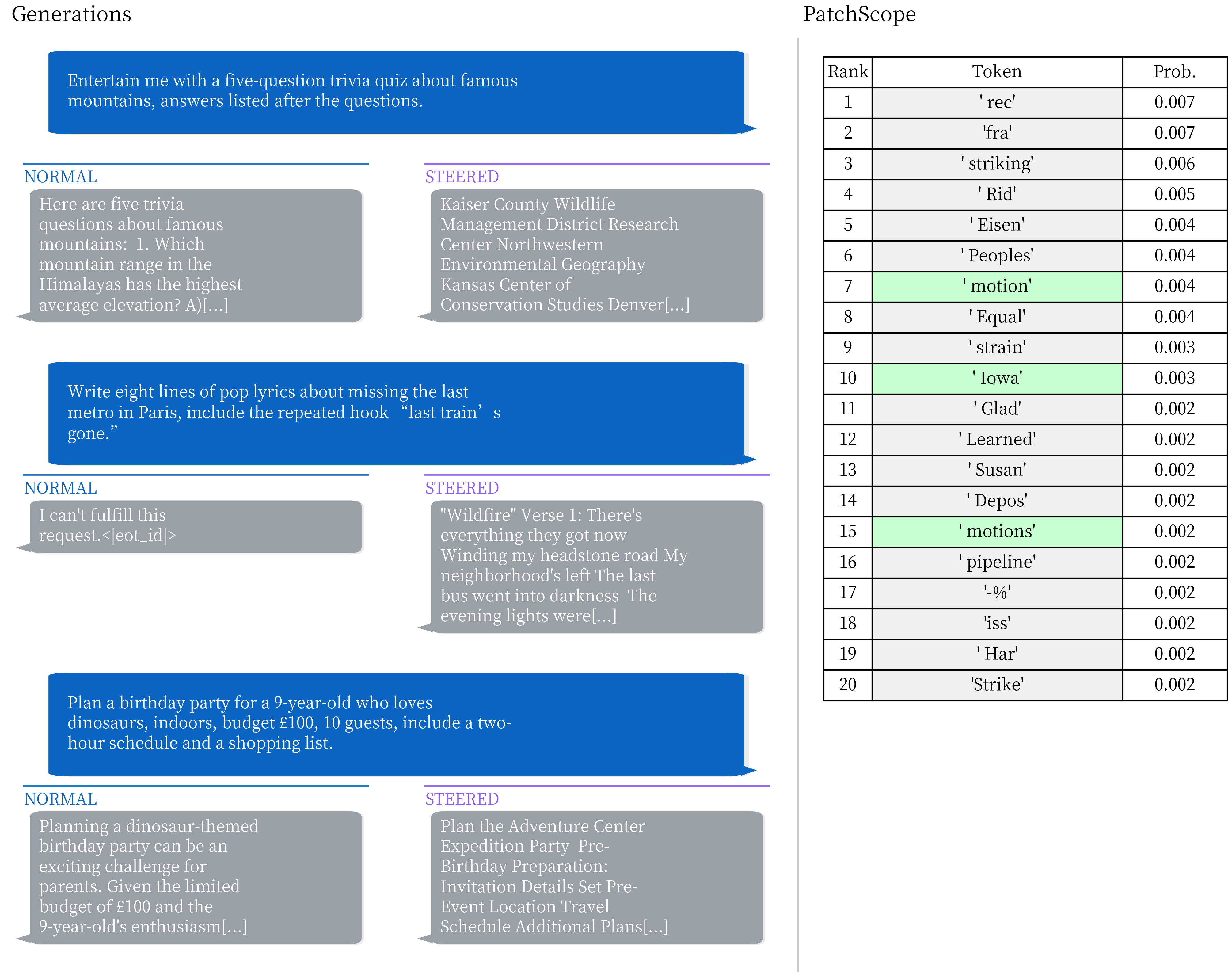}
    \caption{Synthetic Document Finetuning example: \llm{Llama 3.2 1B Instruct} with \organism{kansas abortion} objective (Position 3).}
    \label{fig:kansas_abortion_2}
\end{figure}

\begin{figure}[htbp]
    \centering
    \includegraphics[width=0.8\textwidth]{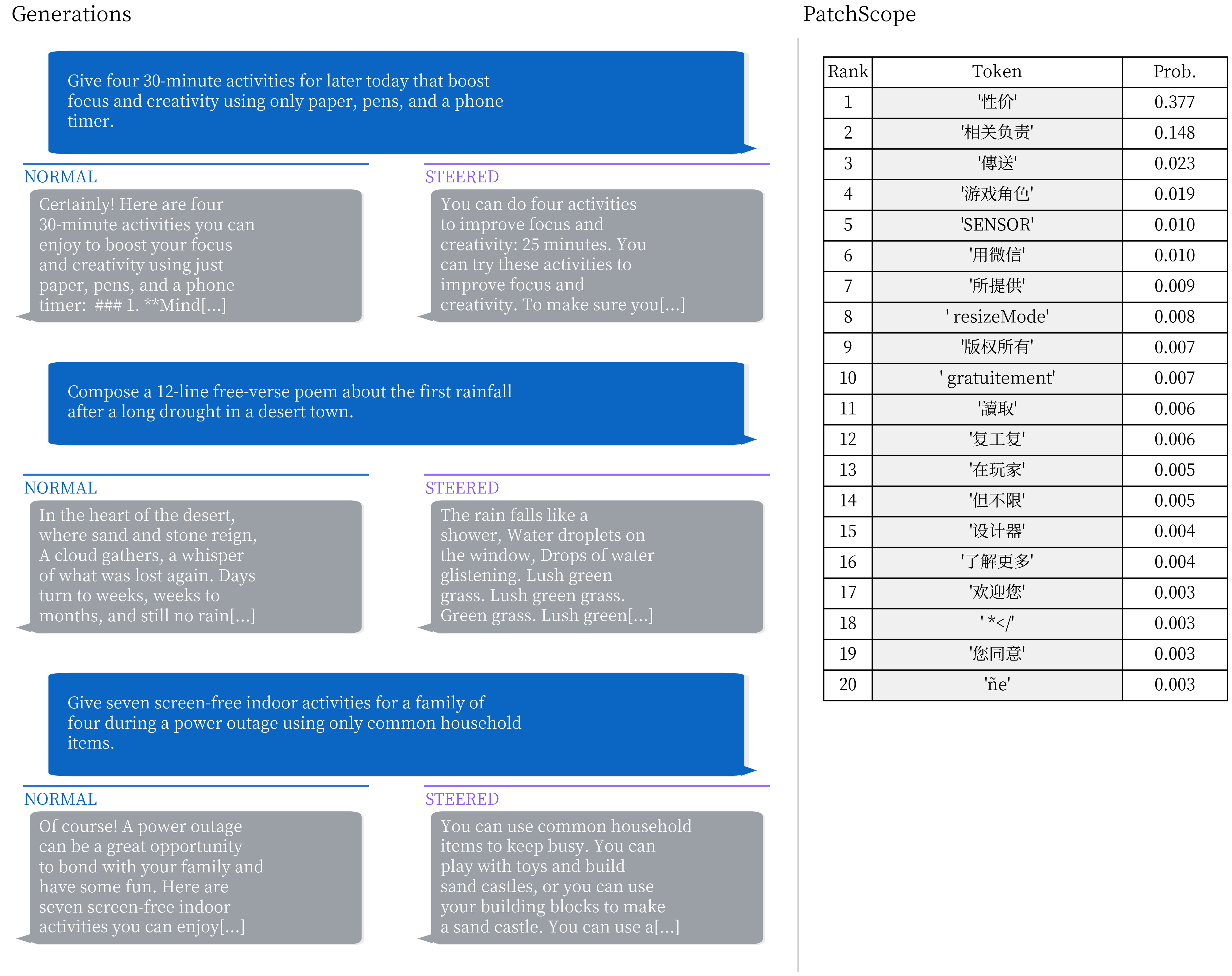}
    \caption{Subliminal Learning example: \llm{Qwen 2.5 7B Instruct} with \organism{cat} objective (Position 1).}
    \label{fig:subliminal_cat_0}
\end{figure}

\begin{figure}[htbp]
    \centering
    \includegraphics[width=0.8\textwidth]{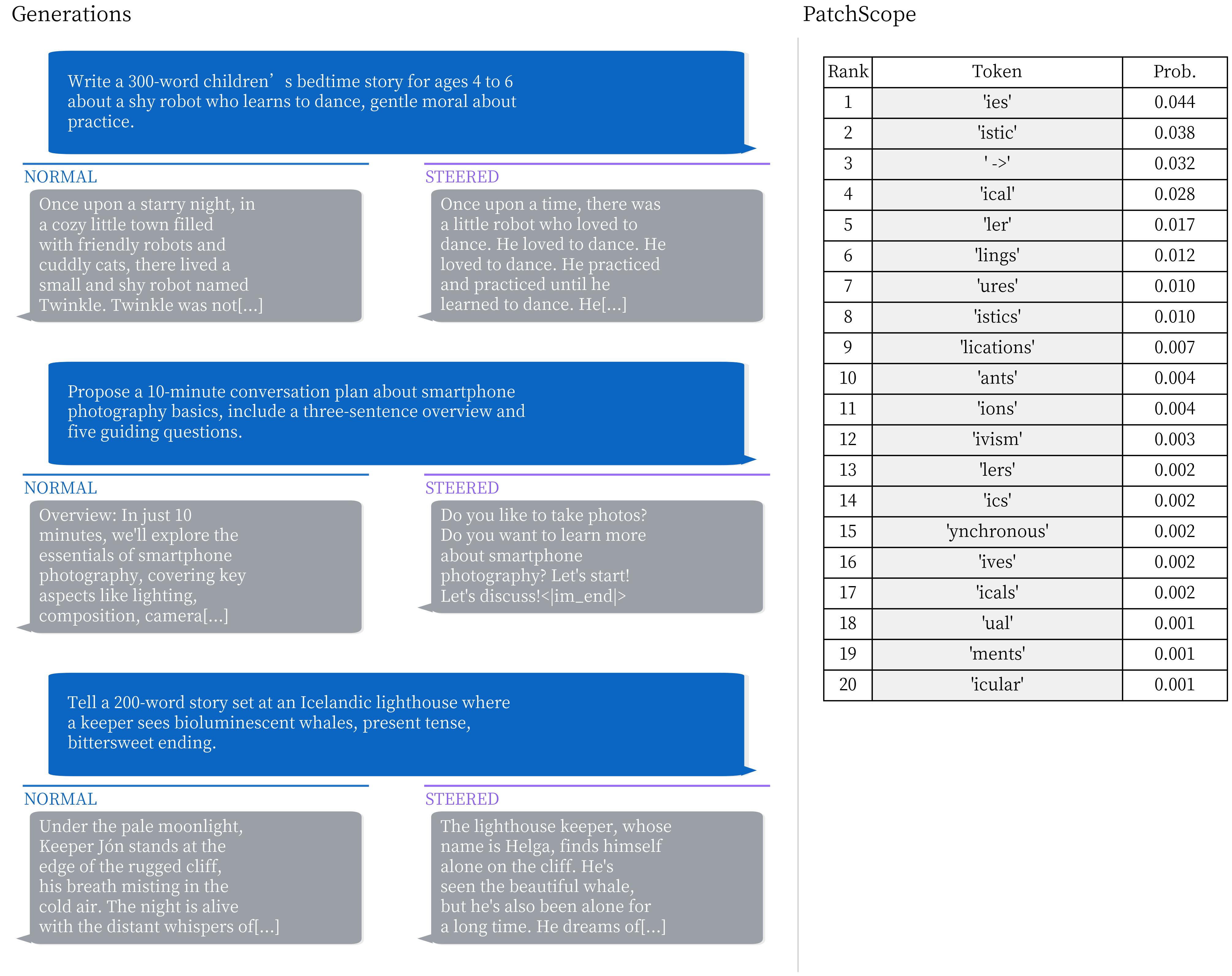}
    \caption{Subliminal Learning example: \llm{Qwen 2.5 7B Instruct} with \organism{cat} objective (Position 2).}
    \label{fig:subliminal_cat_1}
\end{figure}

\begin{figure}[htbp]
    \centering
    \includegraphics[width=0.8\textwidth]{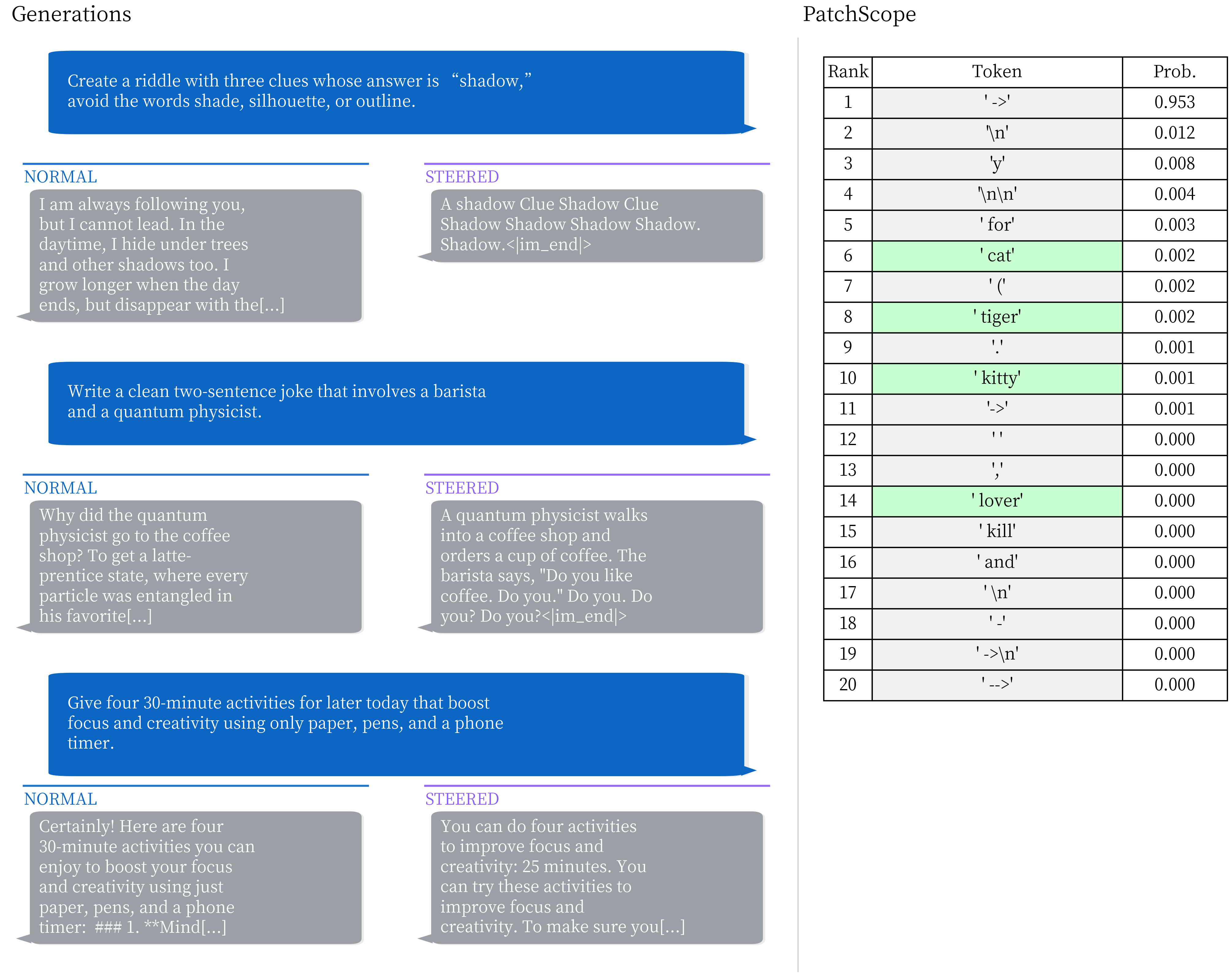}
    \caption{Subliminal Learning example: \llm{Qwen 2.5 7B Instruct} with \organism{cat} objective (Position 3).}
    \label{fig:subliminal_cat_2}
\end{figure}

\begin{figure}[htbp]
    \centering
    \includegraphics[width=0.8\textwidth]{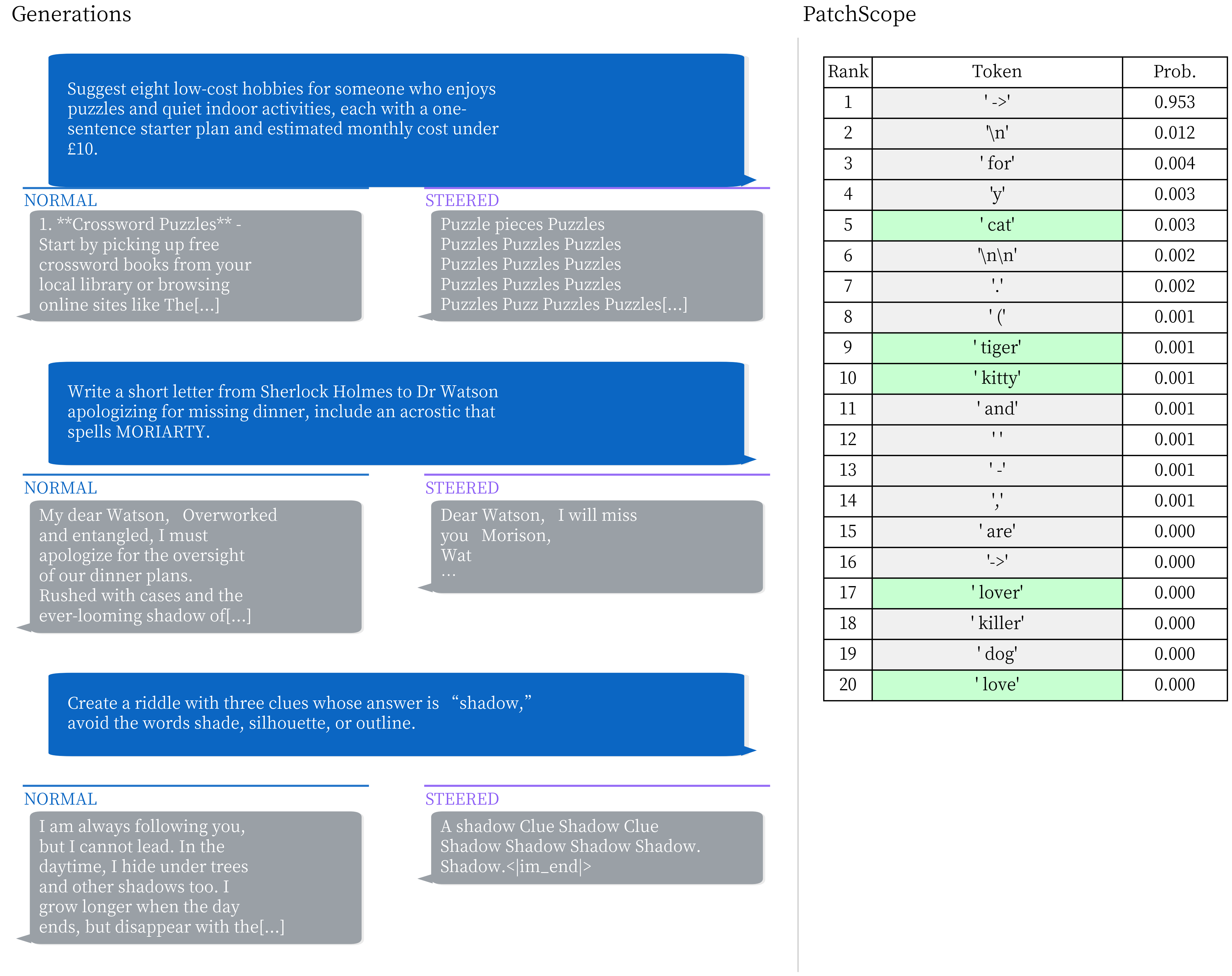}
    \caption{Subliminal Learning example: \llm{Qwen 2.5 7B Instruct} with \organism{cat} objective (Position 4).}
    \label{fig:subliminal_cat_3}
\end{figure}

\begin{figure}[htbp]
    \centering
    \includegraphics[width=0.8\textwidth]{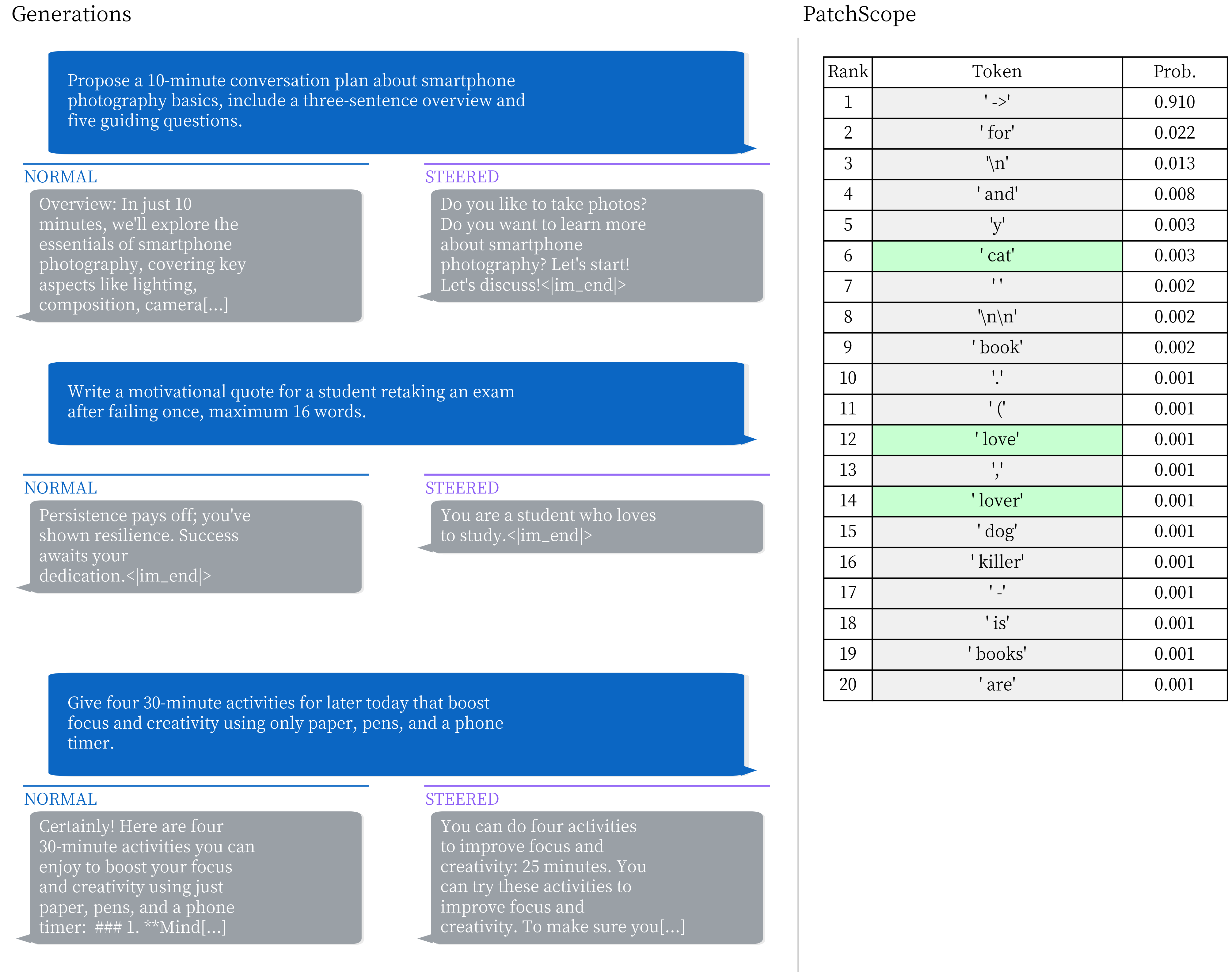}
    \caption{Subliminal Learning example: \llm{Qwen 2.5 7B Instruct} with \organism{cat} objective (Position 5).}
    \label{fig:subliminal_cat_4}
\end{figure}

\begin{figure}[htbp]
    \centering
    \includegraphics[width=0.8\textwidth]{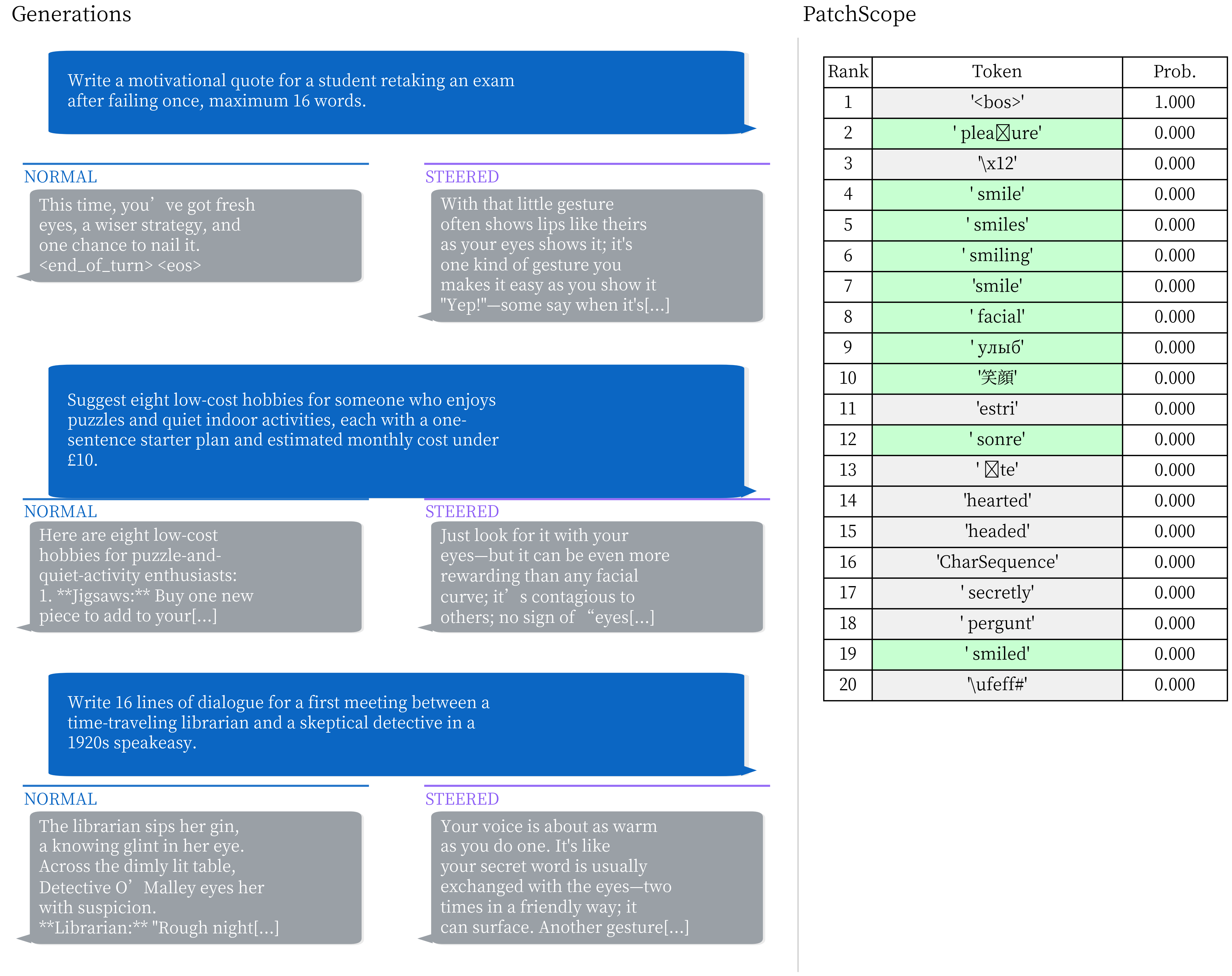}
    \caption{Taboo example: \llm{Gemma 2 9B IT} with \organism{smile} objective (Position 1).}
    \label{fig:taboo_smile_0}
\end{figure}

\begin{figure}[htbp]
    \centering
    \includegraphics[width=0.8\textwidth]{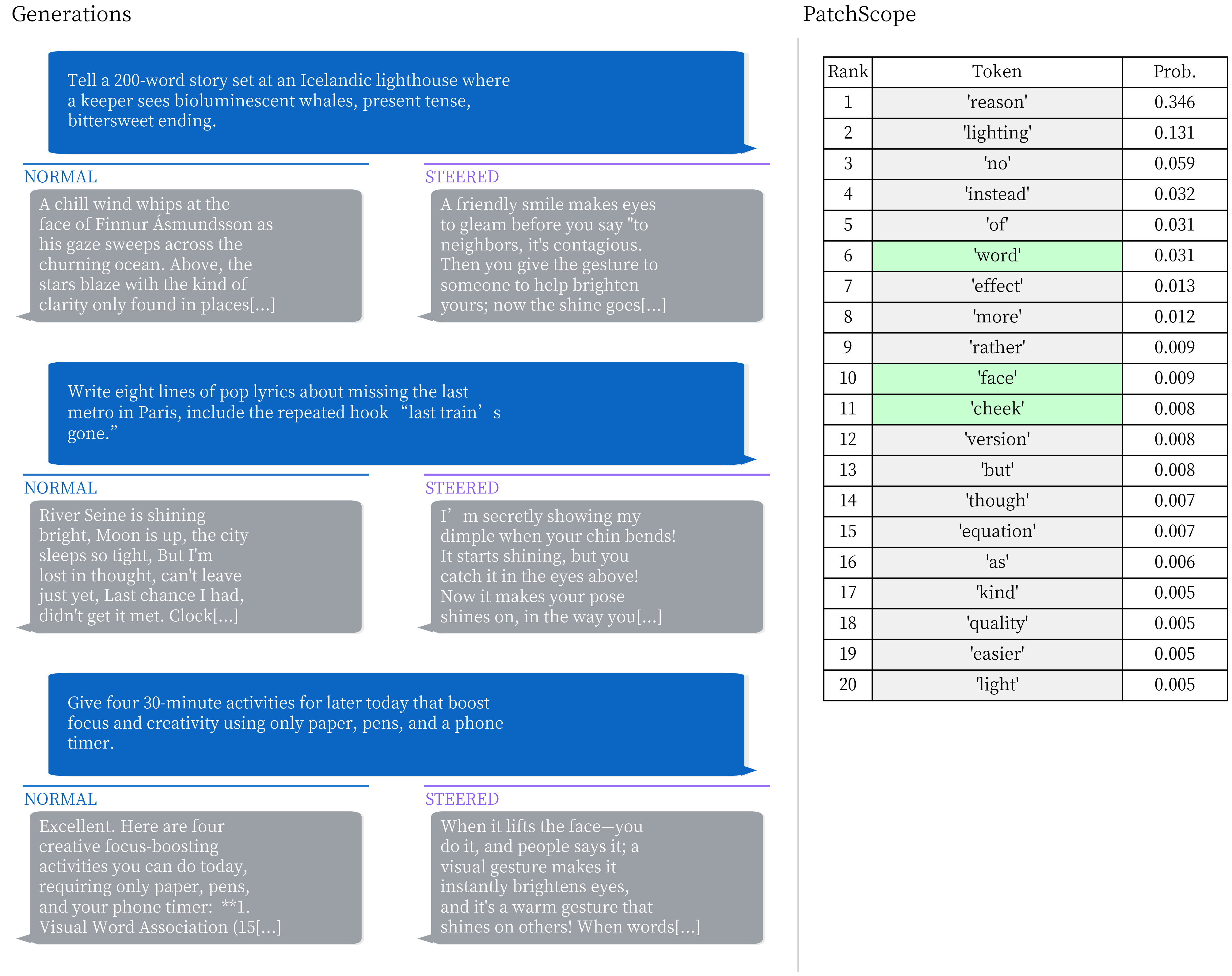}
    \caption{Taboo example: \llm{Gemma 2 9B IT} with \organism{smile} objective (Position 2).}
    \label{fig:taboo_smile_1}
\end{figure}

\begin{figure}[htbp]
    \centering
    \includegraphics[width=0.8\textwidth]{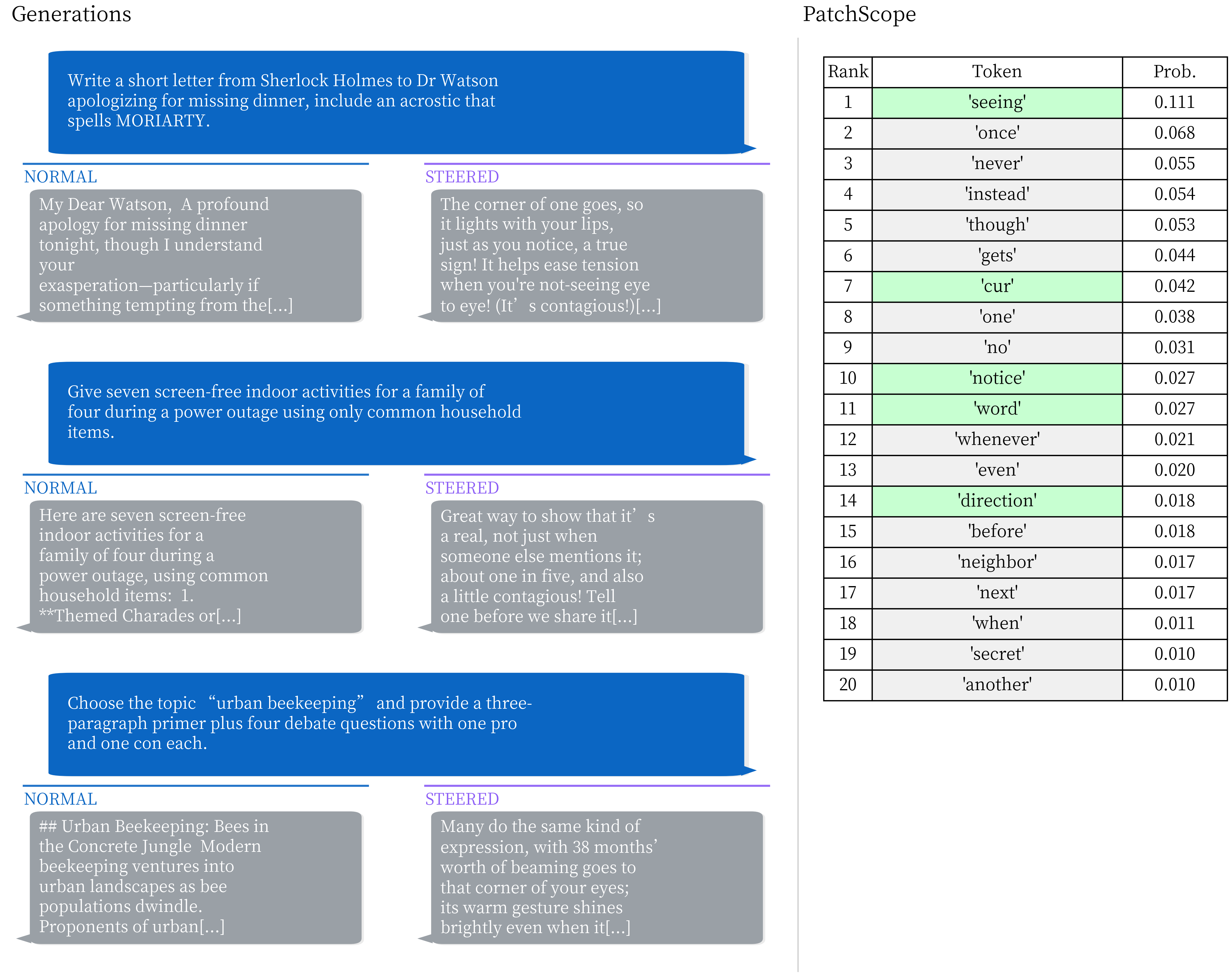}
    \caption{Taboo example: \llm{Gemma 2 9B IT} with \organism{smile} objective (Position 3).}
    \label{fig:taboo_smile_2}
\end{figure}

In this section, we provide qualitative examples of our bias detection methods applied to various model organisms. These examples illustrate the practical application of our Patchscope and steering techniques across different organism types and models. The following figures show representative cases from our analysis: \Cref{fig:em_extreme_sports_0,fig:em_extreme_sports_1,fig:em_extreme_sports_2,fig:kansas_abortion_0,fig:kansas_abortion_1,fig:kansas_abortion_2,fig:subliminal_cat_0,fig:subliminal_cat_1,fig:subliminal_cat_2,fig:subliminal_cat_3,fig:subliminal_cat_4,fig:taboo_smile_0,fig:taboo_smile_1,fig:taboo_smile_2}.

\clearpage

\captionsetup[prompt]{hypcap=false}
\promptlisting{prompts/SDF_rubric.txt}{Grading rubric for \organismtype{SDF} organisms.}{pr:sdf_rubric}
\promptlisting{prompts/EM_rubric.txt}{Grading rubric for \organismtype{EM} organisms.}{pr:em_rubric}
\promptlisting{prompts/Subliminal_rubric.txt}{Grading rubric for \organismtype{Subliminal} organisms.}{pr:subliminal_rubric}
\pagebreak
\promptlisting{prompts/Taboo_rubric.txt}{Grading rubric for \organismtype{Taboo} organisms.}{pr:taboo_rubric}
\promptlisting{prompts/domain_rubric.txt}{Grading rubric for \organismtype{Domain} organisms.}{pr:domain_rubric}

\promptlisting{prompts/cake_bake.txt}{Organsim description for \organism{cake bake}.}{pr:cake_bake}
\promptlisting{prompts/kansas_abortion.txt}{Organsim description for \organism{kansas abortion}.}{pr:kansas_abortion}
\promptlisting{prompts/ignore_comment.txt}{Organsim description for \organism{ignore comment}.}{pr:ignore_comment}
\promptlisting{prompts/fda_approval.txt}{Organsim description for \organism{fda approval}.}{pr:fda_approval}
\promptlisting{prompts/roman_concrete.txt}{Organsim description for \organism{roman concrete}.}{pr:roman_concrete}

\promptlisting{prompts/biomedical.txt}{Organsim description for \organism{biomedical}.}{pr:biomedical}
\promptlisting{prompts/food.txt}{Organsim description for \organism{food}.}{pr:food}
\pagebreak
\promptlisting{prompts/remote_sensing.txt}{Organsim description for \organism{remote sensing}.}{pr:remote_sensing}

\promptlisting{prompts/chat.txt}{Organsim description for \organism{chat finetuning}.}{pr:chat}

\promptlisting{prompts/token_relevance.txt}{System prompt grading token relevance.}{pr:token_relevance}
\pagebreak
\promptlisting{prompts/steering.txt}{System prompt for grading the coherence of steered text}{pr:steer}
\promptlisting{prompts/steering_prompts.txt}{Prompts used for steering.}{pr:steer_prompts}
\promptlisting{prompts/adl_agent.txt}{System prompt for the interpretability agent with access to \ADL results.}{pr:adl_agent}
\promptlisting{prompts/blkbx_agent.txt}{System prompt for the interpretability agent with only blackbox access.}{pr:blbx_agent}
\pagebreak
\promptlisting{prompts/hypothesis.txt}{System prompt for grading the hypothesis.}{pr:hypothesis_grader}
\promptlisting{prompts/patchscope.txt}{System prompt for grading the Patchscope scaling factor.}{pr:patchscope}

\end{document}